\documentclass{article}

\usepackage{xcolor}   

\usepackage[final]{corl_2026} 

\title{Combining Constrained Sampling and Reinforcement Learning for Manipulation}

\author{
  Marc Toussaint\textsuperscript{1,2,4},~
  Cornelius V.~Braun\textsuperscript{1},~
  Armand Jordana\textsuperscript{3},~
  Sayantan Auddy\textsuperscript{1},~\\
\textbf{Eckart Cobo-Briesewitz\textsuperscript{1},~
  Denis Shcherba\textsuperscript{1,4},~
  Tilman Burghoff\textsuperscript{1},~
  Justin Carpentier\textsuperscript{5}}\\
\textsuperscript{1}Technical University Berlin \quad
\textsuperscript{2}Robotics Institute Germany \\
\textsuperscript{3}LAAS-CNRS \quad
\textsuperscript{4}Fraunhofer HHI \qquad
\textsuperscript{5}Inria \& DI-ENS, PSL Research University\\
\texttt{toussaint@tu-berlin.de} \\
}

\usepackage{amsmath}
\usepackage{amssymb}
\usepackage{amsfonts}
\usepackage{amsthm}
\usepackage{eucal}
\usepackage{graphicx}
\usepackage{subcaption}

\usepackage{siunitx}
\graphicspath{{pics/}}

  \renewcommand{\a}{\alpha}
  \renewcommand{\b}{\beta}
  \renewcommand{\d}{\delta}
    
    \newcommand{\e}{\epsilon}
    \newcommand{\g}{\gamma}
    
  \renewcommand{\l}{\lambda}
  \renewcommand{\L}{\Lambda}
    \newcommand{\m}{\mu}

  \renewcommand{\k}{\kappa}
  \renewcommand{\o}{\omega}
  \renewcommand{\O}{\Omega}

  \renewcommand{\r}{\varrho}
    \newcommand{\s}{\sigma}
  
  \renewcommand{\t}{\theta}

  \renewcommand{\AA}{\mathcal{A}}

    \newcommand{\DD}{\mathcal{D}}

  \renewcommand{\SS}{\mathcal{S}}

    \newcommand{\UU}{\mathcal{U}}

  \newcommand{\RRR}{{\mathbb{R}}}

  \renewcommand{\[}{\Big[}

  \renewcommand{\|}{\,|\,}
  
  \renewcommand{\=}{\!=\!}

  \newcommand{\Id}{{\rm\bf I}}

  \newcommand{\st}{~~\text{s.t.}~~}

  \newcommand{\half}{\ensuremath{\frac{1}{2}}}

  \renewcommand{\vec}{\boldsymbol}

  \DeclareMathOperator*{\argmin}{argmin}

  \newcommand{\Exp}[2][]{\mathbb{E}_{#1}\!\left\{#2\right\}}

  \providecommand{\href}[2]{{\color{blue}USE PDFLATEX!}}
  \providecommand{\url}[2]{\href{#1}{{\color{blue}#2}}}
  \newcommand{\anchor}[2]{\begin{picture}(0,0)\put(#1){#2}\end{picture}}

  \newcommand{\mytitle}[1]{\title{#1}\newcommand{\thetitle}{#1}}
  \newcommand{\header}{\begin{document}\mytitle\cleardefs}
  \newcommand{\contents}{{\tableofcontents}\renewcommand{\contents}{}}
  \newcommand{\footer}{\small\bibliography{marc,bibs}\end{document}}
  \newcommand{\widepaper}{\usepackage{geometry}\geometry{a4paper,hdivide={25mm,*,25mm},vdivide={25mm,*,25mm}}}
  \newcommand{\moviex}[2]{\movie[externalviewer]{#1}{#2}} 
  \newcommand{\rbox}[1]{\fboxrule2mm\fcolorbox[rgb]{1,.85,.85}{1,.85,.85}{#1}}
  \newcommand{\mpage}[2]{{\begin{minipage}{#1\columnwidth}#2\end{minipage}}}
  \newcommand{\redbox}[2]{\fboxrule1mm\fcolorbox[rgb]{1,.7,.7}{1,.7,.7}{\begin{minipage}{#1\columnwidth}\center#2\end{minipage}}}
  \newcommand{\onecol}[2]{
    \begin{minipage}[c]{#1\columnwidth}#2\end{minipage}}
  \newcommand{\twocol}[5][0]{
    \begin{minipage}[c]{#2\columnwidth}#4\end{minipage}\hspace*{#1\columnwidth}%
    \begin{minipage}[c]{#3\columnwidth}#5\end{minipage}}
  \newcommand{\threecol}[7][0]{%
    \begin{minipage}[c]{#2\columnwidth}#5\end{minipage}\hspace*{#1\columnwidth}%
    \begin{minipage}[c]{#3\columnwidth}#6\end{minipage}\hspace*{#1\columnwidth}%
    \begin{minipage}[c]{#4\columnwidth}#7\end{minipage}}
  \newcommand{\threecoltext}[7][c]{
    \begin{minipage}[#1]{#2\textwidth}#5\end{minipage}%
    \begin{minipage}[#1]{#3\textwidth}#6\end{minipage}%
    \begin{minipage}[#1]{#4\textwidth}#7\end{minipage}}
  \newcommand{\threecoltop}[7][0]{%
   \begin{minipage}[t]{#2\columnwidth}#5\end{minipage}\hspace*{#1\columnwidth}%
   \begin{minipage}[t]{#3\columnwidth}#6\end{minipage}\hspace*{#1\columnwidth}%
   \begin{minipage}[t]{#4\columnwidth}#7\end{minipage}}
  \newcommand{\fourcol}[9][0]{%
   \begin{minipage}[c]{#2\columnwidth}#6\end{minipage}\hspace*{#1\columnwidth}%
   \begin{minipage}[c]{#3\columnwidth}#7\end{minipage}\hspace*{#1\columnwidth}%
   \begin{minipage}[c]{#4\columnwidth}#8\end{minipage}\hspace*{#1\columnwidth}%
   \begin{minipage}[c]{#5\columnwidth}#9\end{minipage}}
  \newcommand{\helvetica}[4]{\setlength{\unitlength}{1pt}\fontsize{#1}{#1}\linespread{#2}\usefont{OT1}{phv}{#3}{#4}}
  \newcommand{\helve}[1]{\helvetica{#1}{1.5}{m}{n}}
  \newcommand{\german}{\usepackage[german]{babel}\usepackage[utf8]{inputenc}}
\newcommand{\sans}{\renewcommand{\familydefault}{\sfdefault}}

\newcommand{\norm}[1]{|\!|#1|\!|}
\newcommand{\expr}[1]{[\hspace{-.2ex}[#1]\hspace{-.2ex}]}

\newcommand{\Jwi}{J^\sharp_W}
\newcommand{\THi}{T^\sharp_H}
\newcommand{\Jci}{J^\natural_C}
\newcommand{\hJi}{{\bar J}^\sharp}
\renewcommand{\|}{\,|\,}
\renewcommand{\=}{\!=\!}
\newcommand{\myminus}{{\hspace*{-.0pt}\text{\rm -}\hspace*{-.5pt}}}
\newcommand{\myplus}{{\hspace*{-.0pt}\text{\rm +}\hspace*{-.5pt}}}
\newcommand{\1}{{\myminus1}}
\newcommand{\2}{{\myminus2}}
\newcommand{\3}{{\myminus3}}
\newcommand{\mT}{{\text{\rm -}\hspace*{-1pt}\top}}
\newcommand{\po}{{\myplus1}}
\newcommand{\pt}{{\myplus2}}
\newcommand{\T}{{\!\top\!}}
\newcommand{\xT}{{\underline x}}
\newcommand{\uT}{{\underline u}}
\newcommand{\zT}{{\underline z}}
\newcommand{\Sum}{\textstyle\sum}
\newcommand{\Int}{\textstyle\int}
\newcommand{\Frac}{\textstyle\frac}
\newcommand{\Max}{\textstyle\max}
\newcommand{\Prod}{\textstyle\prod}

\newenvironment{centy}{
\vspace{15mm}
\large
\hspace*{5mm}
\begin{minipage}{8cm}\it\color{blue}
}{
\end{minipage}
}

\newcommand{\old}{{\text{old}}}
\newcommand{\new}{{\text{new}}}
\newcommand{\MAP}{{\text{MAP}}}
\newcommand{\ML}{{\text{ML}}}

\newcommand{\redArrow}{\quad\anchor{0,-1}{\includegraphics[scale=.5]{figs/redArrow}}}
\newcommand{\pub}[1]{{\color{green}\helvetica{8}{1.3}{m}{n}#1\\}}
\DeclareMathOperator{\opKL}{KL}
\newcommand{\KL}[2]{\opKL\big(#1\,\big|\!\big|\,#2\big)} 

\renewcommand{\show}[2][.7]{\centerline{\includegraphics[width=#1\columnwidth]{#2}}}
\newcommand{\showh}[2][.8]{\includegraphics[width=#1\columnwidth]{#2}}
\newcommand{\shows}[2][.8]{\centerline{\includegraphics[scale=#1]{#2}}}
\newcommand{\showhs}[2][.8]{\includegraphics[scale=#1]{#2}}
\newcommand{\mov}[2]{\movie[externalviewer]{{\color{blue}\small #1}}{movies/#2}}
\newcommand{\movex}[2]{\movie[externalviewer]{#1}{#2}} 
\newcommand{\movh}[3][autostart,loop]{
\movie[#1]{\showh[#2]{#3.jpg}}{#3.mp4}%
\movie[externalviewer]{$\circ$}{#3.mp4}
}
\newcommand{\movc}[3][autostart,loop]{\centerline{\movh[#1]{#2}{#3}}}
\newcommand{\mymovie}[3][]{\centerline{\movie[autostart,loop,#1]{\includegraphics[#1]{#2}}{#3}}}

\newcommand{\cen}[1]{\centerline{#1}}

\newcommand{\citing}[1]{
{\color{citcol}\ttiny#1\par}
}

\newcommand{\cit}[3]{
\par\smallskip
{\color{citcol}\ttiny #1: \emph{#2}. #3 \par}
}

\newcommand{\citurl}[4]{
\par\smallskip
{\color{citcol}\ttiny #1: \protect{\href{#4}{\color{blue}{#2.}}} #3 \par}
}

\newcommand{\cito}[3]{
\par\smallskip
{\color{bluecol}\ttiny #1: \emph{#2}. #3 \par}
}

\newcommand{\redoMacrosInProof}{
  \renewcommand{\d}{\delta}
  \renewcommand{\=}{\!=\!}
}

\newcommand{\pdflatex}{
  \definecolor{bluecol}{rgb}{0,0,.5}
  \usepackage[
    colorlinks,
    urlcolor=bluecol,
    citecolor=black,
    linkcolor=bluecol,
    pdfborder={0 0 0},
    pdfpagemode=UseOutlines, 
    pdfauthor={Marc Toussaint}
  ]{hyperref}
  \DeclareGraphicsExtensions{.pdf,.png,.jpg,.eps}
  \renewcommand{\r}{\varrho}
  \renewcommand{\l}{\lambda}
  \renewcommand{\L}{\Lambda}
  \renewcommand{\s}{\sigma}
  \renewcommand{\b}{\beta}
  \renewcommand{\d}{\delta}
  \renewcommand{\k}{\kappa}
  \renewcommand{\t}{\theta}
  \renewcommand{\O}{\Omega}
  \renewcommand{\o}{\omega}
  \renewcommand{\SS}{\mathcal{S}}
  \renewcommand{\=}{\!=\!}
}

\newenvironment{code}{%
\codefont
\begin{shaded}
}{
\end{shaded}
}

\newcommand{\signature}{
  \includegraphics[height=6ex]{signatureColor} 
}

\usepackage{algorithm}
\usepackage[noend]{algpseudocode}
\algrenewcommand{\algorithmicrequire}{\textbf{Input:~~}}
\algrenewcommand{\algorithmicensure}{\textbf{Output:}}
\algrenewcommand{\algorithmiccomment}[1]{\qquad\hfill~\hspace*{-5ex}\textit{// #1}}
\algrenewcommand{\alglinenumber}[1]{\helvetica{6}{1.3}{m}{n}#1:}

\newenvironment{algo}[1][8]{
\quad\begin{minipage}{.8\columnwidth}\helvetica{#1}{1.4}{m}{n}
\medskip\hrule\medskip
\begin{algorithmic}[1]
}{
\end{algorithmic}
\medskip\hrule\medskip
\end{minipage}
}

\usepackage{etoolbox}

\usepackage{multirow}
\usepackage{colortbl}
\usepackage{empheq}

\newcommand{\mypause}{\pause}
\newcommand{\defi}[1]{\textbf{#1}}
\newcommand{\red}[1]{\emph{\color{red}#1}}
\newcommand{\pos}{{\textsf{pos}}}
\newcommand{\eff}{{\textsf{eff}}}
\newcommand{\rot}{{\textsf{rot}}}
\newcommand{\veC}{{\textsf{vec}}}
\newcommand{\quat}{{\textsf{quat}}}
\newcommand{\col}{{\textsf{col}}}
\newcommand{\de}[2]{\frac{\partial #1}{\partial #2}}
\newcommand{\target}{{\text{target}}}
\newcommand{\near}{{\text{near}}}
\newcommand{\qfree}{Q_{\text{free}}}
\renewcommand{\vec}{\boldsymbol}
\newcommand{\lft}{\text{left}}
\newcommand{\rgh}{\text{right}}
\DeclareMathOperator{\real}{real}
\newcommand{\prev}{{\text{prev}}}
\newcommand{\TR}[2]{T_{{#1}\to{#2}}}
\newcommand{\RO}[2]{R_{{#1}\to{#2}}}
\newcommand{\liter}{\helvetica{8}{1.1}{m}{n}\parskip 1ex}
\newcommand{\Fc}{\color{green}F}
\newcommand{\muc}{\color{blue}\mu}
\newcommand{\Astar}{A$^*$}

\providecommand{\defn}[1]{\textbf{#1}}

\newcommand{\adec}{\r_\a^-}
\newcommand{\ainc}{\r_\a^+}
\newcommand{\ldec}{\r_\l^-}
\newcommand{\linc}{\r_\l^+}
\newcommand{\minc}{\r_\m^+}
\newcommand{\mdec}{\r_\m^-}
\newcommand{\step}{{\delta}}
\newcommand{\lsstop}{\r_{\text{ls}}}
\newcommand{\stepmax}{\step_{\text{max}}}

\definecolor{boxcol}{rgb}{.85,.9,.92}
\newcommand{\eqbox}[1]{\centerline{\fboxrule0mm\fcolorbox{boxcol}{boxcol}{#1}}}
\newcommand{\movgb}[1]{\hfill\movie[externalviewer]{\small[movie]}{/home/mtoussai/movies/10-goalDirectedBehavior/#1}}
\newcommand{\demo}[1]{{{\color{blue}[\small #1]}}}

\graphicspath{{../pics-robotics/}{../pics-ML/}{../pics-all/}{../pics-all2/}{../pics-Optim/}}
\DeclareGraphicsExtensions{.pdf,.png,.jpg}

\newcommand{\SUM}{\texttt{sum}}
\usepackage{float}

\makeatletter
\newcommand{\NewParNoBreak}[1][\parskip]{\par\vspace*{-\parskip}\vspace*{#1}\nobreak\@afterheading}
\makeatother

\usepackage[textsize=tiny,disable]{todonotes} 
\newcommand{\marc}[1]{\todo[color=blue!25]{MT: #1}}
\newcommand{\cvb}[1]{\todo[color=red!25]{CB: #1}}
\newcommand{\sa}[1]{\todo[color=magenta!25]{SA: #1}}
\newcommand{\armand}[1]{\todo[color=green!25]{AJ: #1}}
\newcommand{\tb}[1]{\todo[color=orange!25]{tb: #1}}

\begin{document}
\maketitle


\begin{abstract}
Training manipulation policies in contact-rich settings is a core challenge in robotics.
While Reinforcement Learning (RL) has demonstrated its strength in such settings, it may struggle to sufficiently explore and discover complex manipulation strategies.
To address this, we combine two basic ideas:
First, designing appropriate reset strategies (the start state distribution of episodes) has shown promise in improving RL exploration and effectiveness.
Second, while model-based approaches to finding \emph{trajectories} through manipulation are hard, recent work showed that model-based approaches to \emph{sampling states} on constrained manifolds can be highly efficient.
Based on these observations, we propose a novel state sampler that enables efficient goal-conditioned RL in complex contact-rich manipulation tasks.
Our sampler explicitly takes into account the structure of contact in order to provide a rich covering of diverse contact modes.
By combining constrained sampling resets with projected interpolation and curriculum learning, our novel approach outperforms RL without constrained sampling and alternative reset methods, and effectively trains universal, prehensile and non-prehensile manipulation policies in contact-rich whole-body and in-hand settings.
See the project page\footnote{https://www.user.tu-berlin.de/mtoussai/26-CSRL/} for videos and full source code access.
\end{abstract}



\section{Introduction}

Contact-rich manipulation is a core open challenge in robotics, where the aim are general skills to create the necessary contacts to achieve any goal \cite{1985-mason-MechanicsManipulation}.
Applying model-based solvers directly to optimize robotic manipulation sequences through contact mode switches is considered challenging, in particular due to the non-linearities and the complementarity structure inherent to contact dynamics, implying a basic discontinuity in controllability over the object when creating and breaking contacts
\cite{2014-posa-DirectMethodTrajectorya,2024-aydinoglu-ConsensusComplementarityControl,2014-dai-WholebodyMotionPlanning,2020-toussaint-DescribingPhysicsPhysical,Grandia-RSS-24}.
Reinforcement Learning (RL) has demonstrated its strength in contact-rich cases \cite{rubiks2019}, arguably due to its implicit smoothing of these underlying discontinuities \cite{2024-lelidec-LeveragingRandomizedSmoothing,2022-suh-BundledGradientsContact}.
However, RL still faces a severe exploration problem, since the relevant interaction modes often require reaching rare contact states before receiving informative reward~\cite{2016-osband-DeepExplorationBootstrappeda,2025-urpi-EpistemicallyguidedForwardbackwardExploration}.

A common strategy to overcome these challenges is modifying the environment.
Prior work has shaped rewards toward interaction-relevant events such as touching or grasping~\cite{2019-merzic-LeveragingContactForces}, or added exploration bonuses~\cite{2018-haarnoja-SoftActorcriticOffpolicy}.
An alternative approach is \emph{reset design}, i.e.\ modifying the reset distribution $p(s_0)$ from which episodes are initialized.
This is especially important because it determines which parts of the state space the policy interacts with, which contact modes can be practiced, and how challenging exploration is for the algorithm.
Existing approaches exploit this using reverse curriculum generation~\cite{2017-florensa-ReverseCurriculumGeneration}, resetting to randomly generated or previously visited states \cite{2018-zeng-LearningSynergiesPushing,2021-ecoffet-FirstReturnThen}, or learned reset samplers specialized to reaching and grasping~\cite{yin2026emergentdexteritydiverseresets, zhang2026balanced}.
While these methods demonstrate that the reset distribution is a key lever for RL on challenging manipulation tasks, they construct it through heuristics or task-specific mechanisms, such as forward simulation from random states, random scene generation, or grasp-specific samplers.
As a result, they do not provide a general way to specify the diverse contact configurations needed for non-prehensile manipulation.

In this work, our contributions are
(i) introducing a novel constrained state sampler for contact-rich manipulation that provides a reset and goal distribution for goal-conditioned RL,
(ii) a novel interpolation and curriculum reset strategy,
and (iii) experimental validation that our approach improves the performance of RL for contact-rich manipulation tasks.

\begin{figure}[tb]\centering\small
  (a)\hspace{-5mm}%
  \begin{minipage}[b]{.45\columnwidth}\vspace{-8pt}
    \includegraphics[width=1.\columnwidth]{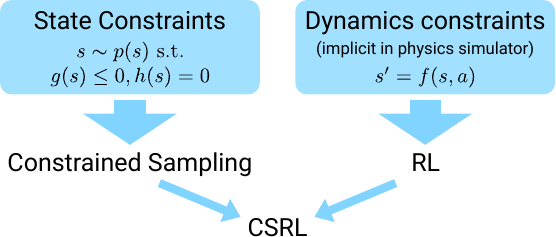}
  \end{minipage}
  \quad(b)~%
  \begin{minipage}[b]{.45\columnwidth}\vspace{-8pt}
\newcommand{\pic}[1]{\includegraphics[width=.33\columnwidth,trim={0 0 0 0pt},clip]{data/configs/images/panda_sphere_sample_0000#1.jpg}}
\pic{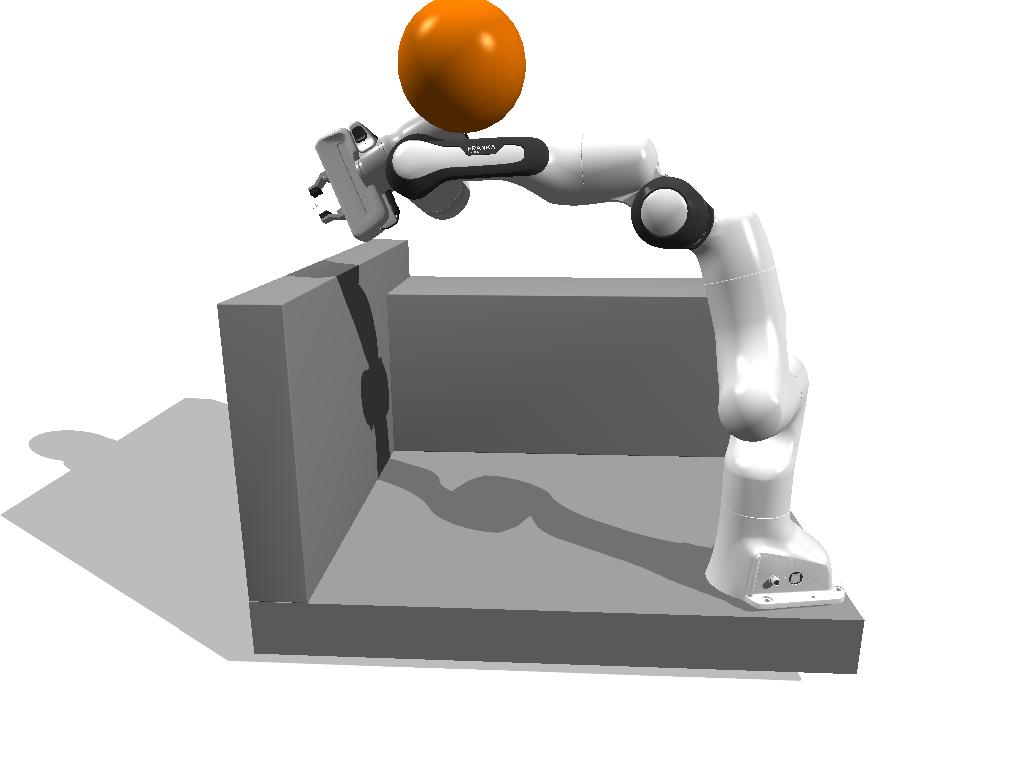}
\pic{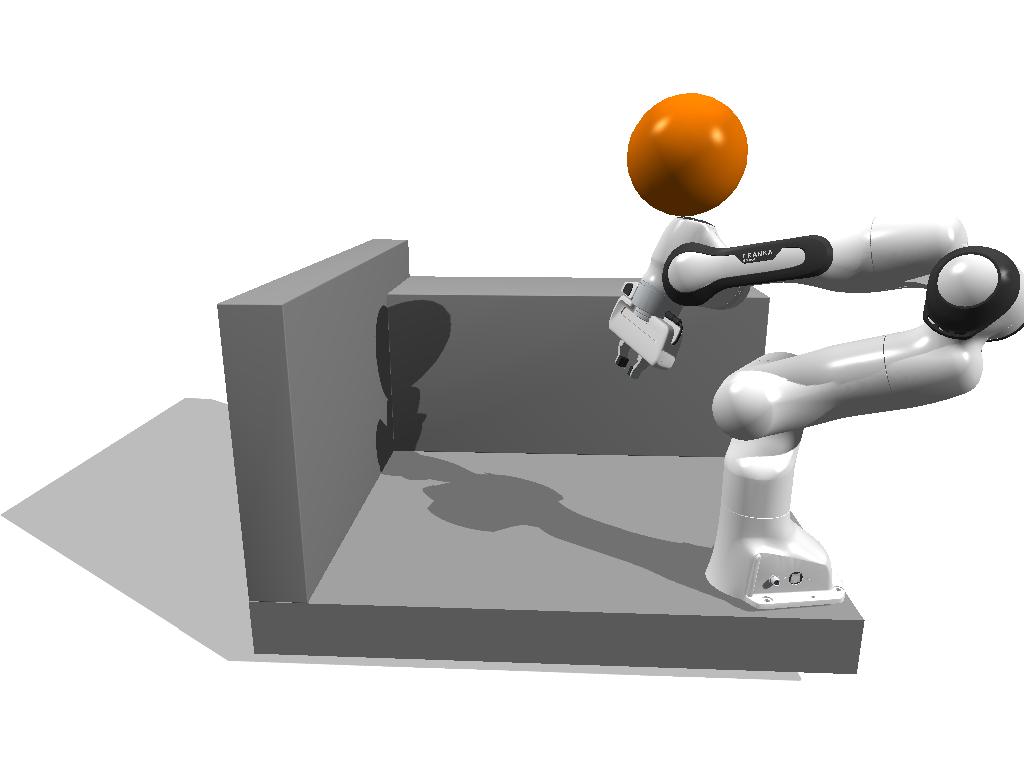}\pic{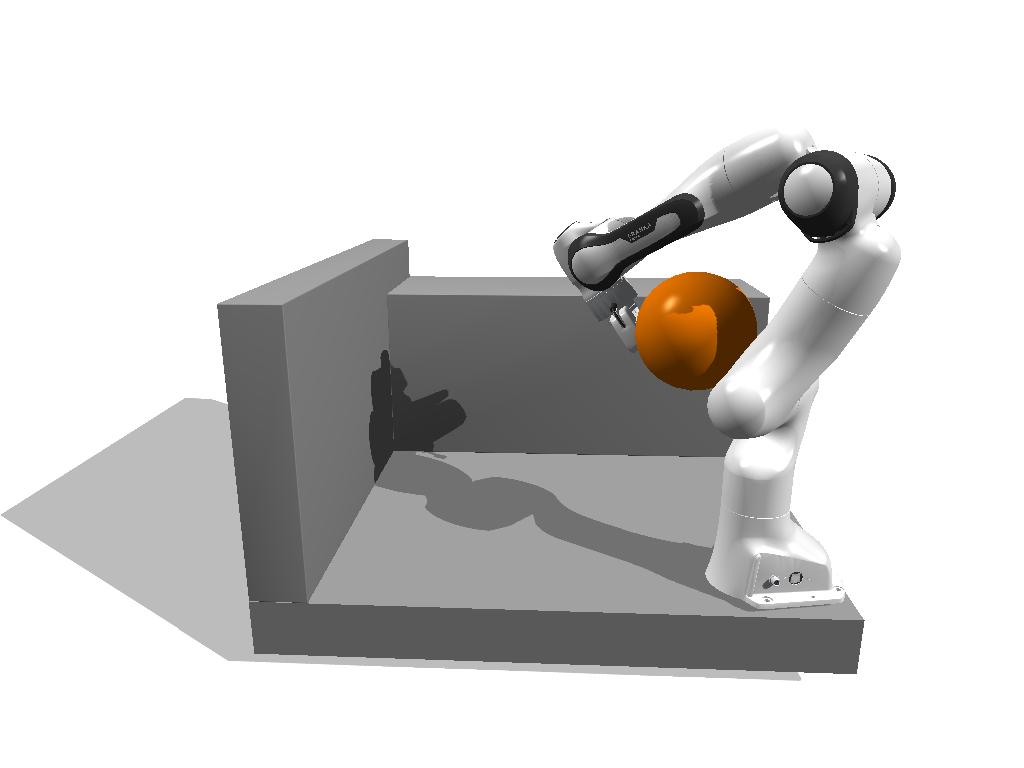}

\renewcommand{\pic}[1]{\includegraphics[width=.33\columnwidth,trim={0 0 0 25pt},clip]{data/configs/images/panda_cube_sample_0000#1.jpg}}
\pic{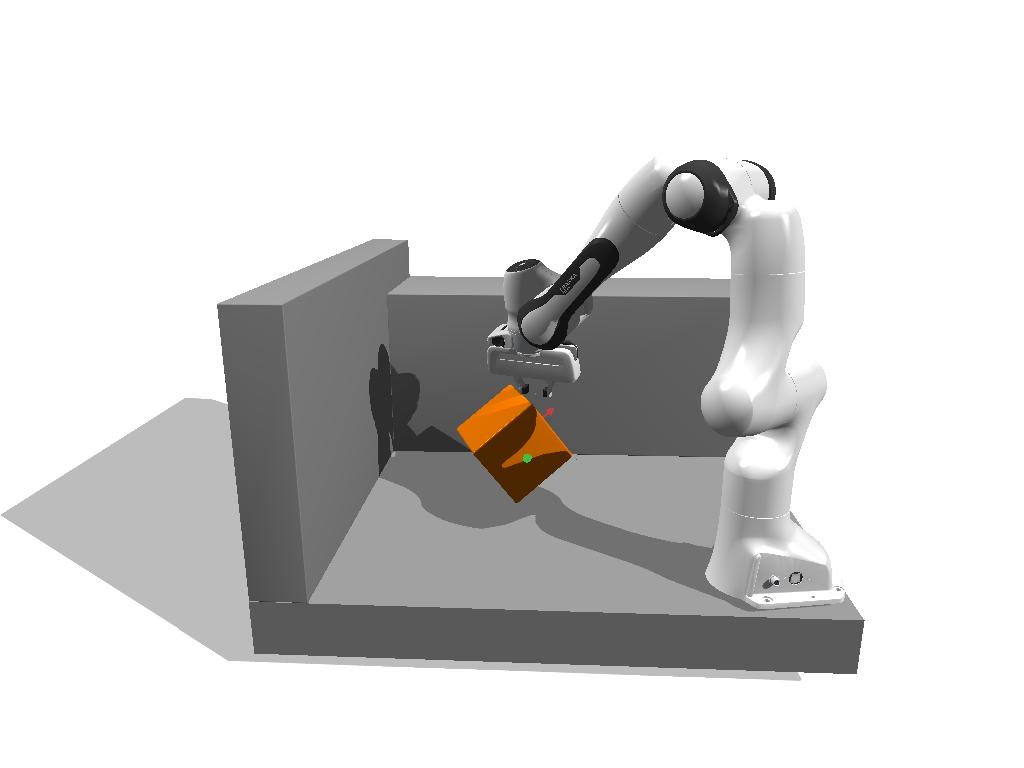}\pic{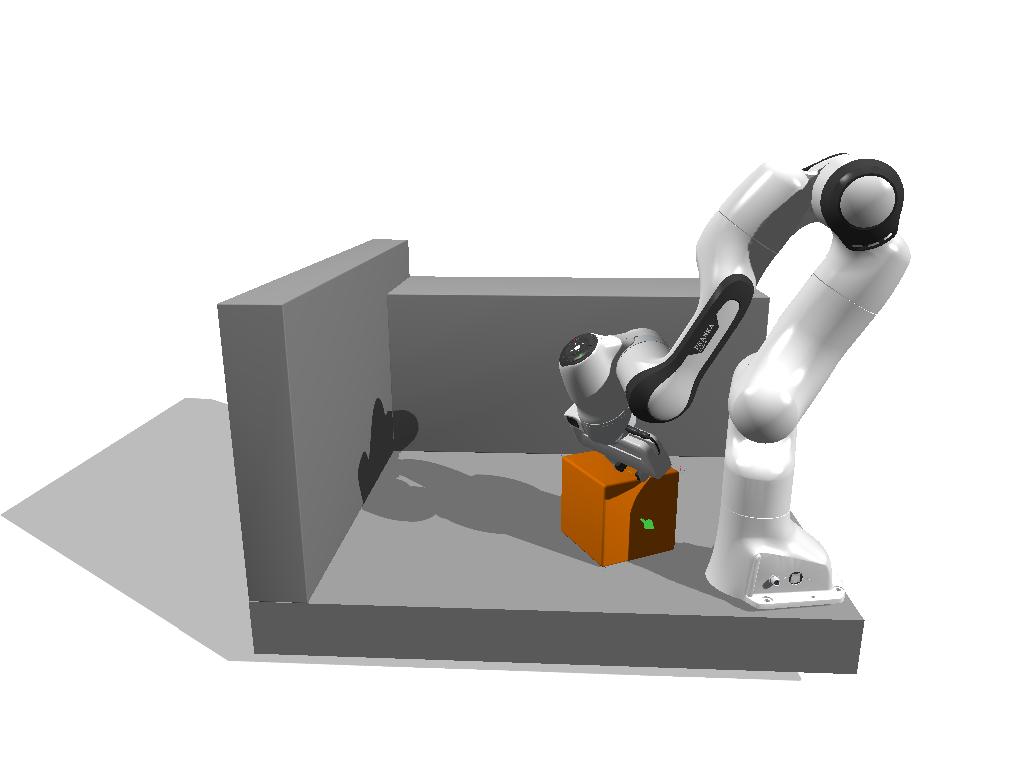}
\pic{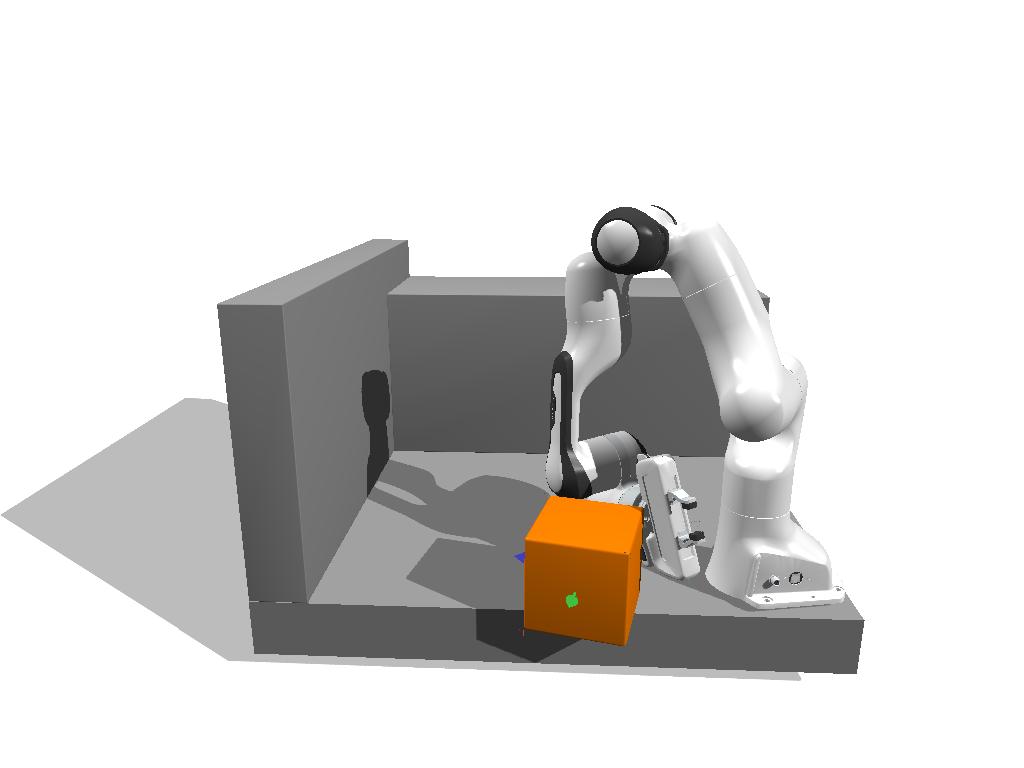}
  \end{minipage}%
  \caption{\label{fig:CSRL}
(a) Complementary to a physics simulator that samples dynamics, we leverage a physics-informed state sampler to globally sample starts and goals. State constraints account for the structure of contact, including coverage of diverse contact modes. (b) Example constrained samples. }
\end{figure}

More specifically, we address training goal-conditioned (or universal \cite{2015-schaul-UniversalValueFunction,2017-andrychowicz-HindsightExperienceReplay}) policies $\pi:s,g \mapsto a$ that control from any starting state $s\sim p(s)$ to any goal $g \sim p(g)$ of the same distribution.
We define that distribution to diversely cover contact modes and manipulation states using a constraint formulation.
The constrained sampler can be viewed as complementary to a physics simulator (Fig.~\ref{fig:CSRL}):
While a physics simulator represents first principles of \emph{dynamics} which is sampled by RL, our sampler encodes first principles of diverse manipulation \emph{states} that may be reachable.
By letting the sampler impose the goal and reset distribution, we can use off-the-shelf RL methods to learn a universal policy to connect such states while mitigating the exploration challenge.
To further improve efficiency in sparse-reward tasks, we augment sampled states using constraint-projected interpolation and combine this with curriculum learning.

We evaluate and demonstrate our approach on a series of goal-conditioned manipulation problems of increasing difficulty, training skills to reach any goal state that can be held static.
The policies we discover are highly dynamic, robust, and capable of using whole-body and environmental contacts to reach a broad range of goals. 
We confirm the effectiveness of our approach to training manipulation skills by comparing it to alternative reset strategies.
For reference, Appendix~\ref{app:BC} reports on negative results on an alternative approach using behavior cloning from diverse open-loop trajectories in our setting.

\section{Related Work}\label{sec:relatedWork}


\paragraph{Model-based Manipulation Planning.}

The literature on model-based robotic manipulation planning reflects a rich understanding of the structure of manipulation in rigid body domains.
This includes the exploitation of environmental constraints \cite{lozano2003simple, simeon2004manipulation},
explicit methods (which explicitly search over contact modes or introduce them as discrete decision variables) \cite{mastalli2020crocoddyl, deits2014footstep, 2020-toussaint-DescribingPhysicsPhysical, lamiraux2021prehensile}, and implicit methods (formulations without discrete variables) \cite{2014-posa-DirectMethodTrajectorya, mordatch2012discovery}.
However, contact dynamics are inherently stiff and non-smooth and involve complementarities, and therefore pose a fundamentally hard challenge to gradient-based solvers.
Pure shooting methods \cite{jordana2025introductionzeroorderoptimizationtechniques, simeon2004manipulation, tassa2012synthesis} as well as RL \cite{rubiks2019} instead treat dynamics as a black-box simulation and avoid these challenges, but they do not exploit the inherent structure of rigid body and contact dynamics.
Our approach aims to bridge this gap by integrating the aforementioned classical formulations of contact interaction to define a constrained sampler of reset states to guide RL.

\paragraph{Goal-conditioned RL \& Universal Policies.}

In universal value function learning the aim is to learn policies that maximize rewards under any task in a fixed MDP (i.e., with fixed objects)~\cite{2015-schaul-UniversalValueFunction,2017-andrychowicz-HindsightExperienceReplay, gong2024goal}.
We follow this approach as we train a universal $Q$-function and policy to reach any goal state from any start state.
Several approaches have been proposed to exploit the structure of goal-conditioned RL and improve sample efficiency, including reward relabeling~\cite{2017-andrychowicz-HindsightExperienceReplay, bai2019guided, huang2024mrher}, reward shaping~\cite{trott2019keeping}, hierarchical goals~\cite{nachum2018data, levy2017hierarchical, nair2018visual, paul2019learning, campero2021amigo, chane2021goal}, intrinsic motivation~\cite{liu2022learn, lintunen2024diversity}, and curriculum-based training~\cite{fang2019curriculum, gong2024goal}.
All of these works have in common that they assume that start and goal states can trivially be sampled as part of the problem definition, which is limiting in the case of non-prehensile and contact-rich manipulation.
To address this we propose non-linear constraint solvers to sample from the model-based constraint manifold that defines feasible starts and goals.

\paragraph{RL for Robot Manipulation.}
Robot manipulation is a challenging application of RL due to the multi-contact long-horizon nature of the tasks, the uncertainty resulting from multiple contact interactions, and perception challenges related to most tasks.
Several works have demonstrated the effectiveness of RL in addressing these challenges.
\citet{ferrandis2023nonprehensile} and \citet{braun2026trajectory} focus on exploration to explore through contacts, thereby enabling object reorientation with a robot gripper.
Another line of work investigates representation learning for non-prehensible manipulation~\cite{kim2023pre, zhou2023hacman, cho2024corn}. 
More recently, \cite{shirai2026learning} leveraged contact-implicit trajectory-optimization to generate demonstrations for an RL policy. 
While these prior approaches facilitate learning for manipulation, our work takes a complementary approach by focusing on the construction of diverse reset and goal distributions.
In addition, we do not rely on dense rewards and consider whole-body object interactions without any human reference motion.


\paragraph{Goal and Reset Distributions in RL.}

A line of work has shown that the distribution over starts and goals can be as important as the RL algorithm itself.
Reverse curriculum generation expands the initial-state distribution outward from the goal by sampling states from which the current policy can still make progress~\cite{2017-florensa-ReverseCurriculumGeneration}.
Similarly, other methods propose to sample starts or goals from models that are trained on the replay buffer~\cite{pong2020skew}.
Related exploration methods such as Go-Explore build an archive of previously visited states and repeatedly reset to promising cells to explore further from them~\cite{2021-ecoffet-FirstReturnThen}.
Asymmetric Self-Play~\citep{openai2021asymmetric} proposes an adversarial approach to robot manipulation learning, which constructs a curriculum of goal states in an unsupervised fashion.
While these methods demonstrate the importance of resets in RL, they only expand the state visitation distribution locally, which can be insufficient for challenging manipulation problems.
Hence, works on robot manipulation propose to construct reset and goal distributions more directly, for example, by randomly generating cluttered scenes through simulation~\cite{2018-zeng-LearningSynergiesPushing, lum2026play2perfect} or by using grasp samplers to obtain states useful for prehensile manipulation~\cite{khandate2023sampling, yin2026emergentdexteritydiverseresets, zhang2026balanced}, with limited applicability for non-prehensile and whole-body manipulation.
In contrast, our work constructs the reset and goal distribution from general constraints on manipulation states, allowing us to expose the policy to diverse contact configurations without committing to a specific interaction primitive.



\section{Methods}

\subsection{Problem Formulation}

We define a \emph{constrained goal-conditioned MDP} as a variant of a standard goal-conditioned MDP \cite{2015-schaul-UniversalValueFunction,2017-andrychowicz-HindsightExperienceReplay}, where we have a state space $\SS \subseteq \RRR^d$, an action space $\AA$, dynamics $p(s'|s,a)$, a goal-conditioned reward function $R(s',s,g)$, discount factor $\g\in (0,1)$, as well as constraint functions $g_c, h_c$ which define the constraint space
\begin{align}
  \SS_c = \{ s\in\SS : g_c(s) \le 0, h_c(s)=0 \}, \label{eqCon}
\end{align}
where $g_c: \RRR^d \to \RRR^{g_c}$ and $h_c: \RRR^d \to \RRR^{h_c}$ and their Jacobians can be queried point-wise.
Given $\SS_c$, we define a distribution $p_0(s) = \UU(\SS_c)$ equally over start and goal states as uniform over $\SS_c$.
The problem is to find a goal-conditioned policy $\pi: s,g \mapsto a$ to maximize
\begin{align}\label{eqOpt}
  & \max_\pi \quad \Exp[s,g \sim p_0]{ V^\pi(s,g) }~, \\
  & \text{with } V^\pi(s,g) = \Exp{ \Sum_t \g^t R(s_{t\po},s_t,g) \| \pi, s_{t=0}=s }
\end{align}
the universal value function \cite{2015-schaul-UniversalValueFunction}. 

Note that to evaluate the objective (\ref{eqOpt}) and to train a policy, we require a sampler of $p_0$.
While in standard goal-conditioned MDPs, sampling from $p_0$ is assumed trivial \cite{2017-andrychowicz-HindsightExperienceReplay}, in our case, sampling from $p_0$ implies solving a non-linear constraint problem, connecting the MDP formulation to standard non-linear programming (NLP) formulations \cite{boyd2004convex}.
Further, introducing the constraint formulation is an opportunity to incorporate prior domain knowledge:
The constrained space $\SS_c$ implies a state visitation prior that we can leverage, e.g., via a reset strategy.
Note that we do not assume $\SS_c$ to capture \emph{all} states that can be visited by the black-box dynamics $p(s'|s,a)$.


\subsection{State Sampler}\label{sec:sampling}


We follow existing manipulation planning approaches \cite{2014-posa-DirectMethodTrajectorya, 2014-dai-WholebodyMotionPlanning, 2020-toussaint-DescribingPhysicsPhysical}
to formulate state constraints.
Specifically, we formulate necessary collision, contact, and force constraints that hold for any state under rigid-body dynamics, as well as assume that goals can be held in quasi-static equilibrium.

\paragraph{State Constraints.} Let $s\in\RRR^n$ provide generalized coordinates for a scene configuration with $m$ rigid shapes.
For each pair $(i,j)$, let $c_{ij} \in\{0,1\}$ be a binary contact mode variable which we will sample randomly (see below).
Further, let $d_{ij}(s)\in\RRR$ be the minimal distance between the $i$th and $j$th shape, and $n_{ij}(s)\in\RRR^3$ the corresponding normal.
For pairs that are in contact (indicated by $c_{ij}=1$), we introduce two auxiliary decision variables: the point of attack (POA) $p_{ij}\in\RRR^3$ of force interaction between $i$ and $j$ \cite{2020-toussaint-DescribingPhysicsPhysical}, and the linear force $f_{ij}\in\RRR^3$.
We have point-wise Jacobians for the pair distance $d_{ij}(s)$, the normal $n_{ij}(s)$, and the distance $d^p_{ij}(s)$ of the POA $p_{ij}$ to the surface of shape $i$ for given generalized coordinates $s$.

Given this notation, for all pairs $(i,j)$ we impose non-collision constraints $d_{ij}(s) \ge 0$, and for pairs in explicit contact $c_{ij}=1$ we impose (omitting dependence on $s$ for brevity)
\begin{align}
  & d_{ij}=0,~ d^p_{ij}=0,~ d^p_{ji} = 0, \label{onS}\\
  & n_{ij}^\T f_{ij} \le 0, \label{posF}\\
  & \norm{(\Id - n_{ij} n_{ij}^\T) f_{ij}}^2 \le \mu^2 \norm{n_{ij}^\T f_{ij}}^2 ~, \label{fric}
\end{align}
i.e., the shapes touch at $p_{ij}$ (\ref{onS}), the normal force is positive (\ref{posF}), and the force lies inside a Coulomb friction cone parameterized by $\mu$ (\ref{fric}).
To describe statically feasible states, we impose quasi-static equilibrium on each unactuated shape via the static Newton-Euler equation:
\begin{equation}\label{grav}
F_i = M_i g_i^\text{grav},
\end{equation}
where $F_i$ is the total 6D wrench on shape $i$ induced by all contact forces, $M_i \in \RRR^{6 \times 6}$ is the inertial matrix of $i$, and $g_i^\text{grav}$ the gravity vector expressed in the local frame.

\paragraph{Non-linear Constrained Sampling.} In summary, our decision variables are the discrete contact mode $c_:$ with $c_{ij}\in\{0,1\}$, and continuous $s\in\RRR^n$, as well as $p_{ij},f_{ij}\in\RRR^3$ for pairs with $c_{ij}=1$.
We use a hierarchical sampling strategy, where we first sample the contact mode $c_:$, and then for fixed $c_:$ sample a solution to the continuous NLP (\ref{onS}-\ref{grav}).
In previous work \cite{toussaint2024nlp, braun2026manifold} we studied NLP Sampling, which combines MCMC methods with differentiable constraints to generate robot configuration samples.
Here we follow a simplified approach and directly use an Augmented Lagrangian method \cite{wright1999numerical} to solve the proximal problem
\begin{align}\label{eq:nlp}
  \min_{s,f_:,p_:} \Frac12\norm{s-\bar s}^2_2 \st g_c(s,f_:,p_:) \le 0,~ h_c(s,f_:,p_:) = 0 ~,
\end{align}
where $\bar s$ is a uniform random sample from box constraints, and $g_c, h_c$ is a short hand for the above constraints (\ref{onS}-\ref{grav}).
The role of $\bar s$ is to imply a uniform box prior, which is then projected to the feasible space by solving the proximal problem.
We also initialize the non-linear solver with $\bar s$.

Concerning sampling $c_{ij}$, in this paper we assume a single free object $i=1$ (which is subject to manipulation) and a finite set of potential support shapes $Z \subseteq \{2,..,m\}$.
We limit the potential number of active support contacts to 3, uniformly first choose between the number 1, 2, or 3 of support contacts, and then uniformly pick the 1-, 2- or 3-tuples of support contacts from $Z$.


For some random $c_{ij}$ and $\bar s$, the resulting NLP (\ref{eq:nlp}) will be infeasible or the solver be stuck in a local optimum.
To generate a dataset $\DD = \{ s_i \}_{i=1}^S \subset \SS_c$, we repeat iterating (for random $c_{ij}$ and box-uniform $\bar s$) until $S$ feasible solutions are found.
Figures~\ref{fig:double_sphere}-\ref{fig:allegroDown_cube} in the Appendix illustrate samples from our constrained state sampler.
  


\begin{figure}[tb]
\newcommand{\pic}[2]{%
\shortstack{\includegraphics[width=.165\columnwidth,trim={0 0 0 25pt},clip]{data/configs/images/#1_sample_000002.jpg}\\\scriptsize #2}%
}
\pic{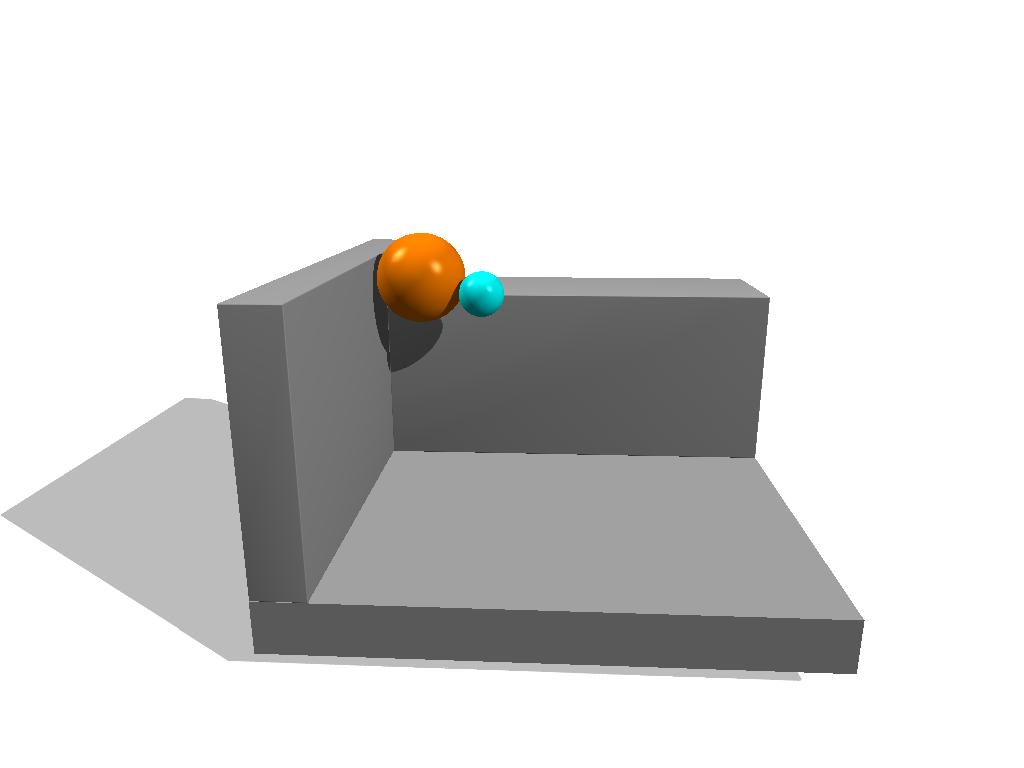}{Double-Sphere}%
\pic{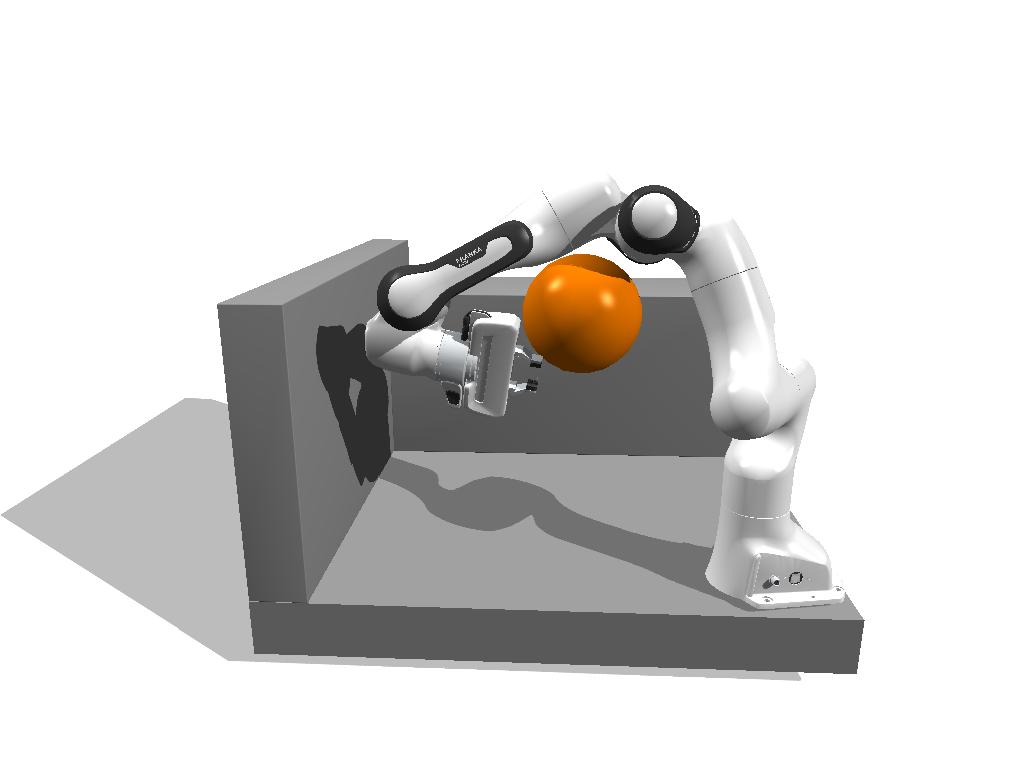}{Panda-Sphere}%
\pic{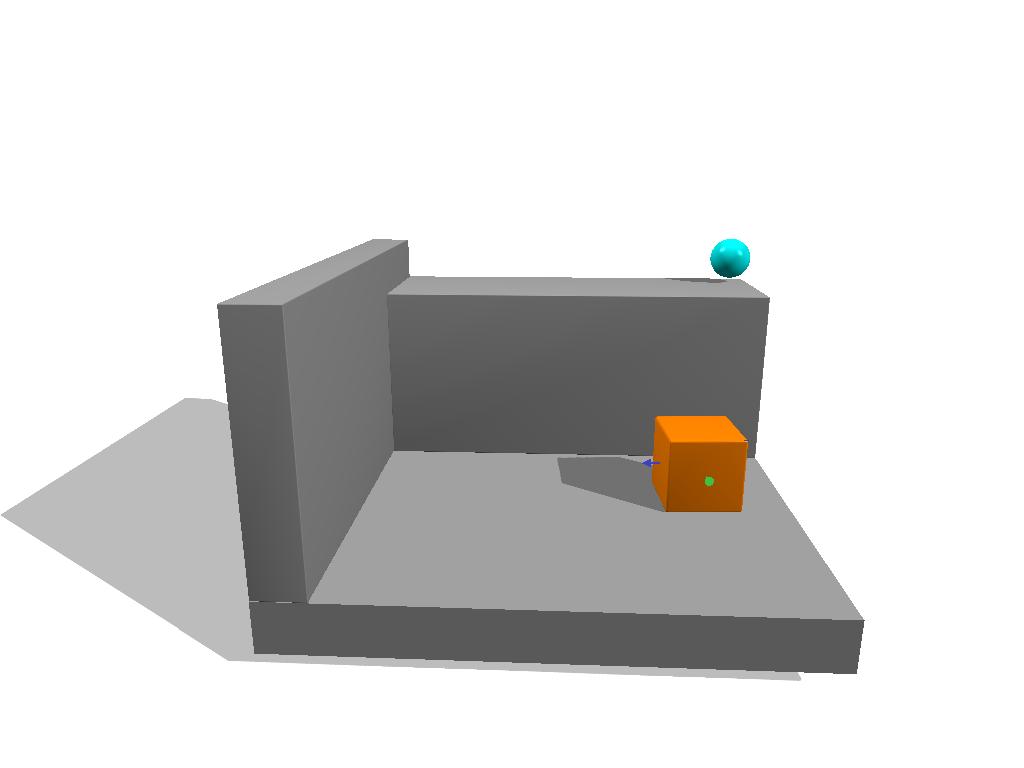}{Sphere-Cube}%
\pic{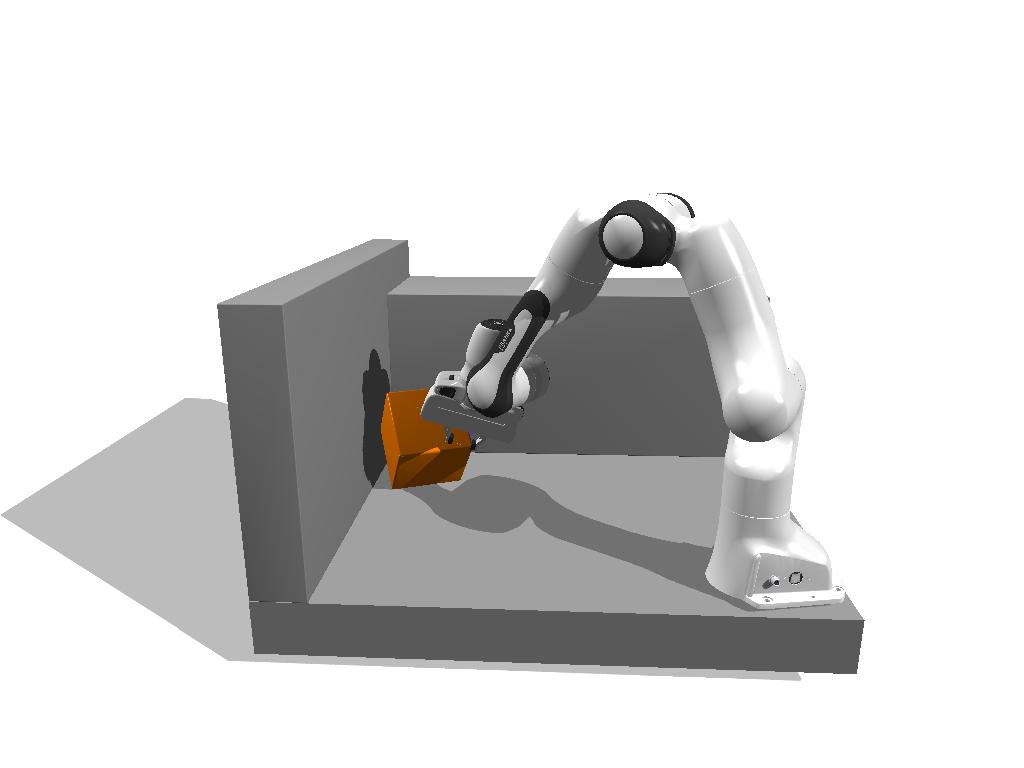}{Panda-Cube}%
\pic{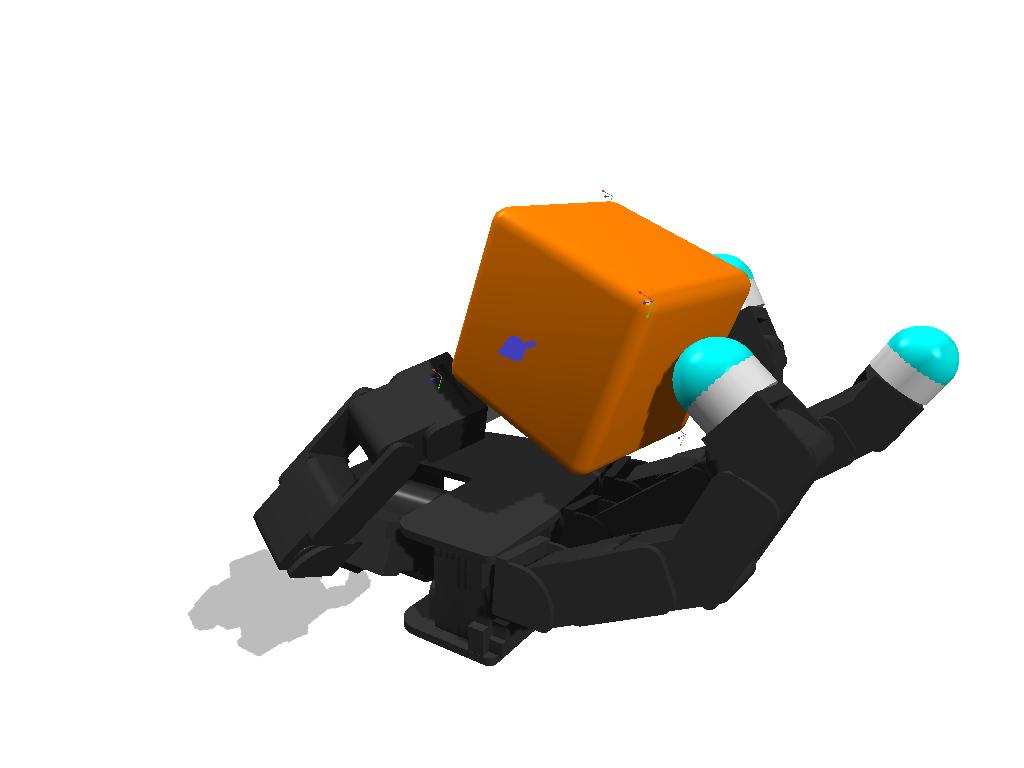}{Allegro-Cube}%
\pic{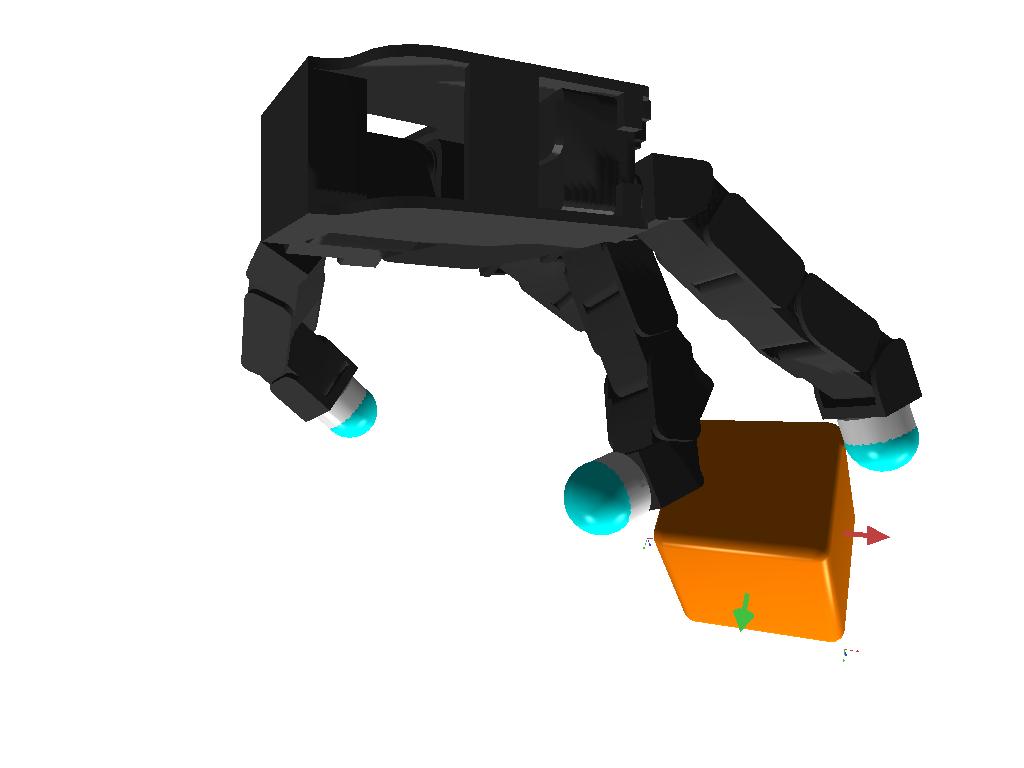}{AllegroDown-Cube}
\caption{\label{fig:domains}The domains for our experiments. For each task the object shall be moved to any feasible given target pose.}  
\end{figure}


\subsection{Additional Reset Strategies}

CSRL combines a given state sampler with an off-the-shelf RL method via the goal and reset distribution.
However, exposing goal-conditioned RL directly to random start and goal states $s,g\sim \DD$ may lead to challenges in high dimensions, where random starts and goals are likely distant and raise a substantial exploration challenge.
Two existing approaches can readily be integrated to address this challenge: Hindsight experience replay (HER) within the RL methods \cite{andrychowicz2017hindsight}, and reverse curriculum resets \cite{2017-florensa-ReverseCurriculumGeneration}.
In the following we propose additional strategies we will study.

First, we propose \textbf{projected interpolation resets}, which exploits our constraint formulation:
At the start of each episode, we sample $s',g\sim \DD$, define $g$ as the goal, and we reset to the start state
\begin{align}\label{eqInterp}
  s=\argmin_{s\in\DD} \norm{\phi(s) - [t\phi(s')+(1-t) \phi(g)]}\quad\text{with}\quad t\sim\UU[0,1] ~.
\end{align}
Here, $\phi:\SS \to \RRR^d$ is a state vector embedding defined in Appendix~\ref{app:envs}, and $t\phi(s')+(1-t)\phi(g)$ their uniform convex interpolation.
Resetting directly to a convex interpolation would be naive, as it could violate basic constraints such as collisions.
We therefore project the interpolation to our constrained space $S_c$ approximately by finding the nearest $s\in\DD$ (using a kd-tree in $\phi$-space).

Second, we propose an optional \textbf{reset curriculum}, where a training phase variable $\a\in(0,1]$ is scheduled linearly from near $0$ to $1$ during an initial phase of RL training, and the interpolation above is modified to use $t\sim\UU[0,\a]$, resetting closer to the goal.
In addition, we adapt the MDP truncation time proportional to $\a$, which implies that RL initially collects short rollouts near to the goal.

To summarize, in our CSRL method we first generate a dataset $\DD$ offline, including vector embeddings and kd-tree.
We then organize RL-training in epochs of $T_\text{epoch}=5000$ training steps:
At the beginning of each epoch, we sample a sufficiently large batch $B$ of start-goal pairs either (i) directly i.i.d.\ uniformly from $\DD$, (ii) using projected interpolation (\ref{eqInterp}), or (iii) using a curriculum with $\a = (t+T_\text{epoch})/T_\text{curr}$, where $t$ are the training steps so far, and we clip $\a$ in $(0,1]$.
For instance, with $T_\text{curr}=100\,000$ we may train for a total of $T_\text{end}=300\,000$ steps, where the first 100\,000 steps are the curriculum phase and after that, $\a=1$ is equivalent to (ii).


\newcommand{\val}[3]{{#1}{\scalebox{.6}{$\pm${#2}}}}
\newcommand{\bfval}[3]{\textbf{#1}{\scalebox{.6}{$\pm${#2}}}}

\section{Experiments}

We evaluate CSRL on six domains (Fig.~\ref{fig:domains}) to evaluate various aspects of contact-rich manipulation skills.
The focus of our empirical evaluations is to quantify the benefit of using constrained sampling to design goal and reset distributions, as well as the effect of the proposed projected interpolation and curriculum strategies.
The study uses the TD7 \cite{2023-fujimoto-SaleStateactionRepresentation} with a HER extension (see Appendix~\ref{app:HER}) as default RL method within CSRL, but we also evaluate omitting the HER extension and a TD3 version.
The project page\footnote{\label{note:webpage}\url{https://www.user.tu-berlin.de/mtoussai/26-CSRL/}}
provides videos to get an impression of the domains and behaviors in these domains, as well as full access to source code.

\subsection{Evaluation Domains}

Fig.~\ref{fig:domains} illustrates the six domains we evaluate on:
(1) The Double-Sphere domain includes a sphere object (orange), a sphere robot (turquoise, 3D position controlled), and three further shapes (floor, walls) to allow for more rich contact modes.
The domain is a basic test for universal manipulation capable to reach any quasi-static state from any other.
(2) The Panda-Sphere domain replaces the sphere robot with a panda arm, and the set of $S$ of potential support shapes includes \emph{all} of panda's links (represented as convex hulls due to MuJoCo's constraints) in addition to the floor.
The domain tests whole-body contact skills for manipulation.
(3) The Sphere-Cube domain replaces the ball object by a cube and includes object orientation as part of the goal feature.
It tests general skills to rotate and control object orientation, including holding the cube with only an edge or a corner on the floor.
(4) The Panda-Cube domain replaces the sphere robot by a panda arm and tests controlling the cube with any contacts on the endeffector and fingers.
(5) The Allegro-Cube domain evaluates dexterous in-hand manipulation.
While in-hand cube orientation has been studied previously \cite{pmlr-v205-qi23a}, in our case the goal includes orientation \emph{and} positioning, which may include balancing it on fingertips.
(6) The AllegroDown-Cube domain turns the previous upside-down, such that random initialization and exploration have high probability of irrevocably dropping the cube.
The domain tests whether CSRL is still effective in generating useful rollout data.

Appendix~\ref{app:envs} provides more details for each domain, Appendix~\ref{app:stats} compute statistics for constrained sampling in each domains, and Appendix~\ref{app:samples} random constrained samples from all domains.

\begin{table}[t]\centering\small
\begin{tabular}{|@{~}c@{~}|@{~}c@{~}|c@{~~}c|c@{~~}c|}
\hline
& & \multicolumn{2}{c|}{Double-Sphere} & \multicolumn{2}{c|}{Panda-Sphere} \\
goals & starts & & sparse & & sparse \\
\hline
\multirow{3}{*}{R}
& R   & \val{0.521}{0.394}{6} & \val{0.596}{0.325}{5} & \val{0.051}{0.073}{6} & \val{0.058}{0.076}{5} \\
& Ri  & \val{0.769}{0.032}{6} & \val{0.772}{0.036}{5} & \val{0.073}{0.075}{6} & \val{0.035}{0.069}{5} \\
& Ric & \val{0.766}{0.017}{6} & \val{0.755}{0.044}{5} & \val{0.112}{0.058}{6} & \val{0.059}{0.076}{5} \\
\hline
\multirow{3}{*}{C}
& R   & \val{0.867}{0.018}{6} & \val{0.685}{0.373}{5} & \val{0.005}{0.004}{6} & \val{0.007}{0.002}{5} \\
& Ri  & \val{0.877}{0.010}{6} & \val{0.697}{0.386}{5} & \val{0.154}{0.175}{6} & \val{0.012}{0.013}{5} \\
& Ric & \val{0.892}{0.016}{6} & \val{0.850}{0.030}{5} & \val{0.287}{0.230}{6} & \val{0.021}{0.023}{5} \\
\hline
\multirow{3}{*}{C}
& C   & \val{0.963}{0.011}{6} & \val{0.953}{0.004}{5} & \val{0.527}{0.255}{6} & \val{0.472}{0.450}{5} \\
& Ci  & \bfval{0.972}{0.007}{6} & \val{0.783}{0.430}{5} & \val{0.630}{0.415}{6} & \val{0.701}{0.466}{4} \\
& Cic & \bfval{0.972}{0.007}{5} & \bfval{0.969}{0.010}{5} & \bfval{0.929}{0.029}{5} & \bfval{0.932}{0.007}{5} \\
\hline
\end{tabular}%
\smallskip
\caption{\label{tab:ex_modes}
  Comparison of alternative goal and reset sampling methods.
  We report the mean success rate of the final policy with std.\ deviation over 5 training runs and highlight the best mean per column.
  Constrained sampling (C) or Random drops (R) is used to sample goals and starts, optionally combined with projected interpolation (i) and a curriculum (c) for sampling starts.
}
\end{table}

\begin{table}[t]\centering\small
\begin{tabular}{|@{~}c@{~}|@{~}c@{~}|c@{~~}c@{~~}c|c@{~~}c@{~~}c|}
\hline
& & \multicolumn{3}{c|}{Double-Sphere} & \multicolumn{3}{c|}{Panda-Sphere} \\
goals & starts & TD7${}_\text{HER}$ & TD7 & TD3${}^\text{sb}$ & TD7${}_\text{HER}$ & TD7 & TD3${}^\text{sb}$ \\
\hline
\multirow{3}{*}{C}
& C   & \val{0.963}{0.011}{6} & \val{0.923}{0.052}{16} & \val{0.131}{0.050}{5} & \val{0.527}{0.255}{6} & \val{0.324}{0.296}{14} & \val{0.008}{0.005}{5} \\
& Ci  & \val{0.972}{0.007}{6} & \bfval{0.982}{0.005}{23} & \val{0.806}{0.024}{5} & \val{0.630}{0.415}{6} & \val{0.837}{0.139}{19} & \val{0.043}{0.014}{5} \\
& Cic & \val{0.972}{0.007}{5} & \bfval{0.983}{0.005}{15} & \val{0.800}{0.006}{5} & \bfval{0.929}{0.029}{5} & \val{0.814}{0.104}{15} & \val{0.066}{0.009}{5} \\
\hline
\end{tabular}

\smallskip
\caption{\label{tab:ex_others}
  Evaluation of alternative RL methods inside CSRL:
  The stablebaselines3 TD3 implementation (with same hyperparameters as TD7), a HER extension of TD7, and the sb3 HER extension of TD3. (Bold: Best per problem.)
}
\end{table}

\begin{table}[t]\centering\small
(a)~%
\begin{tabular}[t]{|@{~}c@{~}|@{~}c@{~}|c|c|c|c|}
\hline
  goals & starts & Sphere-Cube & Panda-Cube & Allegro-Cube & AllegroDown-Cube \\
\hline
\multirow{3}{*}{C}
& C   & \val{0.878}{0.032}{7} & \val{0.949}{0.009}{6} & \val{0.954}{0.011}{4} & \val{0.903}{0.025}{4} \\
& Ci  & \val{0.912}{0.048}{6} & \val{0.966}{0.005}{6} & \val{0.957}{0.006}{3} & \val{0.901}{0.023}{3} \\
& Cic & \val{0.928}{0.056}{5} & \val{0.964}{0.006}{8} & \val{0.966}{0.010}{4} & \val{0.891}{0.016}{5} \\
\hline
\end{tabular}%

\quad(b)~%
\begin{tabular}[t]{|@{~}c@{~}|@{~}c@{~}|c|c|}
\hline
goals & starts & Double-Sphere & Panda-Sphere \\
\hline
R & Rrev & \val{0.846}{0.021}{3} & \val{0.197}{0.042}{4} \\
C & Crev & \val{0.988}{0.023}{3} & \val{0.274}{0.018}{2} \\
\hline
\end{tabular}

\smallskip
\caption{\label{tab:ex_cube}
  (a) Performance on the cube domains.
  (b) Comparison to reverse curriculum.
}
\end{table}

\subsection{Comparison of Alternative State Samplers and Reset Strategies}

\newcommand{\mypar}[1]{\par\emph{#1.}~~}

\mypar{Random Sampling Baseline}
Prior work such as \citet{2018-zeng-LearningSynergiesPushing} generates reset states by sampling random states which are then stepped in simulation to generate feasible reset states for learning robust pushing and grasping policies.
We transfer this approach to our domains as a baseline to generate diverse random configurations.
To sample a random state $s$, we first sample its generalized coordinates uniform from box constraints, followed by simulation until a stable state is reached and reject the sample if the state is out of bounds.
Note that this sampler can equally be used to sample goals and start states, as well as combined with our proposed interpolation and curriculum strategies.

\mypar{Constrained Sampling outperforms Random Sampling}
Table~\ref{tab:ex_modes} compares alternatives combinations of using constrained sampling (C) or random sampling (R) for the goal and start distribution, as well as combining this with projected interpolation (i) and curriculum learning (c).
On the Double-Sphere and Panda-Sphere domains, as well as their sparse variants (see Appendix~\ref{app:envs}), using constrained sampling for starts and goals significantly outperforms random sampling.
On our simplest domain, the Double-Sphere, random resets can achieve considerable but sub-optimal success rates.
On the Panda-Sphere problem, random sampling performs poorly, presumably as it is unlikely to provide reset states where the sphere is held static between limbs.
On the sparse Panda-Sphere variant, this effect becomes more pronounced.

\mypar{Alternative RL methods inside CSRL}
Table~\ref{tab:ex_others} includes additional evaluation when 
using TD7 without HER extension, as well as stablebaselines3's TD3 implementation as alternative RL engines inside CSRL.
We used the same hyperparameters as for TD7, but found stable-baselines3's TD3 implementation less robust, confirming the benefits of TD7's contributions.
The HER extension is not always better, but we found it generally more robust -- especially in the Panda-Sphere domain -- which is why we choose it as default option.

\mypar{Projected interpolation resets and curricular can further improve performance}
Tables~\ref{tab:ex_modes} and \ref{tab:ex_others} confirm that adding projected interpolation and a curriculum into the reset strategy further increases performances, across domains and RL methods used inside CSRL.
This also holds when using random sampling resets.

Table~\ref{tab:ex_cube} provides additional results on the considerably more demanding cube manipulation domains, including Panda-Cube (with 66-dim observation) and Allegro*-Cube domains (with 96-dim observation, see Table~\ref{tab:env_dims} in Appendix~\ref{app:envs}).
We stabilized training in these domains by increasing the replay buffer size up to 5M and using more constrained samples.

\begin{figure}[t]\centering\small \newcommand{\pic}[2]{\includegraphics[width=.24\columnwidth]{data/plots/#1.pdf}}

\pic{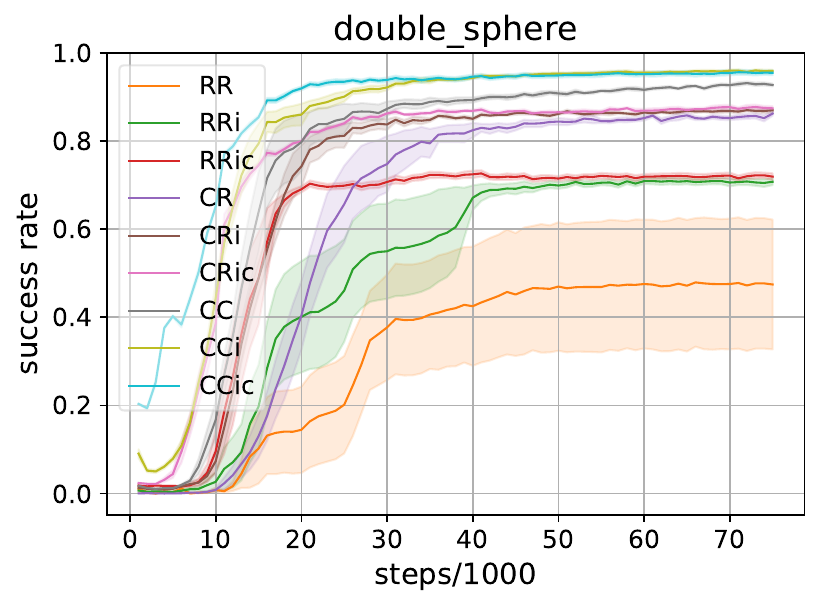}{a}~~%
\pic{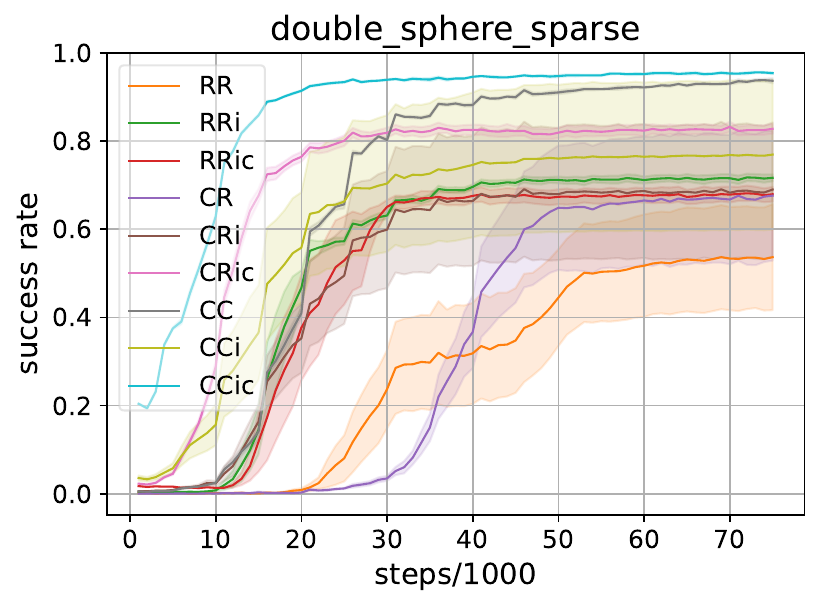}{b}~~%
\pic{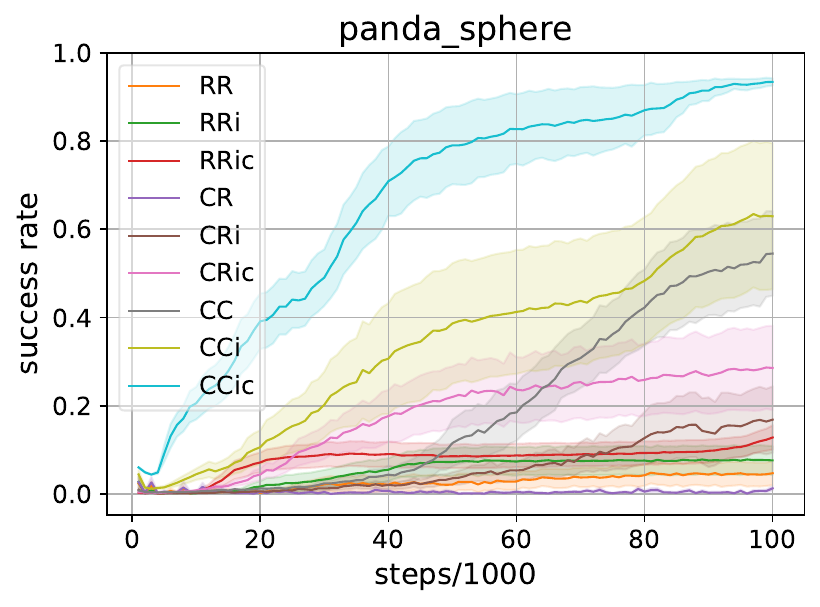}{c}~~%
\pic{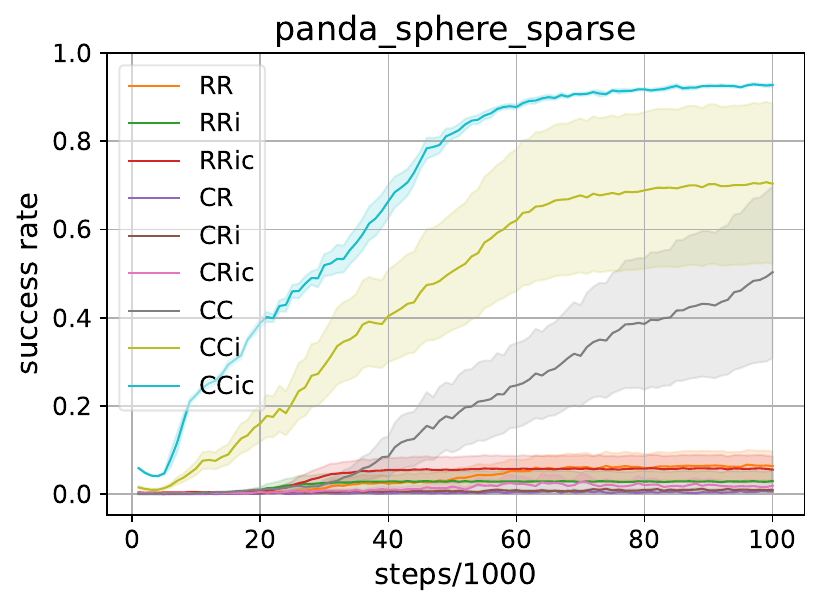}{d}
\pic{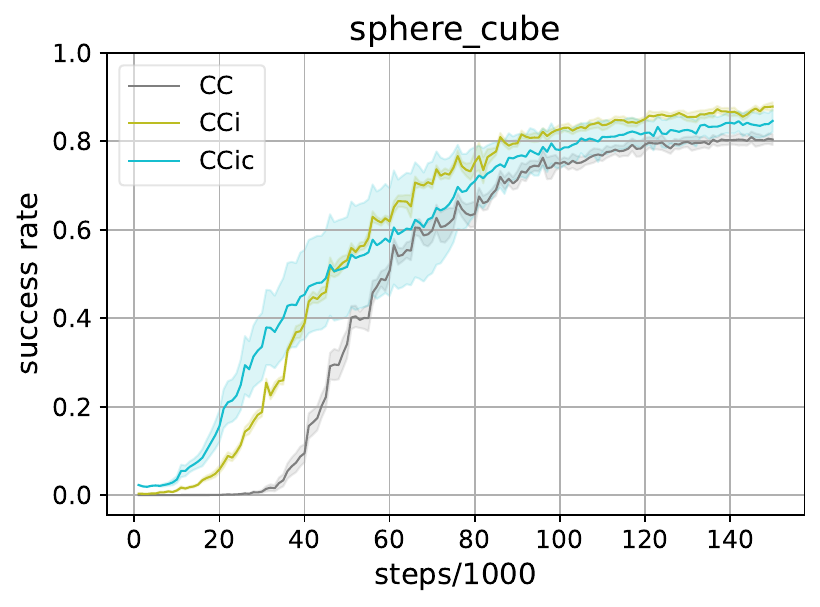}{e}~~%
\pic{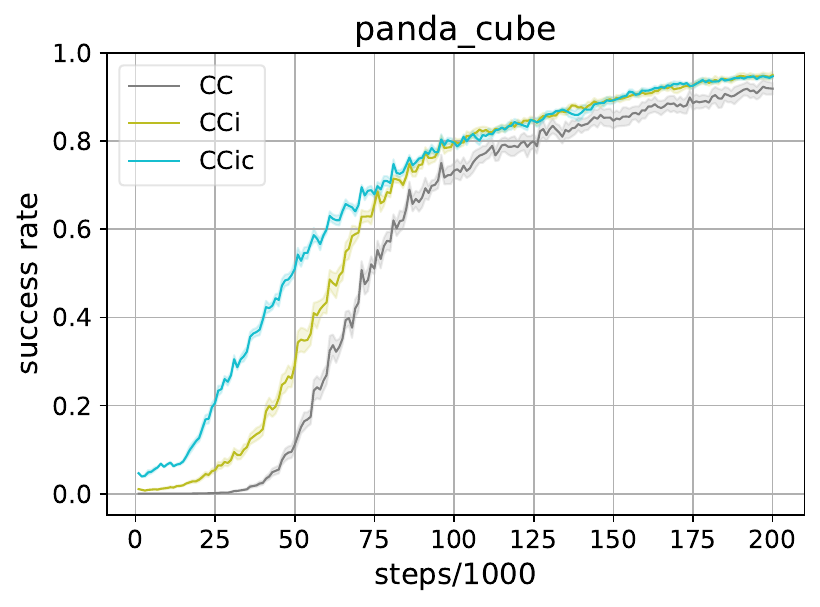}{f}~~%
\pic{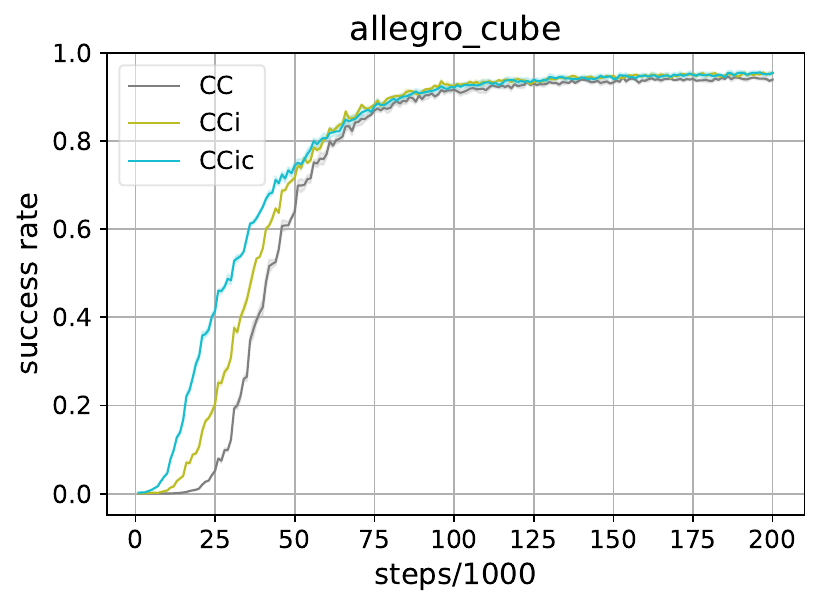}{g}~~%
\pic{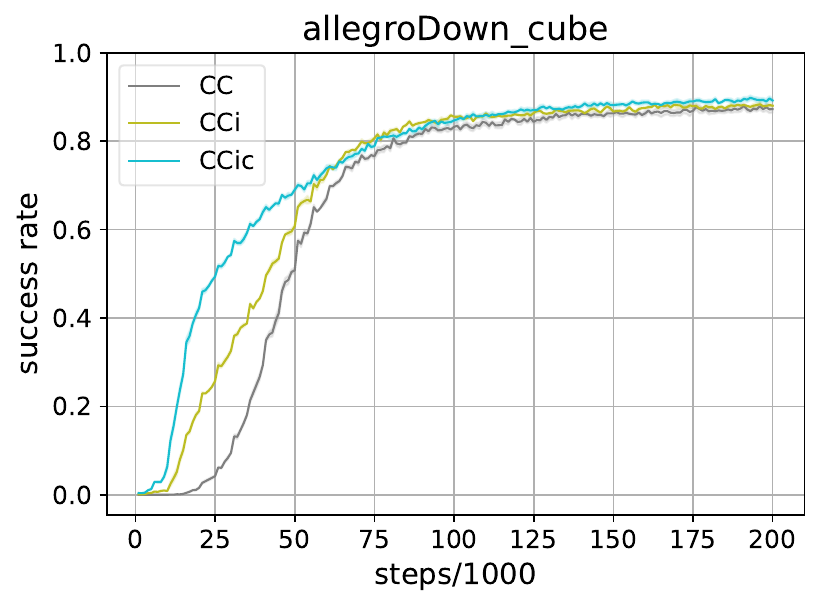}{h}
  \caption{\label{fig:RLtraces}
    Success rate during RL training.
    Mean and std.\ error (shading) over 5 independent runs for each method.
    The x-axis are TD7 training steps, which each corresponds to 80 environment samples in our setting.
  }
\end{figure}

Fig.~\ref{fig:RLtraces} provides success rates during training.
The convergence behaviors confirm the benefit of our proposed methods not only for the final policy performance, but also for convergence speed and thereby sample efficiency.

\mypar{Projected interpolation outperforms reverse curricula}
Reverse reset curricula (RC) \cite{2017-florensa-ReverseCurriculumGeneration} have previously been proposed to design a curriculum of reset states with increasing distance to the goal.
We explored RC also in our goal-directed setting, simulating a random policy from a given goal state for a limited (increasing) amount of time.
However, as we used a single scene-instance Mujoco simulation to generate reset states on the fly, the approach comes with substantial computational overhead and we explored it only on the small domains.
RC need an apriori goal sampler from which reverse actions are executed; we tested using random sampling and our constrained sampler.
Table~\ref{tab:ex_cube}(b) provides results on the Double-Sphere and Panda-Sphere domains.
For both domains, our interpolation approach outperforms reverse curricular, also when using random drop starts.

\subsection{Discussion of Found Behaviors}

The videos on the project page\footref{note:webpage} illustrate the learned behaviors and success rates.
Concerning the Double-Sphere domain, the optimal policy shows versatile skill in reaching any possible quasi-static state, including balancing the ball and highly reactive recovery from external perturbations.
For the Panda-Sphere domain, we find versatile skill in bringing the sphere to be held with diverse limbs, using complex whole-body contact maneuvers to scoop and transport the ball into holding positions.

For the Sphere-Cube domain, our policy is highly effective in turning the cube to desired orientations and balancing it on the edge or corner.
In the latter case it reaches near-goal configurations with highly dynamic and reactive behaviors, though sometimes missing the goal precision threshold slightly.
In the Panda-Cube domain the learned strategies include controlling cube rotation and pushes via a flat endeffector squeezes and complex non-prehensile interaction.

For the Allegro-Cube domain, we discover in-hand cube control to reach feasible orientations \emph{and} positions.
For instance, the optimal policy often first rotates the cube to final orientation before it is then brought into position, including getting it to balance on finger tips.
Very similar manipulation skills are found also for the AllegroDown-Cube domain.
CSRL demonstrates high robustness to transfer to the upside-down in-hand manipulation.

\subsection{Real-World Transfer}

\begin{figure}[t]
  \centering
  \resizebox{\linewidth}{!}{%
  \setlength{\tabcolsep}{1pt}
  \renewcommand{\arraystretch}{0}
  \newcommand{\rcell}[1]{\includegraphics[width=0.2\textwidth]{figs/real_sim_compare_figs/#1_real.jpg}}
  \newcommand{\scell}[1]{\includegraphics[width=0.2\textwidth]{figs/real_sim_compare_figs/#1_sim.jpg}}
  \begin{tabular}{*{5}{c}}
    \rcell{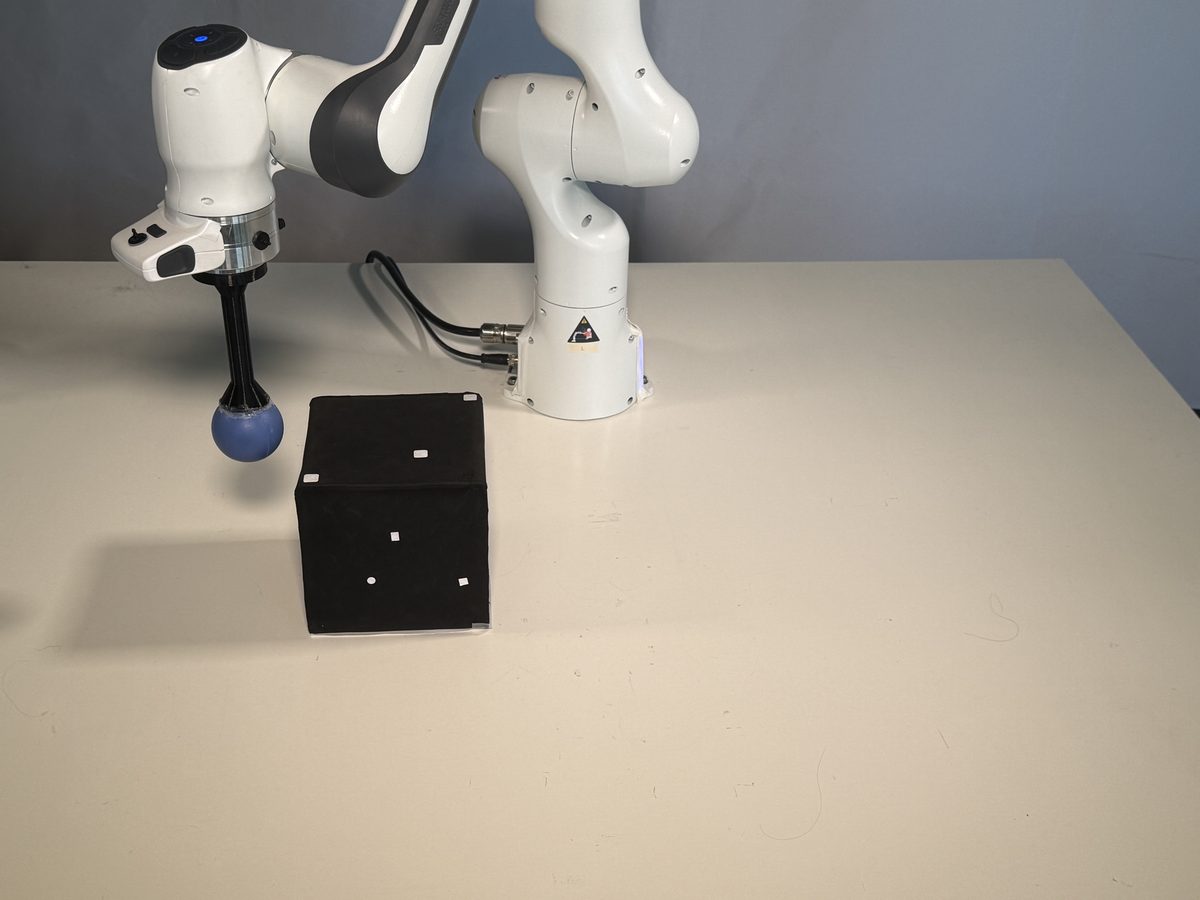} & \rcell{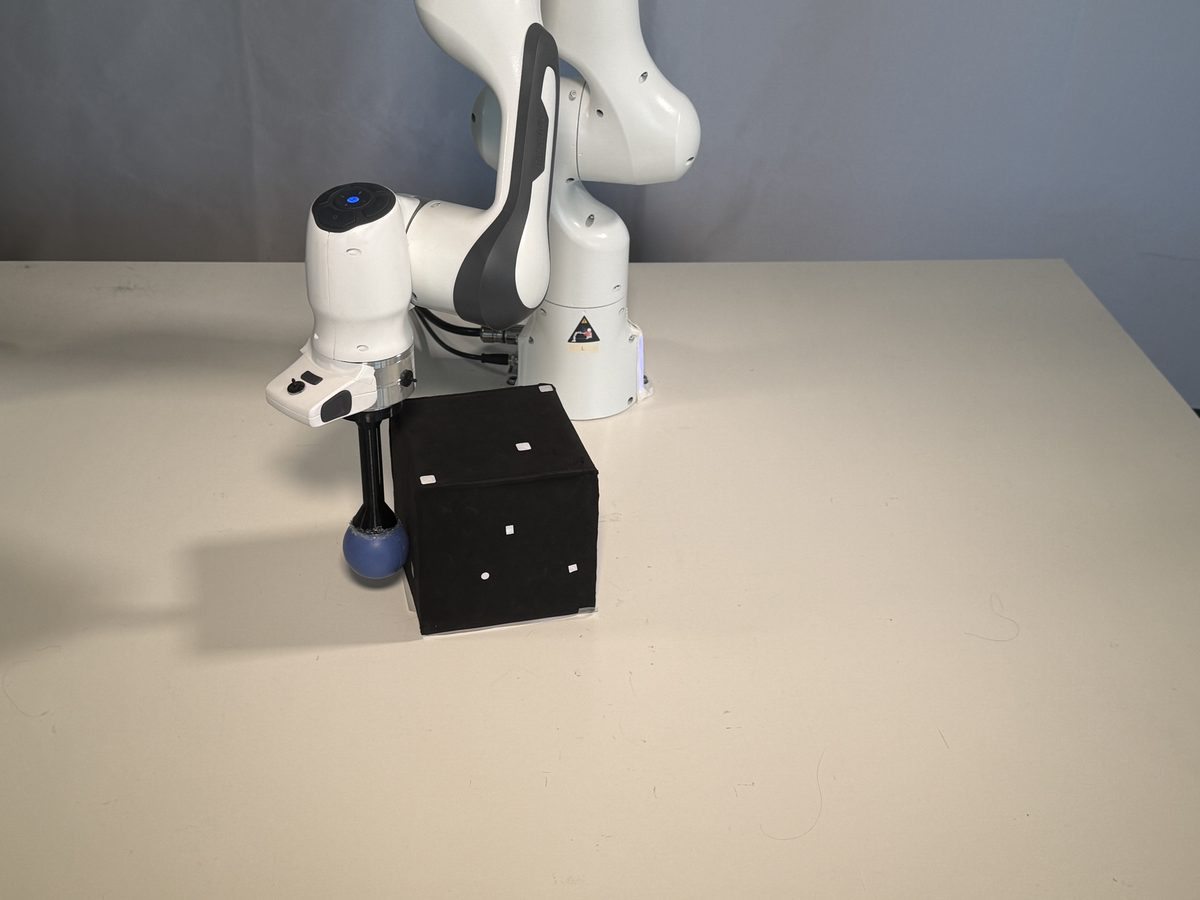} & \rcell{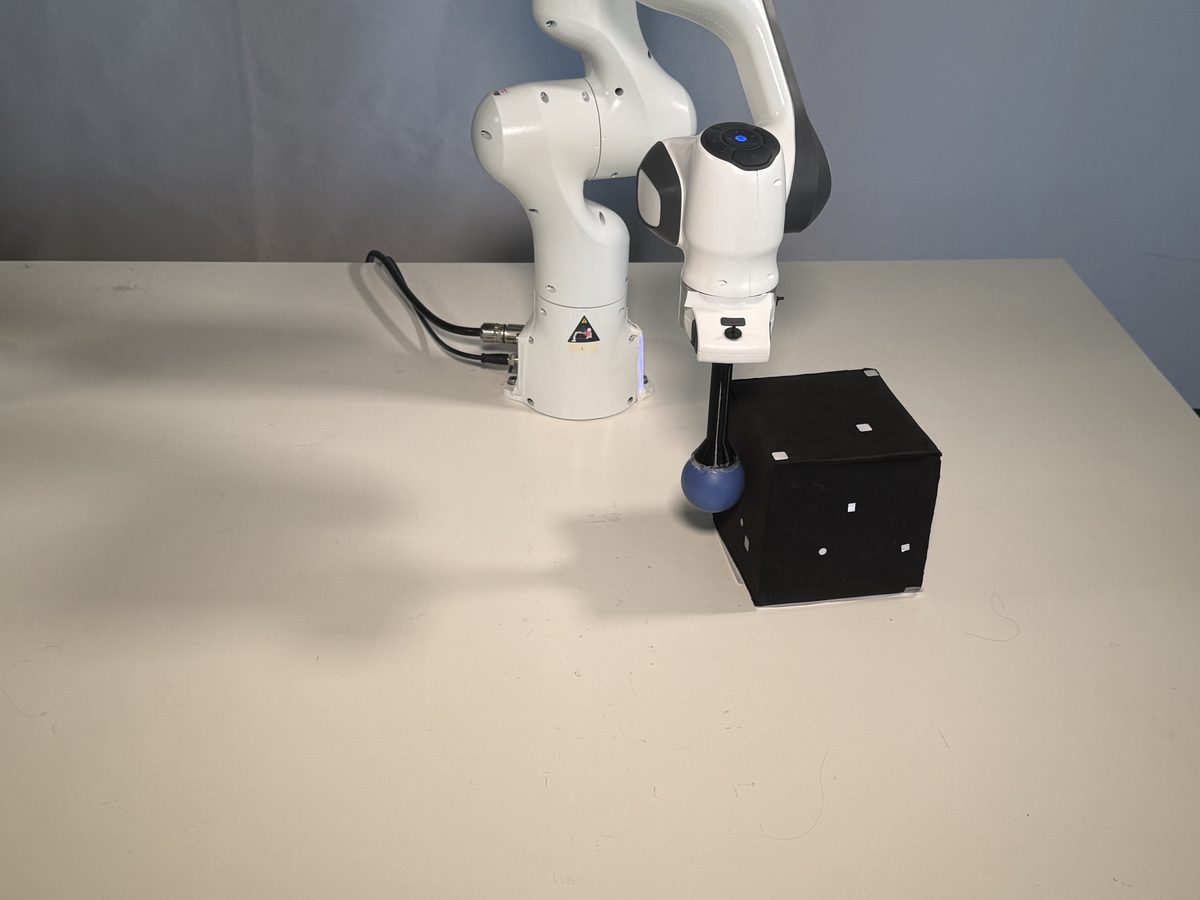} & \rcell{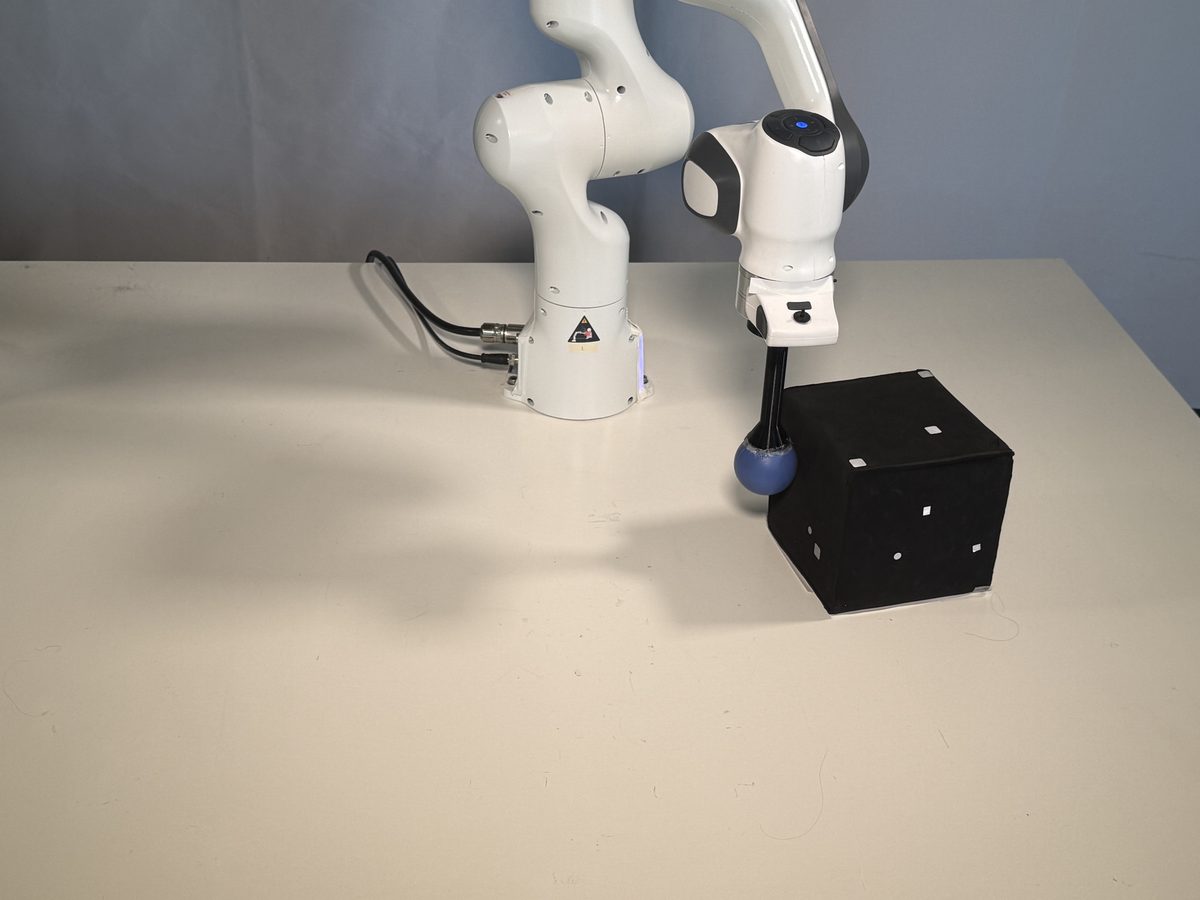} & \rcell{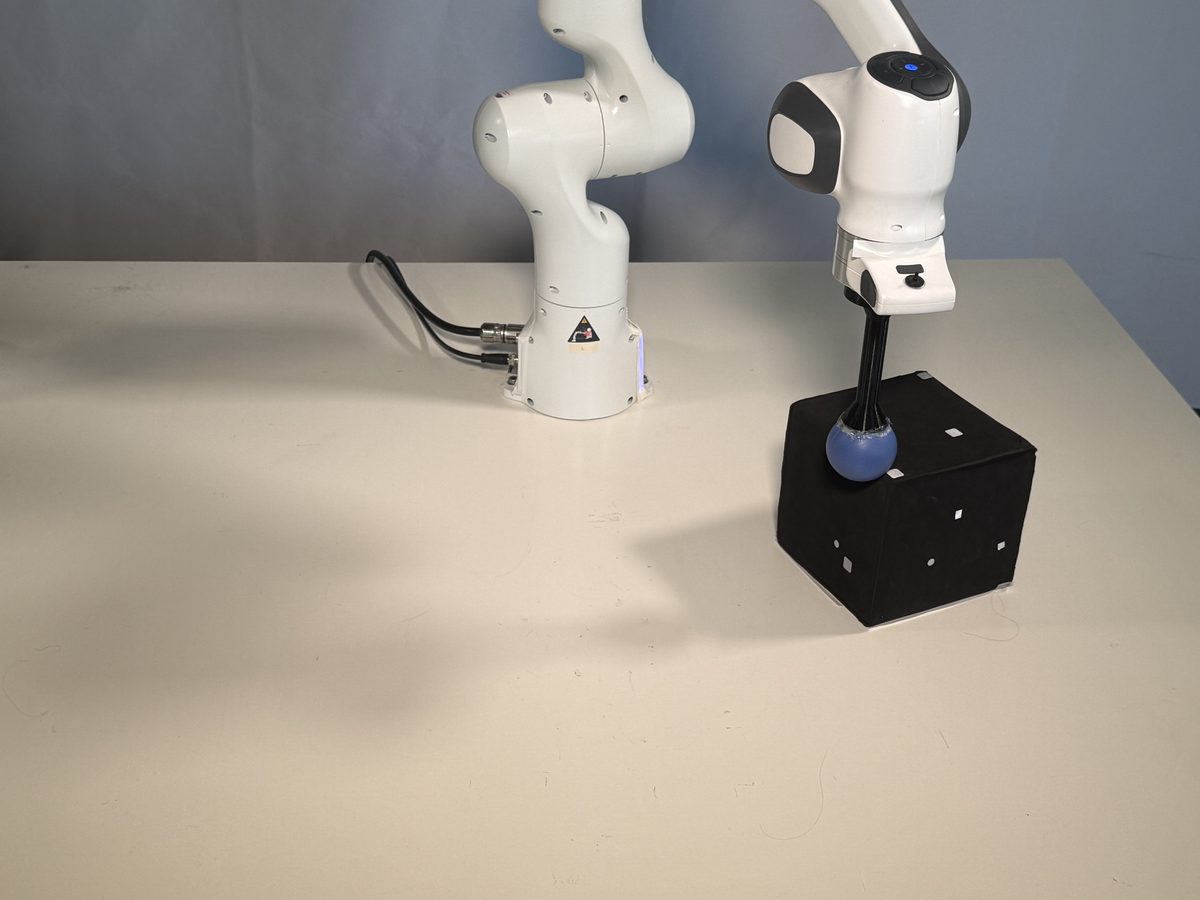} \\
    \scell{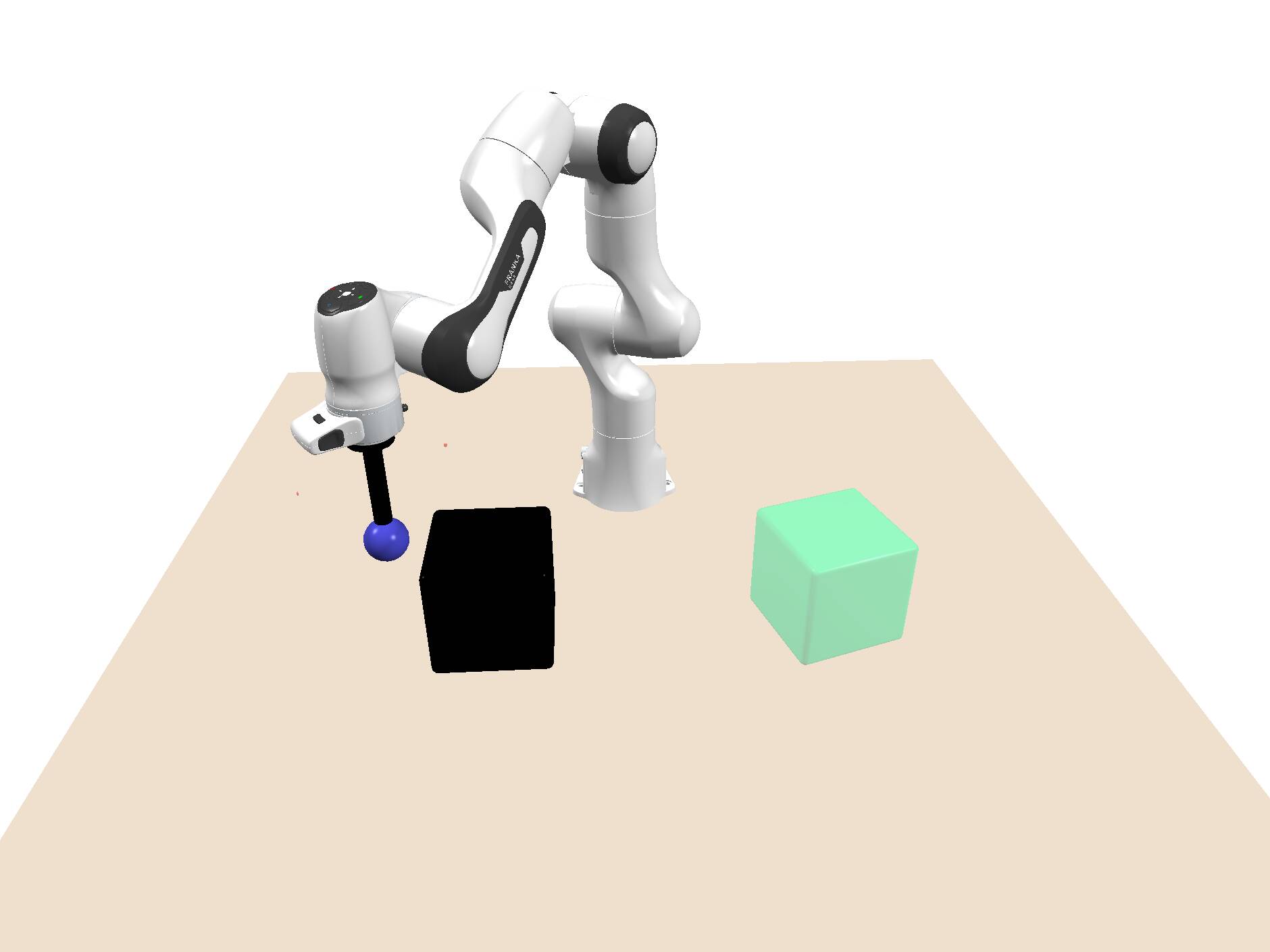} & \scell{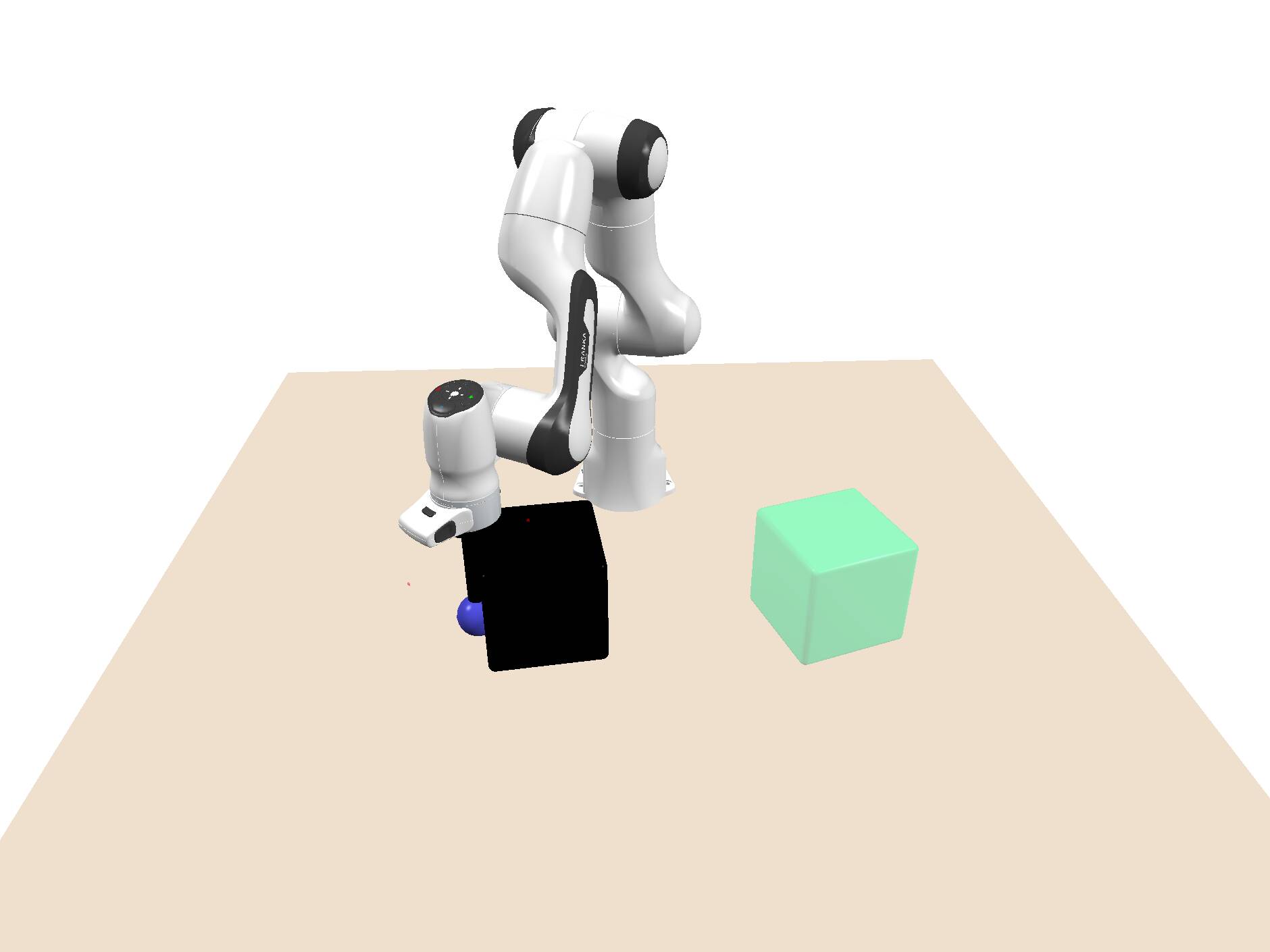} & \scell{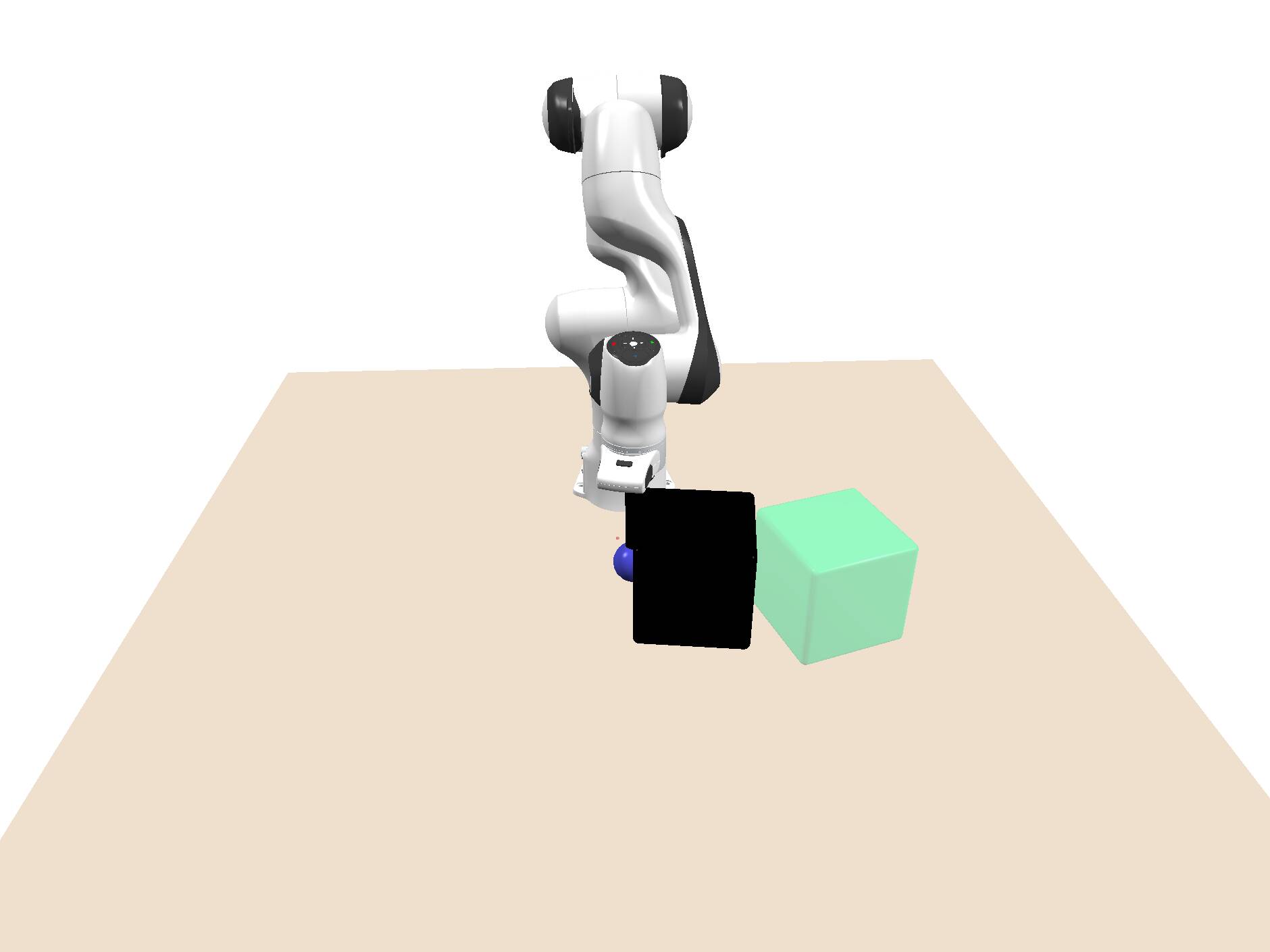} & \scell{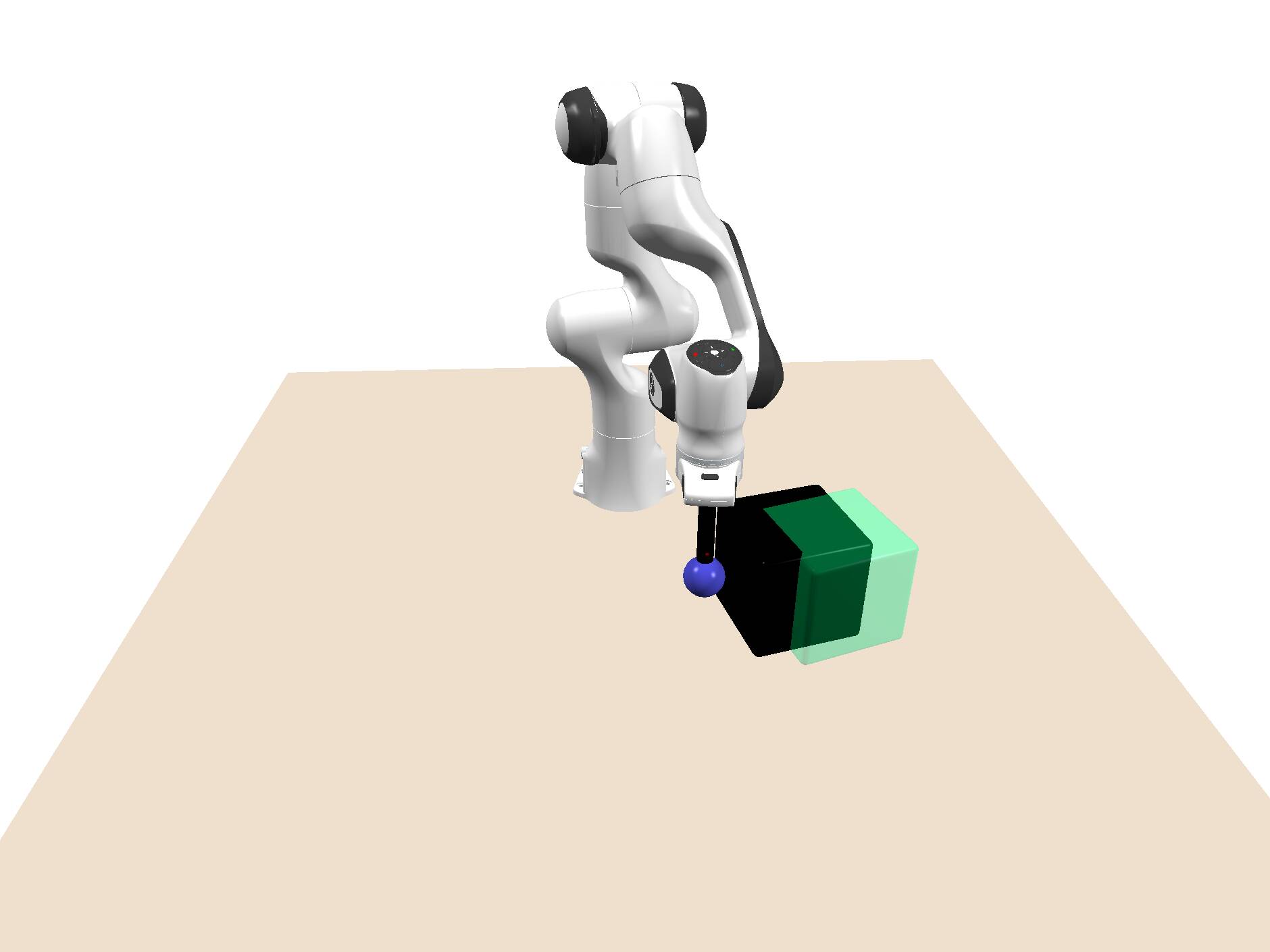} & \scell{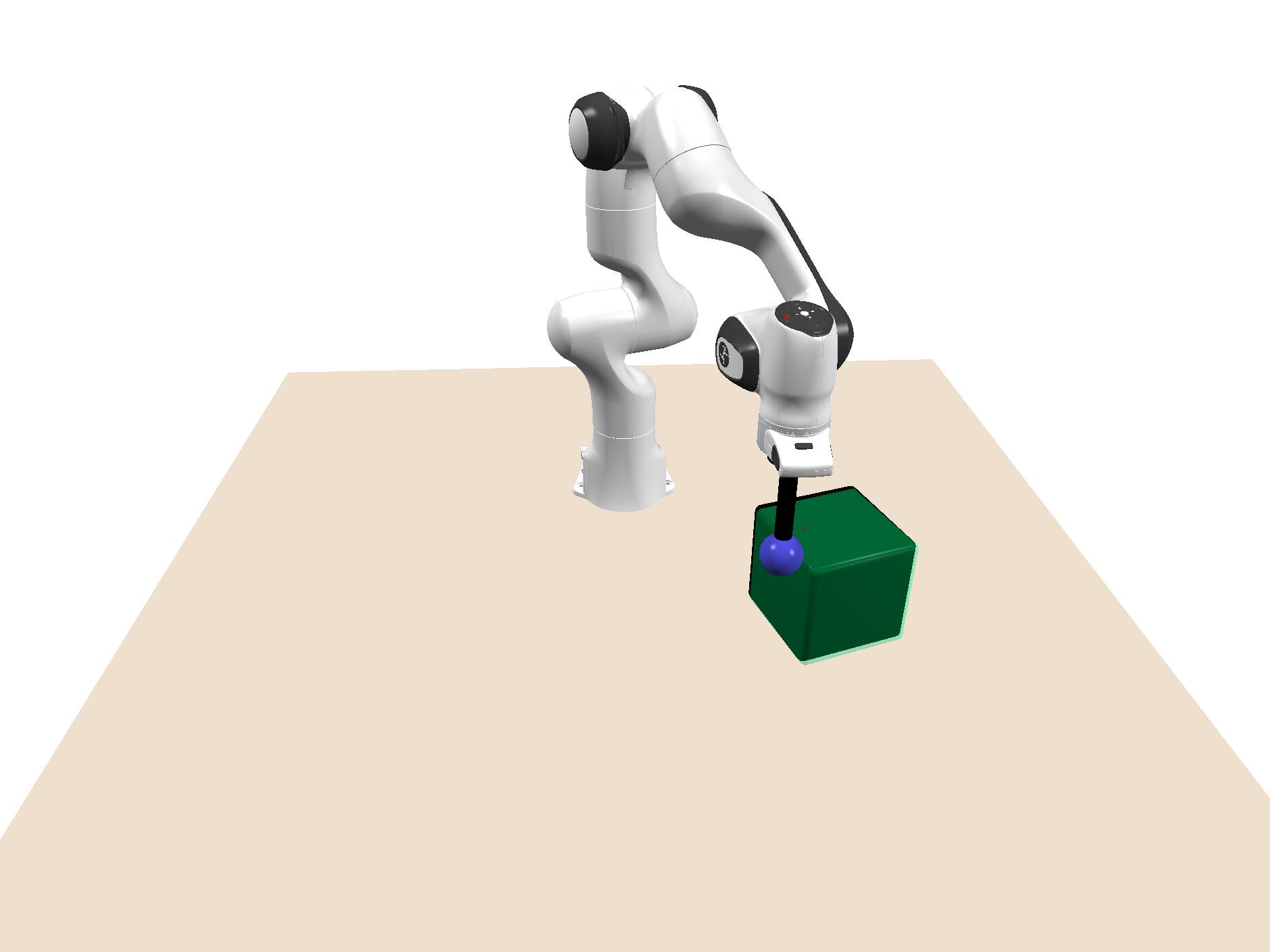} \\
  \end{tabular}
  }
  \includegraphics[width=0.82\linewidth]{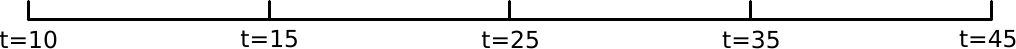}
    \caption{Real (top) vs.\ synced view (bottom), paired by action index $t$ (number of policy actions taken). The synced view mirrors the OptiTrack cube pose and the robot's joint
  state; per-action wall-clock time is set by a blocking IK move.}
  \label{fig:real_sim_compare}
\end{figure}

With CSRL we propose a novel framework for \emph{privileged} training of fundamental manipulation skills.
To evaluate basic real-world transfer we considered cube manipulation with using OptiTrack state tracking.
In particular, we considered transfer of a Sphere-Cube policy, where the simulated sphere robot is replaced by an endeffector-attached ball.
This interprets our simulated sphere robot as abstraction of single contact interaction skills that can be transferred to real world using inverse kinematics.
Fig.~\ref{fig:real_sim_compare} illustrates a pushing behavior. The videos provides additional demonstrations.
For further details, we refer to Appendix~\ref{app:real-world-details}.

\section{Conclusion}

Relating back to Fig.~\ref{fig:CSRL}, while a physics simulator provides a means to sample the dynamics, with this paper we propose to complement this with a physics-based constrained sampler of states.
Our core interest is advancing generalist robotic manipulation -- we therefore proposed a novel state sampler that explicitly takes into account the structure of contact interaction in order to provide a rich covering of diverse contact modes.
Integrating the constrained sampler in a reset strategy and optionally combining this with an interpolation and curriculum strategy we attain an effective method to train universal manipulation policies.
Our evaluations demonstrate contact-rich whole-body manipulation in the Panda-Sphere scenario, highly agile behaviors to bring a cube to balance on the edge or a corner in the Sphere-Cube scenario, as well as dexterous in-hand manipulation to control a cube orientation and positioning also in an upside-down Allegro setting.

\textbf{Limitations.} (i) Our approach targets \emph{privileged} training of universal manipulation skills, where the policy has full access to the state.
This is motivated by the teacher-student paradigm that recently showed great success in the context of locomotion \cite{2020-chen-LearningCheatinga,Lee_2020, yin2026emergentdexteritydiverseresets}.
However, we have not addressed transfer to a sensor-based student policy yet and expect sim2real challenges that are specific to manipulation, e.g.\ as the admittance observation might deviate substantially.
%
%
(ii) In this work we used a constrained sampling method based on a general-purpose NLP solver on CPU.
While sampling for simple scenes such as the Double-Sphere is still efficient, the non-linearities in constraints (\ref{onS}-\ref{grav}) raise substantial computational costs for more complex objects.
To address this limitation, future work should invest in dedicated GPU-parallelized solvers -- following the example of modern physics simulators.

	
\acknowledgments{
This work has received support from the German Federal Ministry of Research, Technology and Space (BMFTR) under the Robotics Institute Germany (RIG);
from the Amazon Fulfillment Technologies and Robotics team;
from the French government, managed by the National Research Agency, under the France 2030 program with the references Organic Robotics Program (PEPR O2R), ``PR[AI]RIE-PSAI'' (ANR-23-IACL-0008) and RODEO (ANR-24-CE23-5886);
and from the European Union through the ARTIFACT project (GA no.101165695) and the AGIMUS project (GA no.101070165).
Views and opinions expressed are those of the author(s) only and do not necessarily reflect those of the funding agencies.
}


\bibliography{26-sgrl, references}


\clearpage
\appendix
\section{Gym Environment Details}\label{app:envs}

The source code to reproduce experiments, including environment definitions and hyperparameter configurations, is accessible at
\url{https://github.com/LIS-TU-Berlin/csrl}.

We choose a Mujoco simulation step time $\tau_\text{sim}=0.001$ seconds, while the environment steps with $\tau_\text{action}=0.05$s.
For a robot with generalized coordinates $q(t)\in\RRR^n$, robot control uses PD motors around a continuous polynomial reference $r(t)\in\RRR^n$ during these $0.05$s.

\textbf{Actions:}
An action continuously overwrites the reference:
Given the current reference position $r_0=r(t_0)$ and robot velocity $v_0=\dot q(t_0)$ at the current control time $t_0$, an action $a\in\RRR^n$ defines the new 2nd-order polynomial reference
\begin{align}\label{eqPolyRef}
    r(t) &= \frac{a-2\lambda v_0}{2\lambda^2} (t-t_0)^2 + v_0 (t-t_0) + r_0 ~.
\end{align}
This implies $r(t_0+\lambda) = r_0 + \half a$, which interprets $a$ as a delta command.
We choose this action definition to ensure continuity in the underlying reference, and thereby limited jerk.
We generally choose $\lambda=2\tau_\text{action}$, i.e.\ twice the environment step time.

\textbf{Goal \& rewards:}
For the cube domains the goal includes reaching a given target object position \emph{and} orientation.
We define a set of $n_G$ scene marker points $G\in\RRR^{n_G\cdot 3}$ which are attached to the object.
For the sphere domains this is only the sphere center $G\in\RRR^3$.
For the cube domains this is the cube center and 4 additional points at opposite corners of the cube, $G\in\RRR^{5\cdot 3}$.
Together these provide information equivalent to SE$(3)$-pose, with the 5 points linearly related to the position and orientation matrix columns.

For current $G_t$ and target $G^*$ (as a flattened vector) we compute the $L_2$-norm $d_t=\norm{G_t-G^*}$ and define success and episode termination when $d_{t\po}\le\e_G$.
In the sparse reward setting, rewards are zero except for $r=1$ on success.
In the non-sparse setting we define $\bar d_t = [1.-d_t/(10\e_G)]_+ \in [0,1]$, which is $1$ at the target and decays linearly to zero at distance $10\e_G$ to the target.
We provide additional rewards $r=\bar d_{t\po} - \bar d_t$, indicating progress towards the target.
Note that by definition the episode sum over shaping reward is $\in[-1,1]$, and successful episode reward sums are $\in[0,2]$.

\textbf{Observations:}
We define the agent observation as
\begin{align}\label{eqObs}
  y &= (q, \dot q, \dot q_o, r-q, P, G^*-G)
\end{align}
with robot generalized coordinates $q, \dot q\in\RRR^n$, object velocity $\dot q_o\in\RRR^{3+3}$, control error $r(t)-q(t)\in\RRR^n$, scene marker points $P\in\RRR^{n_P\cdot3}$, and goal error $G^*-G\in\RRR^{n_G\cdot3}$.
The scene marker points include the goal marker points (so that object position and orientation are observed).
For panda scenarios they additionally include 3 marker points on the gripper, for the Allegro scenarios they include the 5 finger tips.
The observation of control error ensures a Markovian environment (as $r_0$ in (\ref{eqPolyRef}) is adopted) and provides indirect information on interaction forces, as the PD motors relate control error to joint torques.
Therefore, the policy can realize admittance control.

\textbf{Reference reset via contact force:}
Given (\ref{eqPolyRef}), the reference $r$ becomes part of the Markovian state and is observed in (\ref{eqObs}).
At reset, not only the state but also the reference $r_0$ has to be initialized.
For the non-allegro domains, we initialize $r_0 = s_0$ equal to the state, which is the natural choice.
However, our constrained sampler
solves (\ref{eq:nlp}) not only for the state $s$ but also the contact forces $f_{:}$ and point of attack $p_{:}$.
Using the Jacobian transpose law, we can back-project these to control space, i.e., we know which joint torques $\tau$ reproduce the equilibrium forces.
Knowing the $k_p$ of the motors, we can in turn define the joint reference as $r_0 = s_0 + \tau/k_p$ (elem-wise division), which pre-loads the joint motors to exert the equilibrium contact forces.
We adopted this for the Allegro domains, such that sending zero actions after a reset indeed holds the cube approximately in equilibrium (modulo Mujoco's drift and slip), also for the AllegroDown domain.

\textbf{Vector embedding for projected interpolation:}
For interpolation (\ref{eqInterp}) we use the vector embedding $\phi(s) = (w_q q, G) \in \RRR^{n+3n_G}$ that combines the goal features with the weighted robot pose $q$.
We choose $w=0.5$ for the sphere domains, and $w=0.2$ for the cube domains.


\begin{table}[t]\centering\small
\begin{tabular}{|c|cccc|c@{~~}c@{~~}c@{~~}c@{~~}c@{~~}c@{~~}c@{~~}c|}
\hline
& $q$ & $P$ & $G$ & obs.\ dim & $T_\text{curr}$ & $T_\text{end}$ & $t_\text{trunc}$ & $\e_G$ & $a_\text{scale}$ & $B_\text{size}$ & $h_\text{enc}$ & $h_\text{Q,$\pi$}$ \\
    \hline
Double-Sphere & 3 & 3 & 3 & 21 & 50k & 75k & 1.  & 0.01  & 2. & 1M & 256 & 256 \\
Panda-Sphere & 7 & 12 & 3 & 42 & 50k & 100k & 2. & 0.01  & 5. & 2M & 512 & 256 \\
Sphere-Cube & 3 & 15 & 15 & 45 & 100k & 150k & 3. & 0.03  & 2. & 4M & 512 & 256 \\
Panda-Cube & 7 & 24 & 15 & 66 & 100k & 200k & 3. & 0.05  & 5. & 4M & 1024 & 512 \\
Allegro*-Cube & 16 & 27 & 15 & 96 & 100k & 200k & 3. & 0.02 & 2. & 5M & 1024 & 512 \\
\hline
\end{tabular}%
  \smallskip
  \caption{\label{tab:env_dims}
    Dimensionality, training and environment parameters for each environments: robot joint state $q$ (which also gives the action dimension and control error dimension), additional scene point observations $P$, goal scene points $G$, and total observation dimension. RL training steps $T_\text{end}$, curriculum length $T_\text{curr}$, gym truncation time $t_\text{trunc}$, goal threshold $\e_G$, action scale $a_\text{scale}$, replay buffer size $B_\text{size}$, and TD7 zs and enc layer size $h_\text{enc}$.
  }
  \end{table}
  
\section{TD7 \& HER Extension Details}\label{app:HER}

The RL method used within our CSRL framework is exchangeable.
We found TD7 \cite{2023-fujimoto-SaleStateactionRepresentation} to be a constructive method to work with, due to its robust training behavior, scaling to high-dimensional observation spaces, and transparent source code.\footnote{\label{note:td7}\url{https://github.com/sfujim/TD7}}
While we left most hyperparameters of TD7 unchanged, we modified some following the \emph{FastTD3} \cite{2025-seo-FastTD3SimpleFast} paradigm:
In each \emph{iteration}, we collect 2000 transition samples in a GPU-parallelized gym environment, add them to the replay buffer, and perform 20 TD7 training steps.
The training batch size of 10k is comparably large, and the learning rate  increased to $10^{-3}$ for all but the Allegro domains.

When introducing non-sparse shaping rewards, we initially observed instabilities where the $Q$-function predicts negative returns (as shaping rewards for random actions may likely accumulate negative total return).
Training became more stable when clipping the Q-target values to become $\ge0$, based on our apriori knowledge that optimal returns are positive.

Concerning HER \cite{2017-andrychowicz-HindsightExperienceReplay}, note that TD7 does not come with a default HER option, stablebaselines3 wrappers do not correctly relabel the done flag\footnote{\url{https://github.com/DLR-RM/stable-baselines3/issues/627}}, and our goal conditioning does not append to state, but adds to the (last) observation entries (\ref{eqObs}).
To integrate HER in TD7, we complemented the replay buffer with a table of pointers to previous samples in the current episodes of all parallelized environments, allowing to trace back episodes.
After each step we identify environments with episode lengths $\ge3$, declare the current ``next\_state'' as goal, and pick a single random previous transition of the episode to be HER-relabeled (accounting for the change $G^*_\text{new}-G^*_\text{old}$ in (\ref{eqObs}), non-sparse rewards, and the termination flag).
This empirically behaves similar to standard HER conventions but allows efficient GPU parallelization.

\section{Statistics of Constraint Sampling}\label{app:stats}

We perform constrained sampling using a general-purpose NLP solver, included in the source code.
Table~\ref{tab:sampling_stats} provides compute statistics for constraint sampling.
We find that sampling without the panda robot is considerably faster, requiring less NLP queries per sample as well as less wall clock time to evaluate the constraints.
This is largely due to panda's complex shapes and complexity of evaluating the constraints (\ref{onS}-\ref{grav}) for these shapes.
Note that replacing the sphere object by a cube does not severely increase sampling complexity.

\begin{table}\centering\small


\begin{tabular}{|c|c|c|c|c|c|c|}
\hline
  & \shortstack{Double-\\Sphere} & \shortstack{Panda-\\Sphere} & \shortstack{Sphere-\\[-0.5ex]Cube} & \shortstack{Panda-\\Cube} & \shortstack{Allegro-\\[-0.5ex]Cube} & \shortstack{AllegroDown-\\[-0.5ex]Cube} \\
\hline
number $|\DD|$ of samples & 20k & 20k & 20k & 20k & 50k & 50k \\
feasibility rate & 0.356 & 0.116 & 0.214 & 0.051 & 0.488 & 0.121 \\
queries/sample & 578.3 & 2934.9 & 876.0 & 6681.5 & 394.3 & 2645.7 \\
time/sample [msec] & 38.7 & 221.3 & 44.6 & 576.8 & 77.6 & 345.2 \\
total time [h] & 0.22 & 1.23 & 0.24 & 3.20 & 1.08 & 4.80 \\
\hline
\end{tabular}

\smallskip
\caption{\label{tab:sampling_stats}
         Compute statistics for constrained sampling: How many NLP solver runs on (\ref{eq:nlp}) find feasible samples; how often are constraint functions $g,h$ queried per feasible sample, and wall clock time for a single and for all samples.}
\end{table}

\section{Real World Transfer Setup}\label{app:real-world-details}

\begin{figure}[tb]\centering\small
  (a)~%
  \begin{minipage}[b]{.6\columnwidth}\vspace{-8pt}
    \includegraphics[width=.9\linewidth]{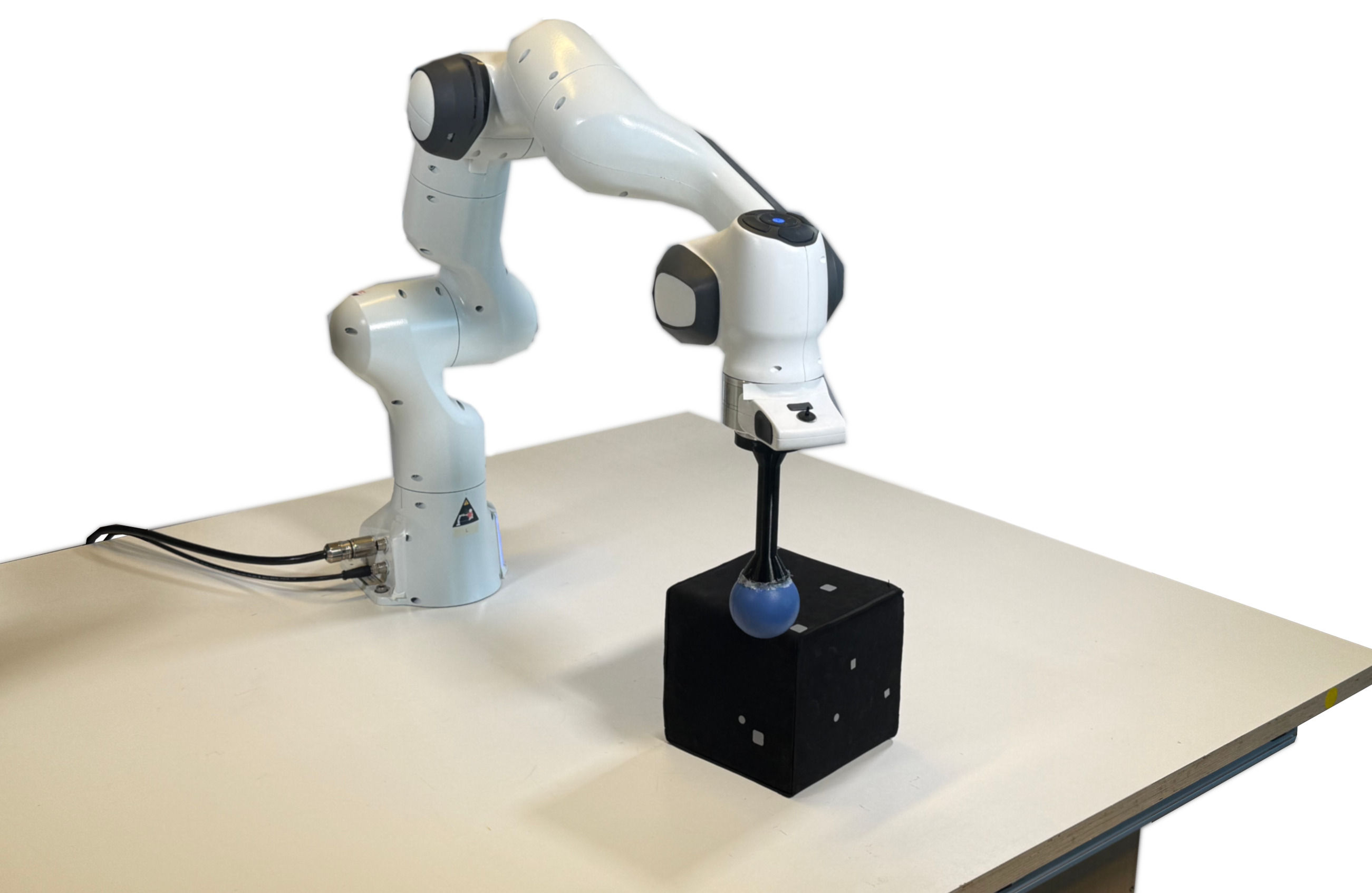}
  \end{minipage}
  \quad(b)~%
  \begin{minipage}[b]{.15\columnwidth}\vspace{-8pt}
  \includegraphics[width=.8\linewidth]{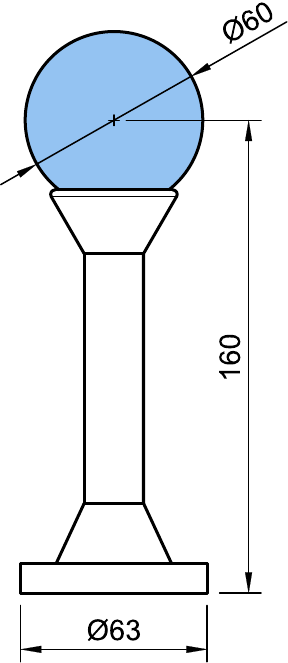}
  \end{minipage}
  \caption{\label{fig:real-world-setup}
(a) Picture of the real world setup, showing the Franka Emika Panda manipulator, the custom end-effector and the cube. (b) Dimensions of the custom end-effector in millimeter.}
\end{figure}

For the real world transfer we equip a Franka Emika Panda manipulator with a custom ball end-effector.
The end-effector consists of a ball with \qty{3}{cm} radius, positioned \qty{16}{cm} away from the end-effector flange of the manipulator.
Fig.~\ref{fig:real-world-setup}(a) shows a picture of the setup and (b) the dimension of the end-effector.
As the policy is state-based, we track the cube's position by using an Optitrack system with 12 Flex13 cameras.


\section{Comparison to Behavior Cloning from Diverse Trajectory Data}\label{app:BC}

In a previous version of this work,\footnote{\label{note:v1}Available on arXiv at \url{https://arxiv.org/abs/2602.08557v1}.} we explored alternative options to train universal policies based on diverse constrained samples.
In particular, we considered leveraging zero-order trajectory optimization to first create a data set of open loop trajectories between random starts and goals from our constrained sampler.
We then studied adding a behavior cloning (BC) objective to RL training -- as is common in offline RL \cite{2021-fujimoto-MinimalistApproachOffline} -- to guide training our policy.
The insights from this study are valuable: namely that BC from such open loop trajectories contributed little more than providing interpolating states to effective RL training.
Given this negative result as well as the substantial computational cost of first generating open loop trajectories via zero-order methods, we decided to remove this approach from the main paper.
However, for the purpose of reporting also negative results we include the method description and result discussion here.
Please note that the quantitative results in the original version are generally much below the quality of policies we report in this paper.
This is due to substantial improvements on our choice of RL hyperparameters.

\subsection{Zero-Order Optimization of Open-Loop Trajectories}\label{sec:TO}

A powerful approach to guide RL is using model-based trajectory data, e.g.\ using it for behavior cloning (BC)~\cite{liu2025opt2skill} or combining BC with RL~\cite{2023-grandesso-CactoContinuousActorcritic}.
However, while we have a model-based description of the constrained space $\SS_c$, our dynamics are black-box, rendering model-based approaches for trajectory generation inapplicable.
We instead investigate using zero-order optimization methods and to this end formulate a low-dimensional spline-based open-loop trajectory optimization problem.

Given random start and goal $(s,g)$, we aim to find an open-loop control and state trajectory $u(t), x(t)$ such that $x(0) = s$ and $x(T)=g$ under \emph{deterministic} dynamics $\dot x = f(x,u)$, which are a deterministic version of our underlying MDP.
We choose a continuous time control formulation here, as the discrete time step of simulating the dynamics differs from the action time step of the MDP.

We assume $u(t)$ is B-spline parameterized by $K$ control points $\t_k \in \RRR^n$, where $n$ is the system's control dimension.
Given $(s,g)$, we solve the optimization problem
\begin{align}\label{traj}
  \min_\t \norm{\phi(g) - \phi(x(T))}^2 \st x(0)=s, \dot x=f(x,u) ~,
\end{align}
where $\phi(x)$ is a state feature that we design so that a low Euclidean distance $\norm{\phi(g) - \phi(x(T))} \le \e$ indicates successfully reaching the goal.
Note that this formulation does not minimize any control costs.
However, our parameterization of controls $u(t)$ as a B-spline with only a few control points automatically implies smooth trajectories.

We define $\phi(x)$ as a weighted concatenation of various state features, namely the object position $p$, robot generalized coordinates $q$, their time derivatives $\dot p$, $\dot q$, and $c \in [0,1]^{m-1}$.
The latter is a continuous indicator for proximity of the object (shape $i=1$) with all potential supports (shapes $i=2,..,m$).
Specifically, $c_i = 1 - [d_{1i}/\s]_{[0,1]}$ where $[\cdot]_{[0,1]}$ clips the value to the interval $[0,1]$.
Therefore, $c_i=1$ if the object is in contact with support $i$, and decreases linearly to $c_i=0$ if support $i$ has distance $d_{1i}\ge \s$.
To be concrete, in our evaluations, we chose weights $\phi(x) = (2 p, 1 q, 0.1 \dot p, 0.1 \dot q, 0.1 c) \in \RRR^{6+2n+m-1}$ to design the state feature and thereby optimization objective.
Note that the $c$ component provides the solver with an (implicit) gradient towards reaching the goal contact modes; while components $p$ and $\dot p$ often have no gradient at all when the initialization has no contact between the robot and the object.

The literature offers a wide spectrum of ideas to solve such a black-box open-loop trajectory optimization problem, including sample-based MPC \cite{2025-jordana-IntroductionZeroOrderOptimization}, kinodynamic motion planning \cite{wahba2024kinodynamic}, or again RL \cite{otto2023deep}.
As it concerns only our secondary problem, it is beyond the scope of the paper to compare these.
We instead choose a robust black-box optimization method, namely CMA-ES \cite{hansen2001completely}, to solve (\ref{traj}). To generate a trajectory dataset
\begin{align}
  \DD_u = \left\{\left(s_i, g_i, u_i(t), x_i(t)\right)\right\}_{i=1}^U ~,
\end{align}
we uniformly sample $s,g \sim \DD_s$, run CMA-ES, and if feasible add $(s,g,u(t),x(t))$ to the generated trajectory dataset until $U$ feasible trajectories (with terminal cost $\le\e$) are found.

\subsection{Behavior Cloning from Trajectory Data}

Our problem (\ref{eqOpt}) is a standard goal-conditional RL problem for which we can leverage any baseline method by defining observations to encode both, the state $s$ and goal $g$.
Rewards $R(s,g)\in\{0,1\}$ indicate the object distance to the goal is below a threshold $\e$.
As our evaluations confirm, the approach of directly sampling random starts and goals $(s,g)\sim p_0$ during training will lead to ineffective RL.
In particular, the likelihood of observing rewards is initially low, leading to a slow or fully muted learning progress.

A well-established approach to guide RL is to integrate behavior cloning (BC): We follow the TD3+BC approach \cite{2021-fujimoto-MinimalistApproachOffline}, but within the more recent TD7 framework \cite{2023-fujimoto-SaleStateactionRepresentation}, which adds a simple regularization to the policy update objective
\begin{align}\label{eq:BC}
  \max_\pi \Exp[s,g,a\sim \DD]{Q(s,g,\pi(s,g)) + \l (\pi(s,g) - a)^2} ~,
\end{align}
where we include goal-conditioning, $\DD$ is a batch of BC data (see below), and we chose the $\l$ to weigh the BC regularization.

To provide technical detail: To realize behavior cloning based on our trajectory dataset $\DD_u$, we need to compile continuous-time control B-splines $u(t)$ to become compatible with the MDP definition.
Note that we use 2nd-order B-splines, which are piecewise 2nd-order polynomials.
We ensured that the knots between pieces align with the time step discretization of the MDP and that a 2nd-order polynomial piece can be translated uniquely to an action $a$ (the 20\,Hz actions are essentially relative position control in joint space; but encoded as single-control-point B-splines. 
By running over $i=1,..,U$ and $t=1,..,T$, the trajectory data can therefore be compiled to a BC dataset
\begin{align}
  \DD_\text{BC} = \{ (s_j, g_j, a_j) \}_{j=0}^{U\cdot T} ~,
\end{align}
with $g_j = g_i$, $s_j=x_i(t)$ and $a(u_i(t))$, with guarantee that these sequences of actions reproduce the exact same state trajectory in a deterministic simulator (such as MuJoCo \cite{todorov2012mujoco}).


Note, however, that mixing a BC loss to the RL updates in (\ref{eq:BC}) is only possible when for each episode we sample $(s,g)$ from the available trajectory data.
The latter, sampling the start state $s$ from the available trajectories, is in itself a very strong bias to guide RL.
This bears the question whether BC is effective because of the BC loss, or because we bias state initialization along the available trajectory data.
In fact, during our research, we gradually shifted from a focus on BC to a focus on biasing state initialization, which turned out crucial.

\subsection{Results: What is the role of BC in effective universal RL?}

\emph{[Please refer to the original version of th paper on arxiv for the specific quantitative results discussed here.]}

Our results indicate that while the state initialization (i.e.\ reset) method has a drastic impact on the training behavior and final performance, adding a BC objective has minor impact.
Therefore, in our settings the core role of trajectories is demonstrating relevant ``interpolating'' states rather than demonstrating actions.
This becomes more intuitive in view of the profound qualitative difference between open-loop trajectories and final policy rollouts (see the accompanying video):
The policies exhibit highly reactive behavior, e.g.\ realizing object transports against the wall with brief reactive push maneuvers and showing recovery behavior, while the spline-based open-loop trajectories exhibit impressive strategies, but seem more conceptual rather than real-world executable.
This view aligns with extensive recent work on the challenge of distilling optimization-based trajectory data into reactive policies \cite{2023-jia-SEILSimulationaugmentedEquivariant,2024-ke-CCILContinuitybasedData,2021-ke-GraspingChopsticksCombating}, e.g.\ suggesting data augmentation to compensate for the lack of reactive control behavior in demonstrations.
In our contact-rich domains, these challenges seem particularly pronounced.

\section{Random Constrained Samples from our Domains}\label{app:samples}

The following tables illustrates the diversity of samples our constrained sampler generated in the four domains.

\begin{figure}[p]\centering
\newcommand{\pic}[1]{\includegraphics[width=.25\columnwidth,trim={0 0 0 25pt},clip]{data/configs/images/double_sphere_sample_0000#1.jpg}}
\pic{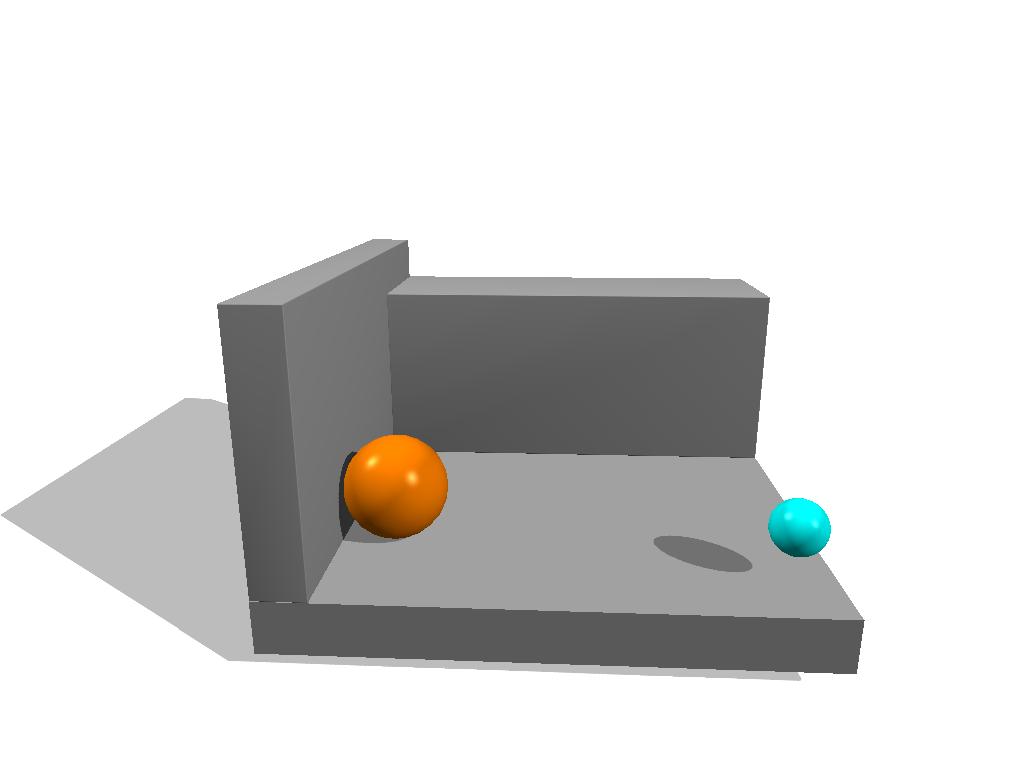}\pic{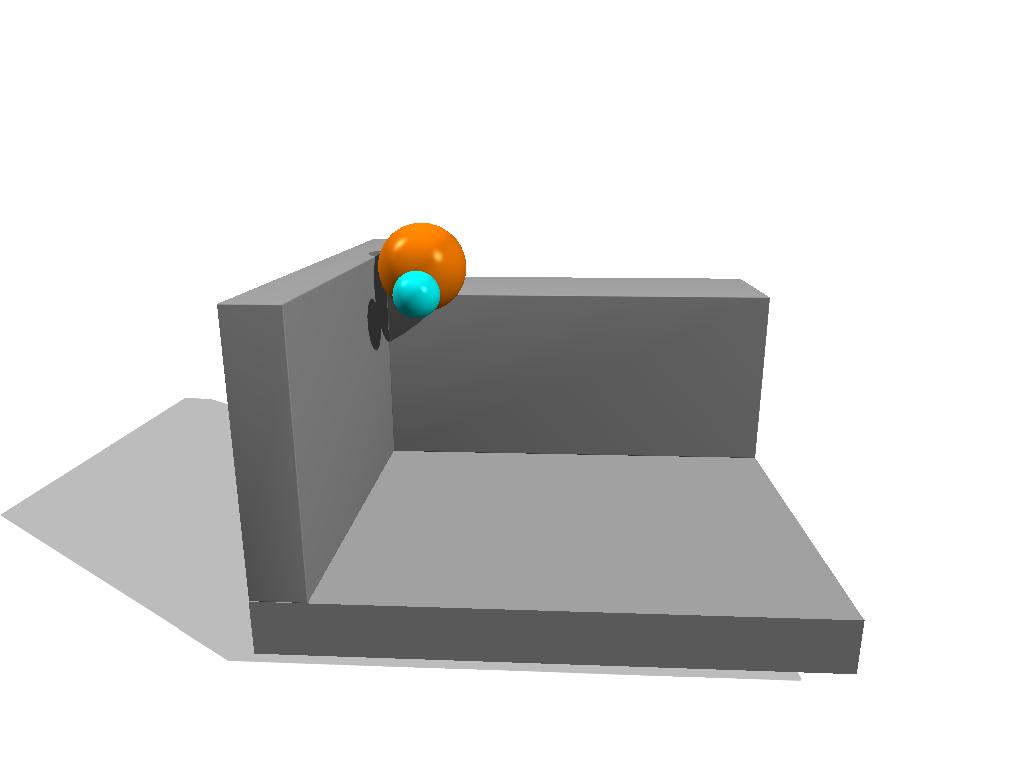}\pic{02}\pic{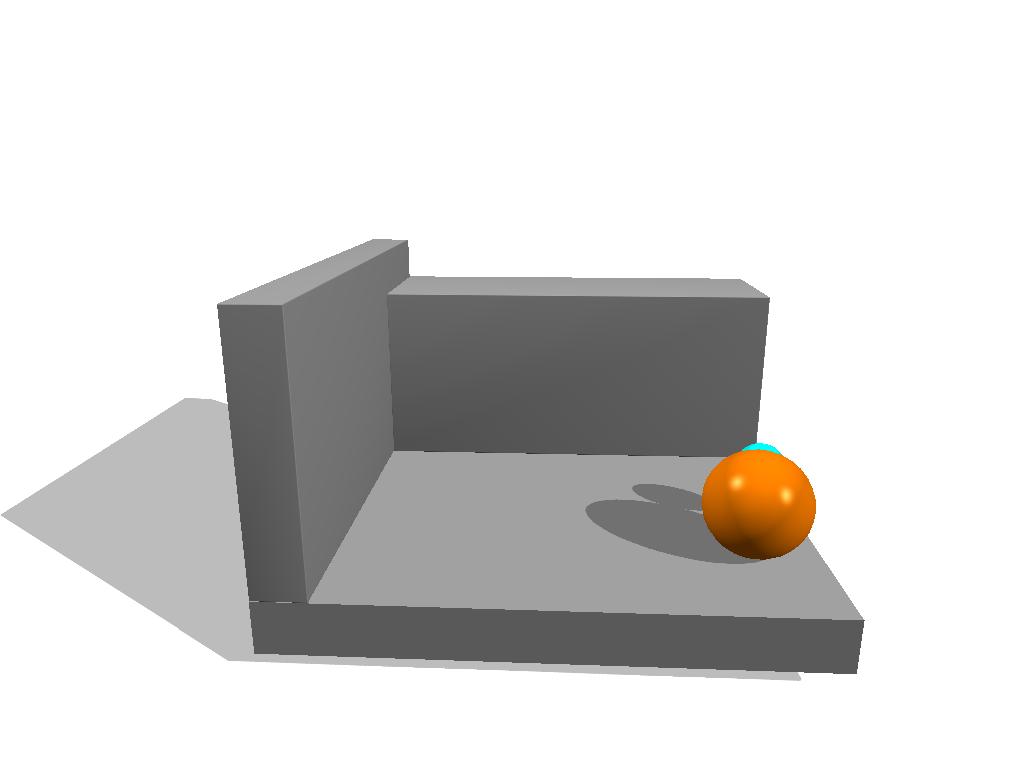}
\pic{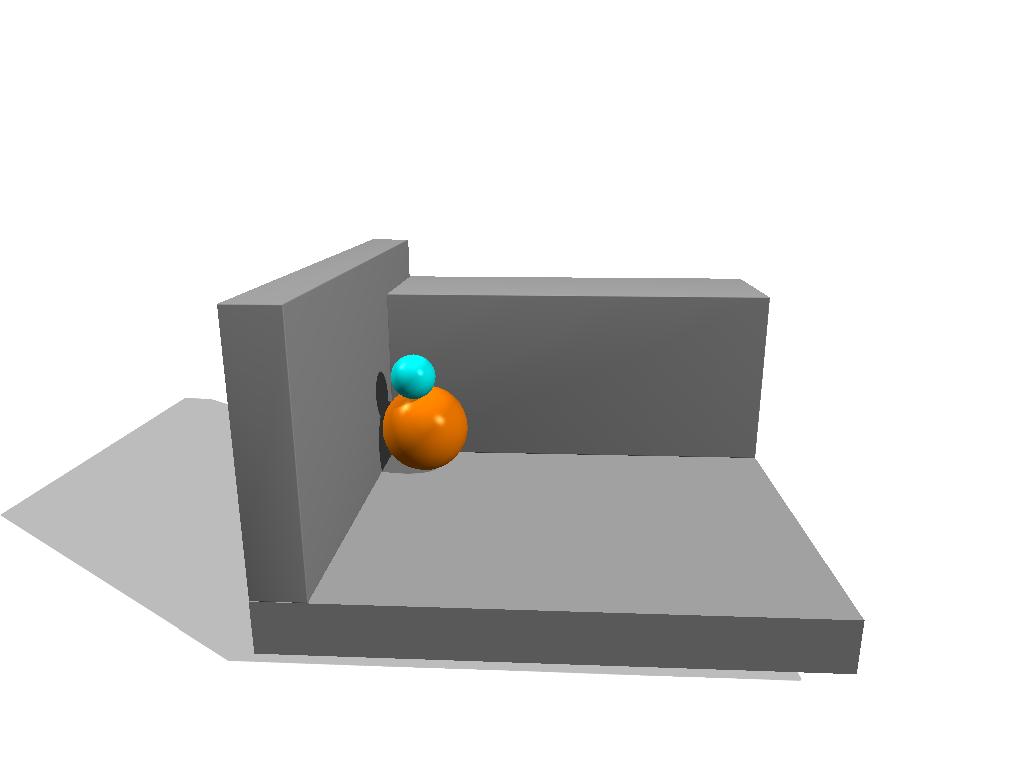}\pic{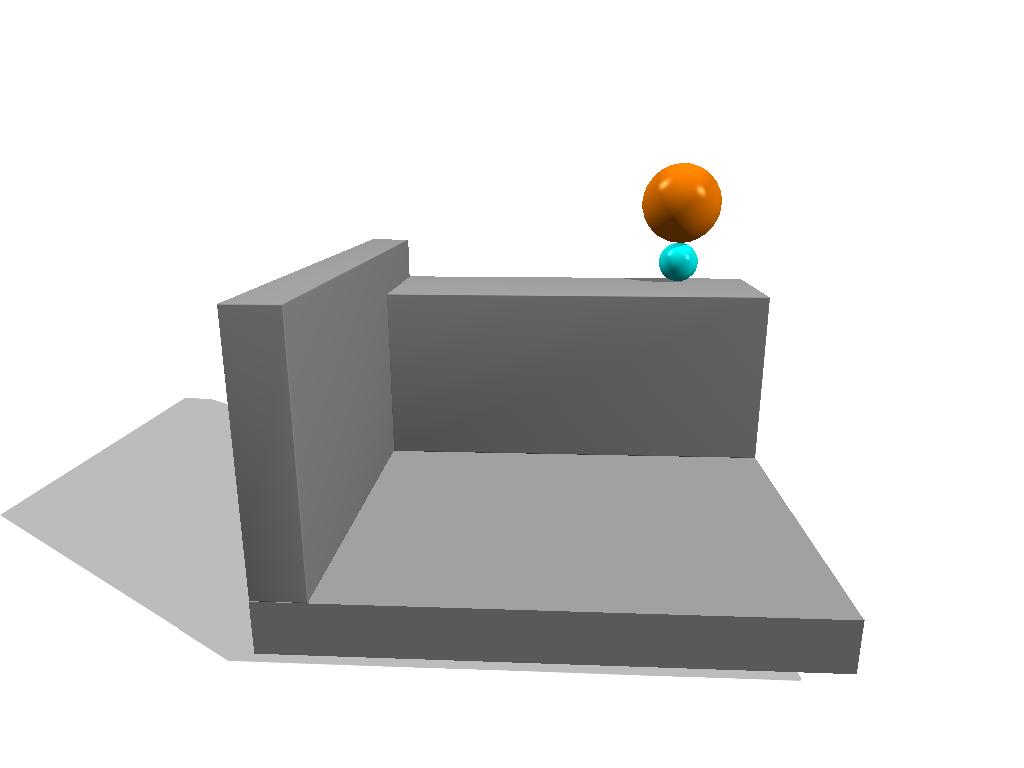}\pic{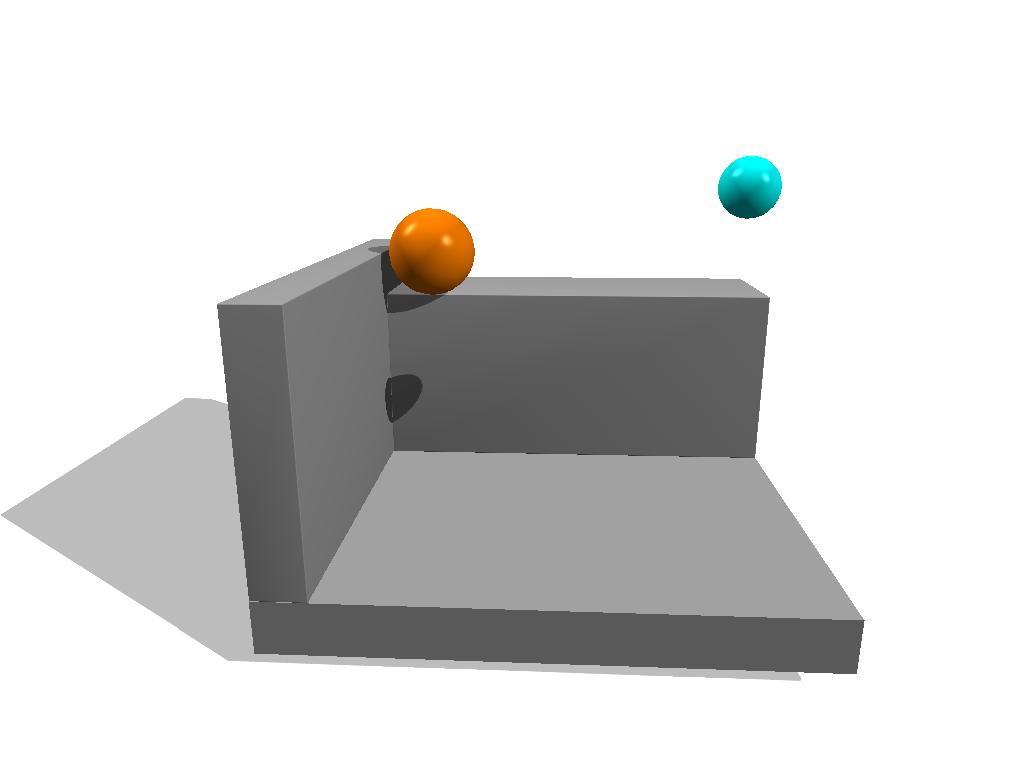}\pic{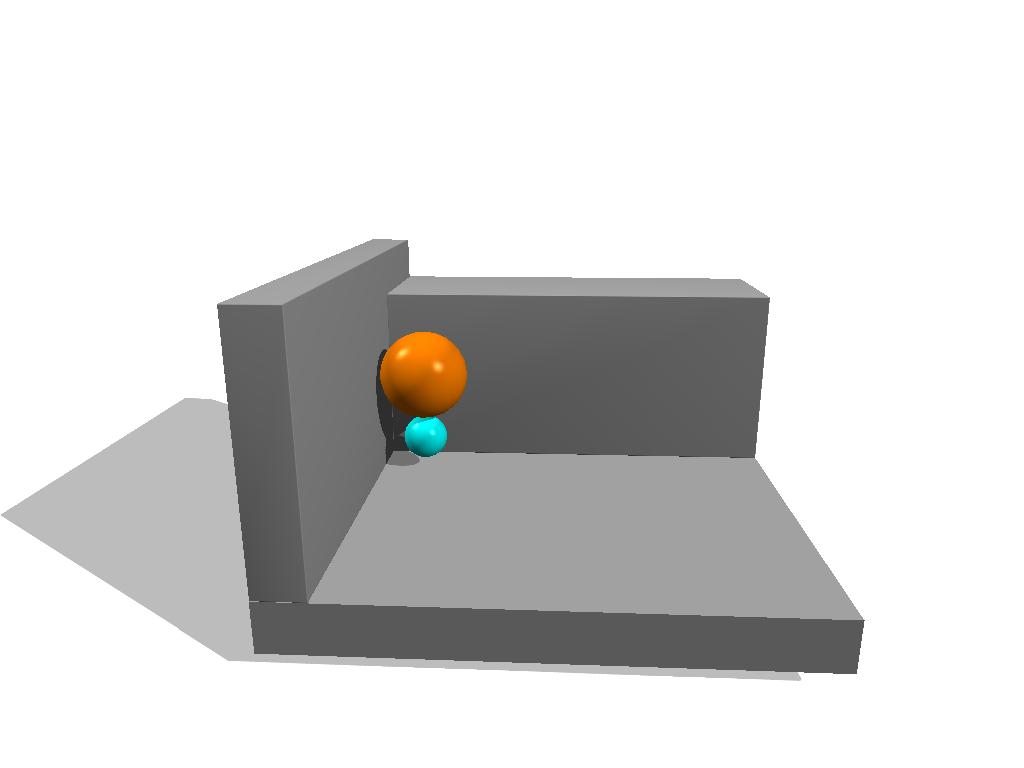}
\pic{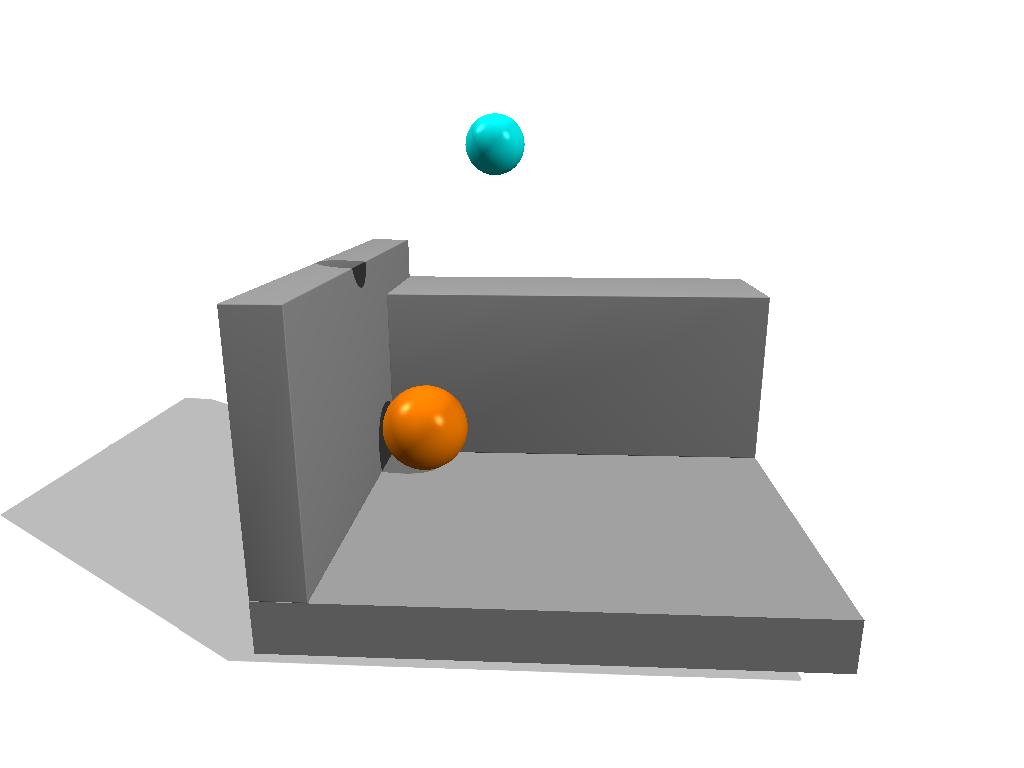}\pic{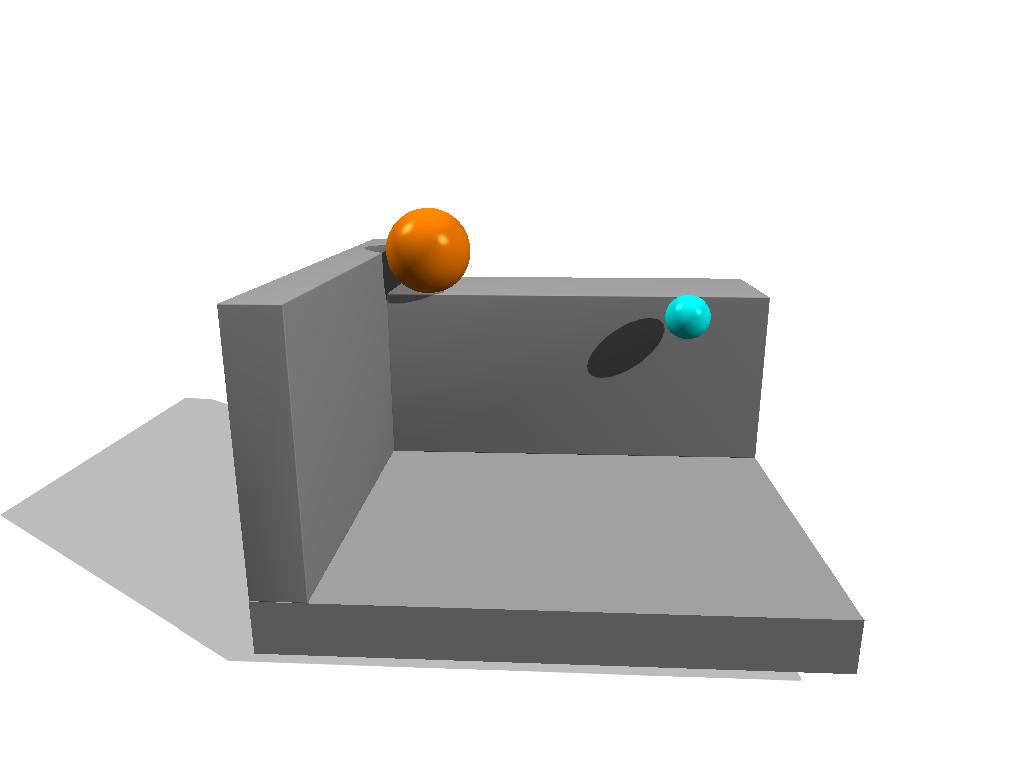}\pic{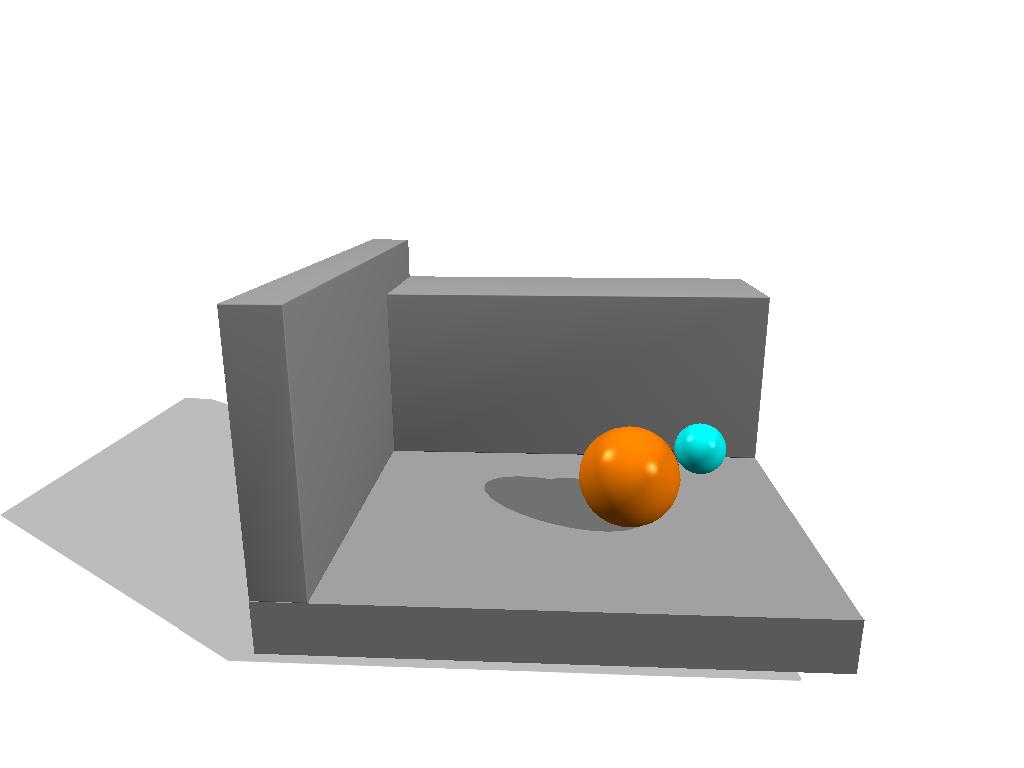}\pic{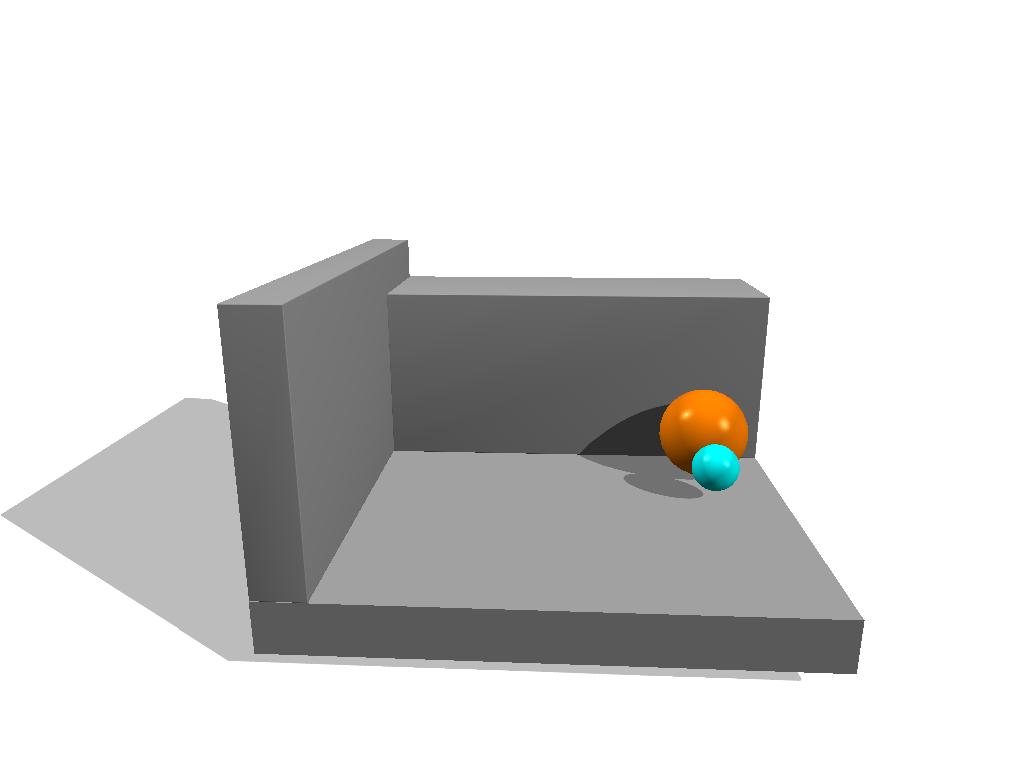}
\pic{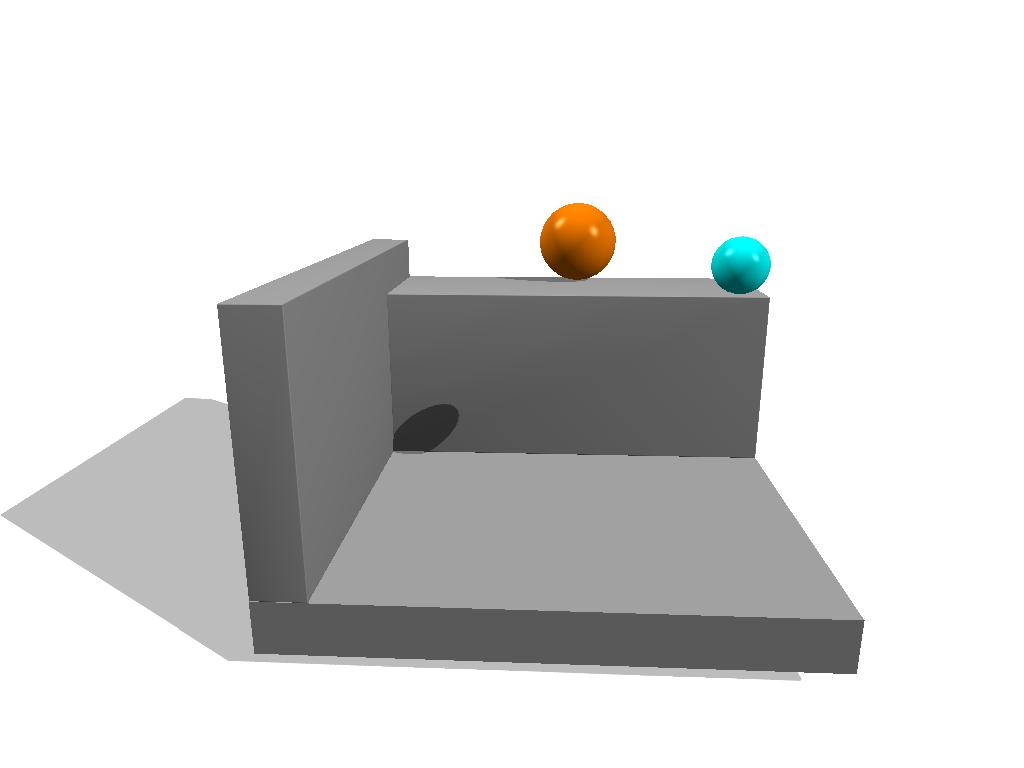}\pic{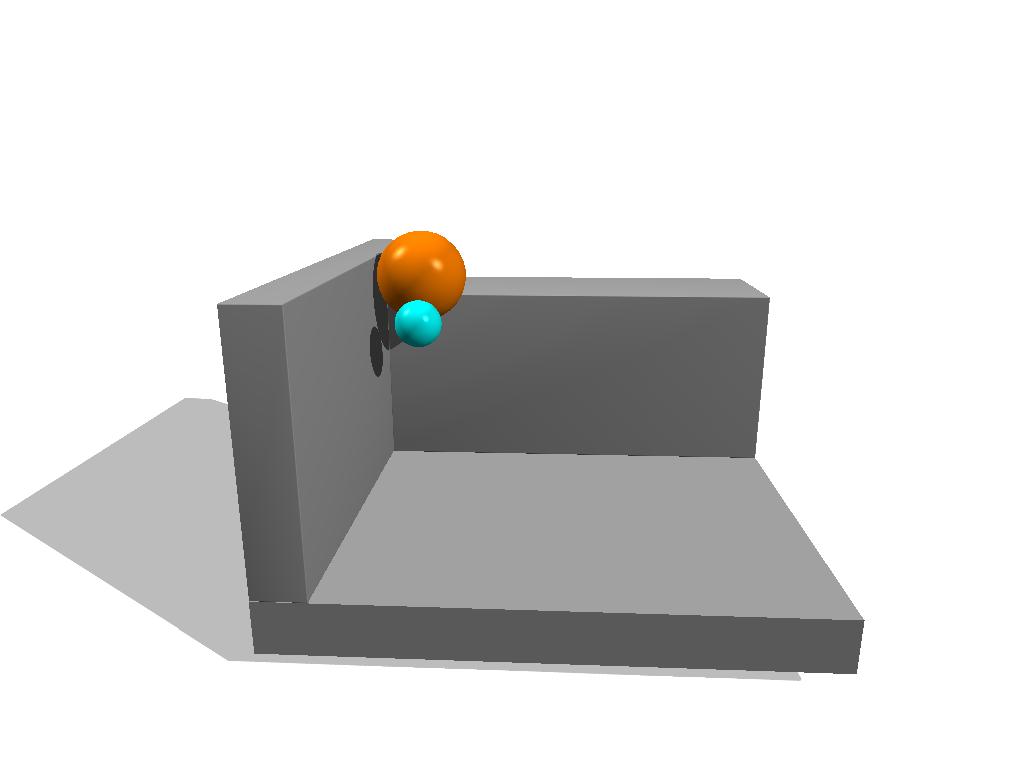}\pic{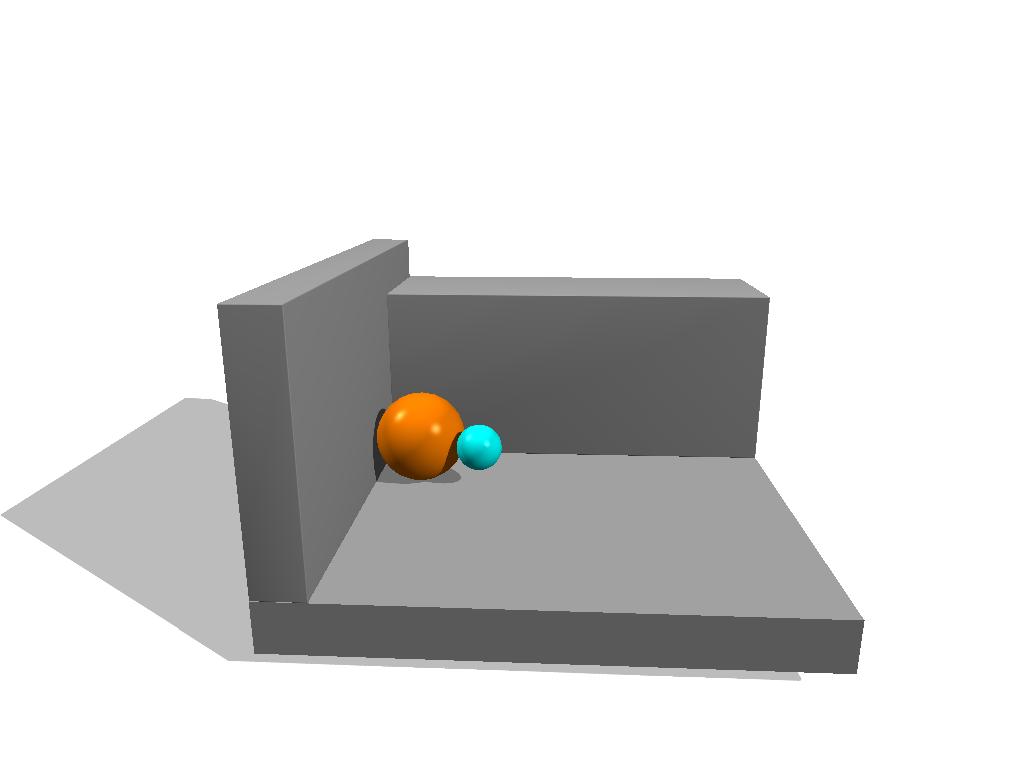}\pic{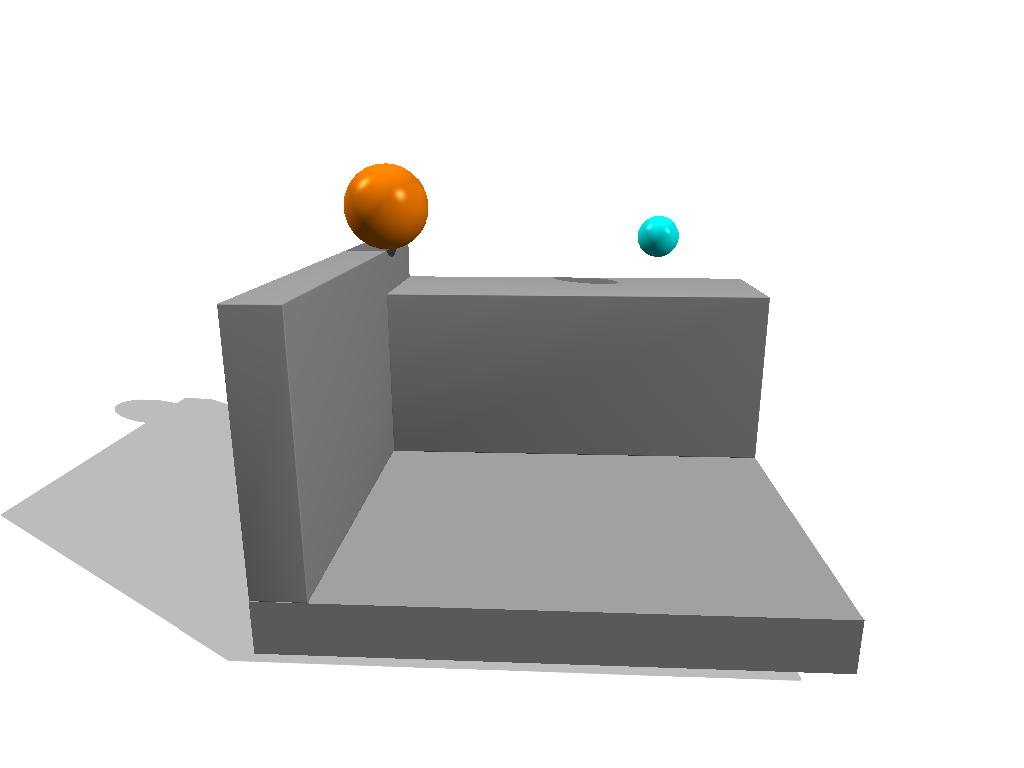}
\pic{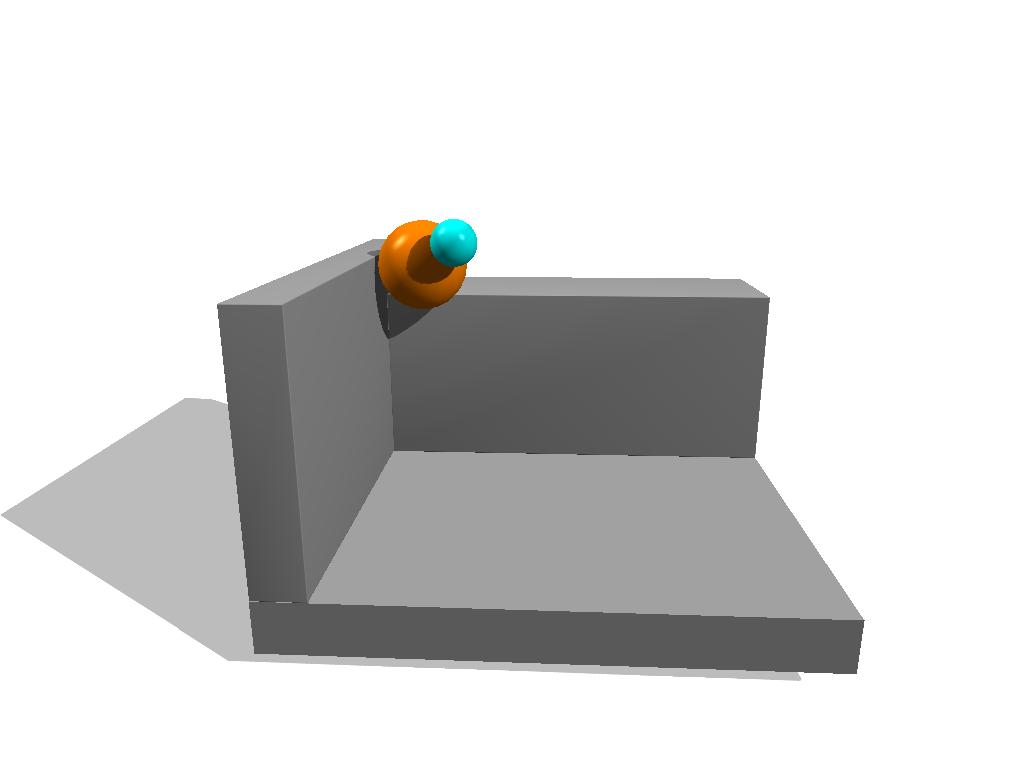}\pic{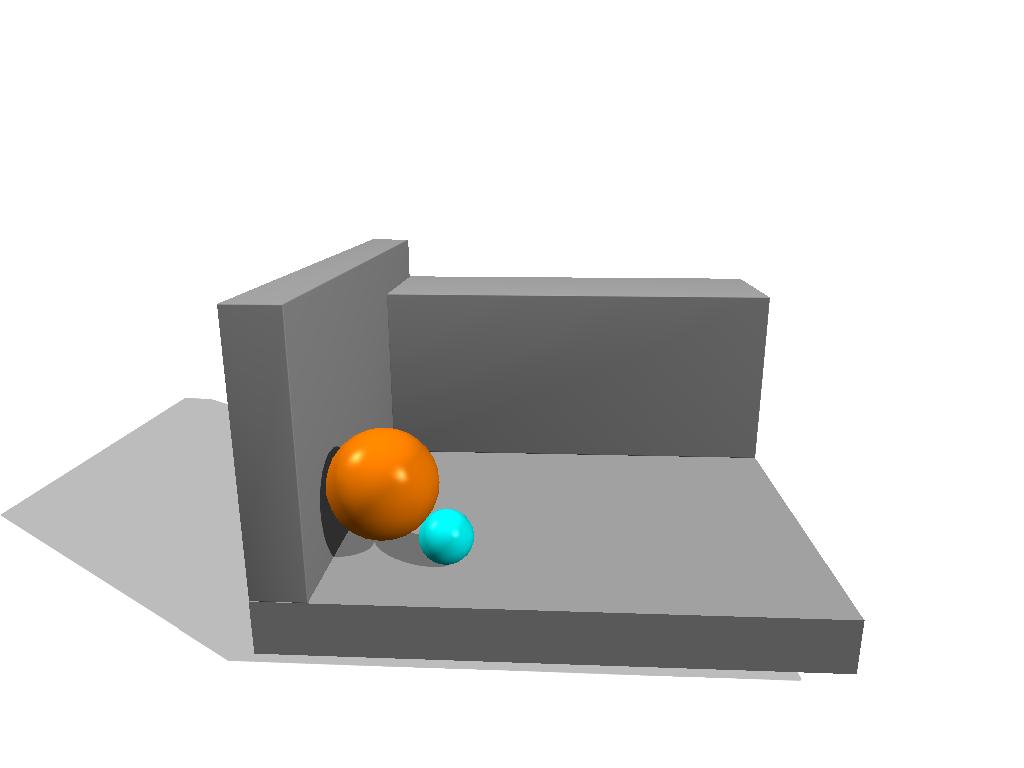}\pic{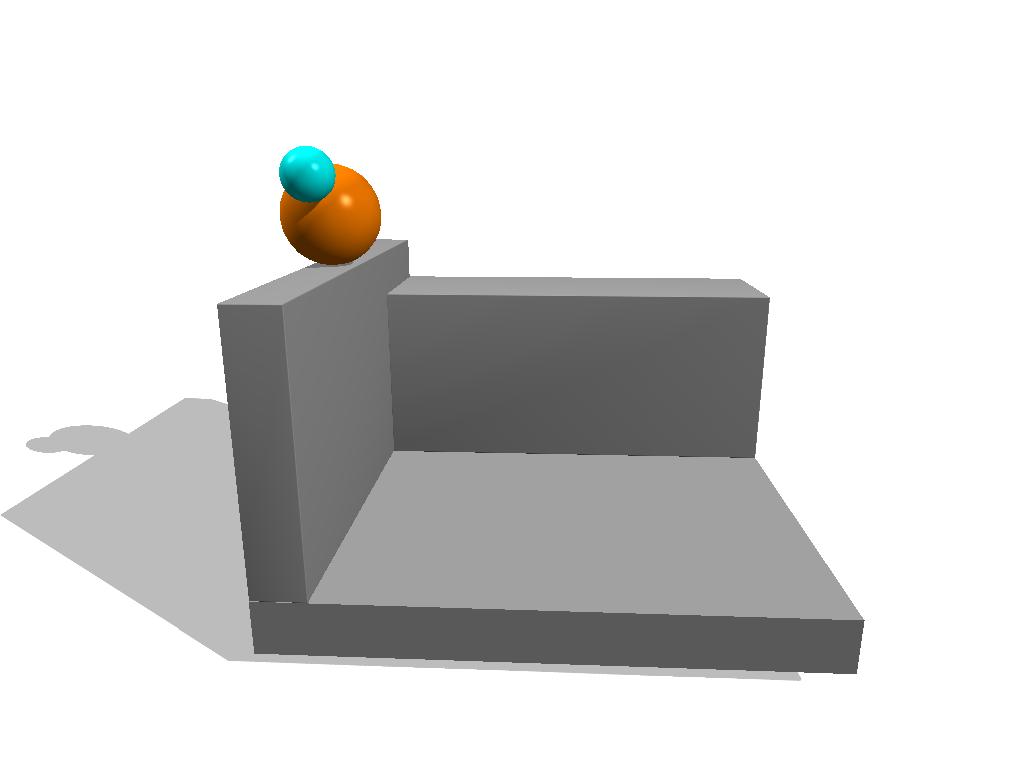}\pic{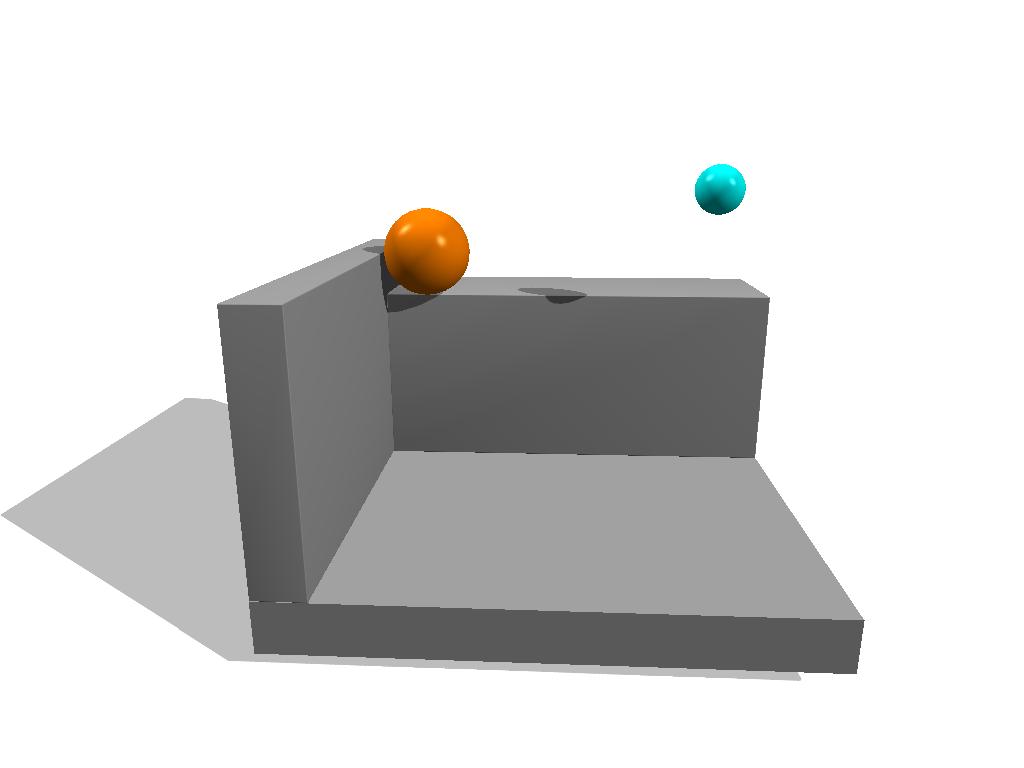}
\pic{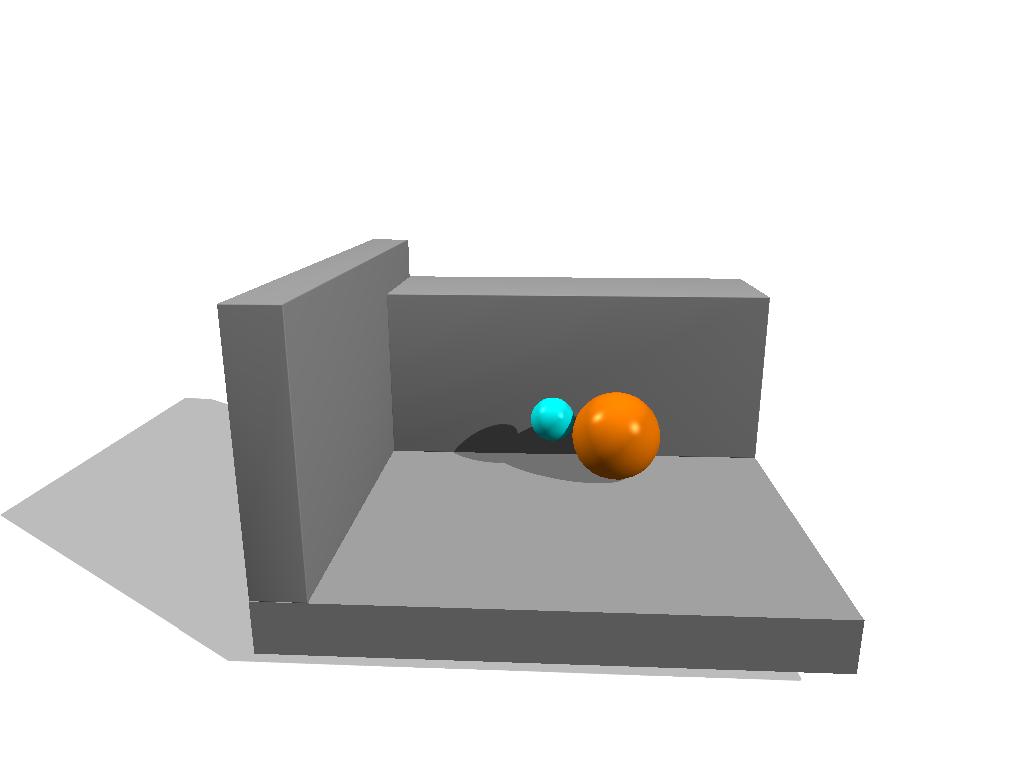}\pic{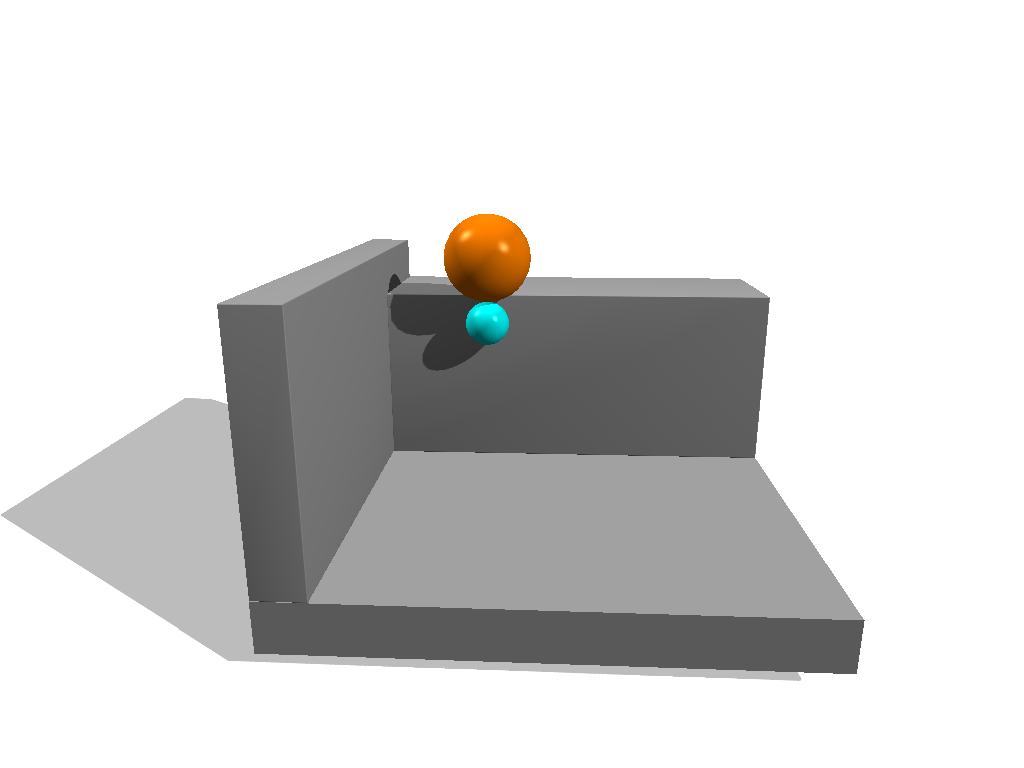}\pic{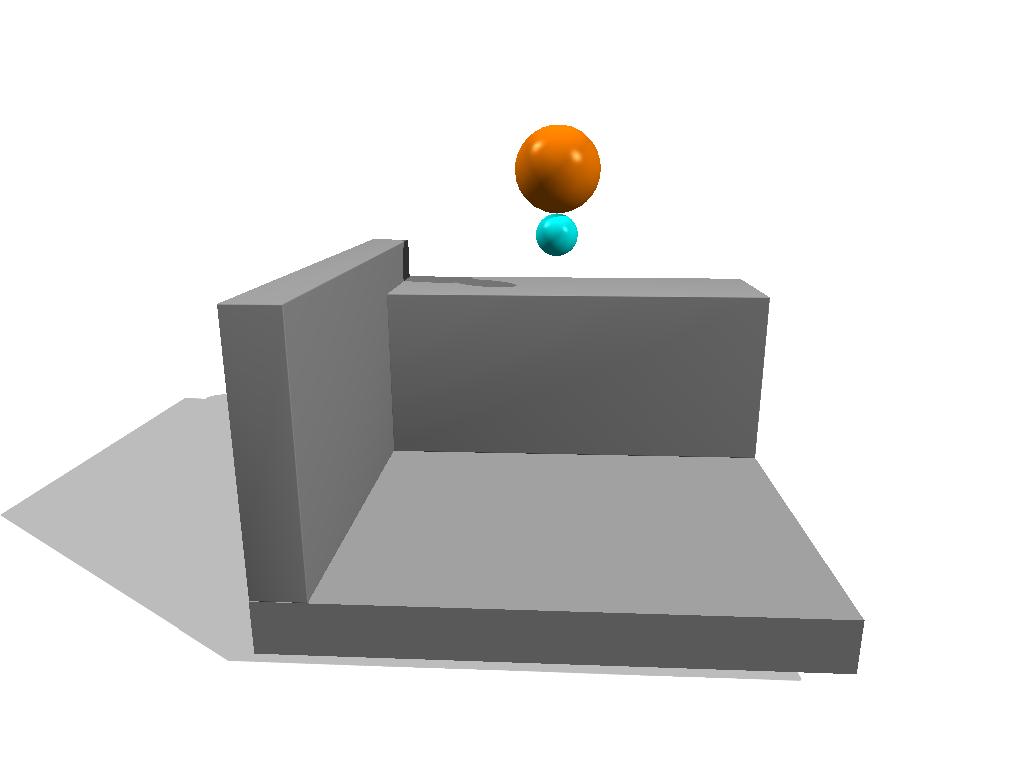}\pic{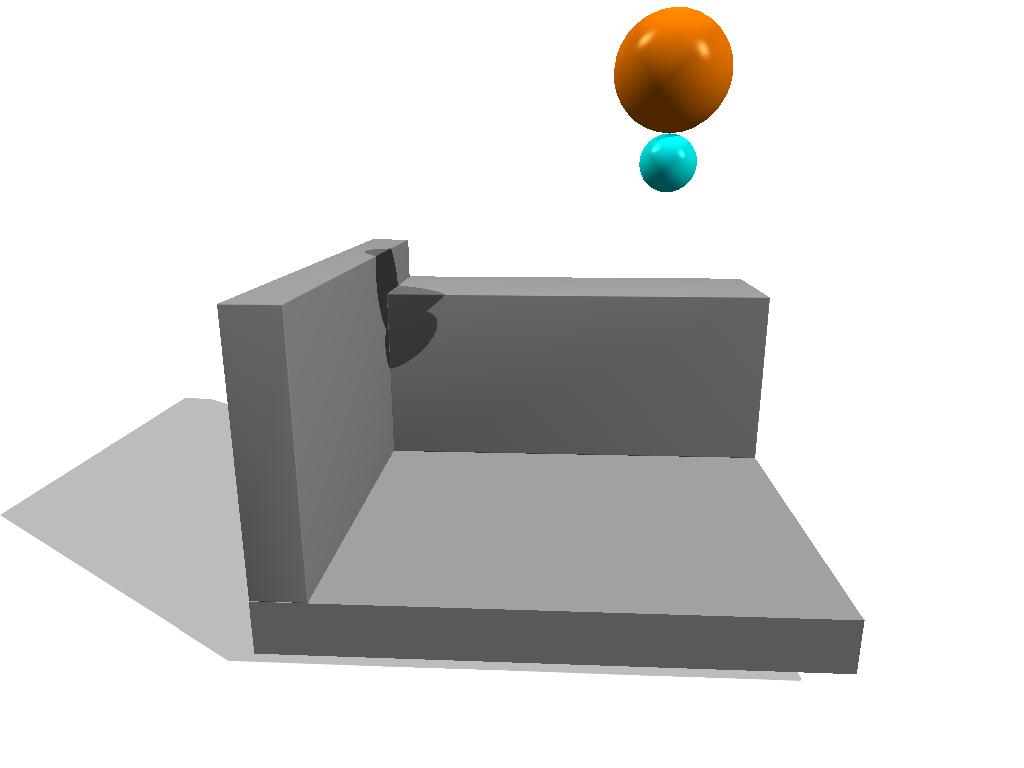}
\pic{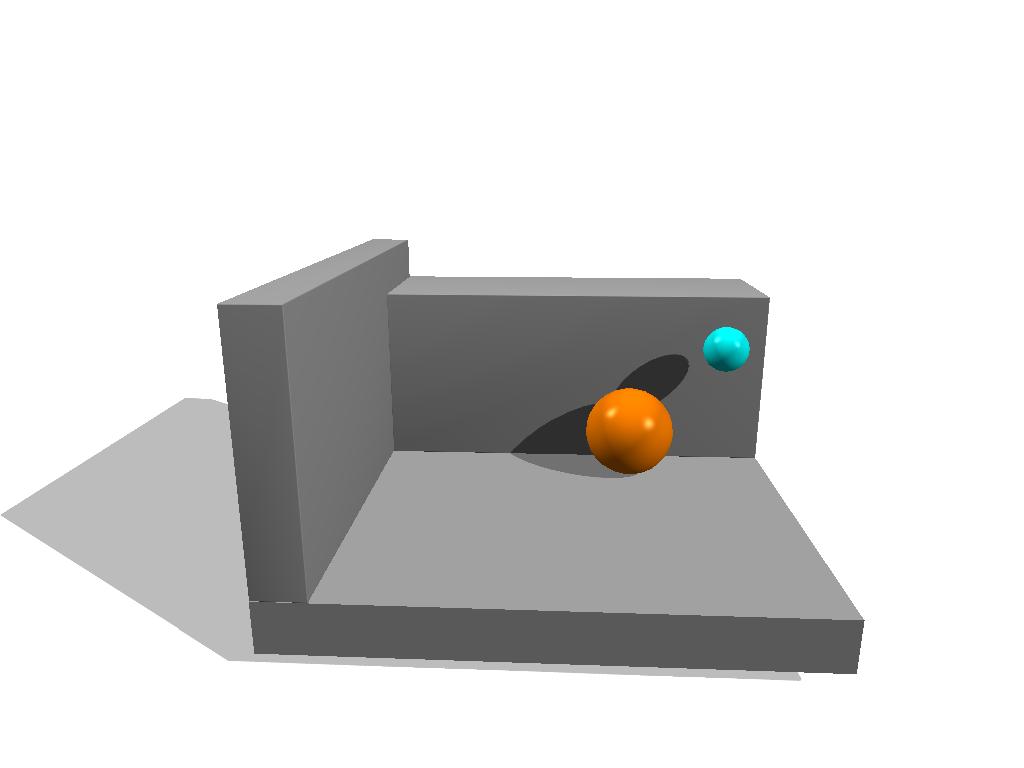}\pic{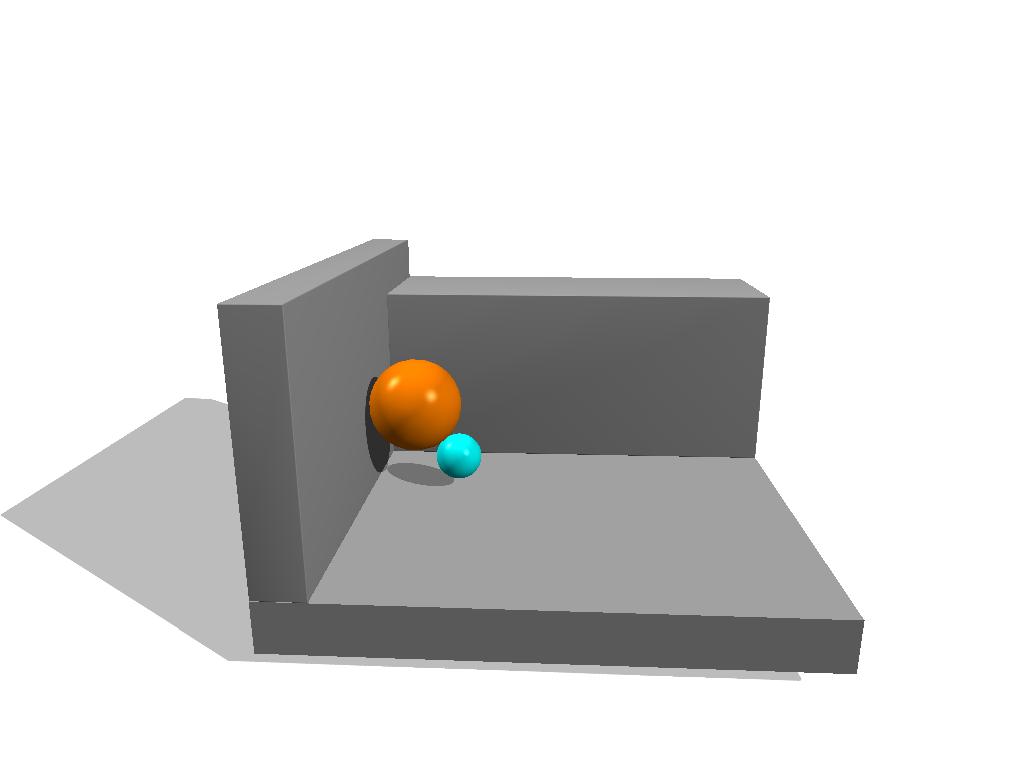}\pic{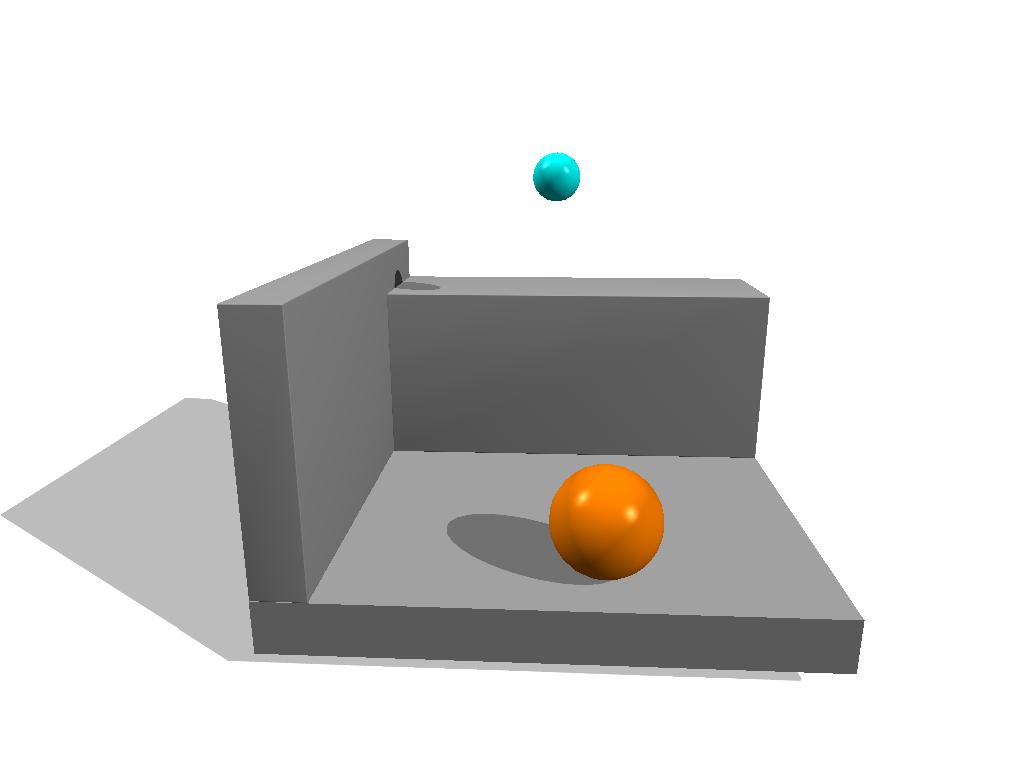}\pic{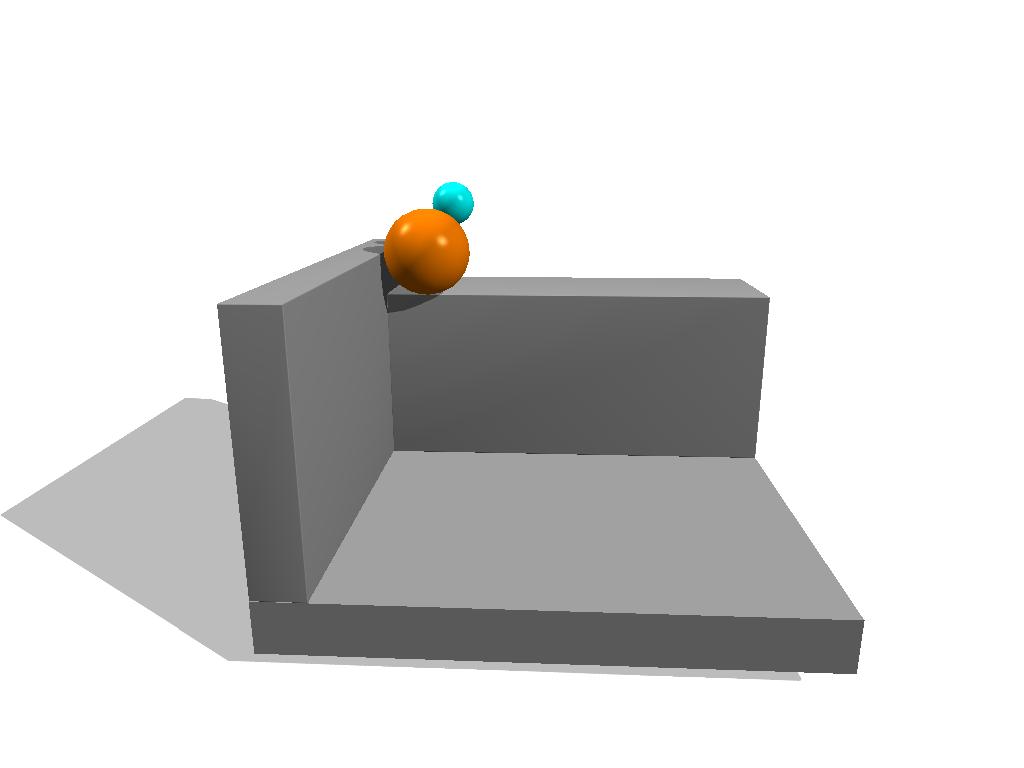}
\caption{\label{fig:double_sphere}
  Constrained samples from the double sphere domain.}
\end{figure}

\begin{figure}[p]\centering
\newcommand{\pic}[1]{\includegraphics[width=.25\columnwidth,trim={0 0 0 25pt},clip]{data/configs/images/panda_sphere_sample_0000#1.jpg}}
\pic{00}\pic{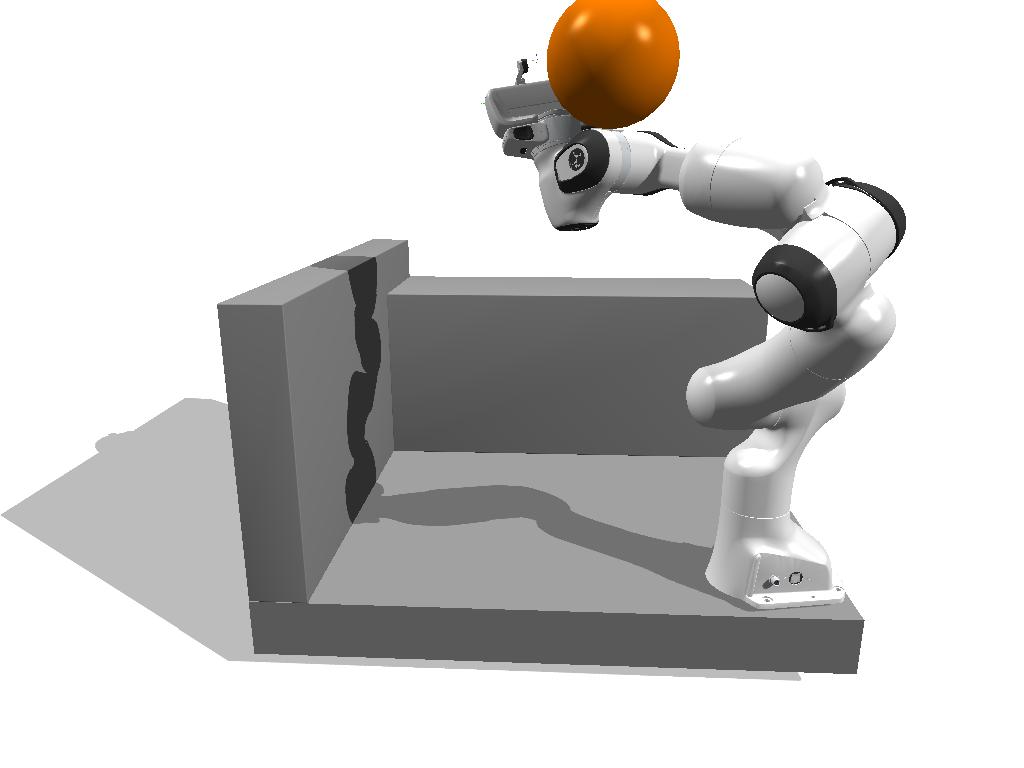}\pic{02}\pic{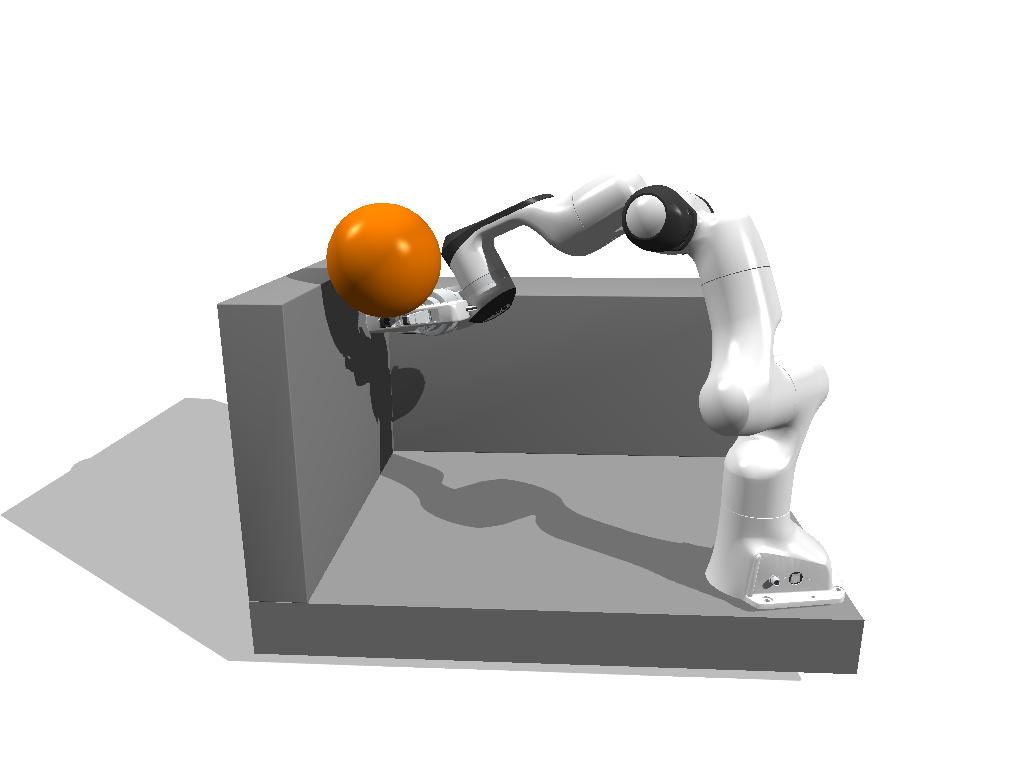}
\pic{04}\pic{05}\pic{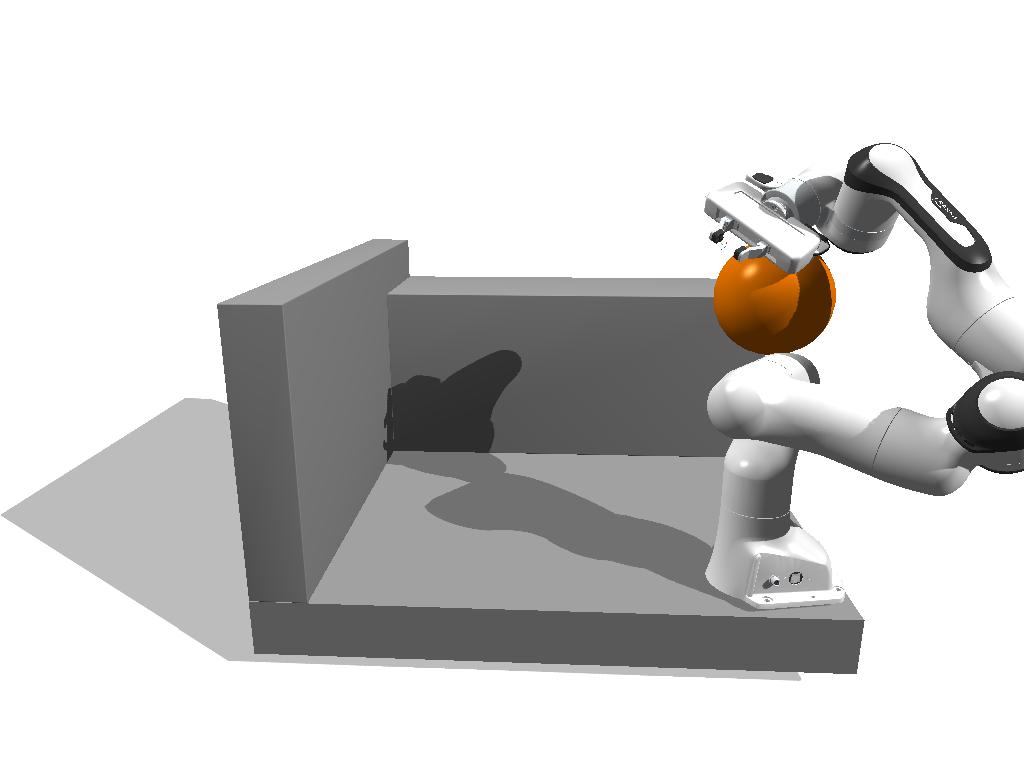}\pic{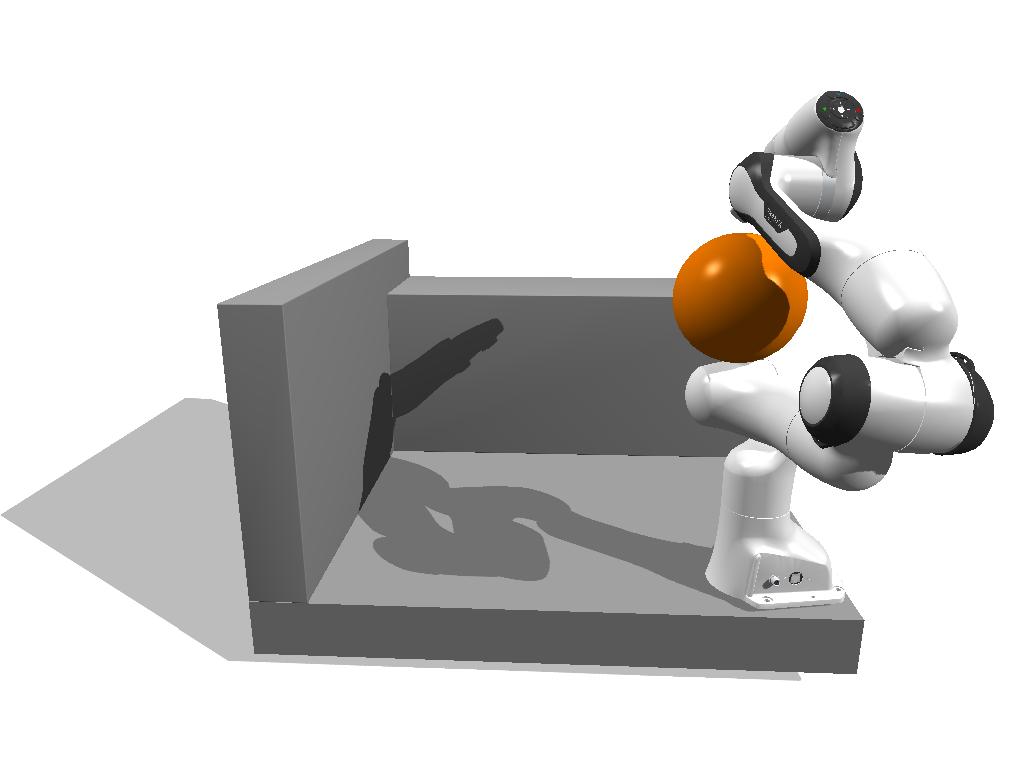}
\pic{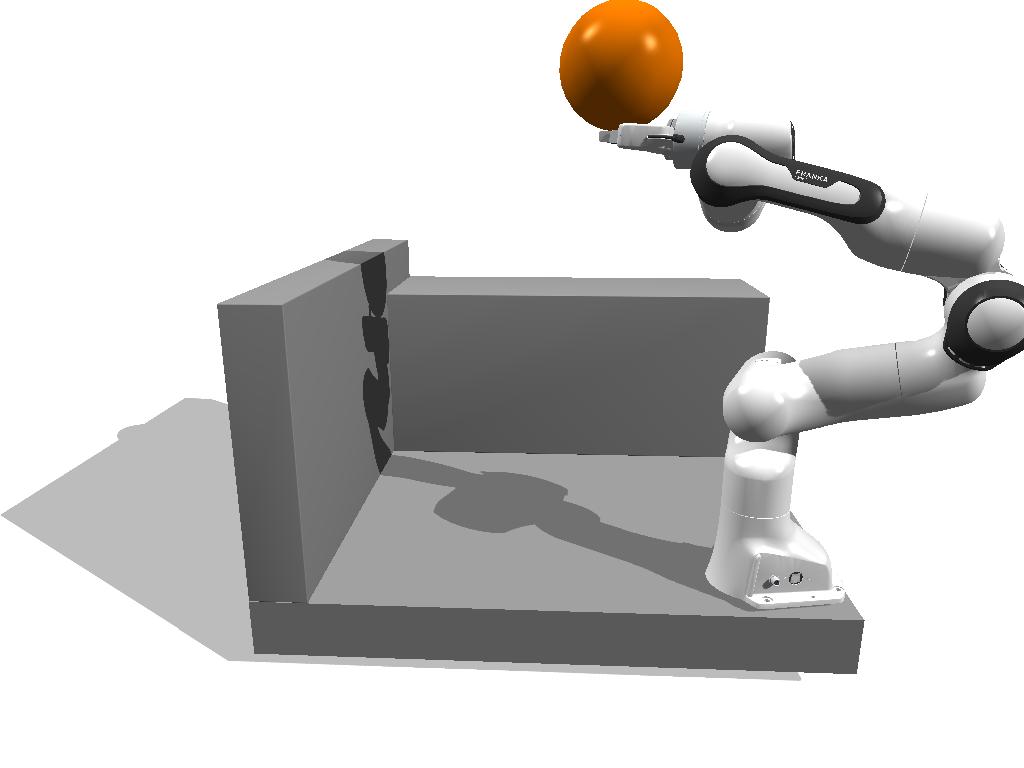}\pic{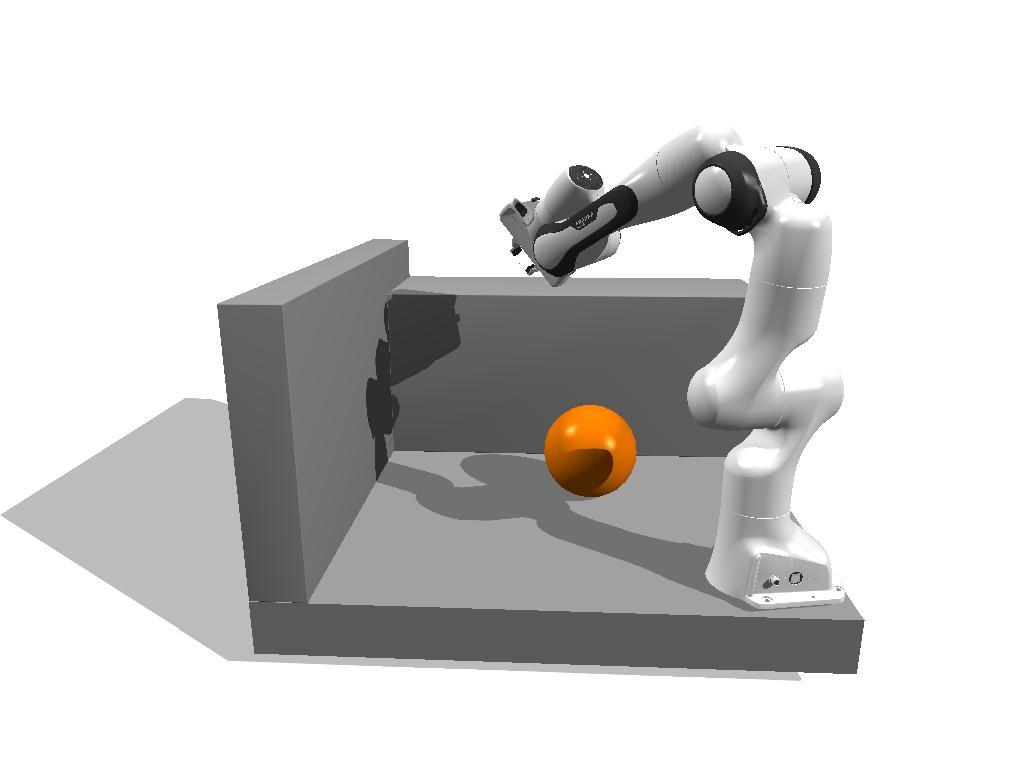}\pic{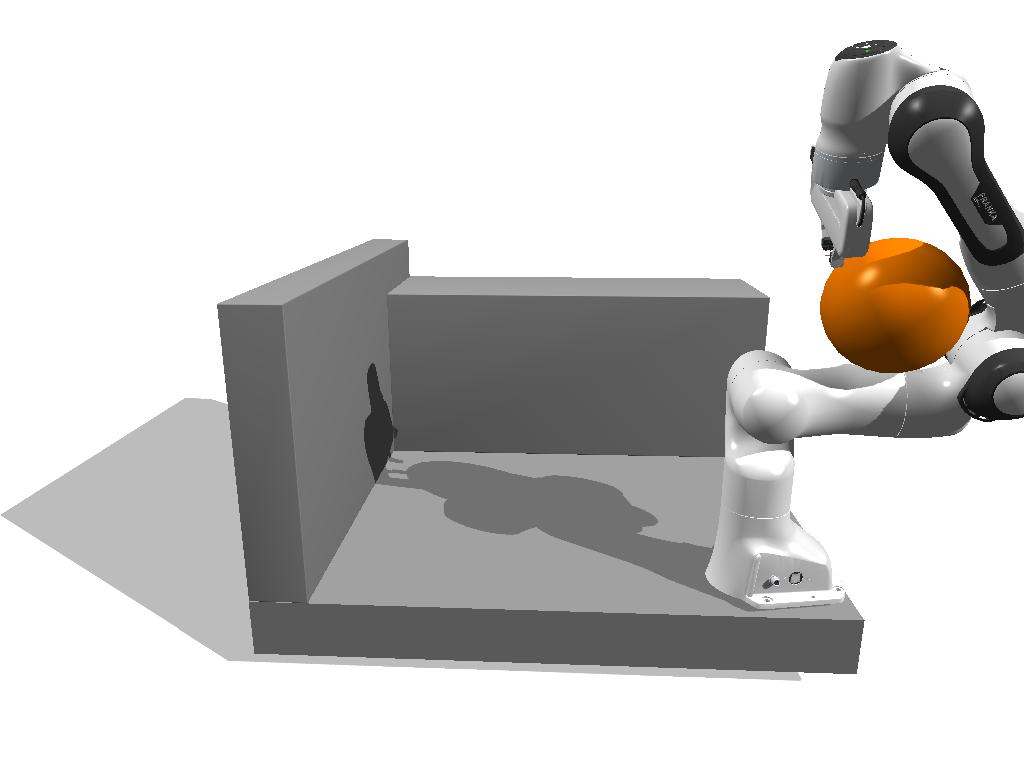}\pic{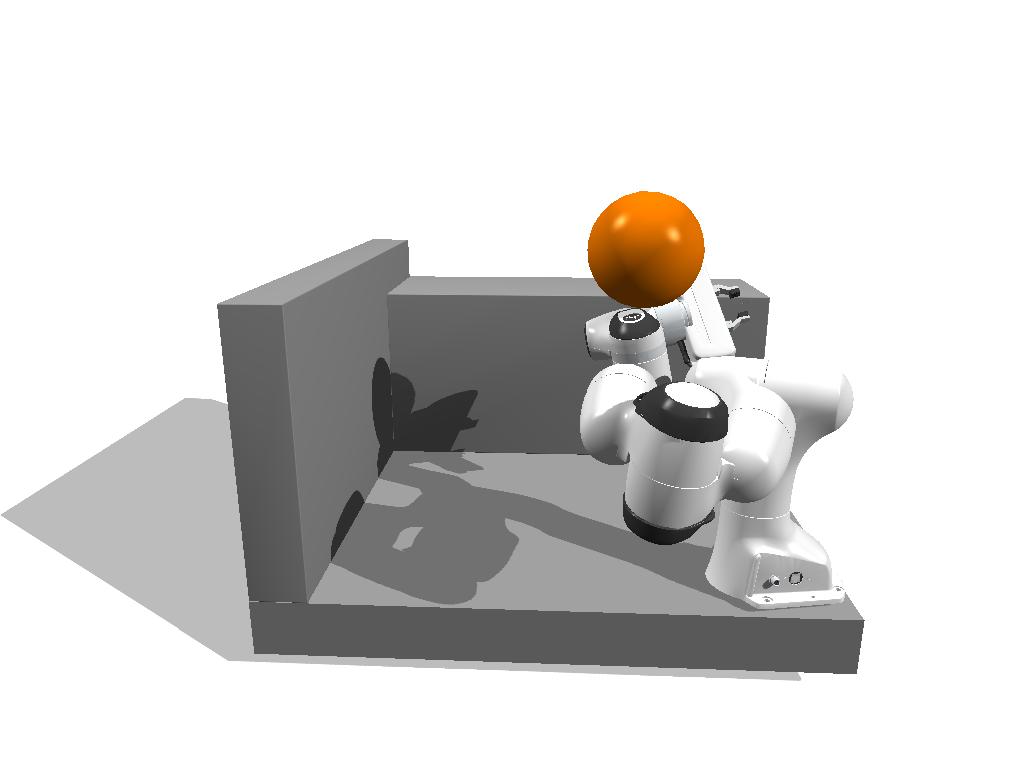}
\pic{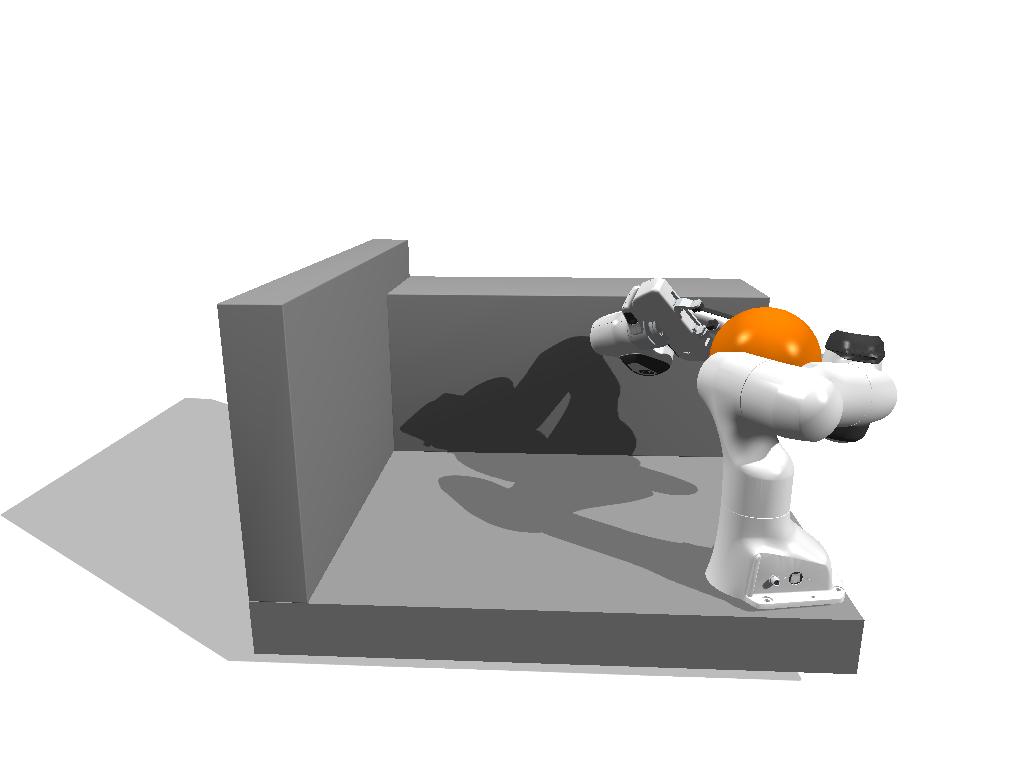}\pic{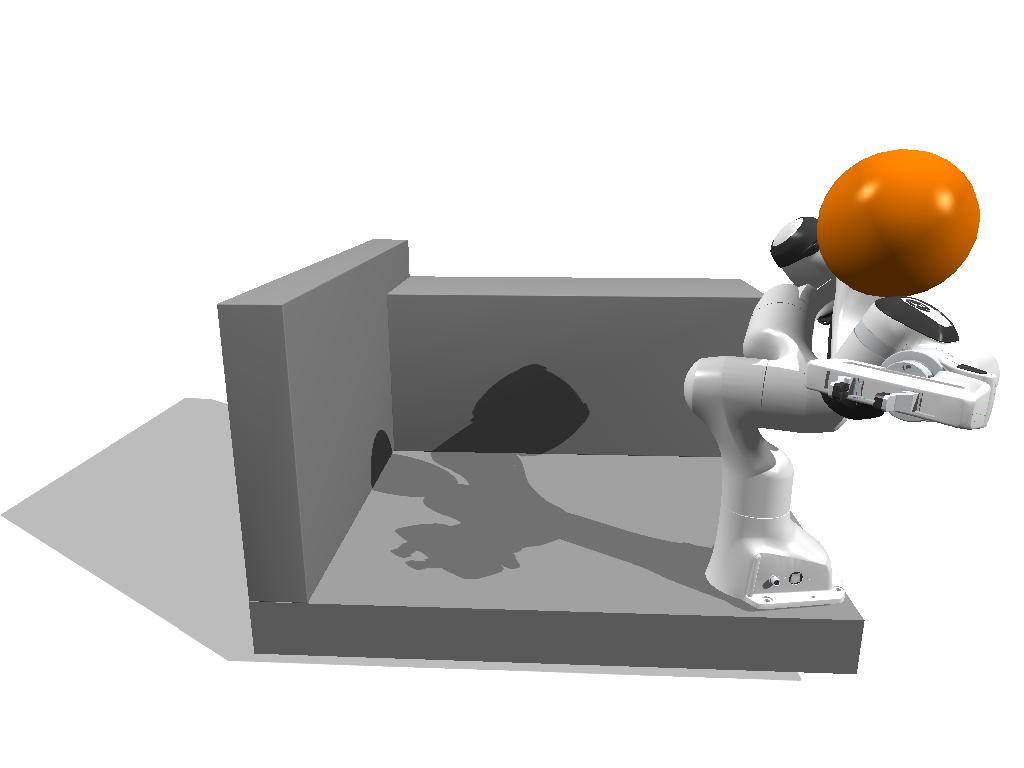}\pic{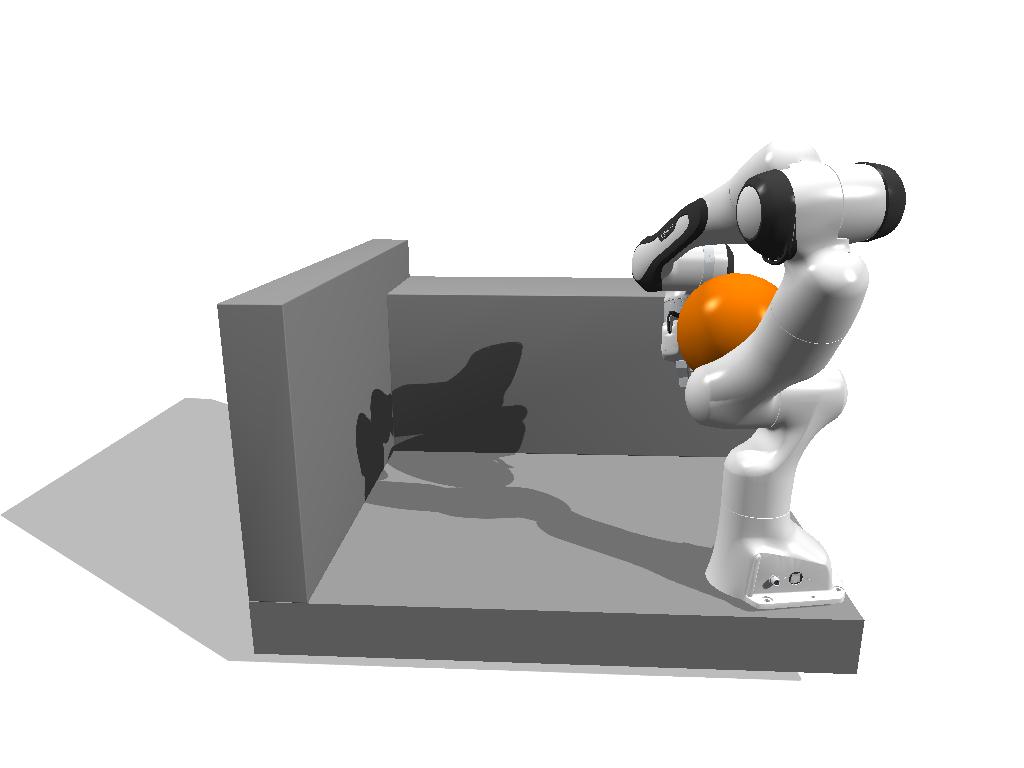}\pic{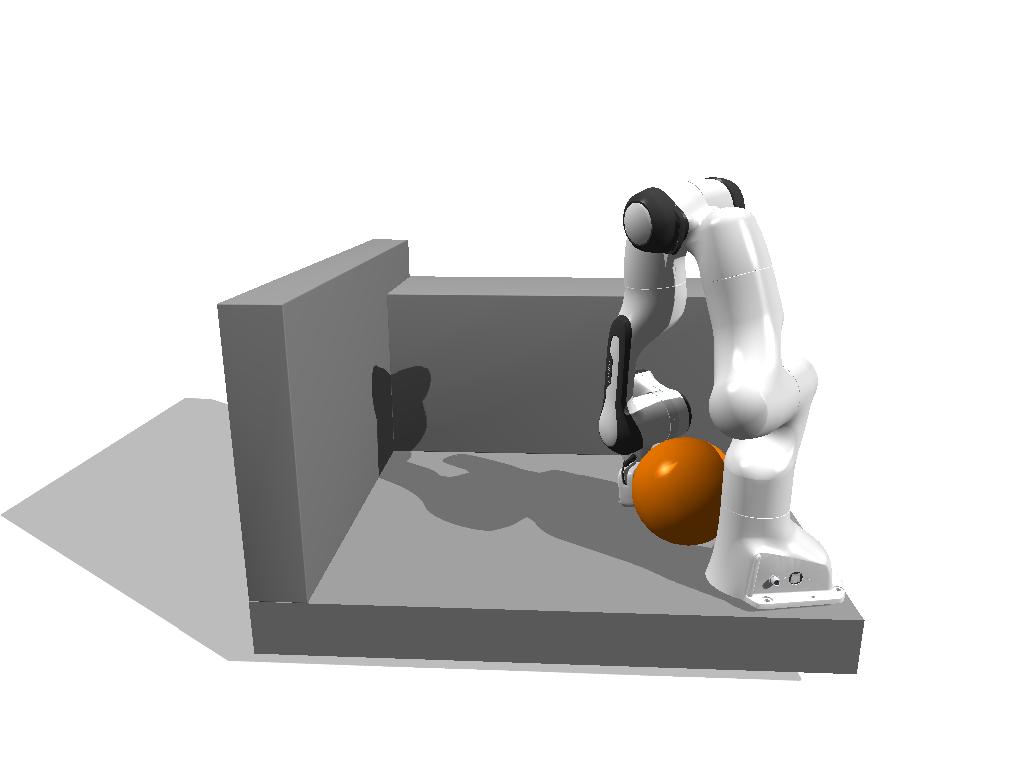}
\pic{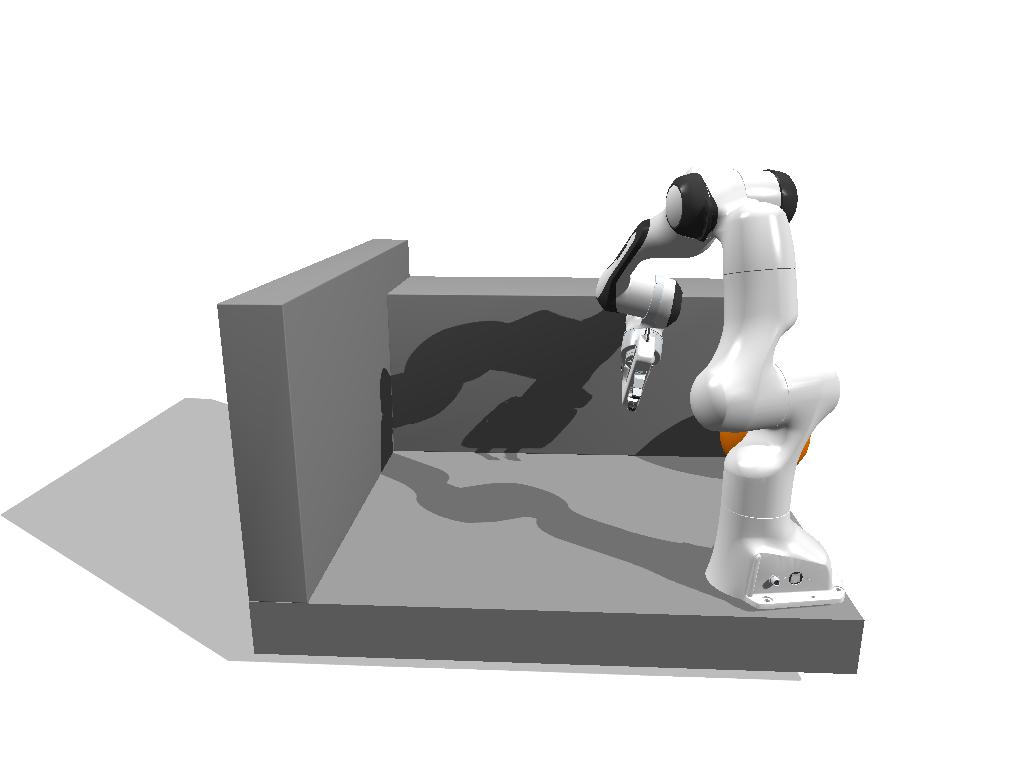}\pic{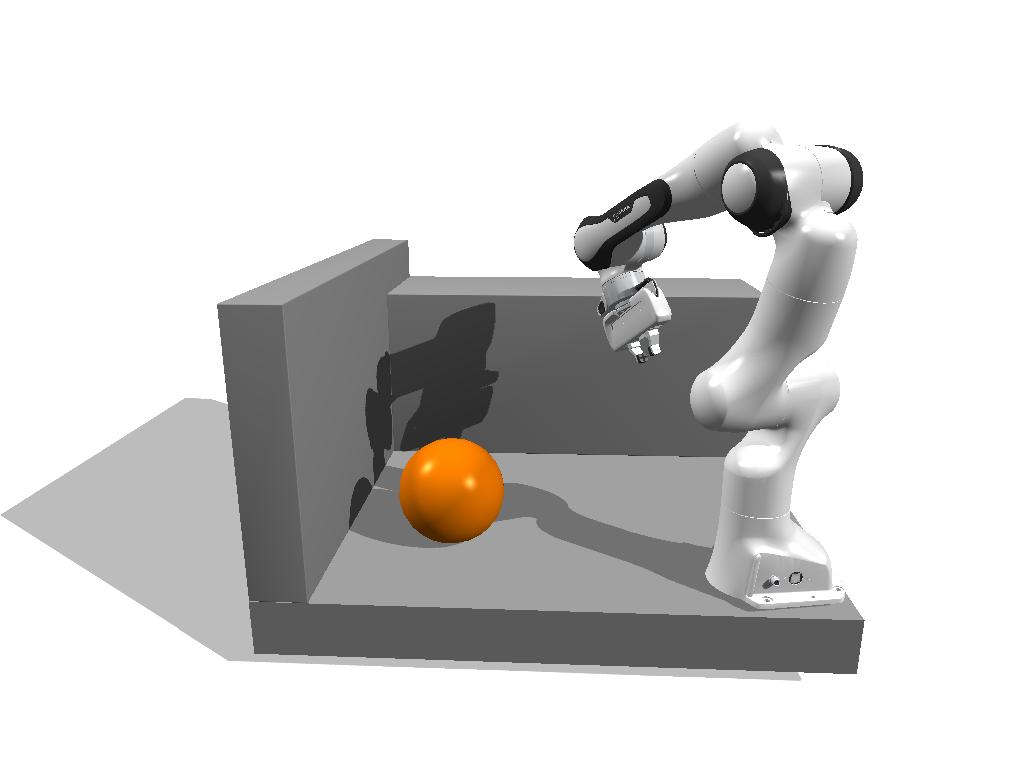}\pic{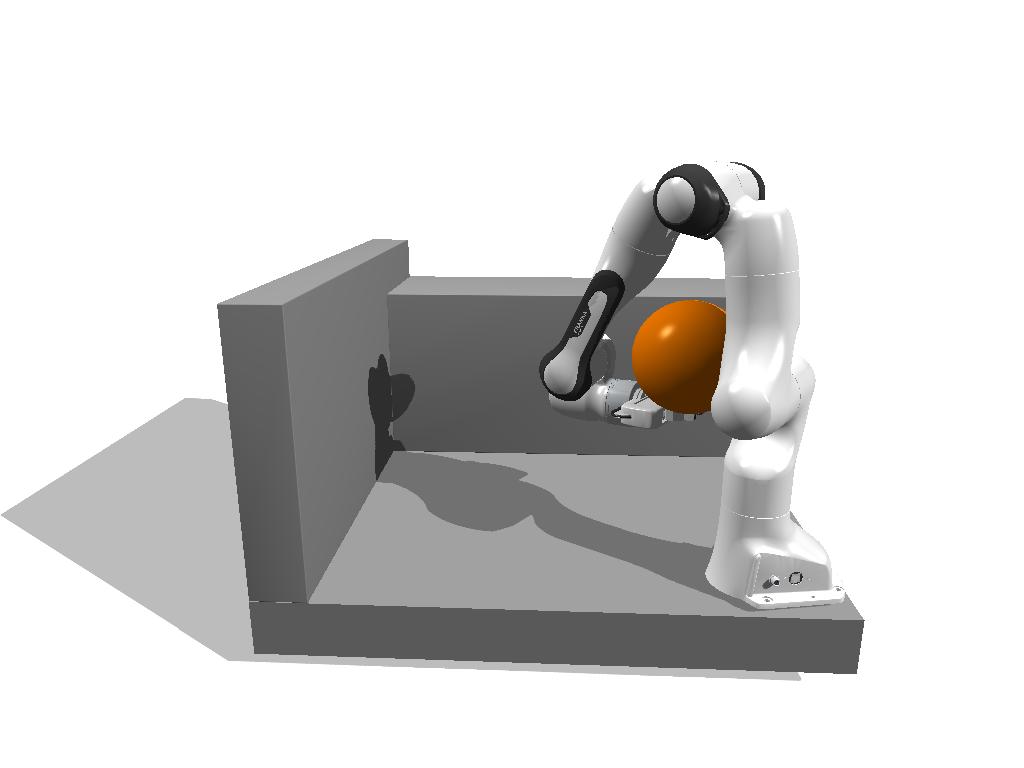}\pic{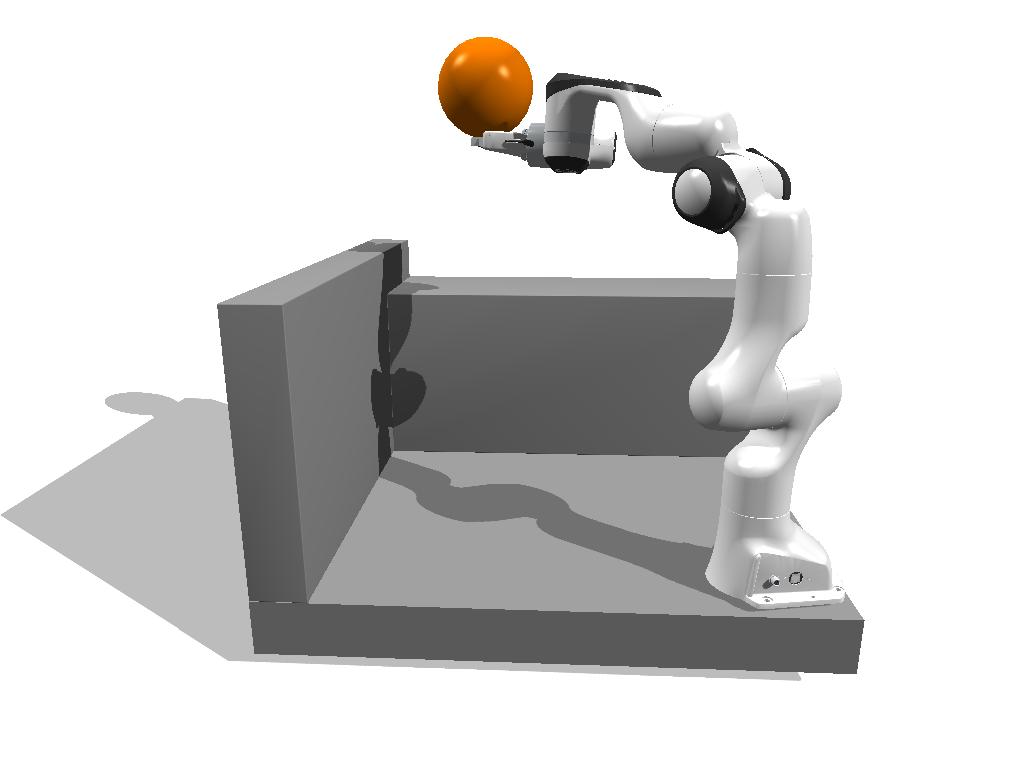}
\pic{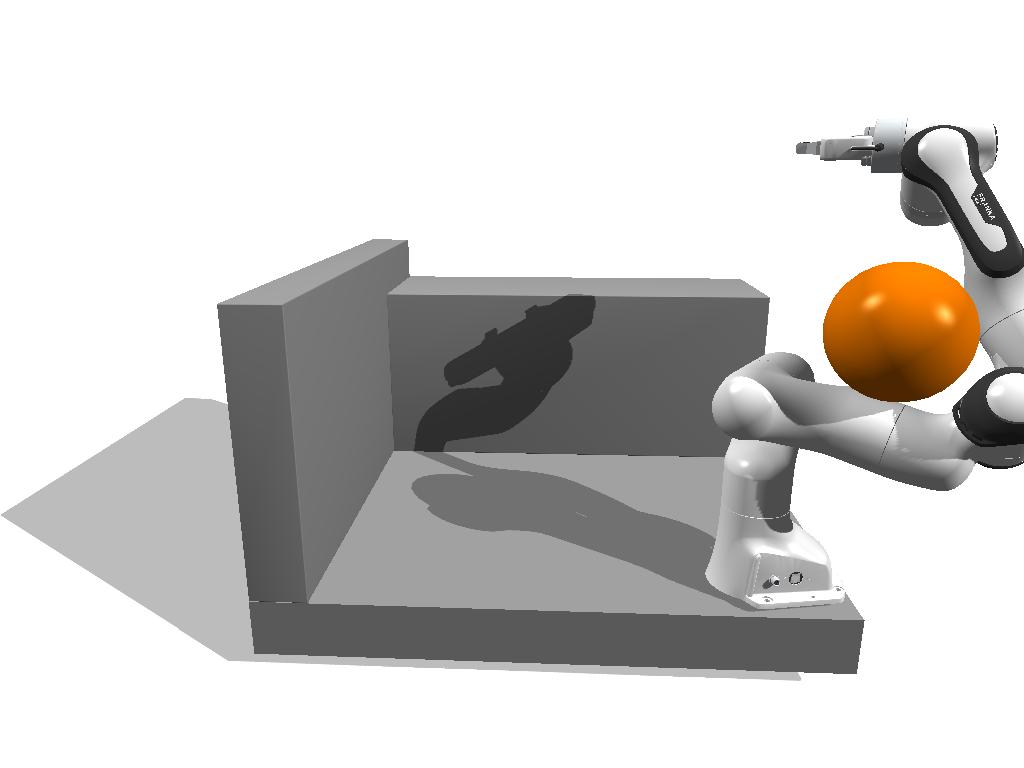}\pic{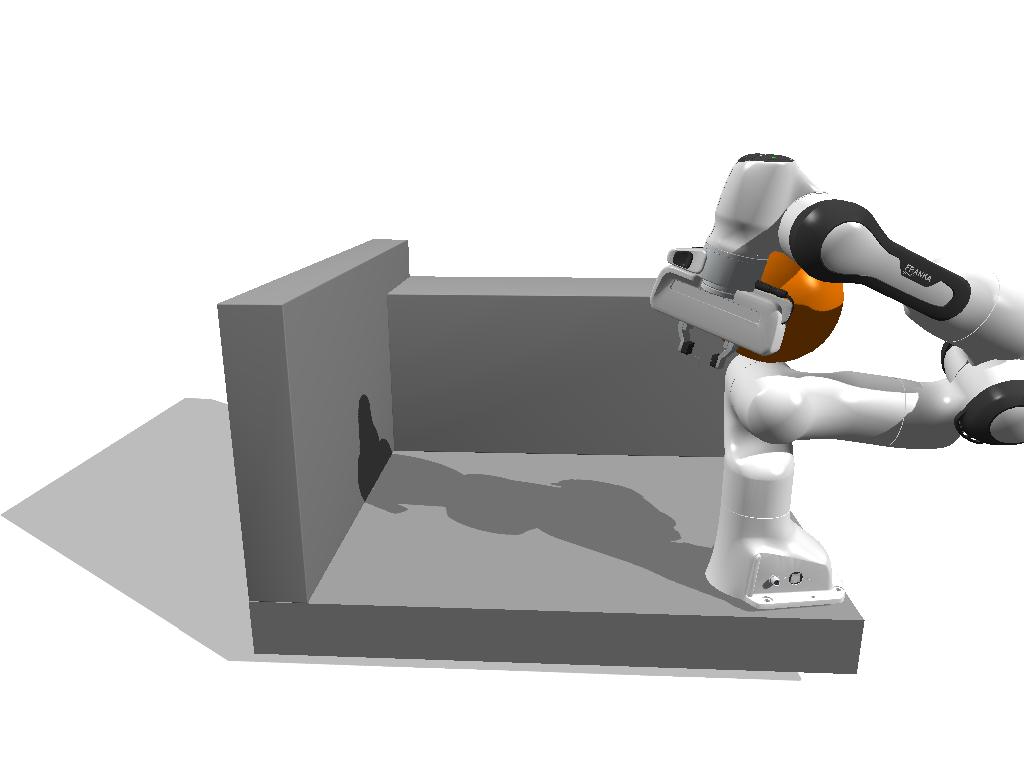}\pic{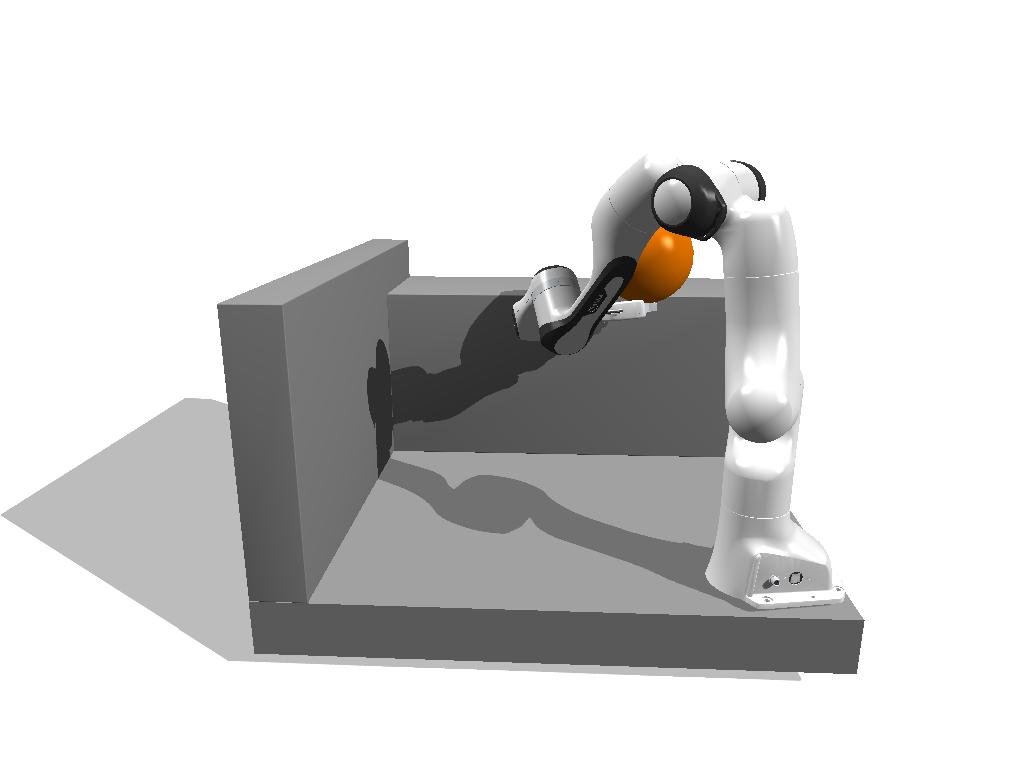}\pic{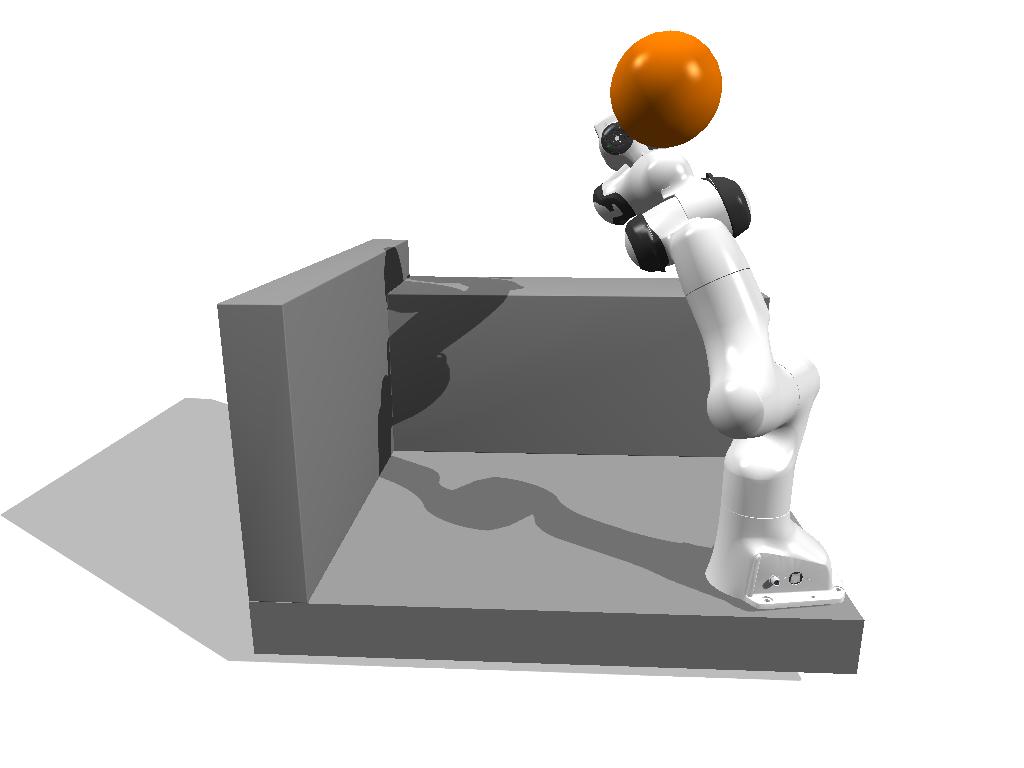}
\pic{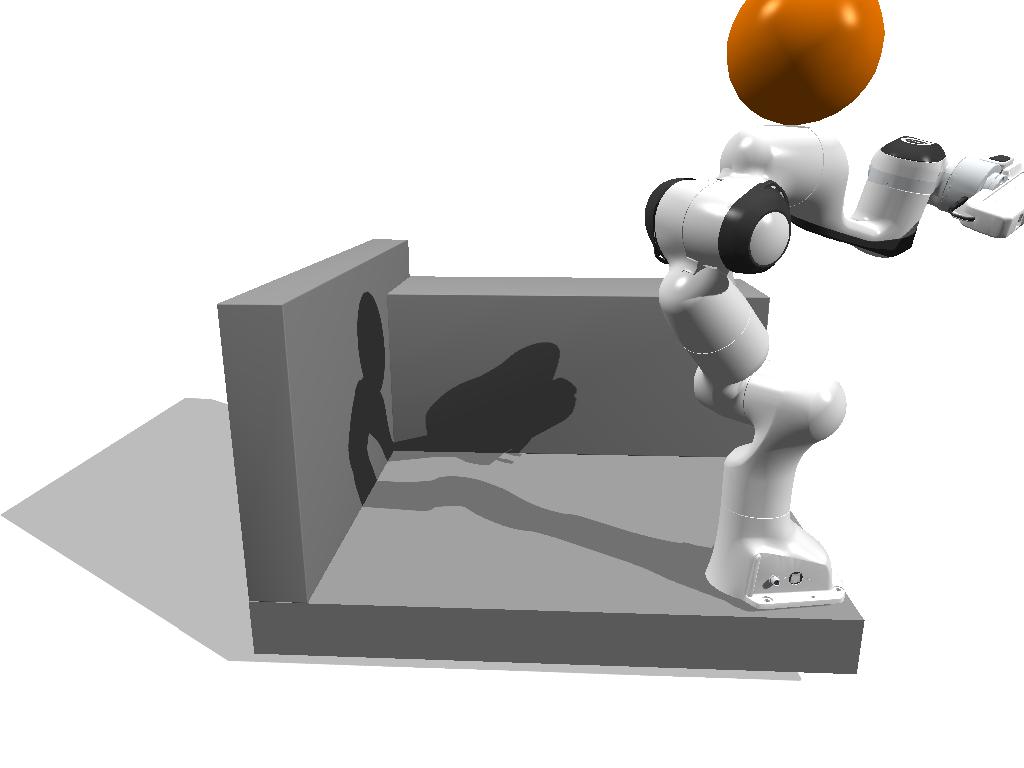}\pic{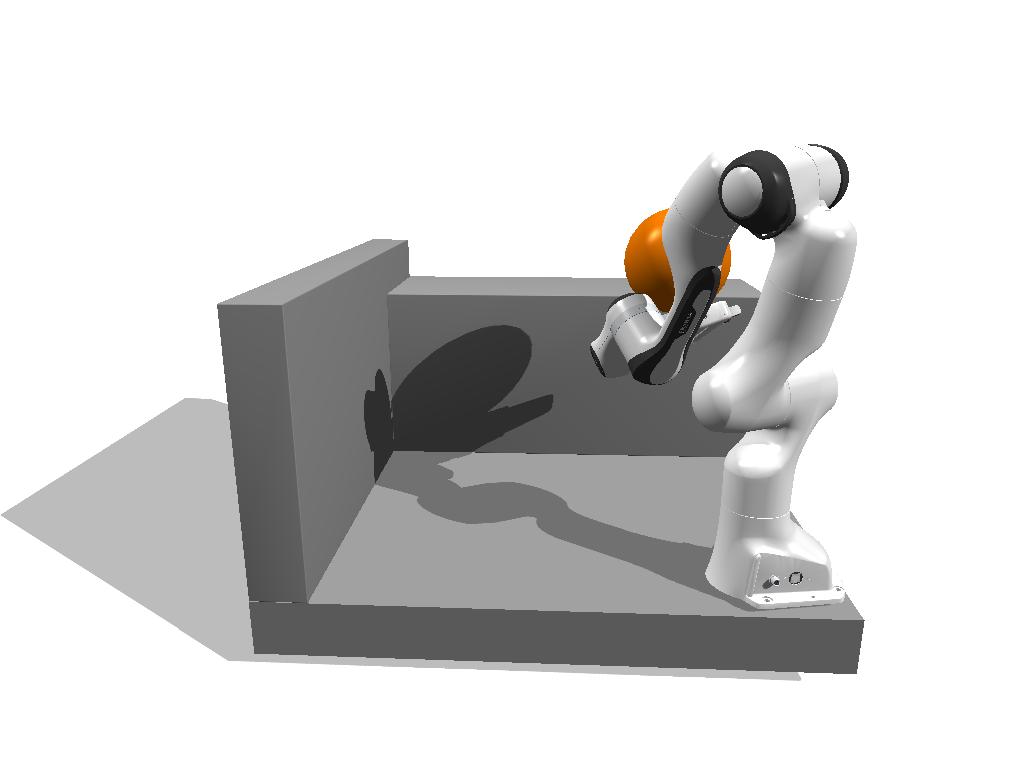}\pic{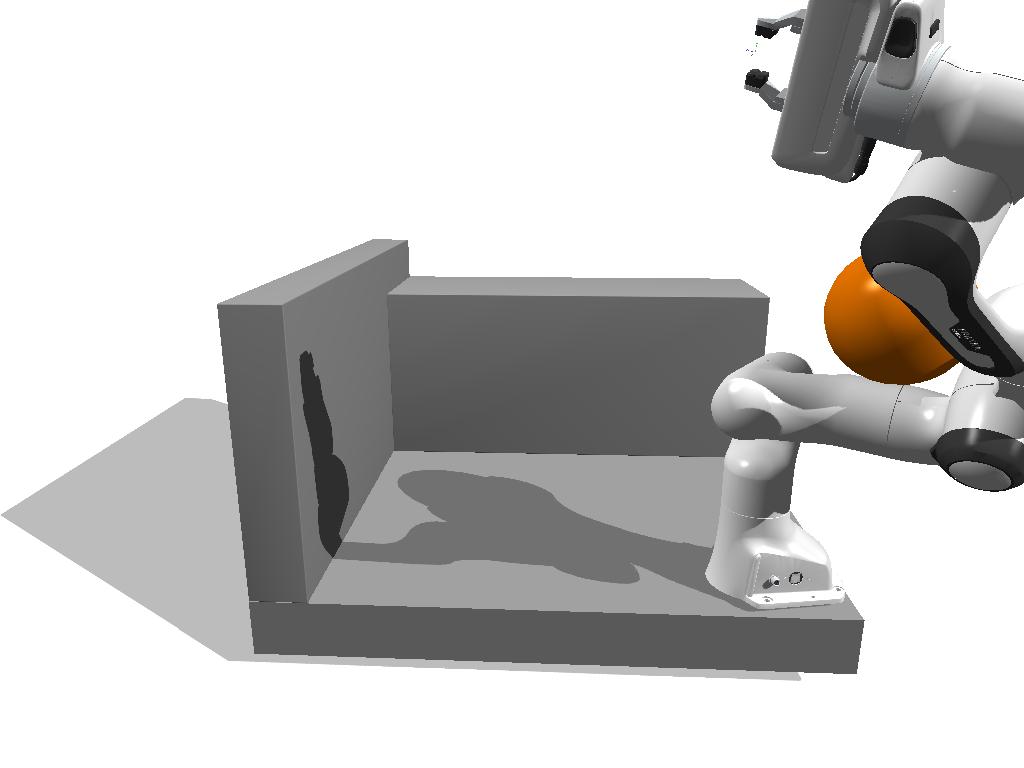}\pic{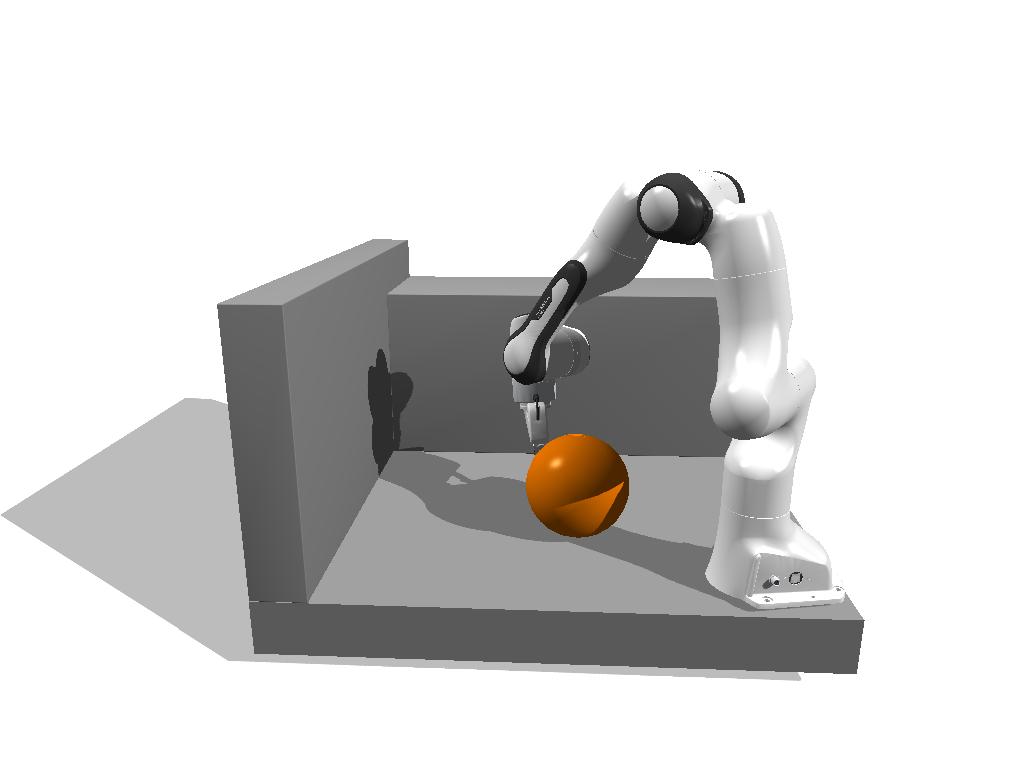}
\caption{\label{fig:panda_sphere}
  Constrained samples from the panda sphere domain.}
\end{figure}

\begin{figure}[p]\centering
\newcommand{\pic}[1]{\includegraphics[width=.25\columnwidth,trim={0 0 0 25pt},clip]{data/configs/images/sphere_cube_sample_0000#1.jpg}}
\pic{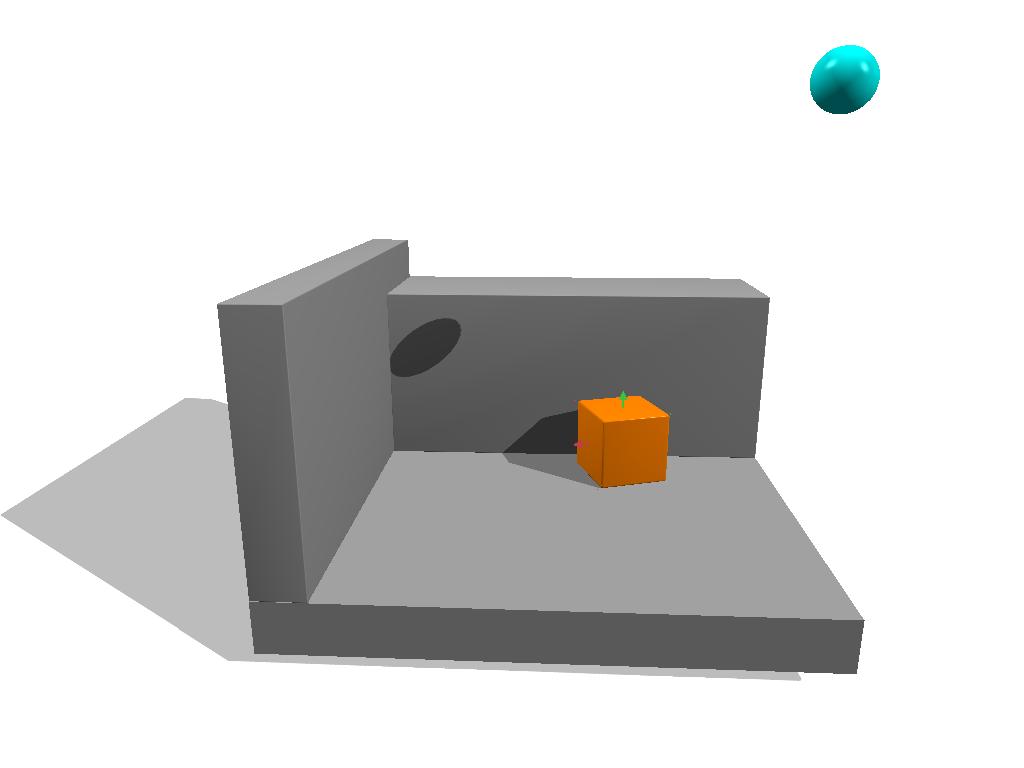}\pic{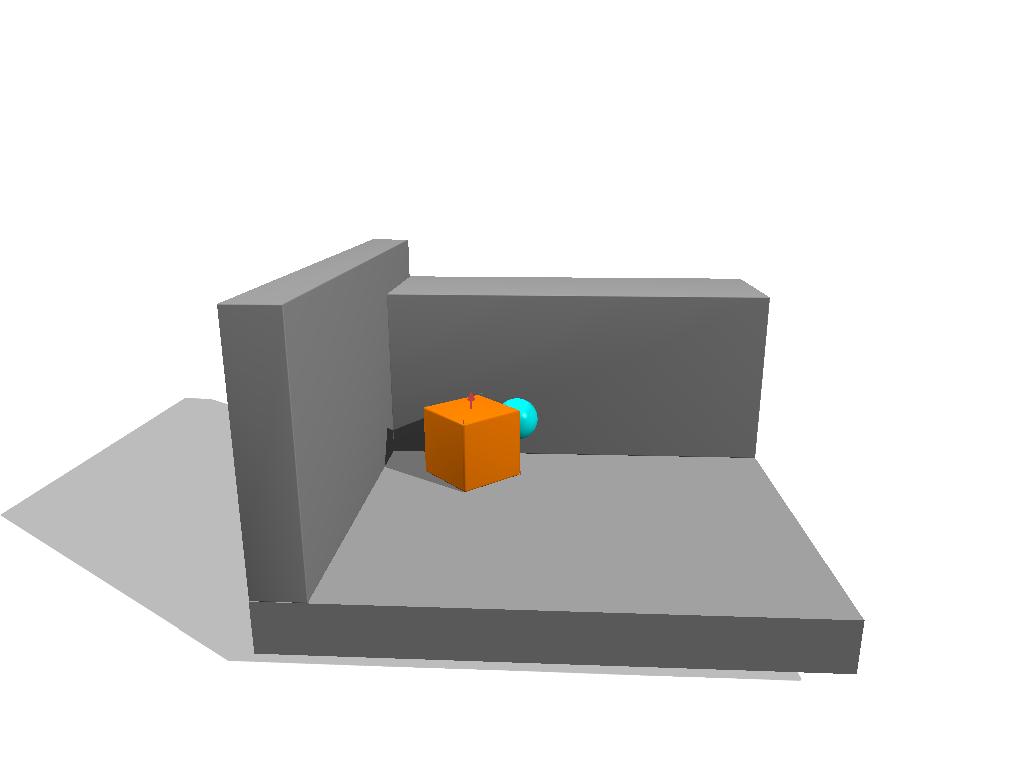}\pic{02}\pic{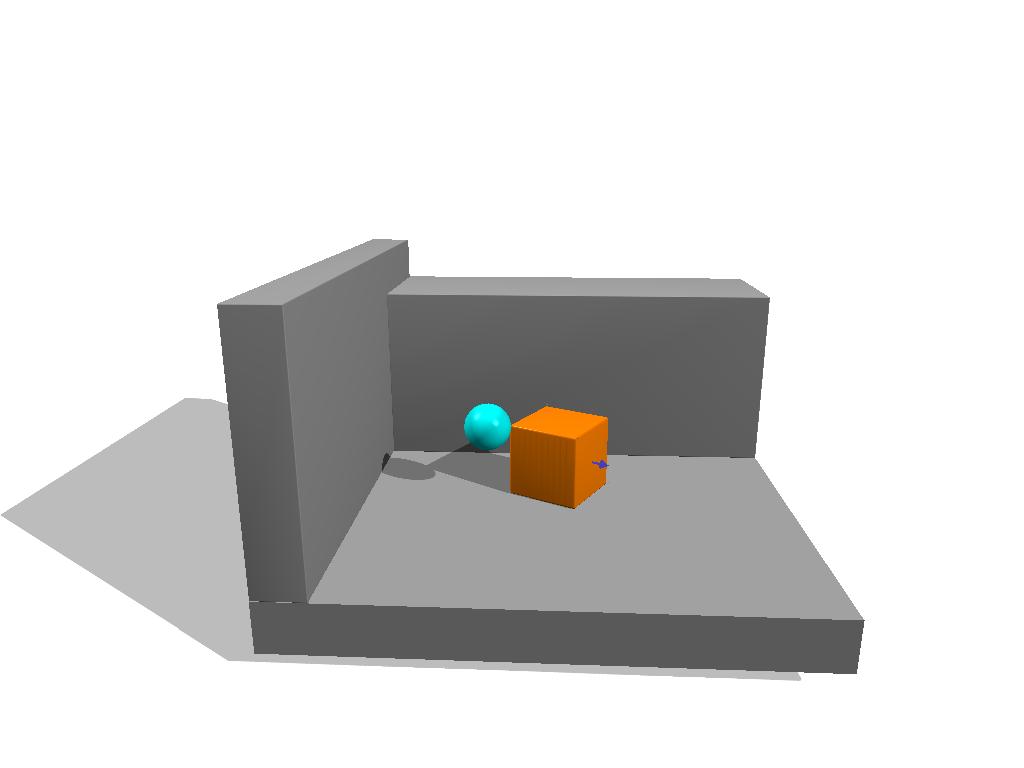}
\pic{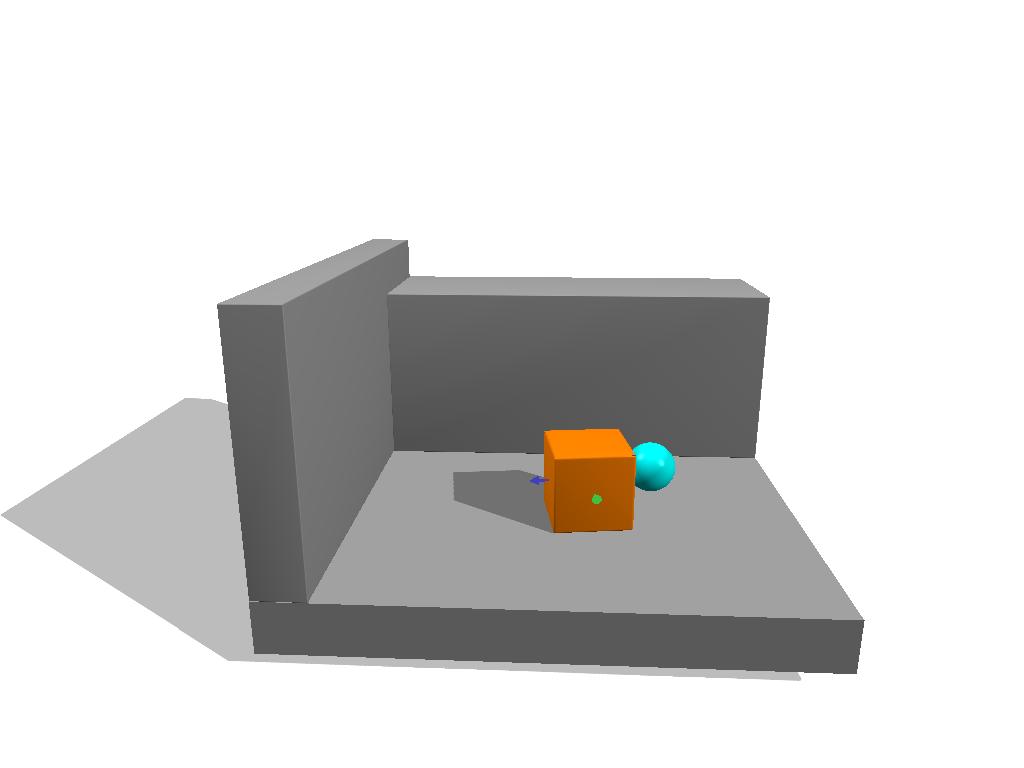}\pic{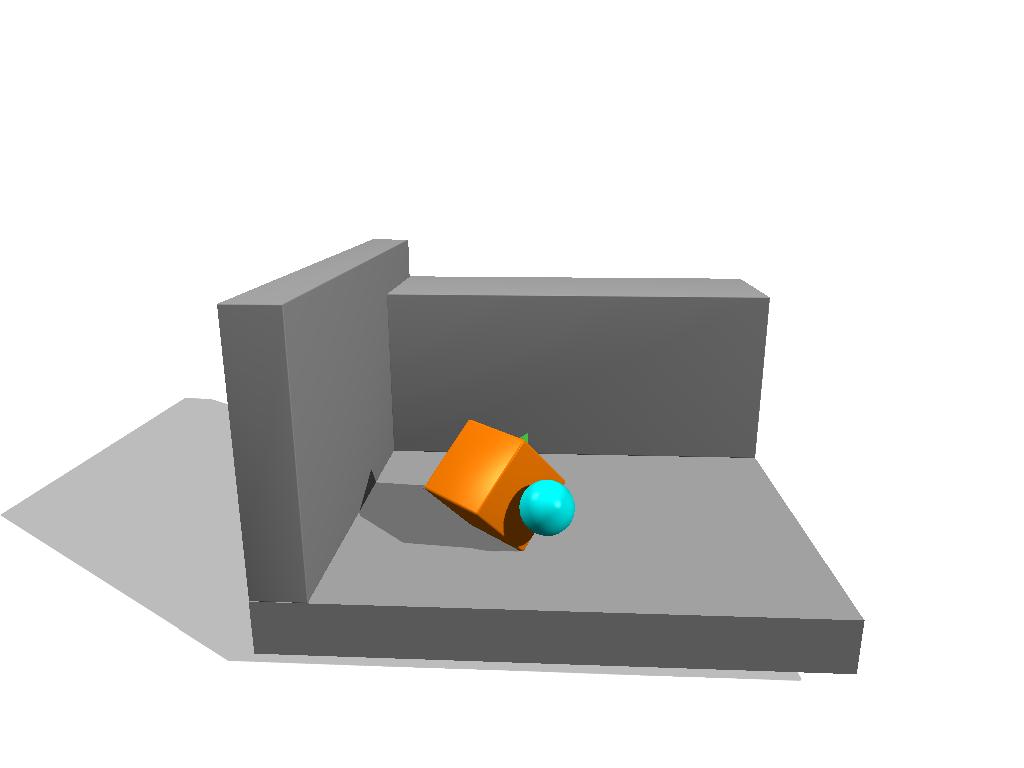}\pic{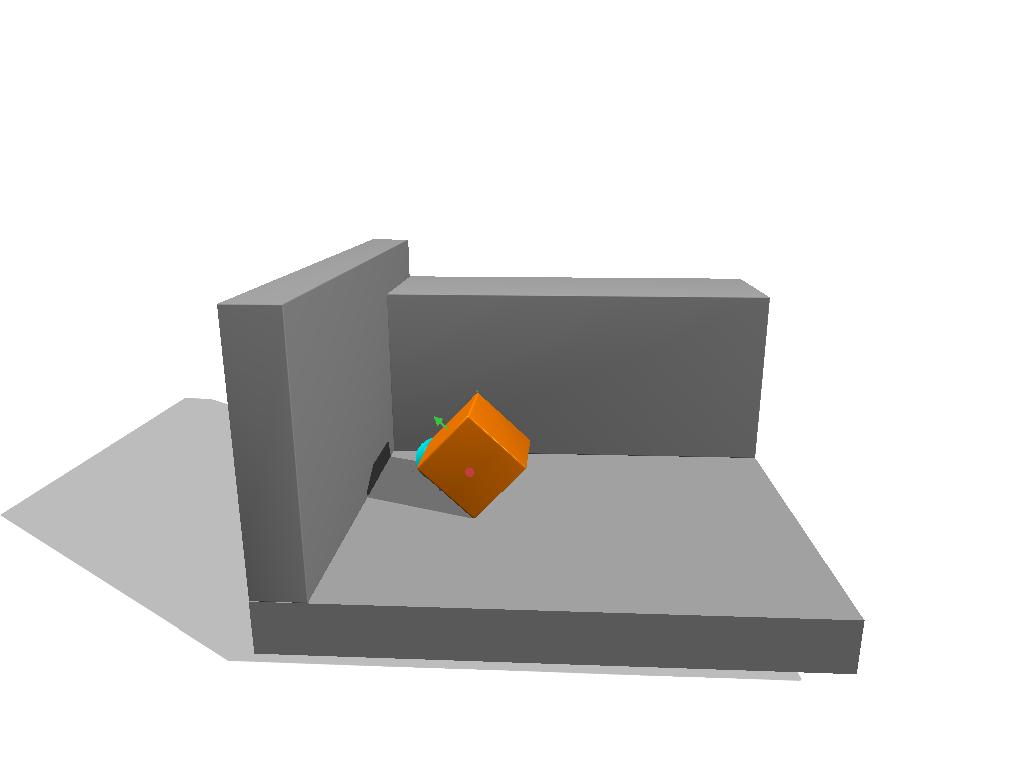}\pic{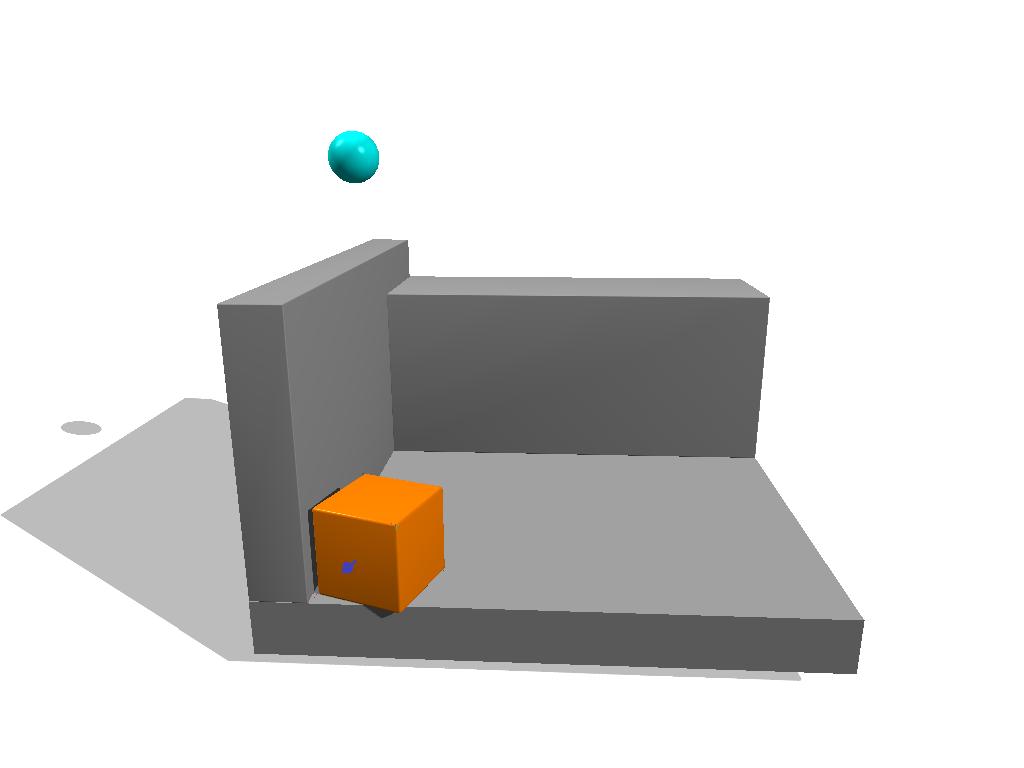}
\pic{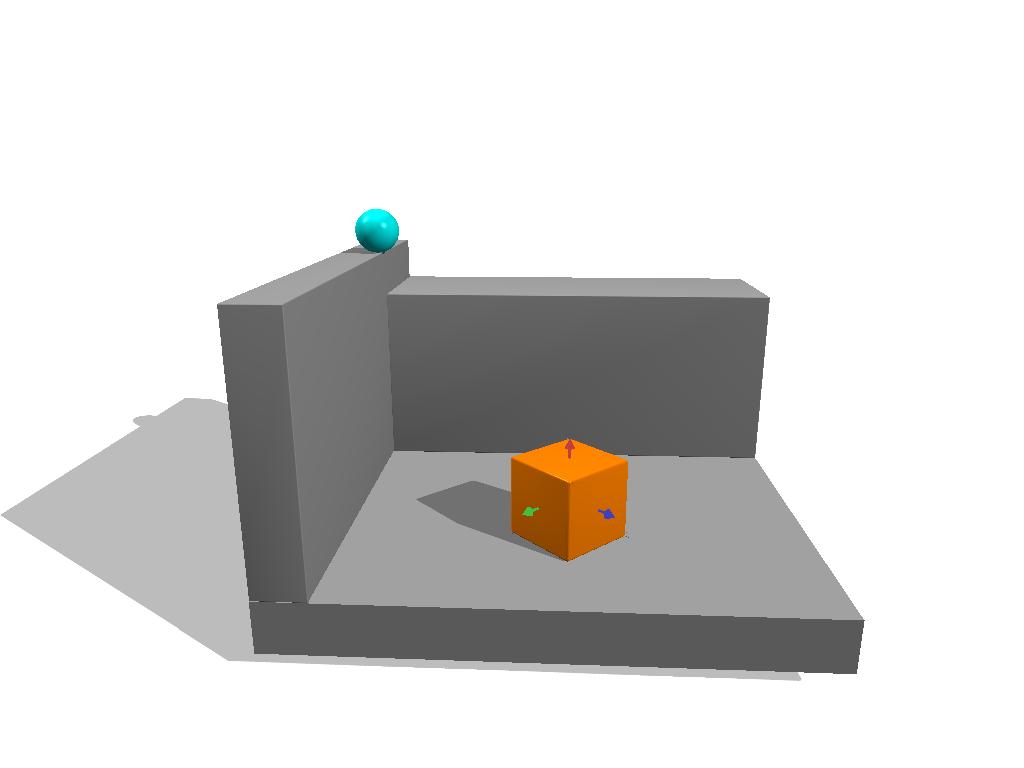}\pic{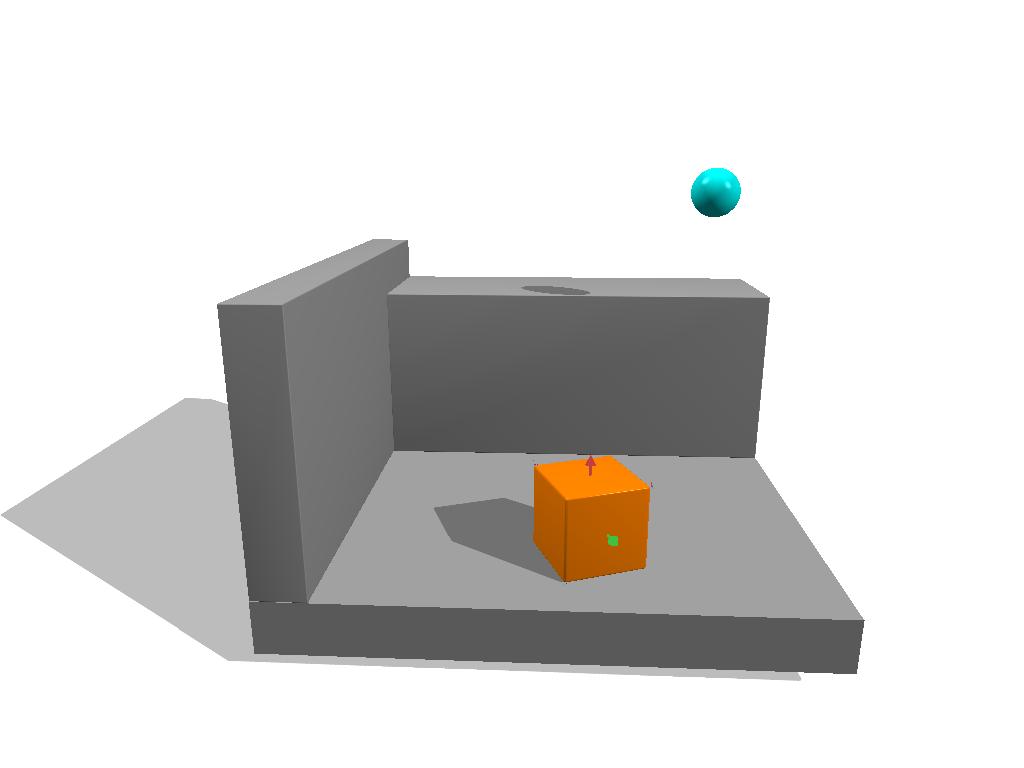}\pic{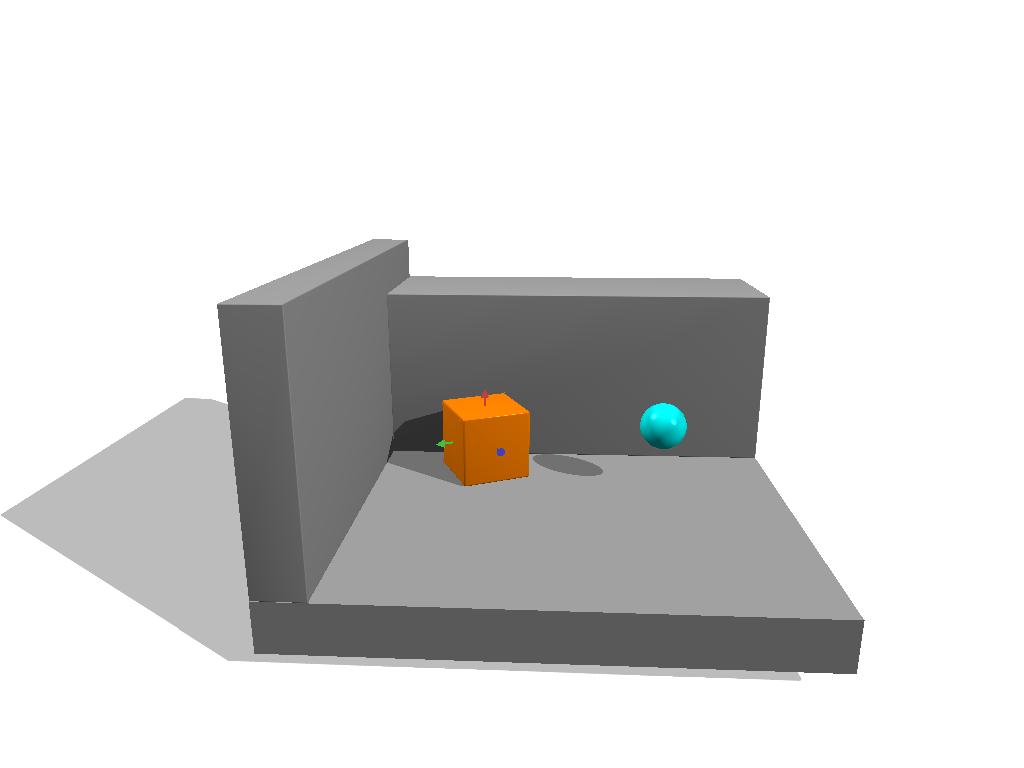}\pic{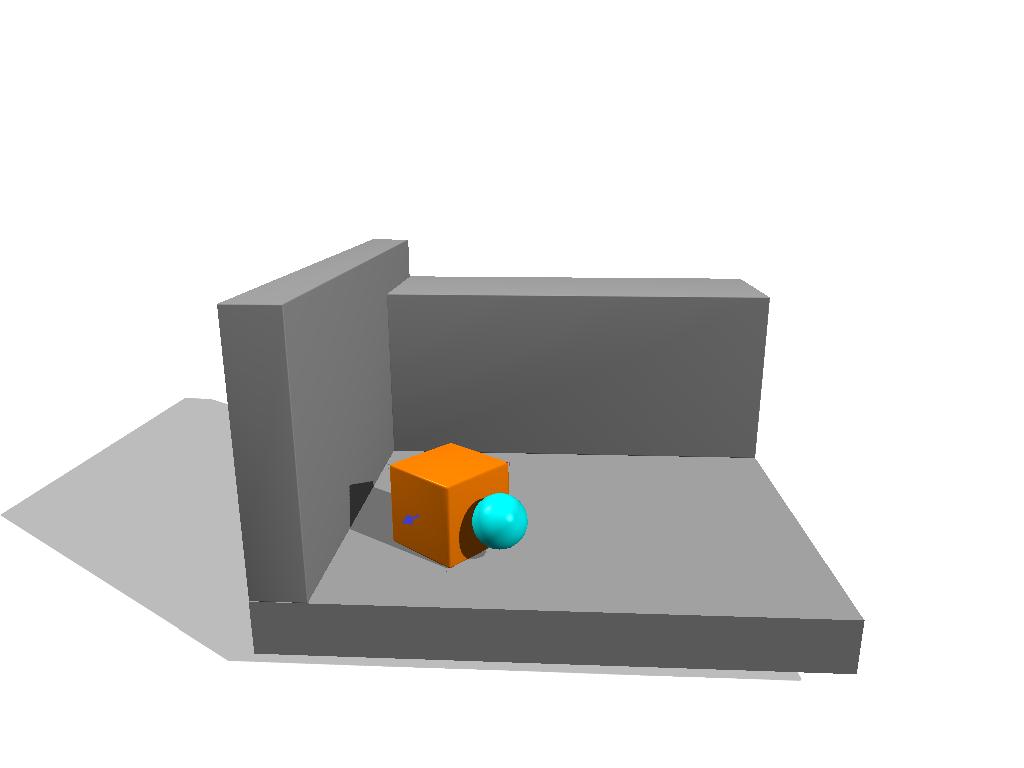}
\pic{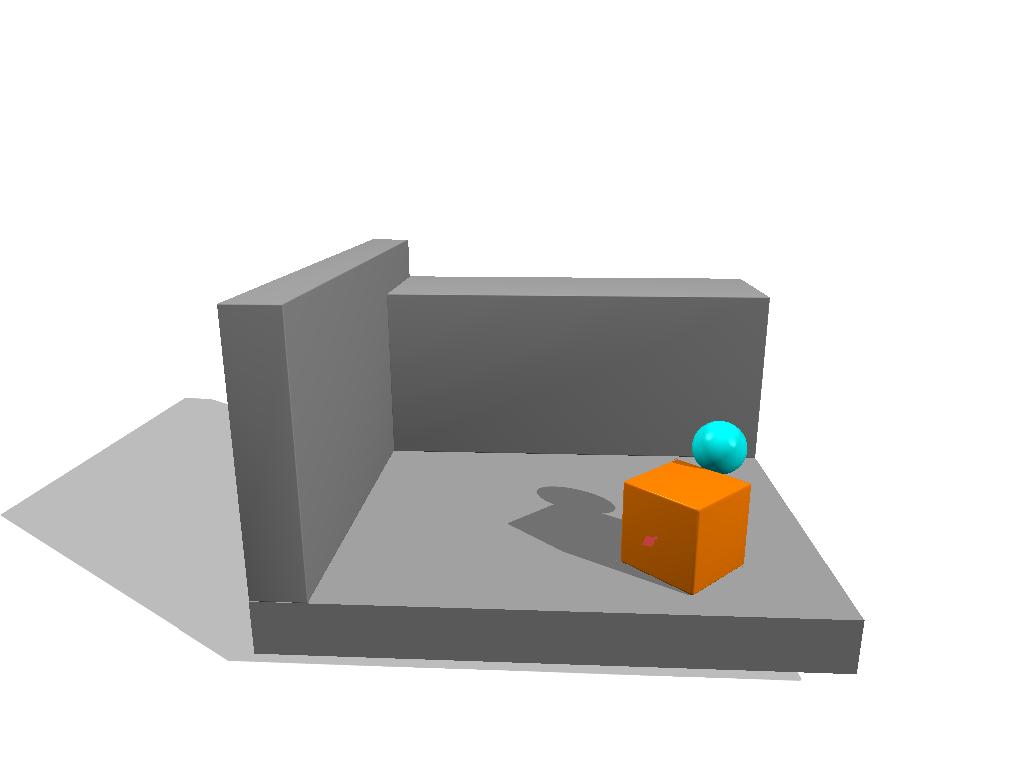}\pic{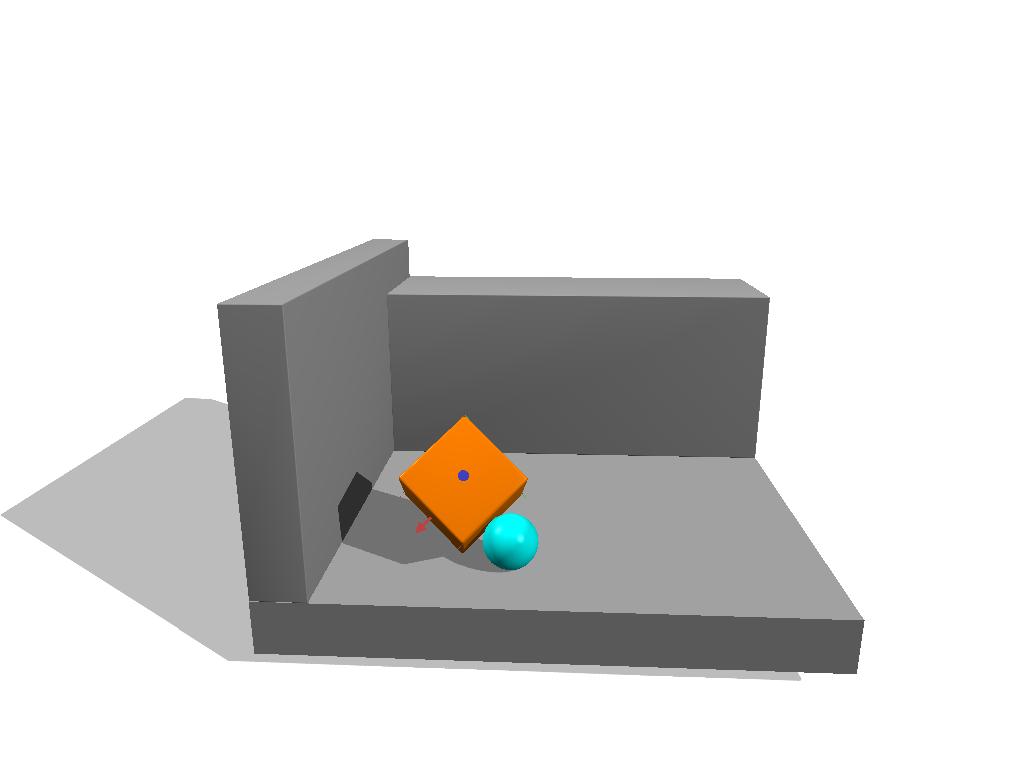}\pic{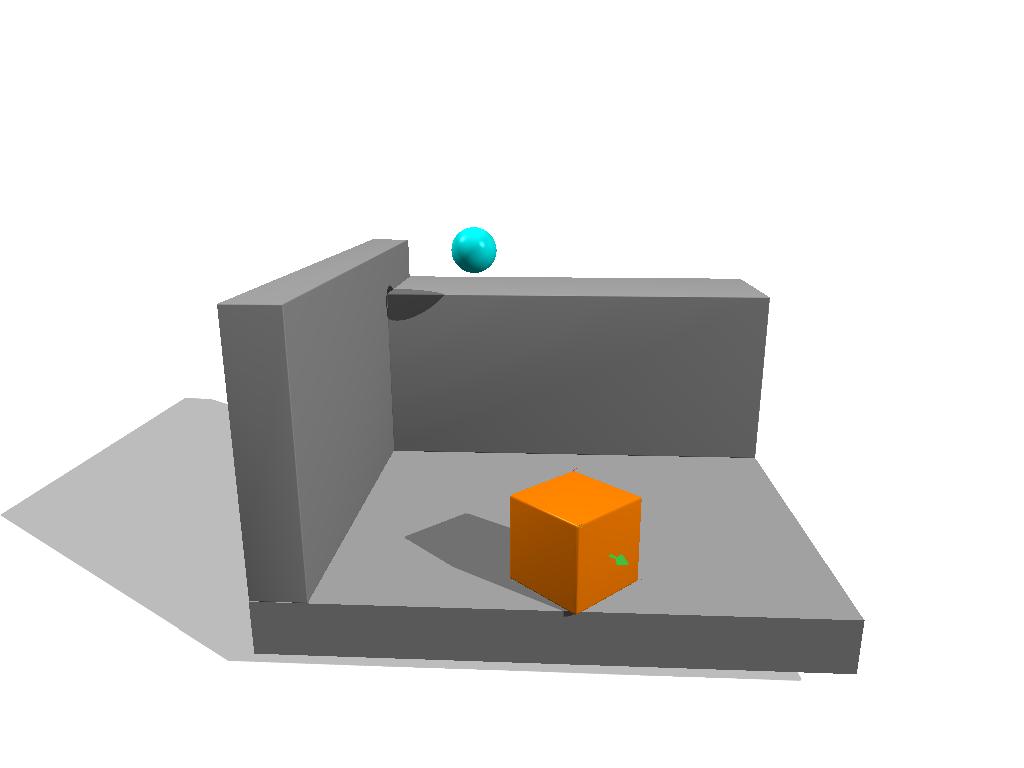}\pic{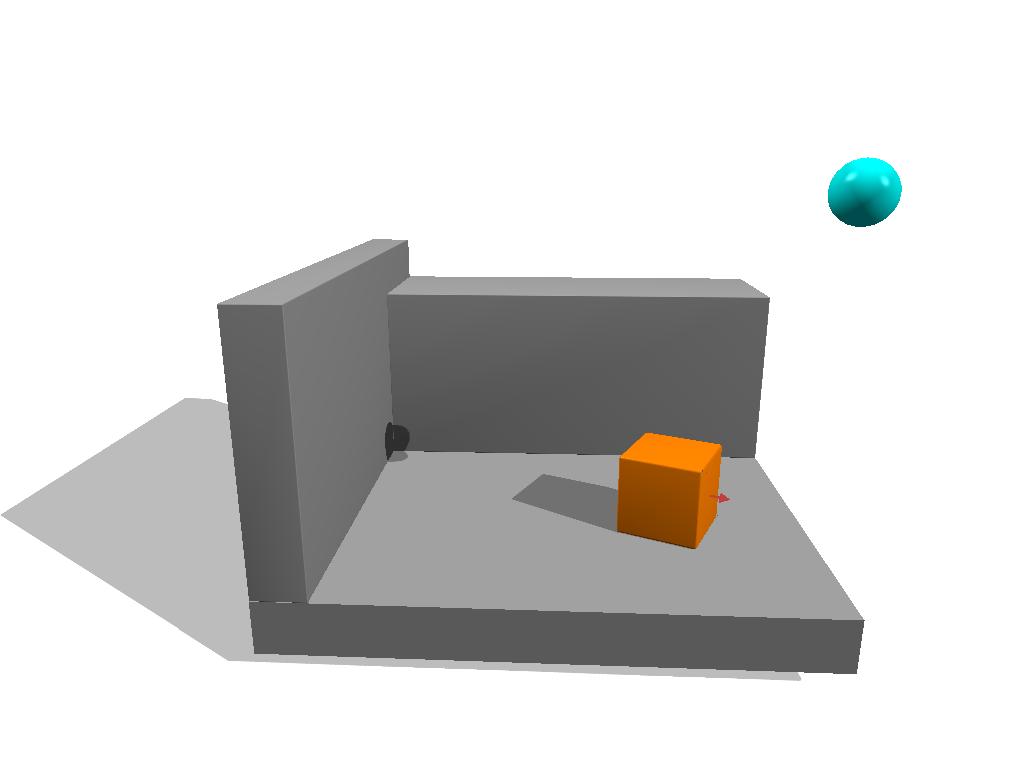}
\pic{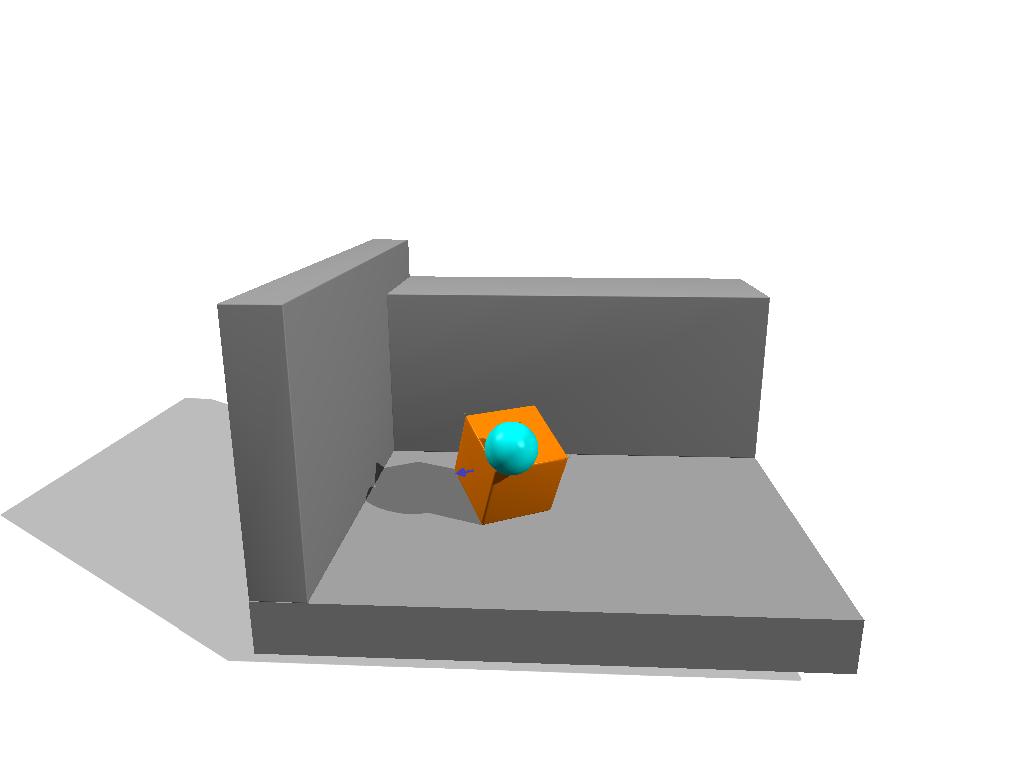}\pic{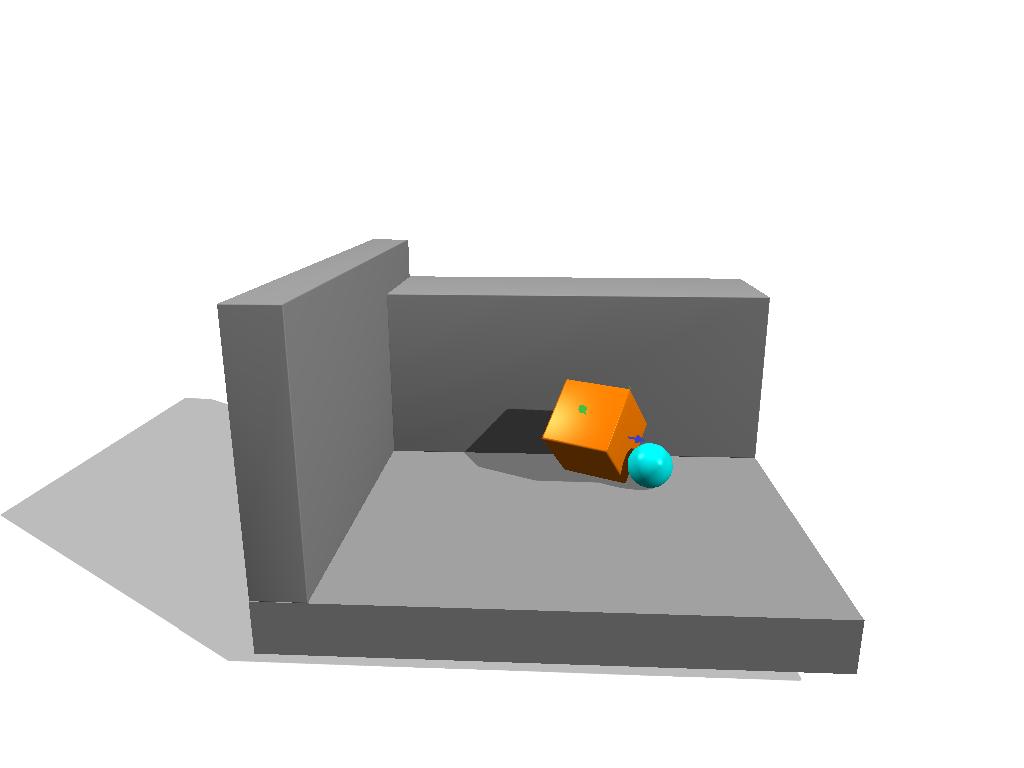}\pic{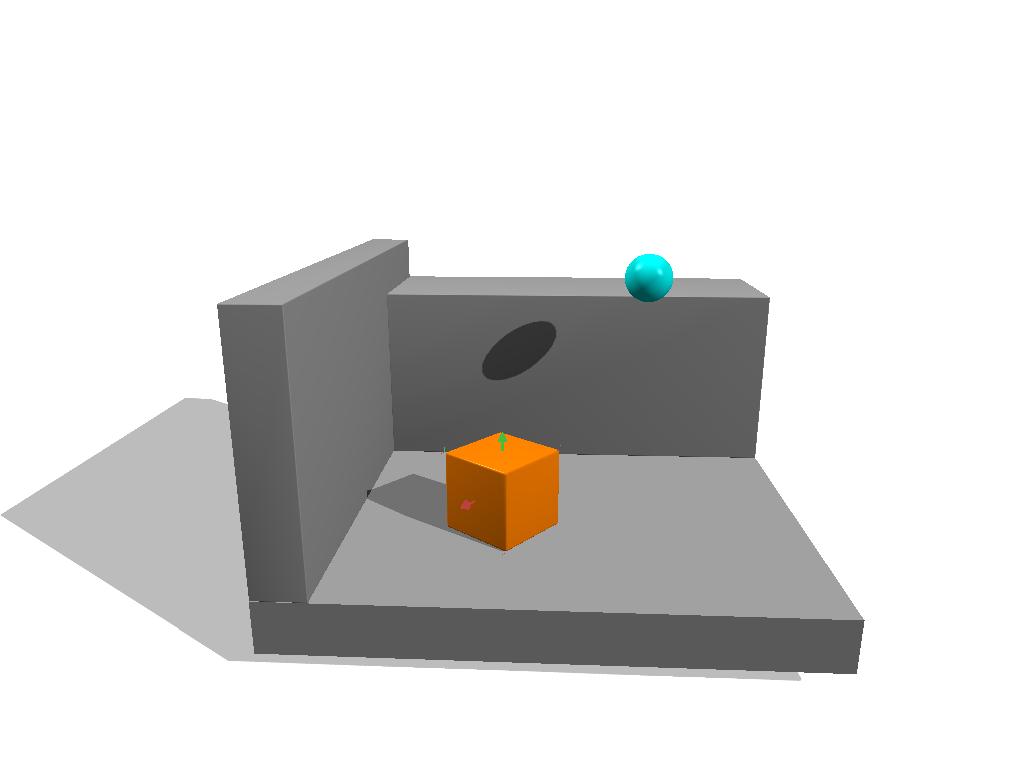}\pic{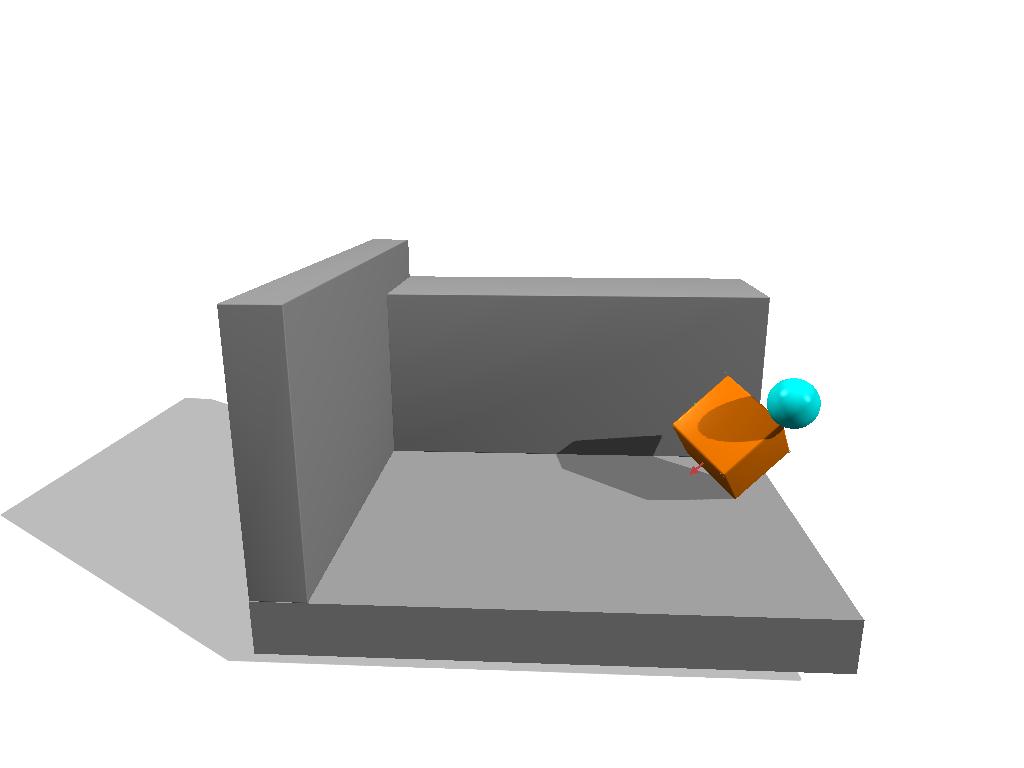}
\pic{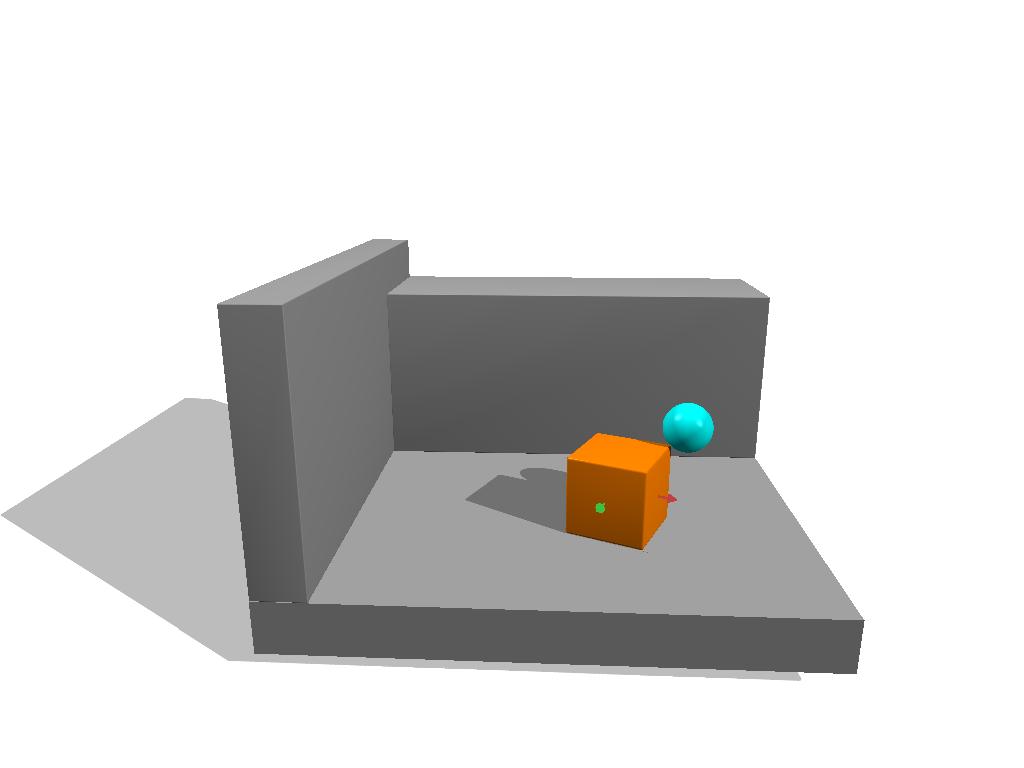}\pic{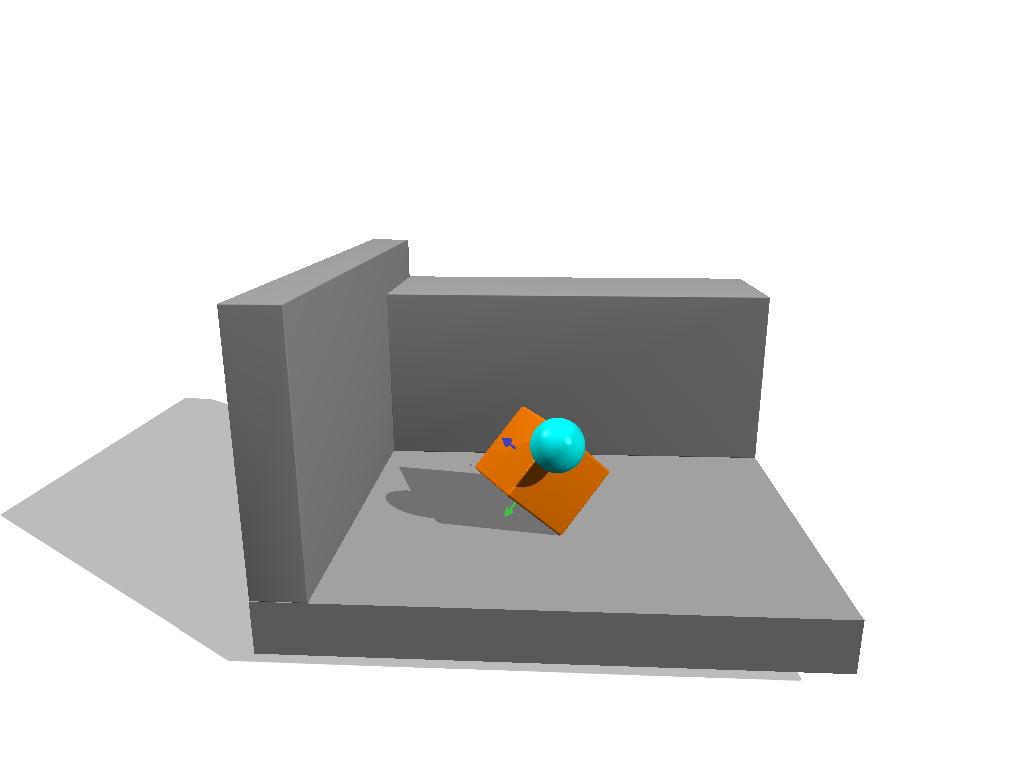}\pic{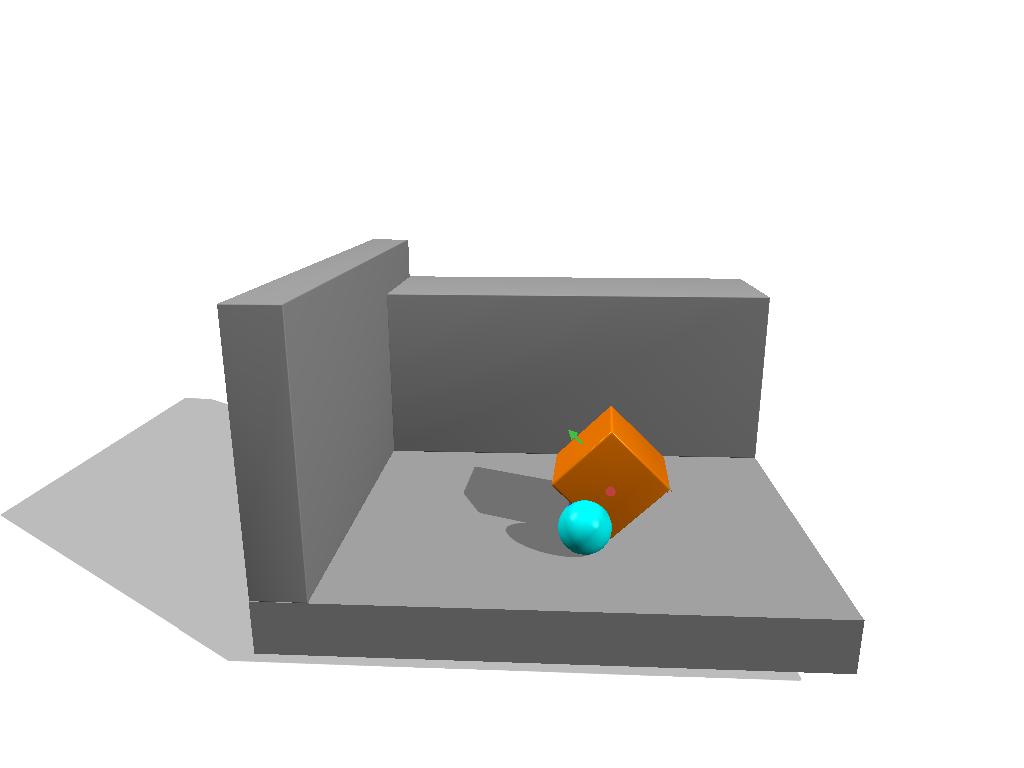}\pic{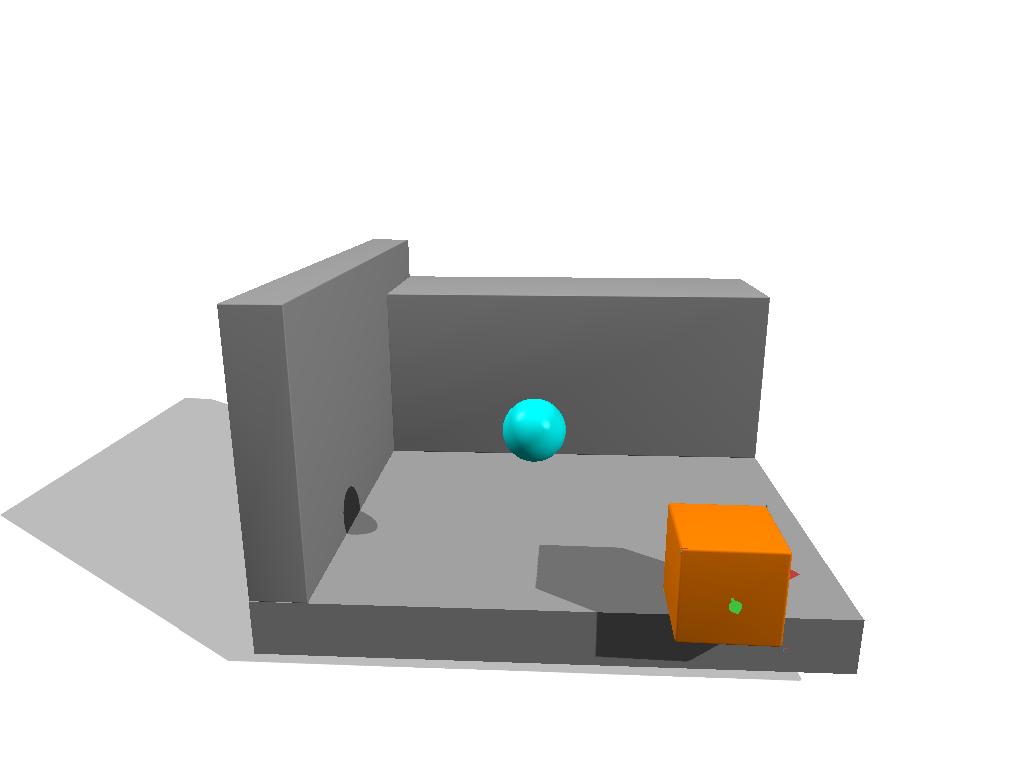}
\pic{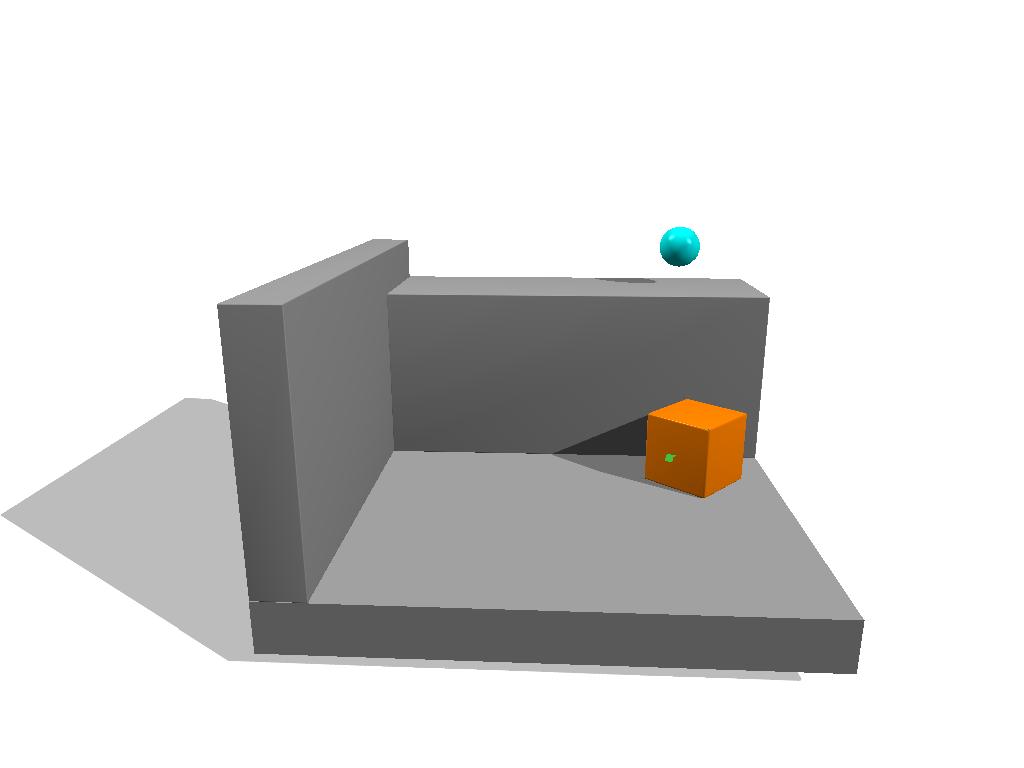}\pic{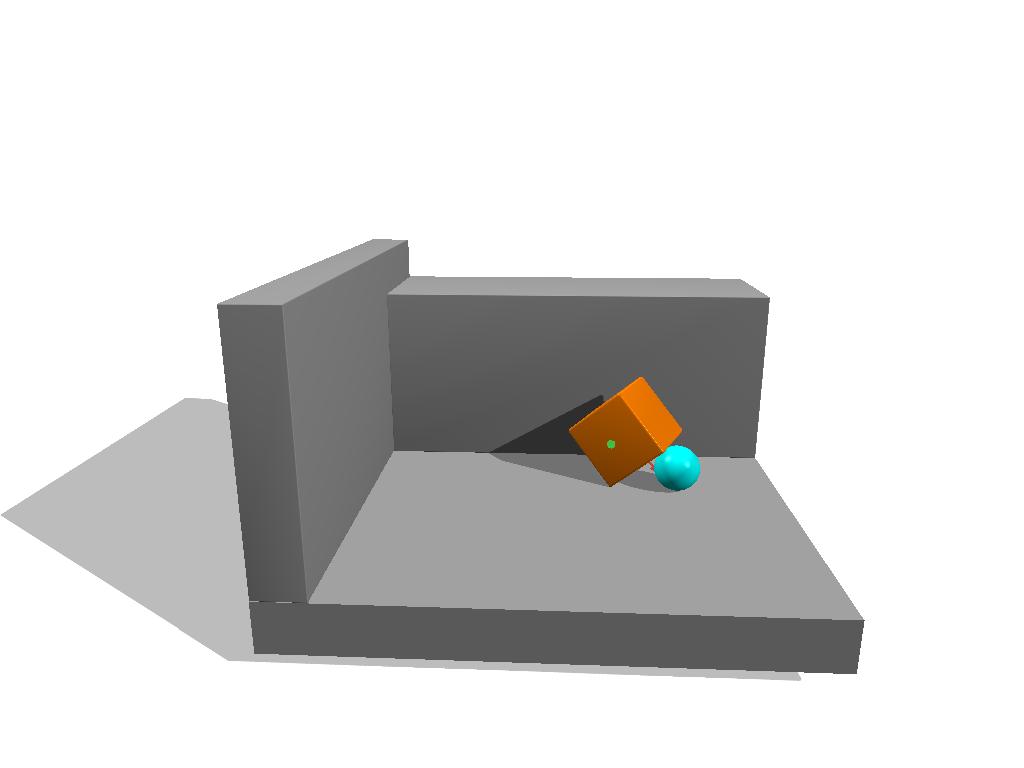}\pic{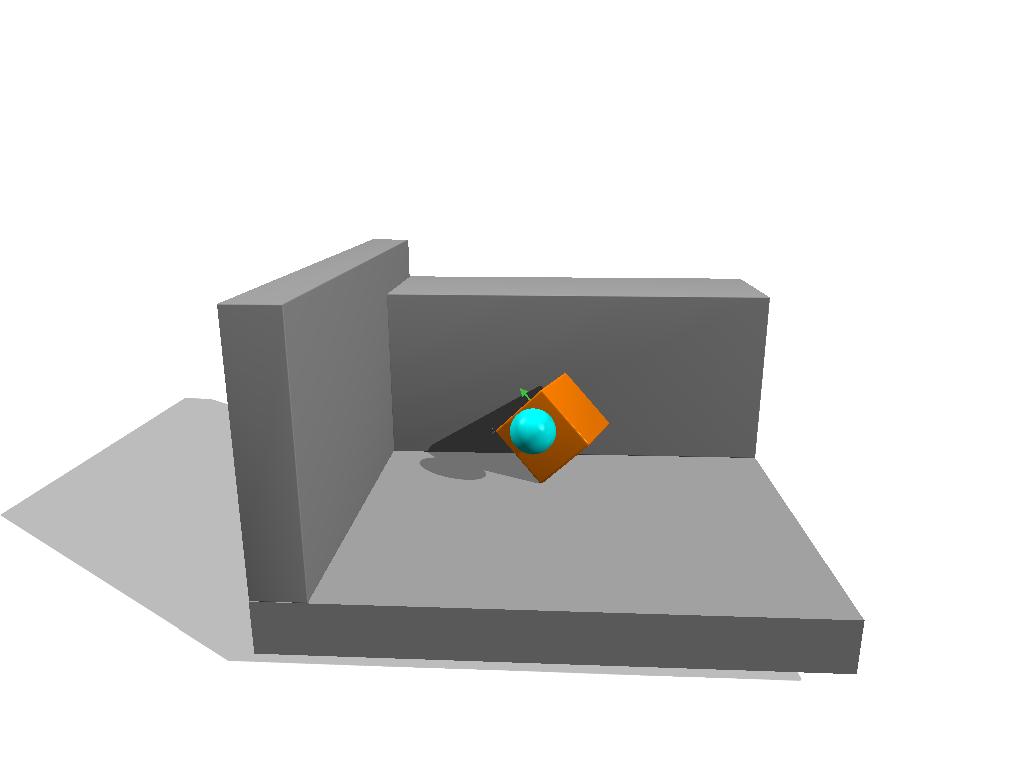}\pic{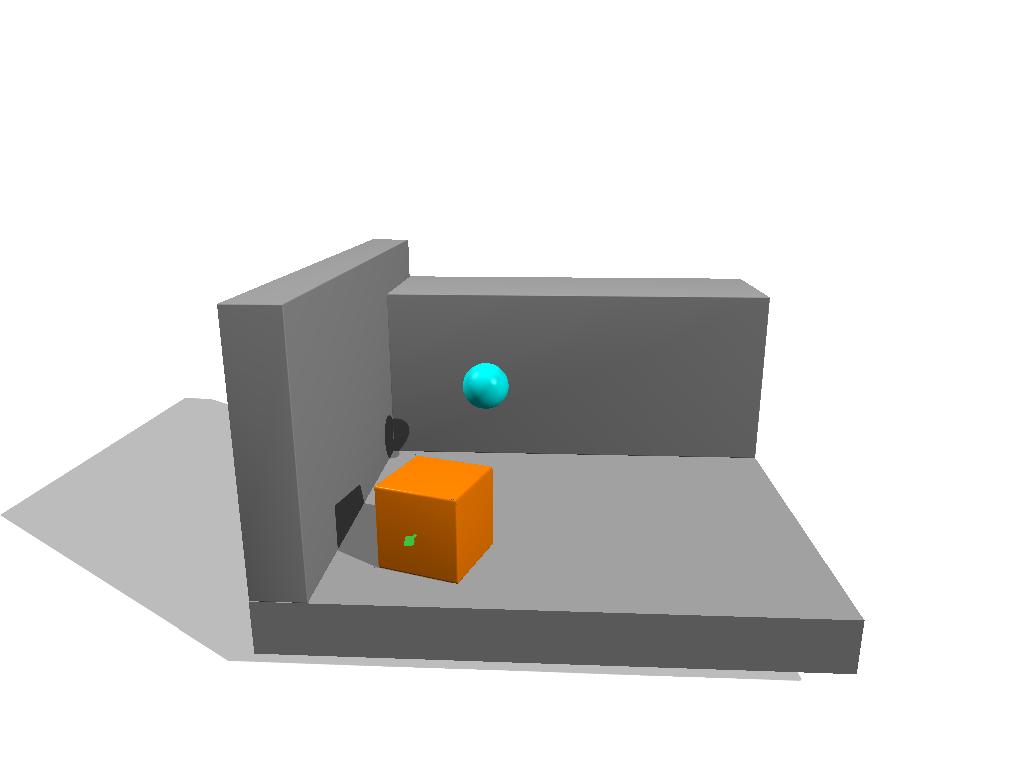}
\caption{\label{fig:sphere_cube}
  Constrained samples from the sphere cube.}
\end{figure}

\begin{figure}[p]\centering
\newcommand{\pic}[1]{\includegraphics[width=.25\columnwidth,trim={0 0 0 25pt},clip]{data/configs/images/panda_cube_sample_0000#1.jpg}}
\pic{00}\pic{01}\pic{02}\pic{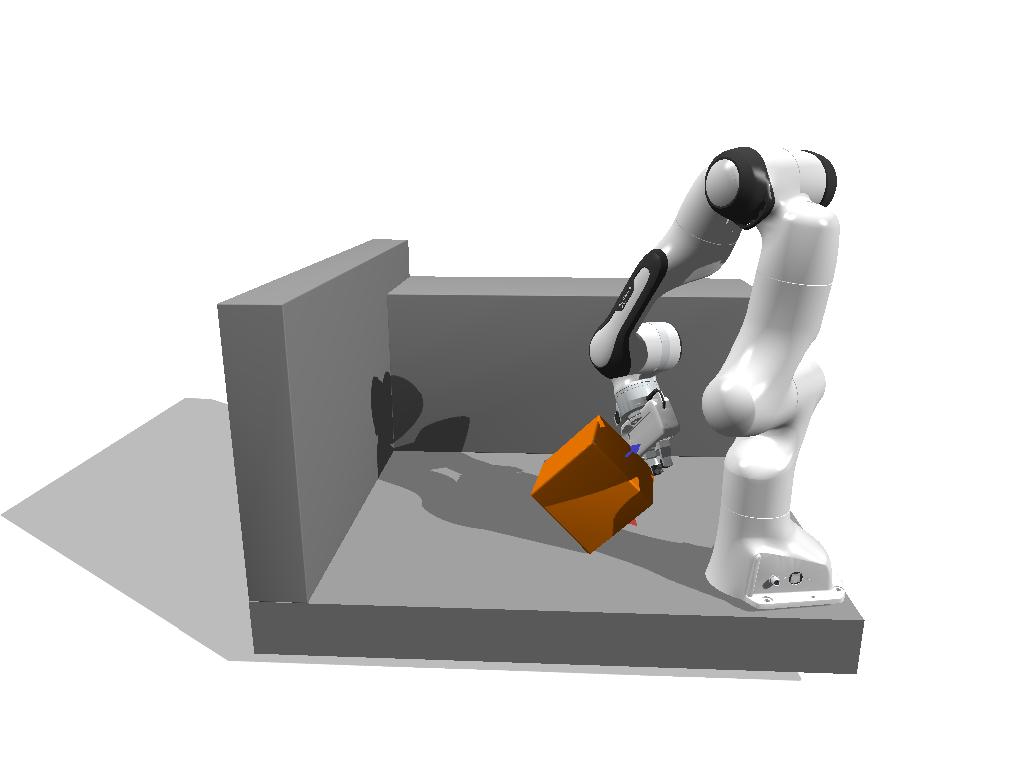}
\pic{04}\pic{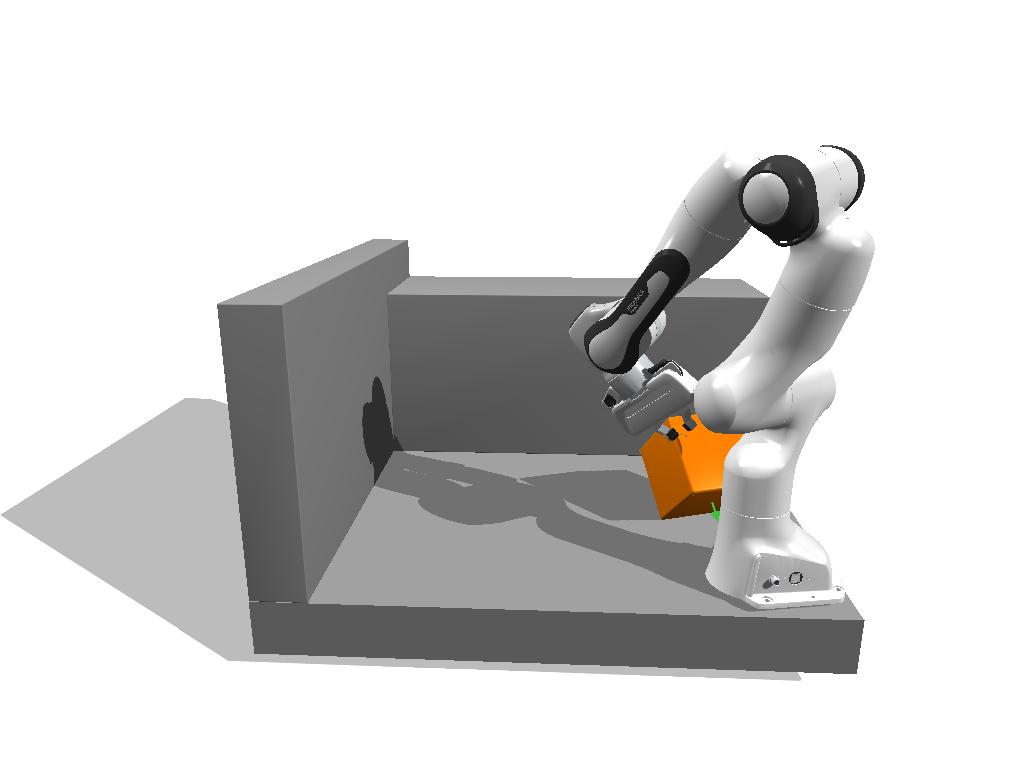}\pic{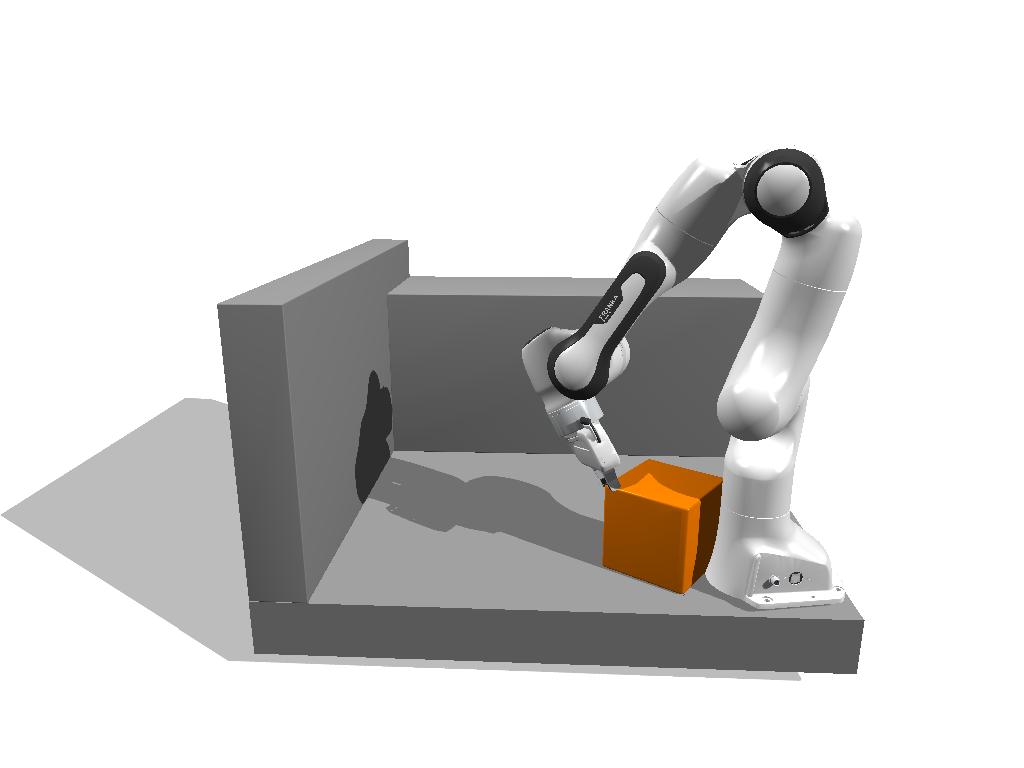}\pic{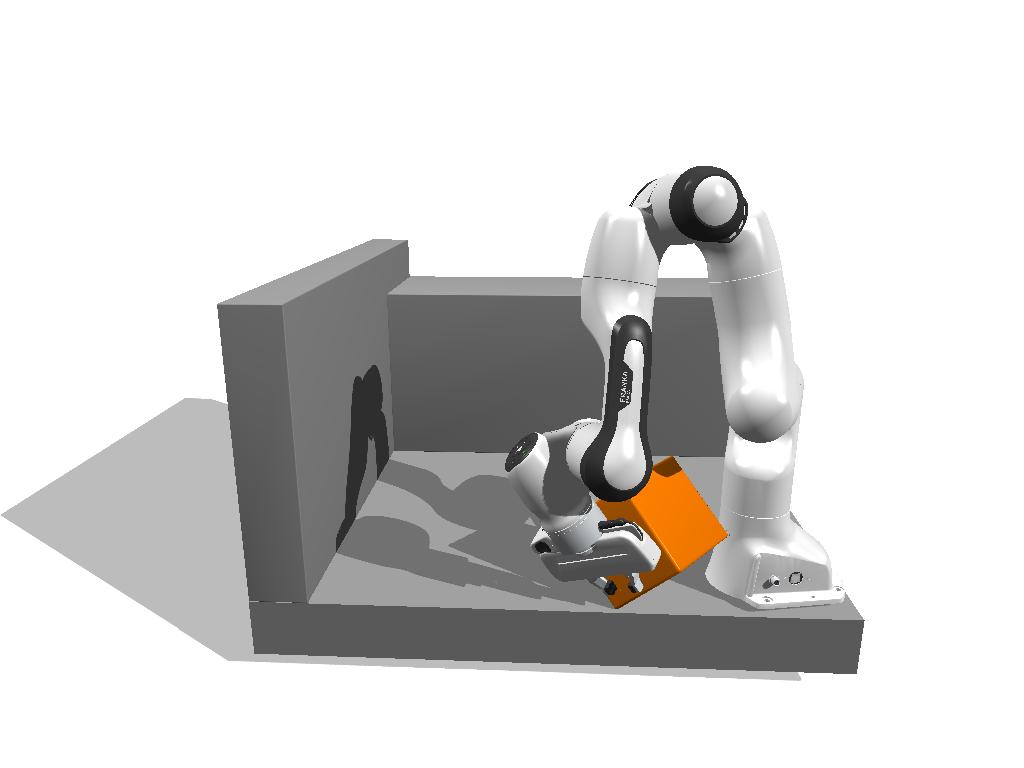}
\pic{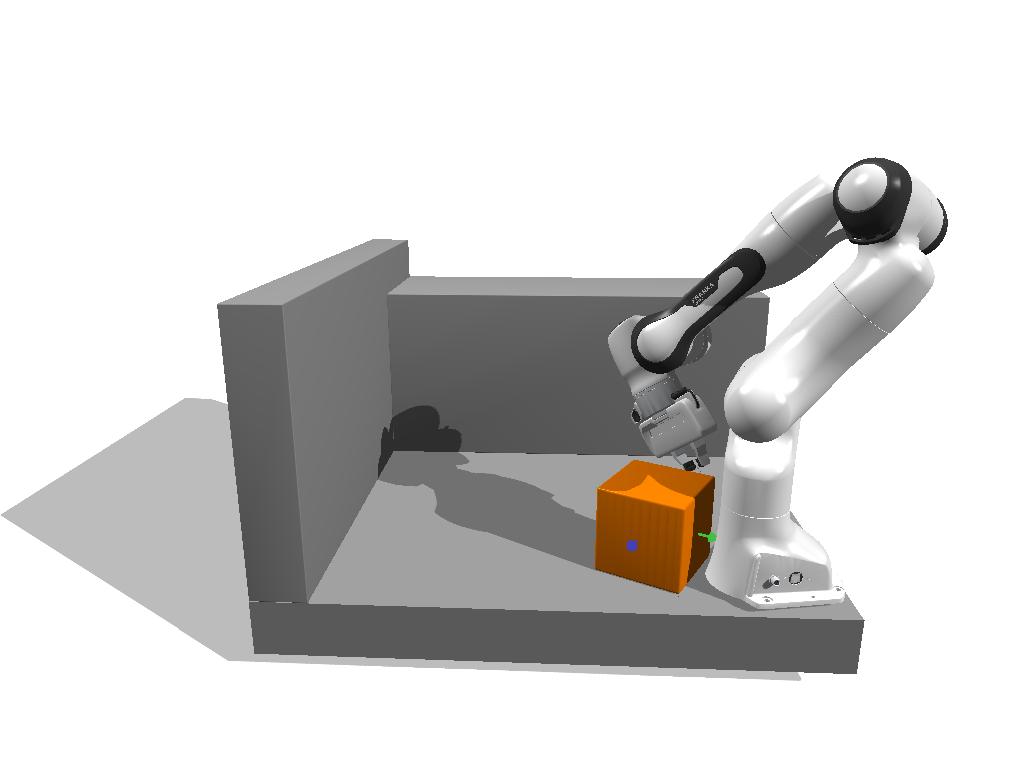}\pic{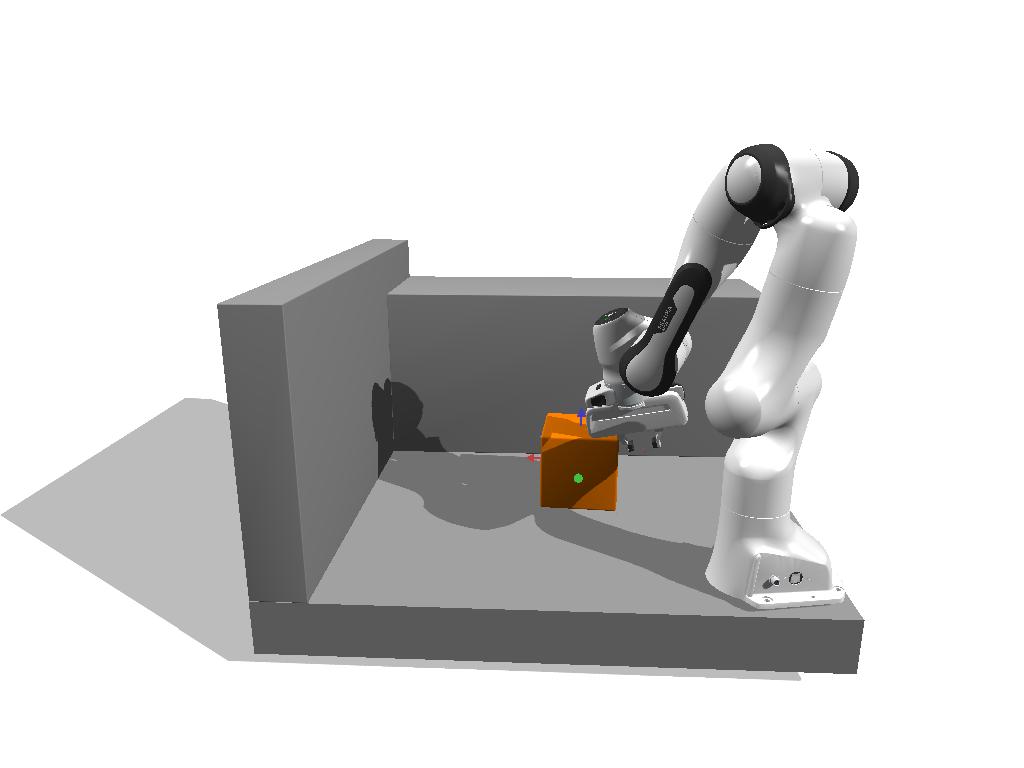}\pic{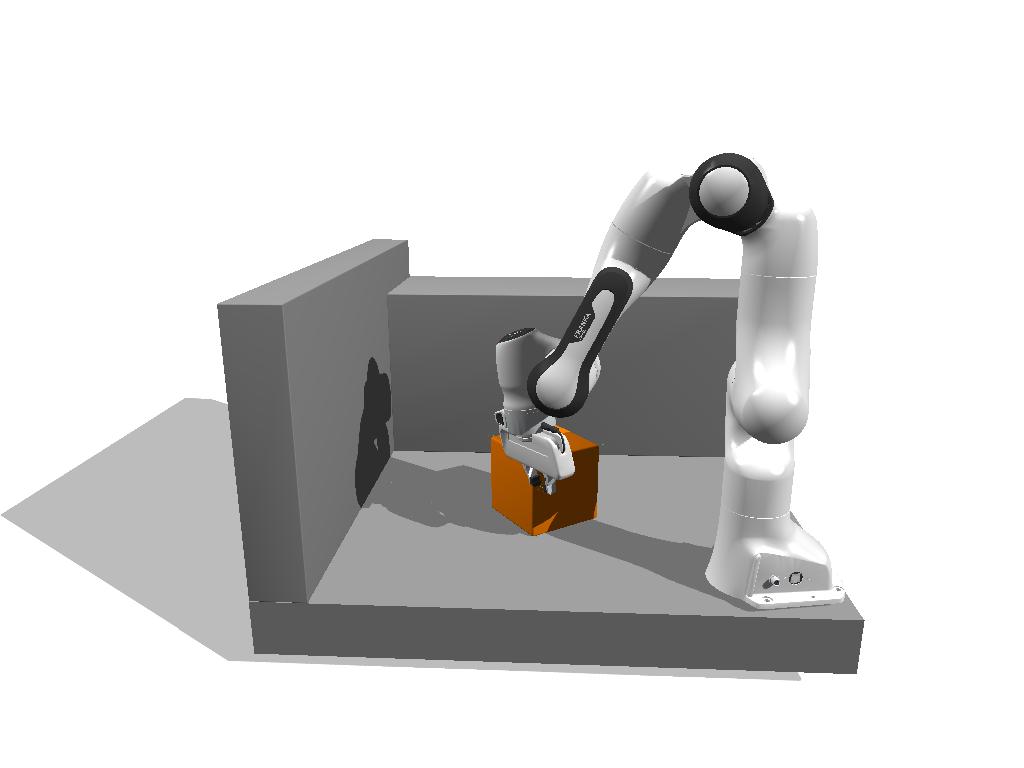}\pic{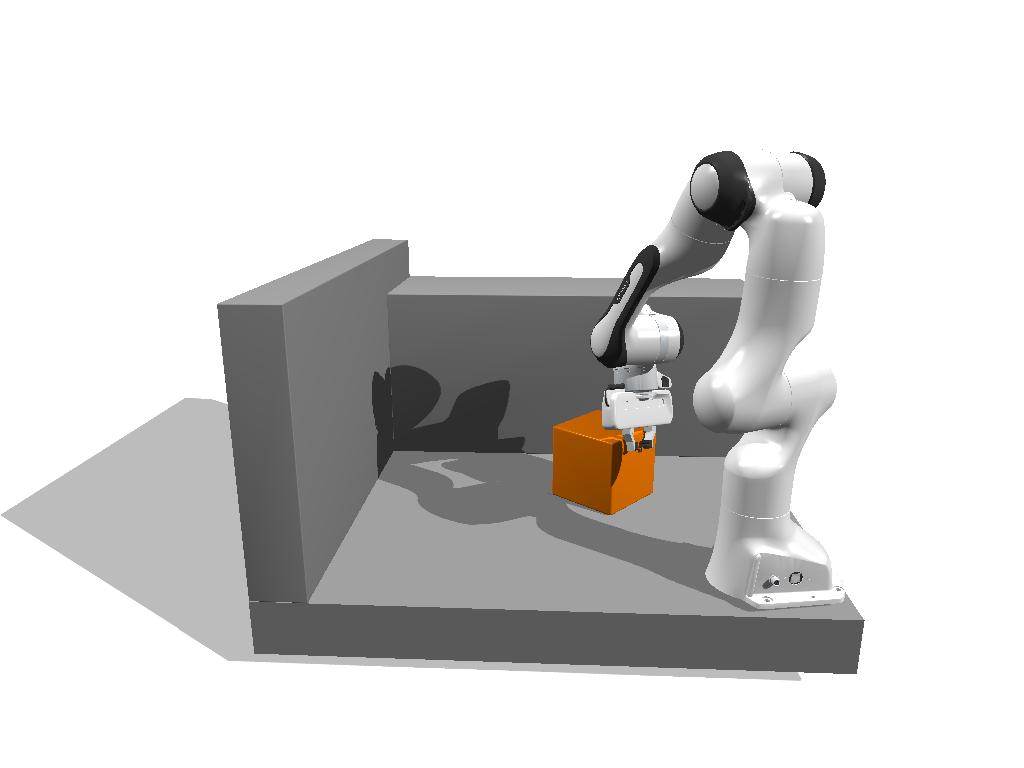}
\pic{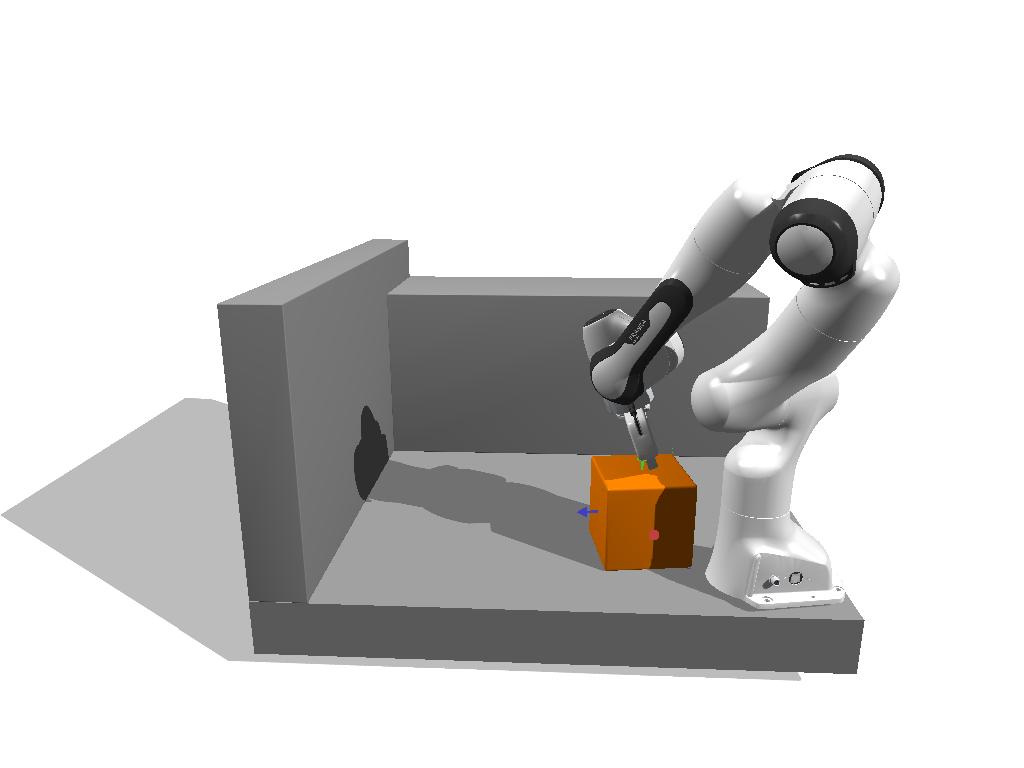}\pic{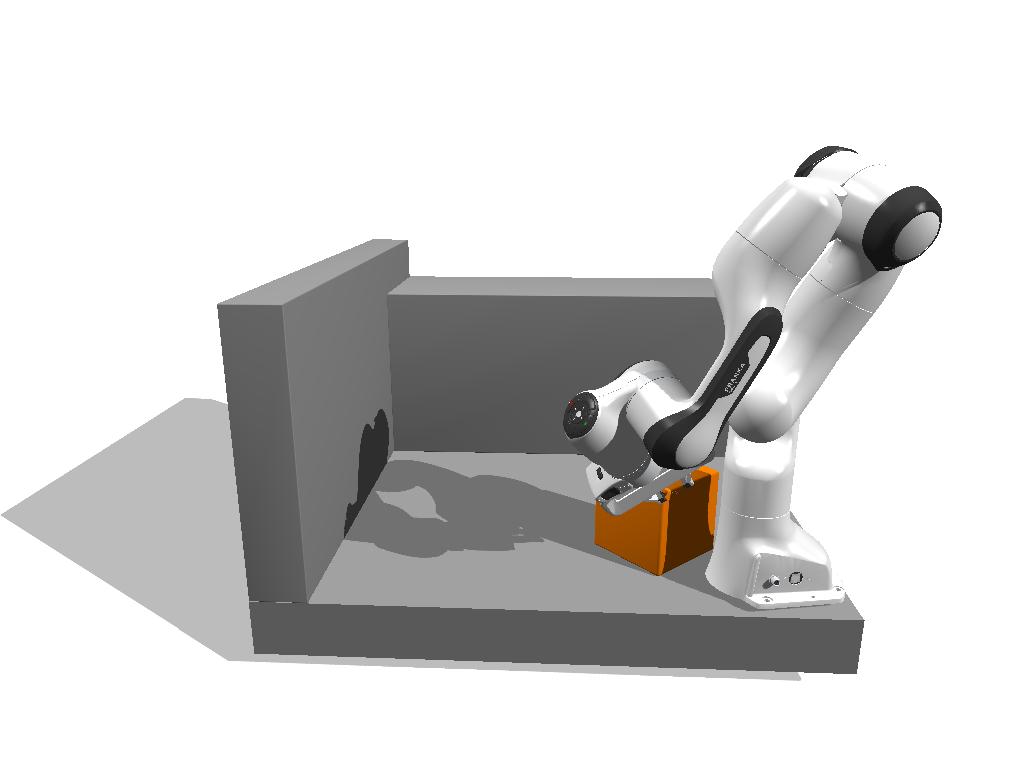}\pic{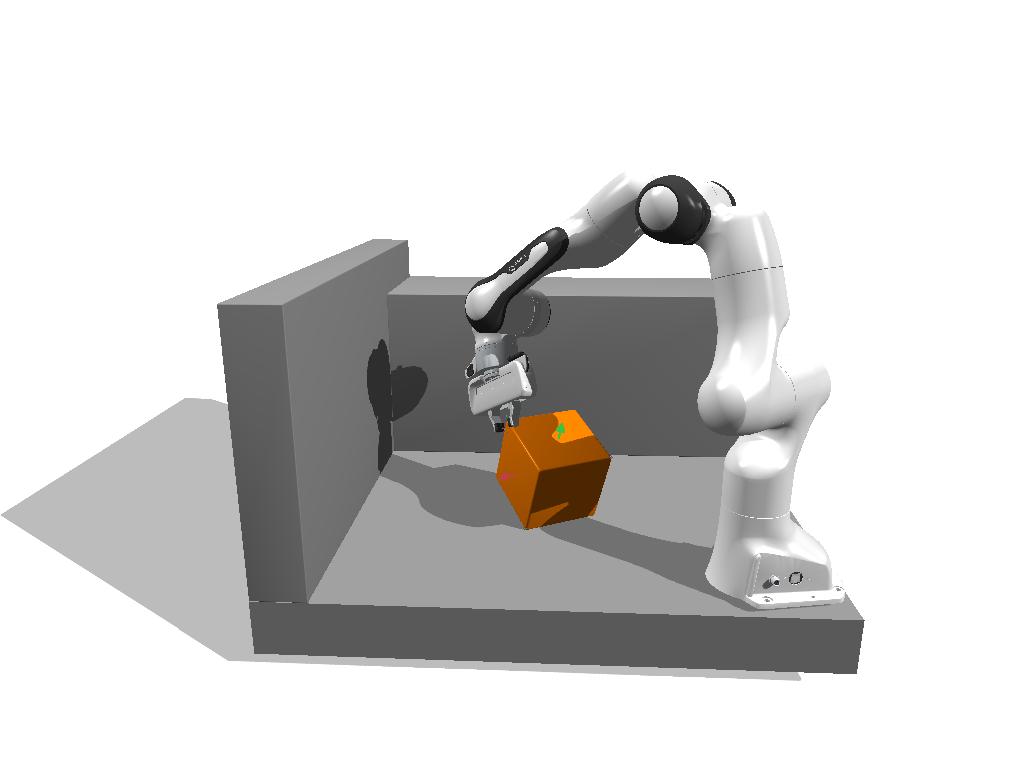}\pic{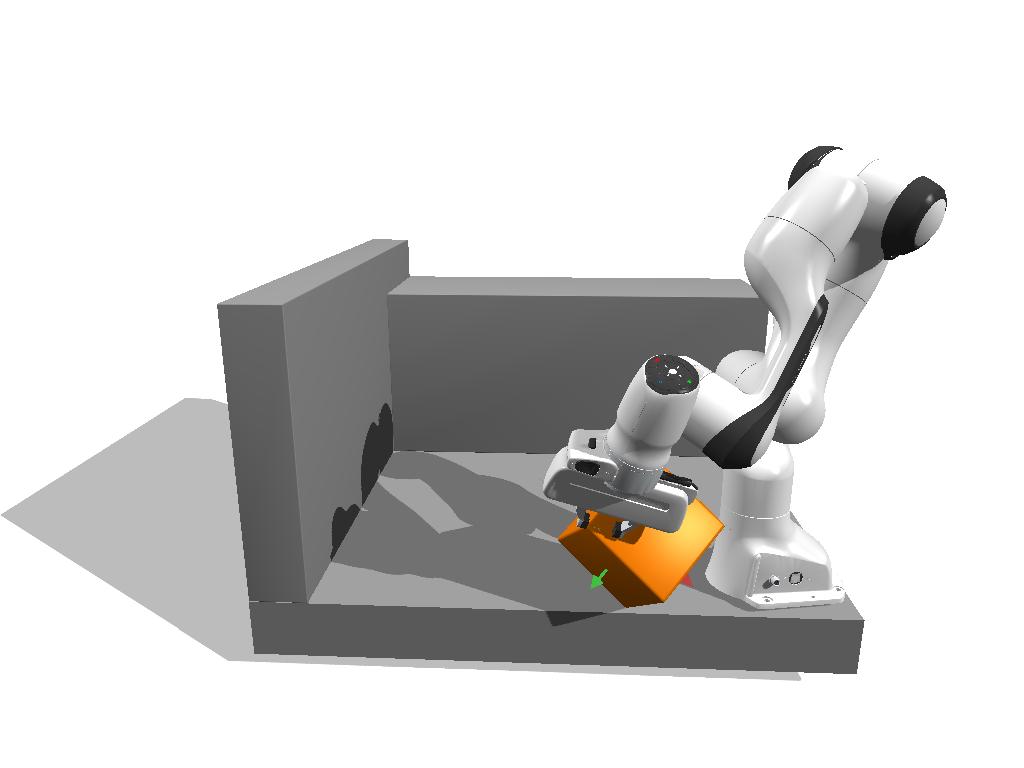}
\pic{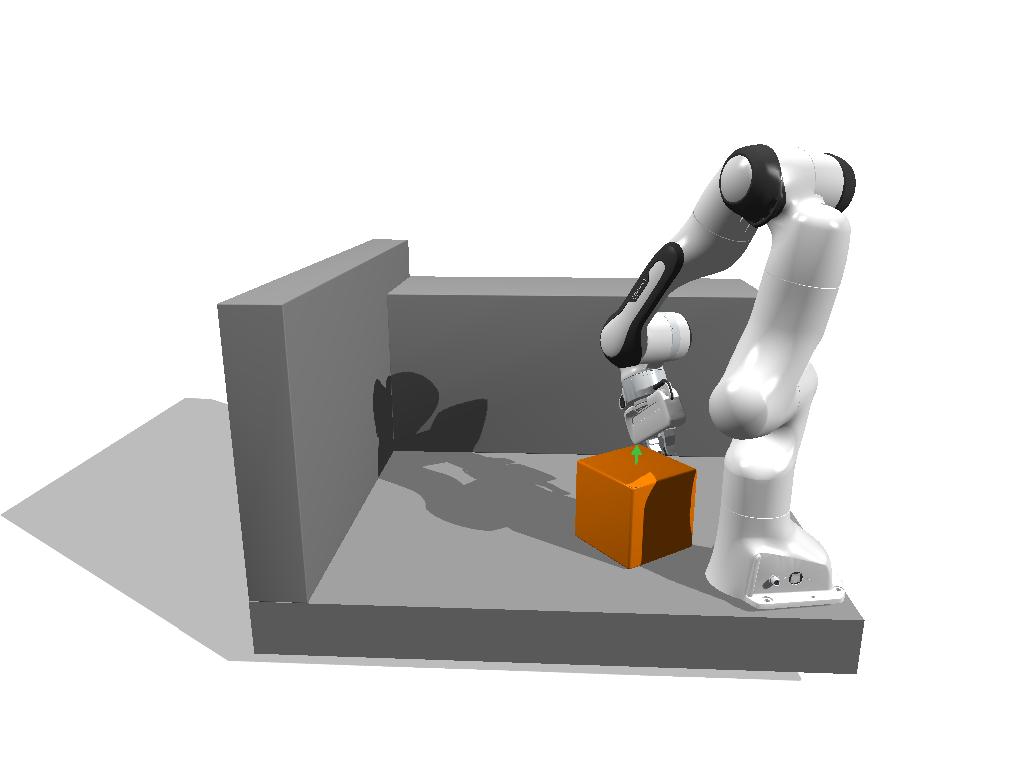}\pic{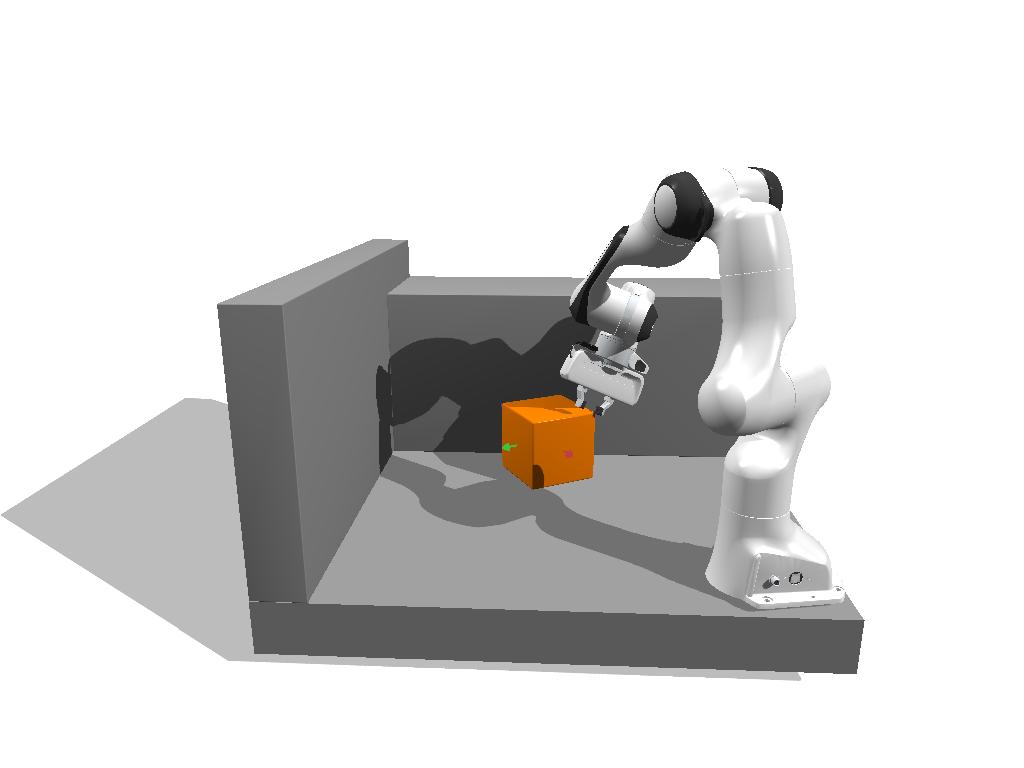}\pic{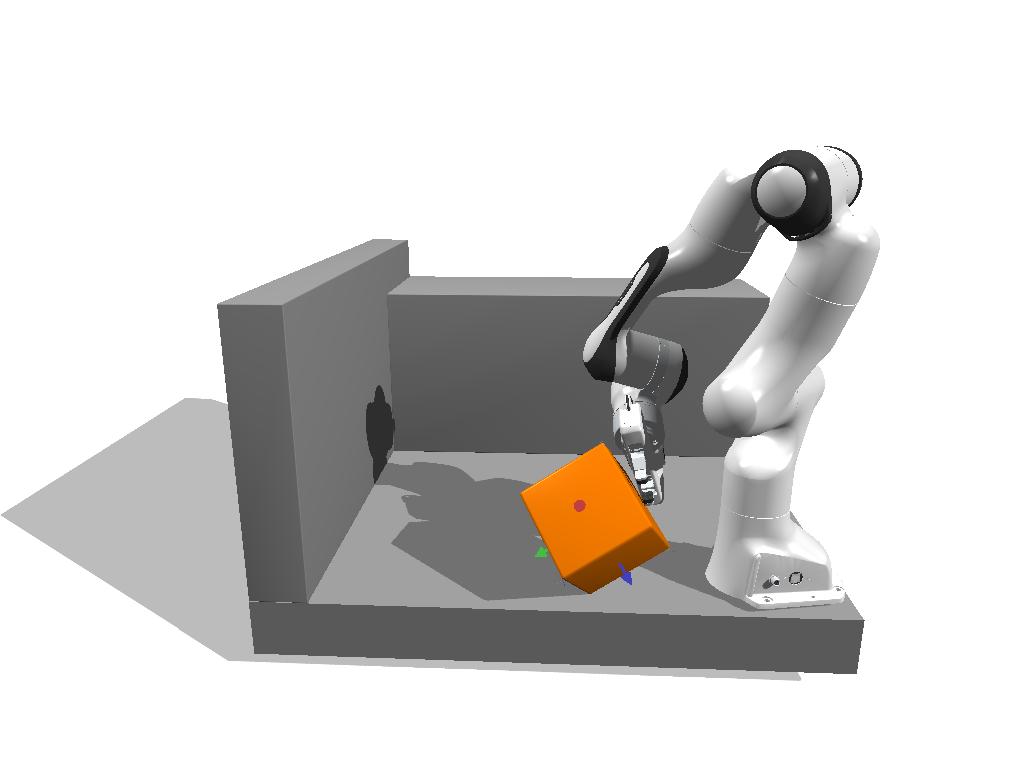}\pic{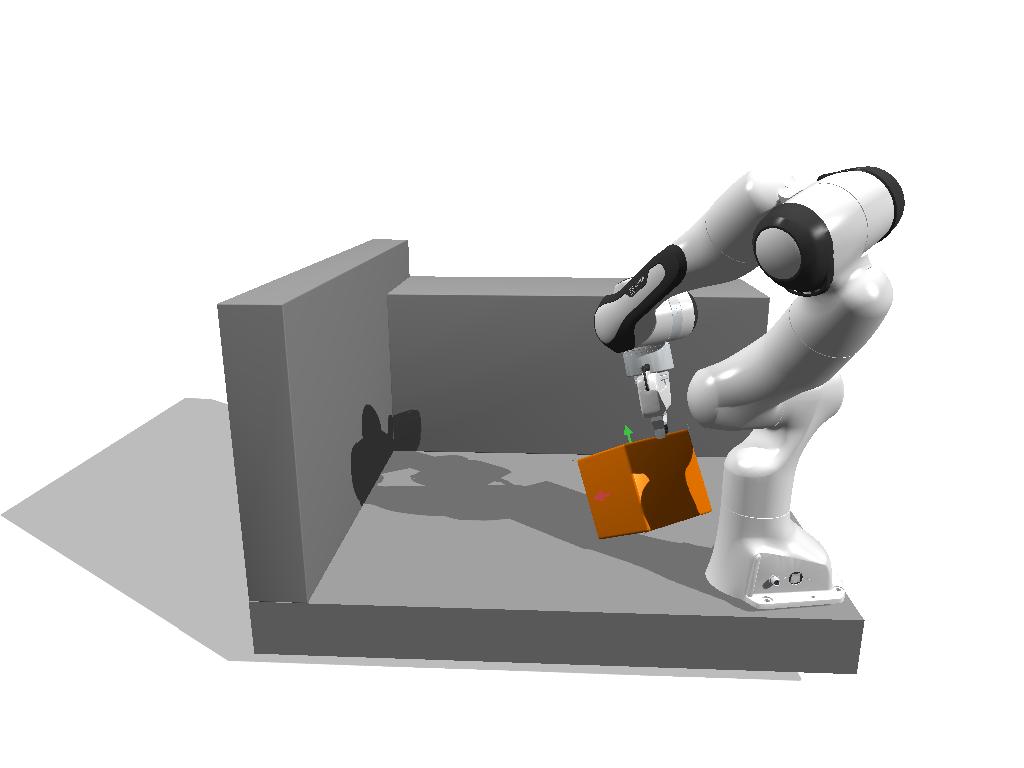}
\pic{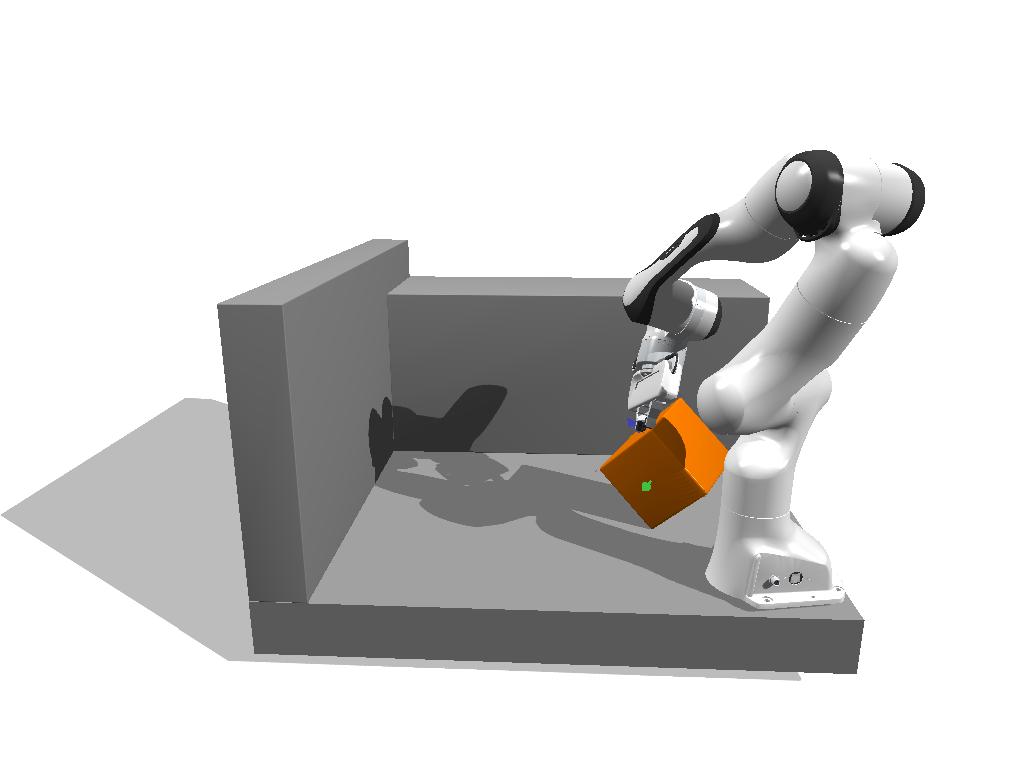}\pic{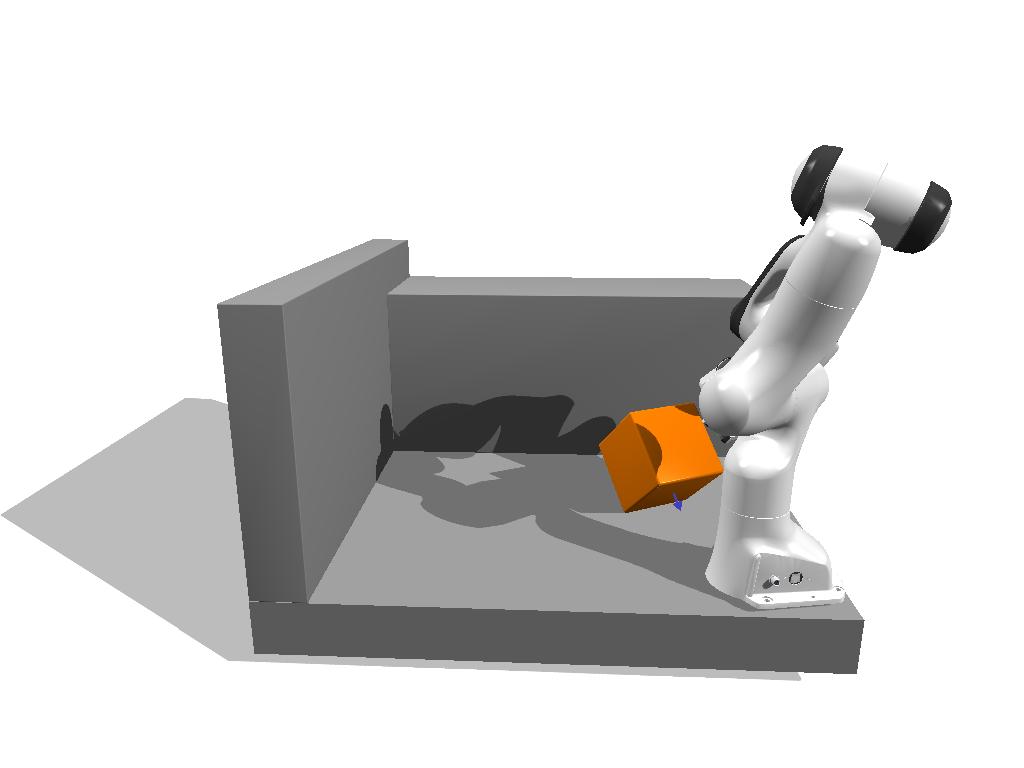}\pic{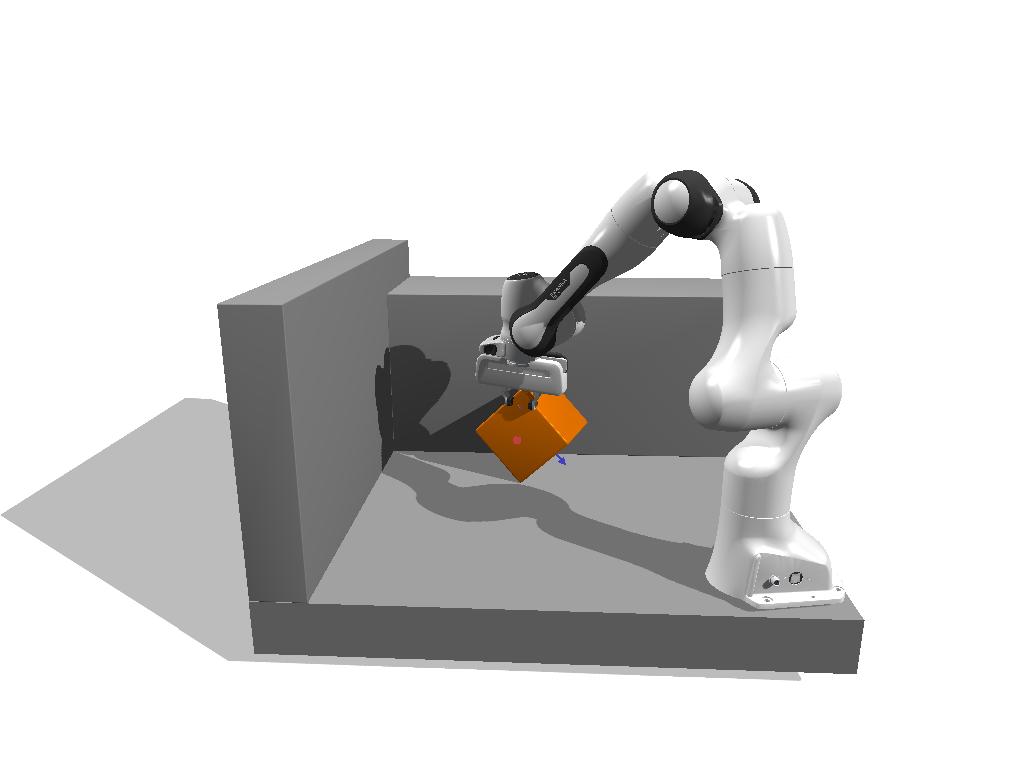}\pic{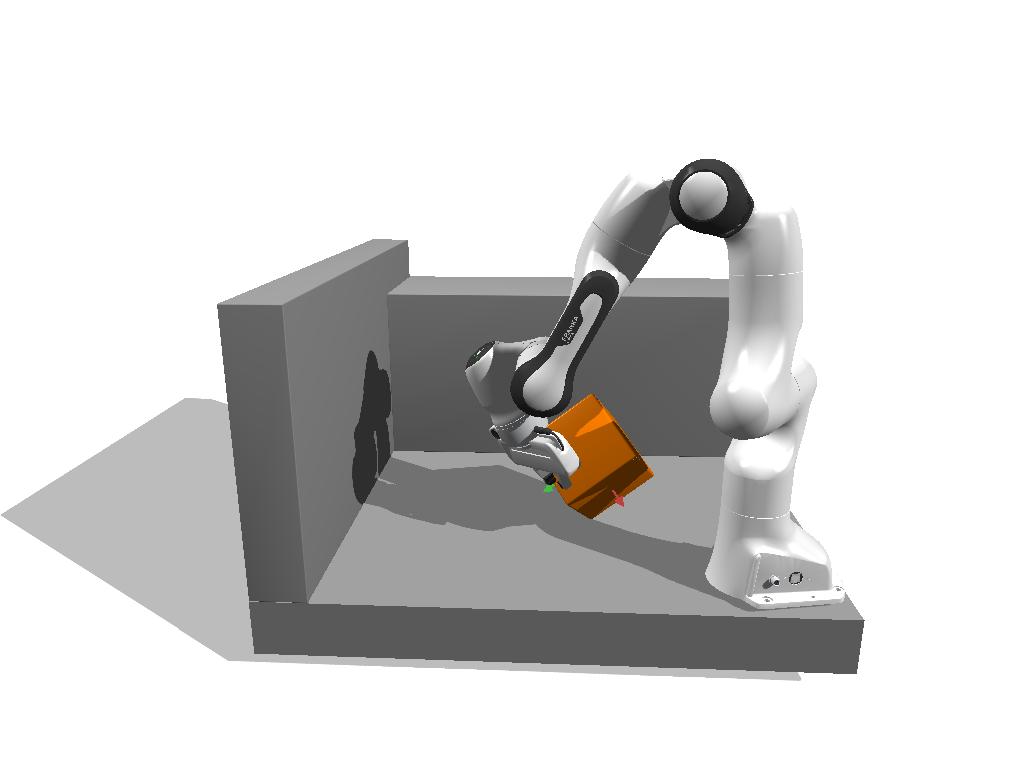}
\pic{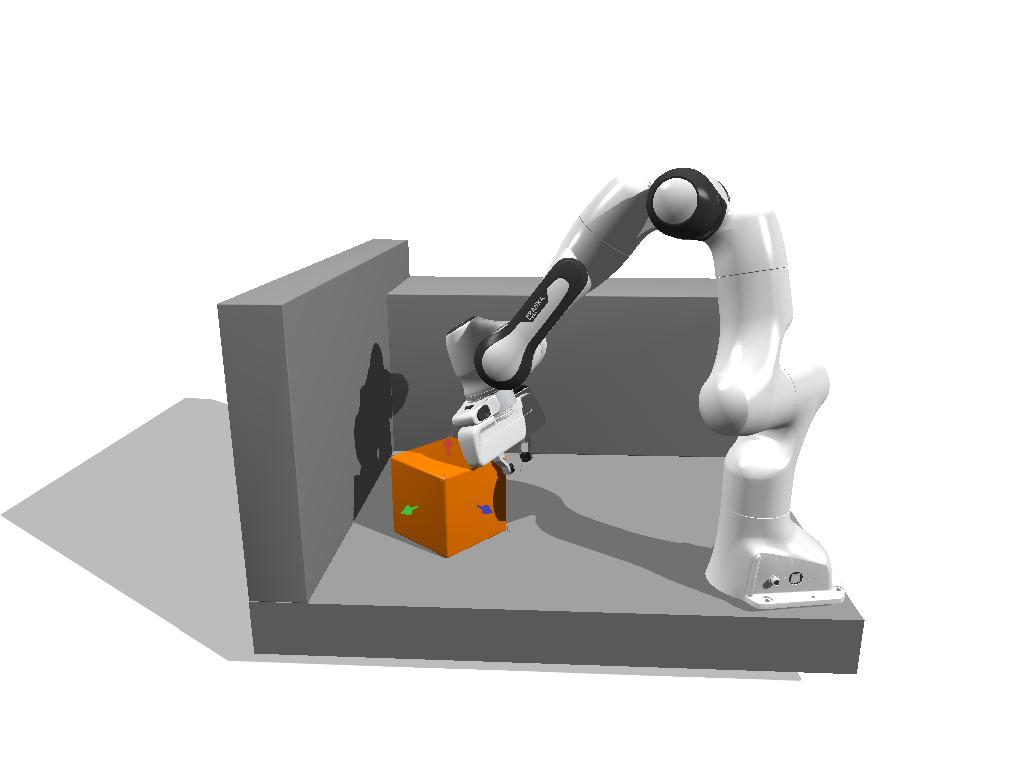}\pic{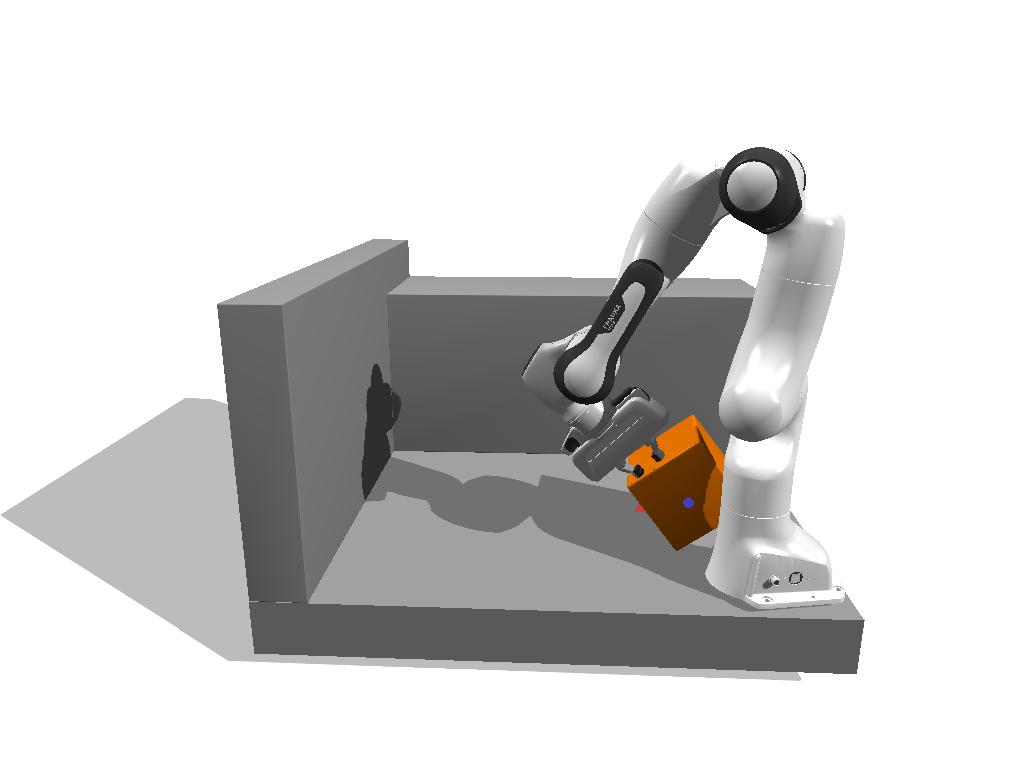}\pic{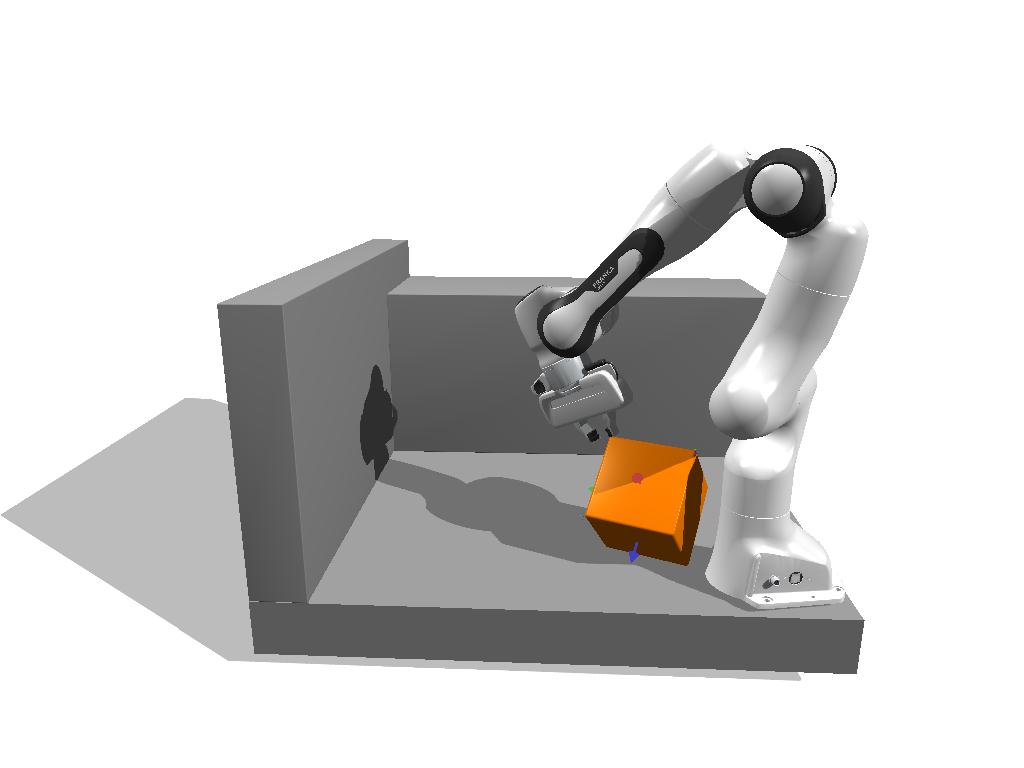}\pic{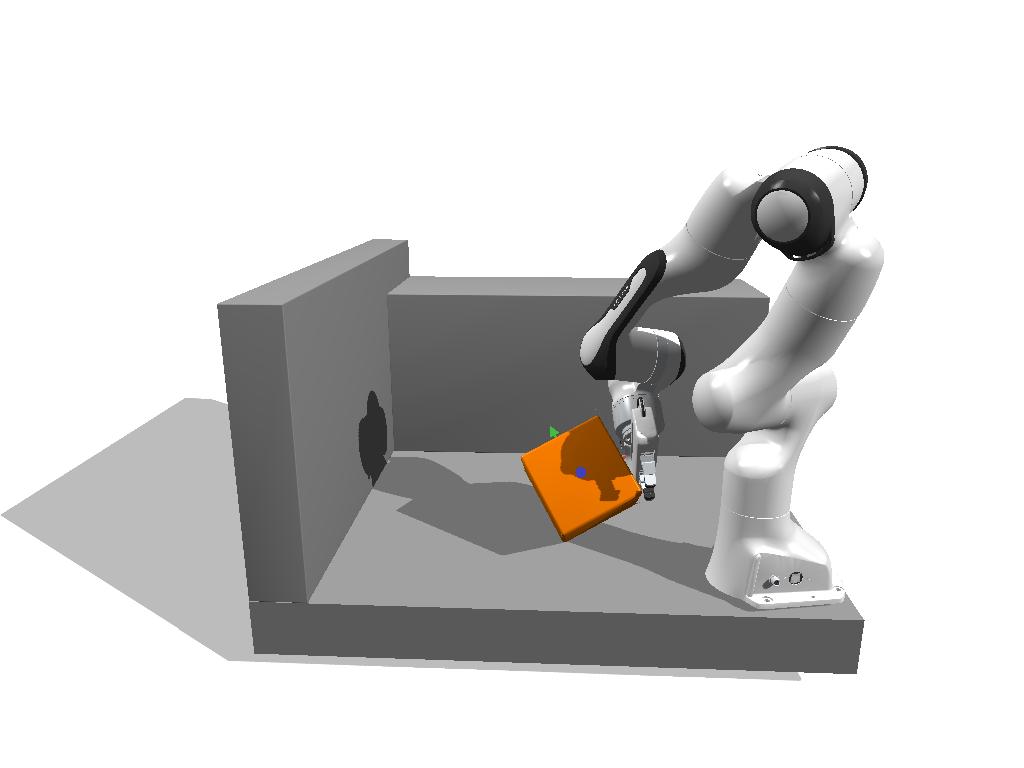}
\caption{\label{fig:panda_cube}
  Constrained samples from the panda cube domain.}
\end{figure}

\begin{figure}[p]\centering
\newcommand{\pic}[1]{\includegraphics[width=.25\columnwidth,trim={0 0 0 25pt},clip]{data/configs/images/allegro_cube_sample_0000#1.jpg}}
\pic{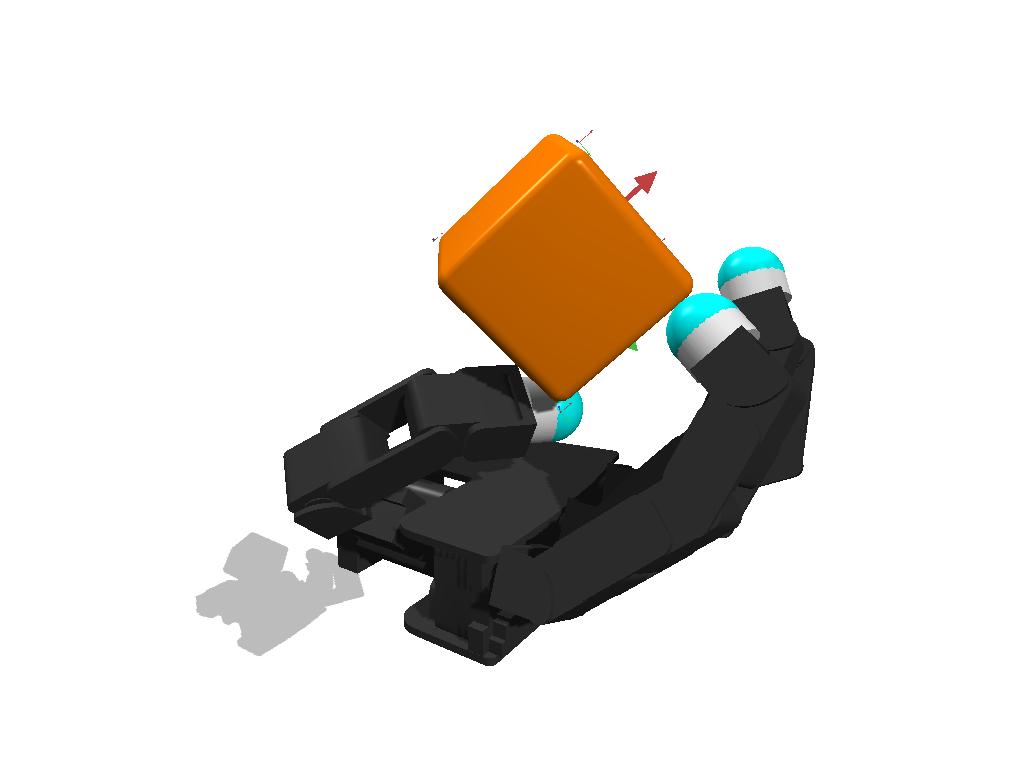}\pic{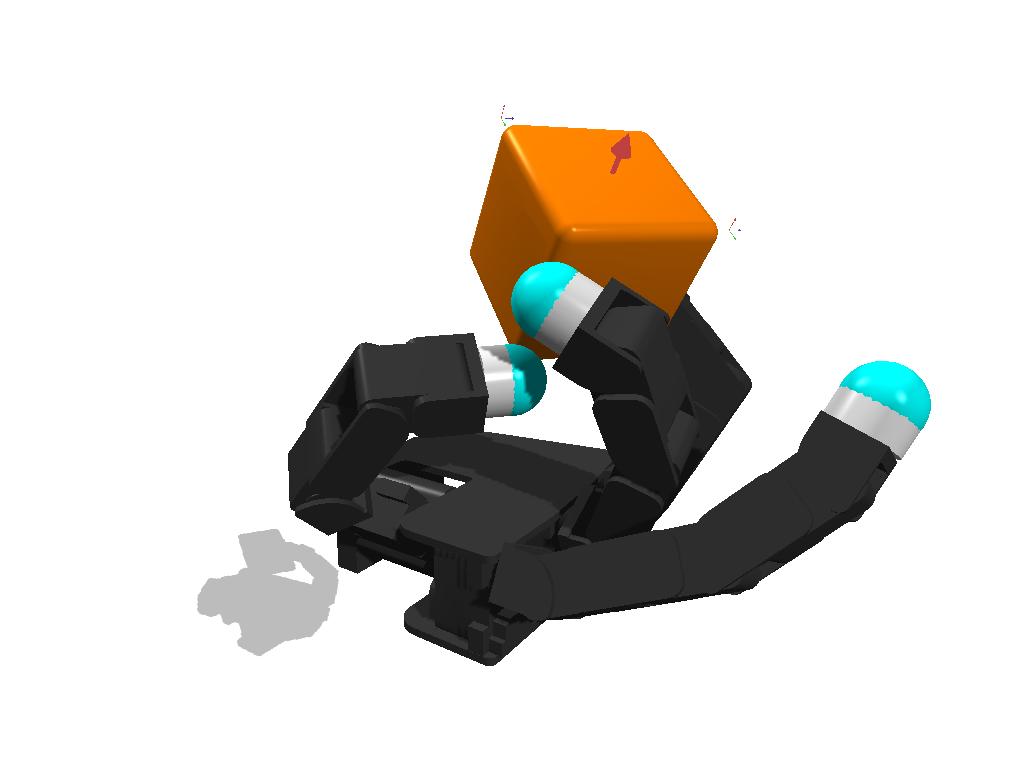}\pic{02}\pic{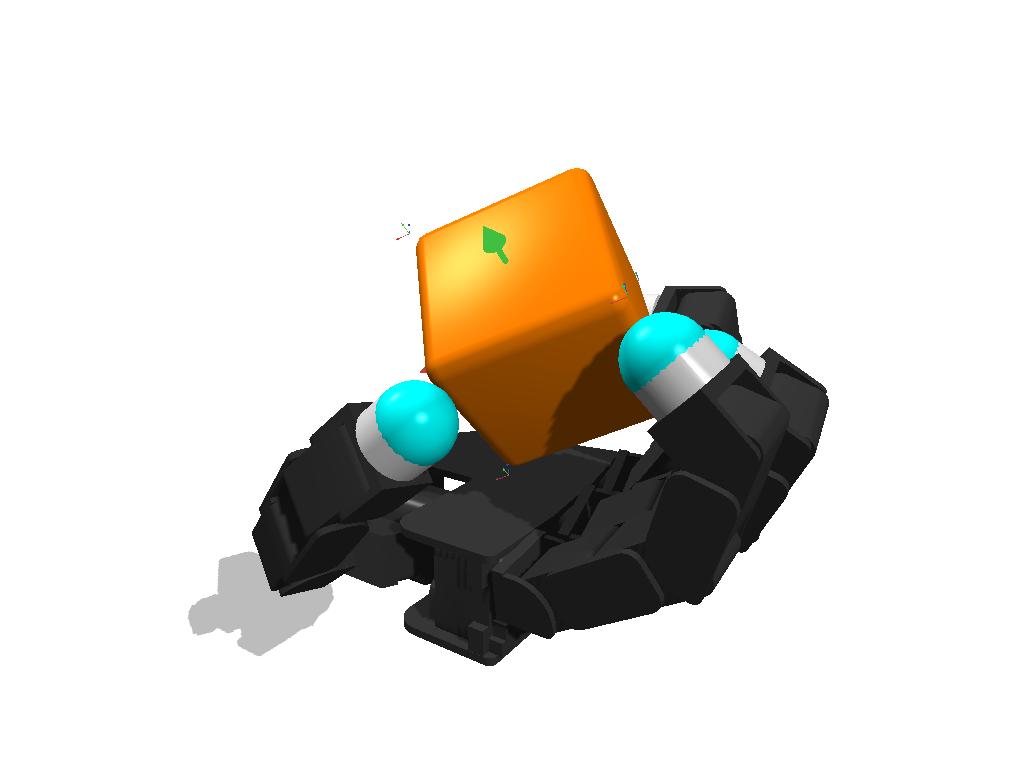}
\pic{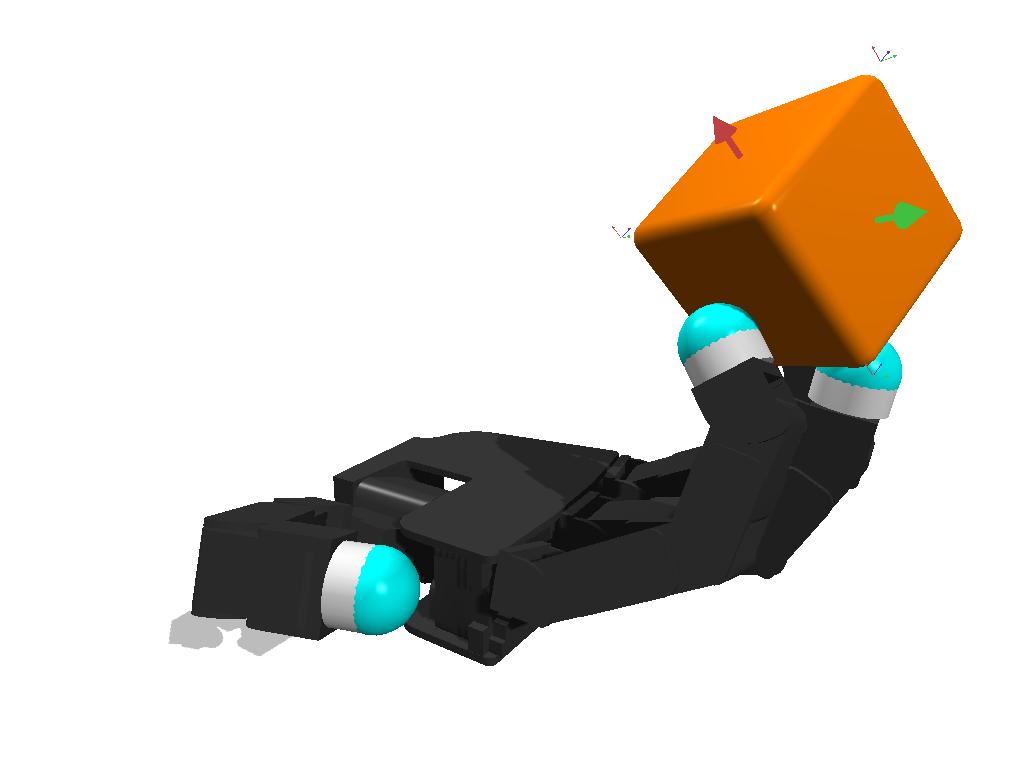}\pic{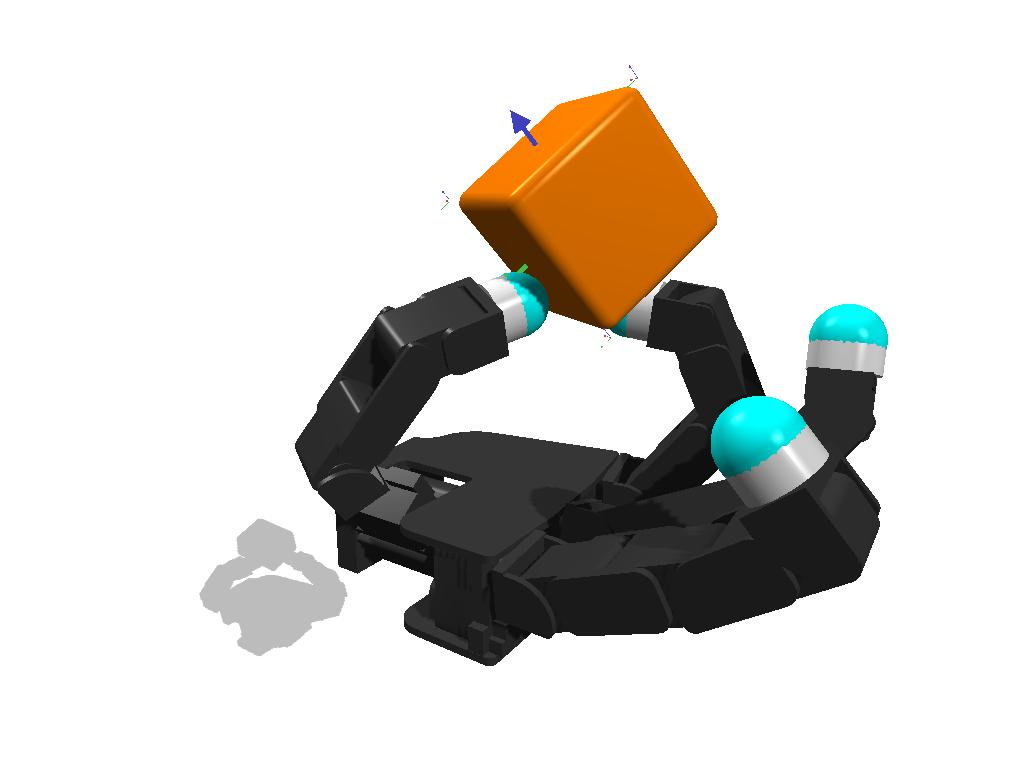}\pic{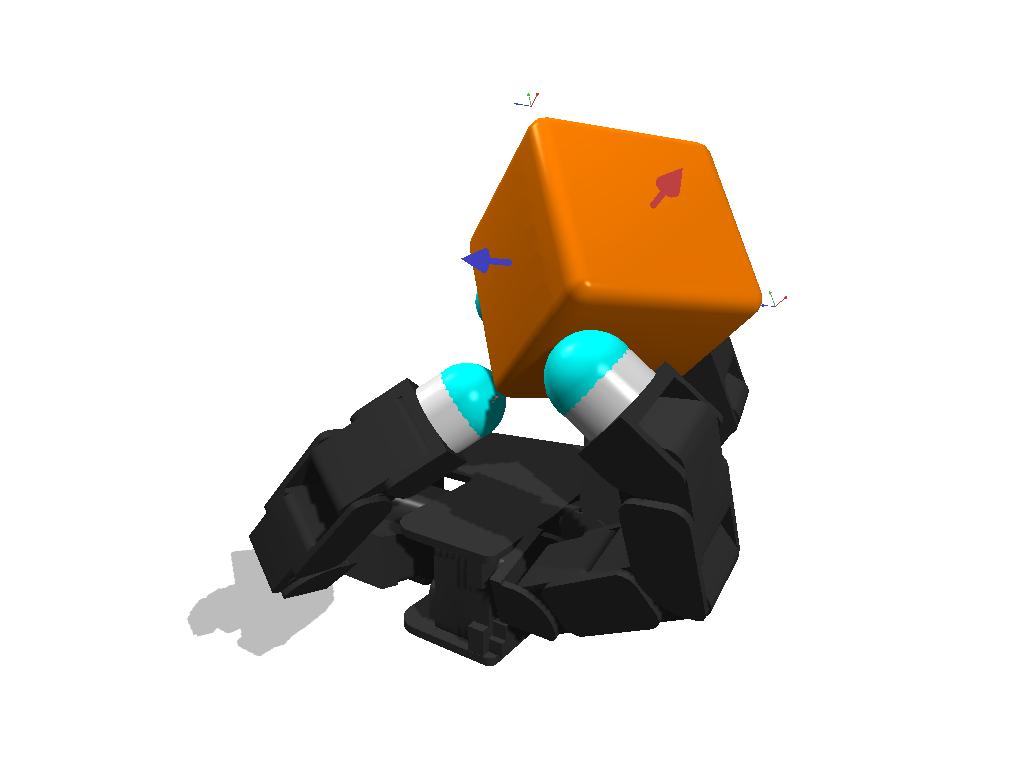}\pic{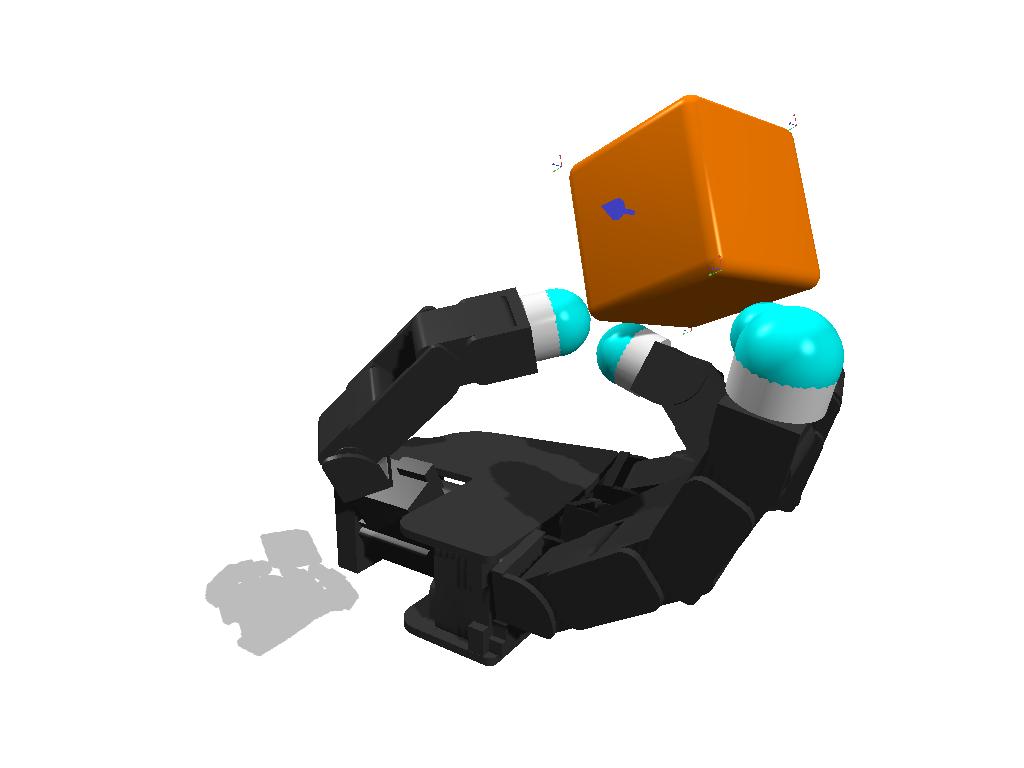}
\pic{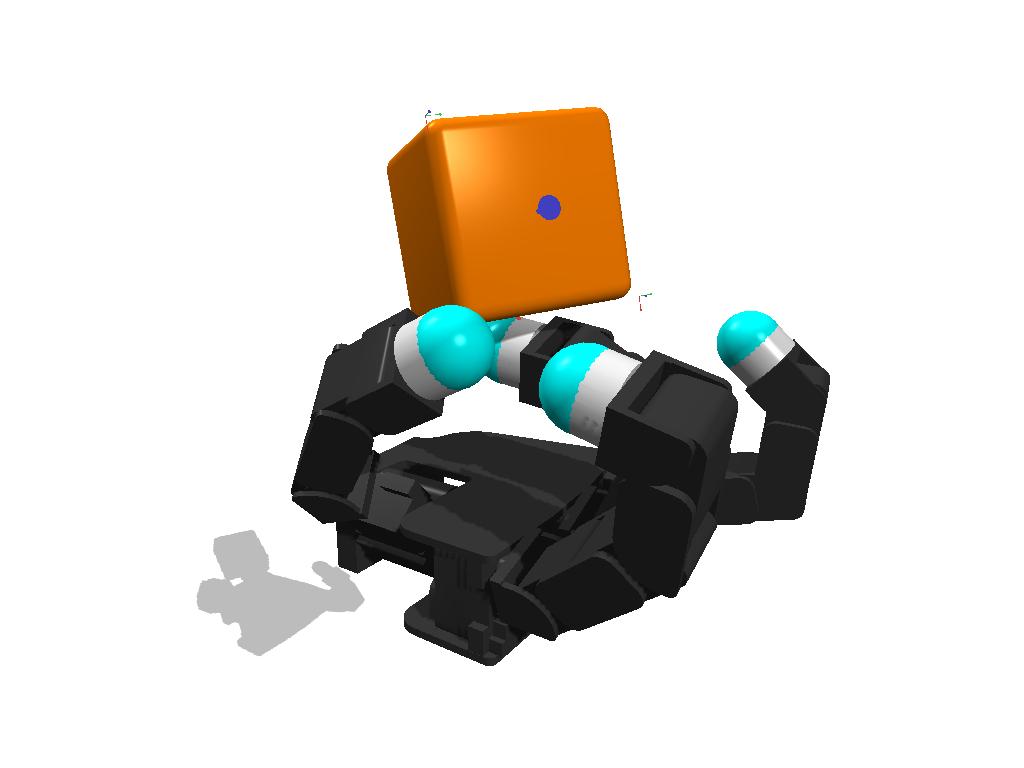}\pic{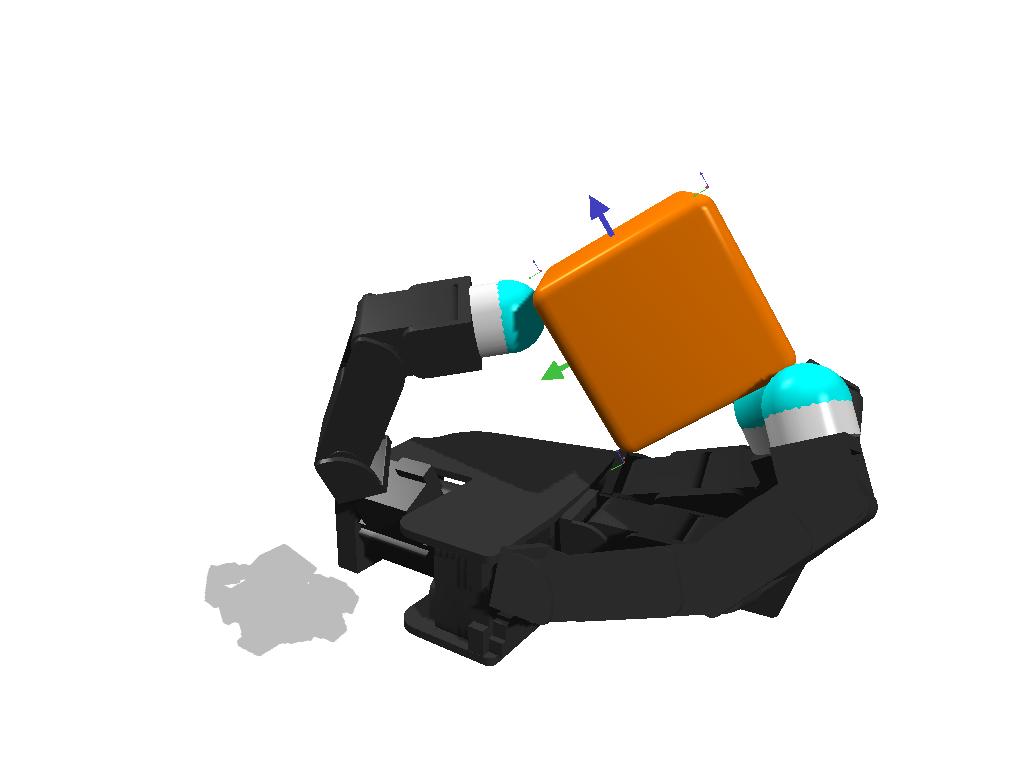}\pic{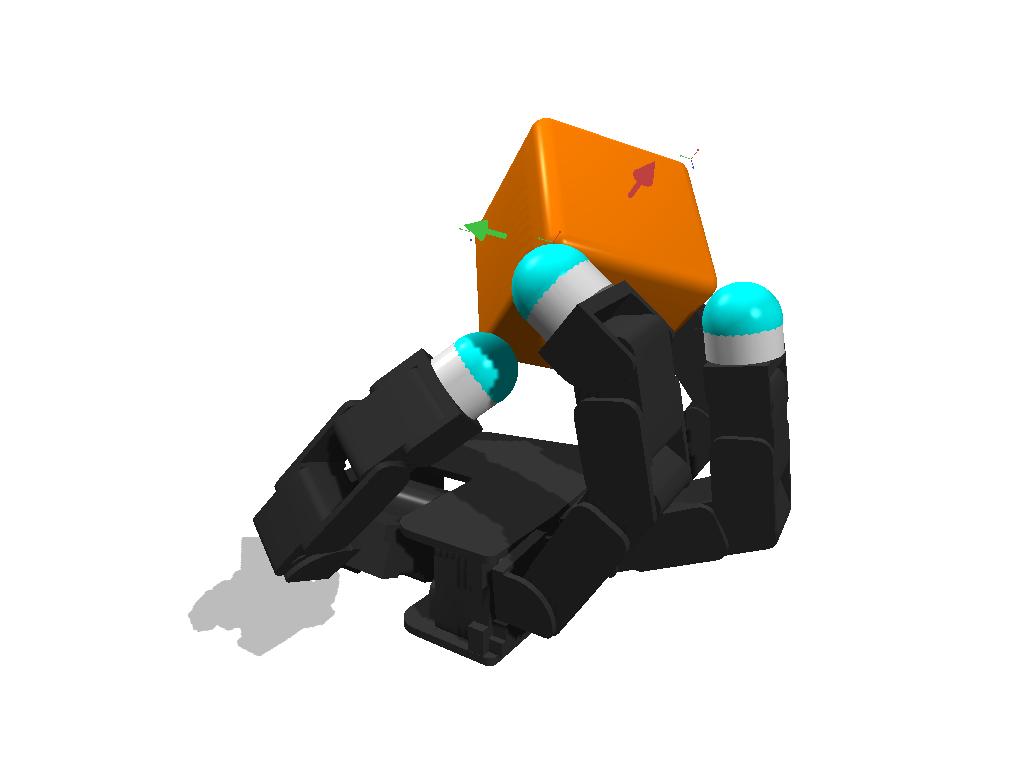}\pic{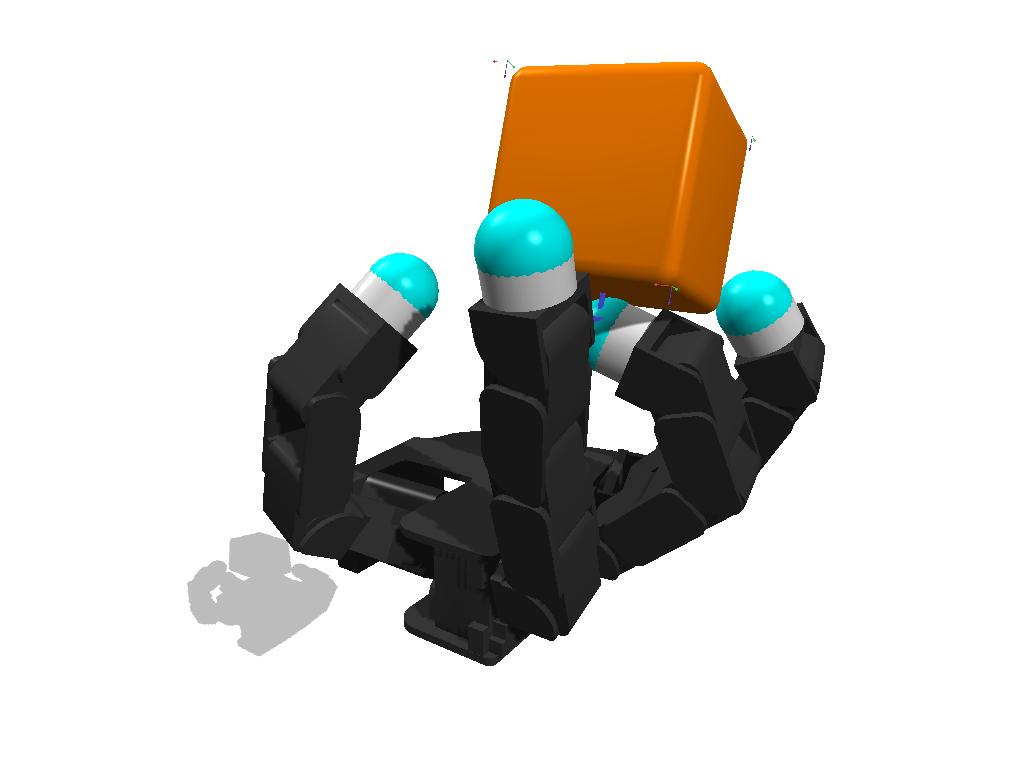}
\pic{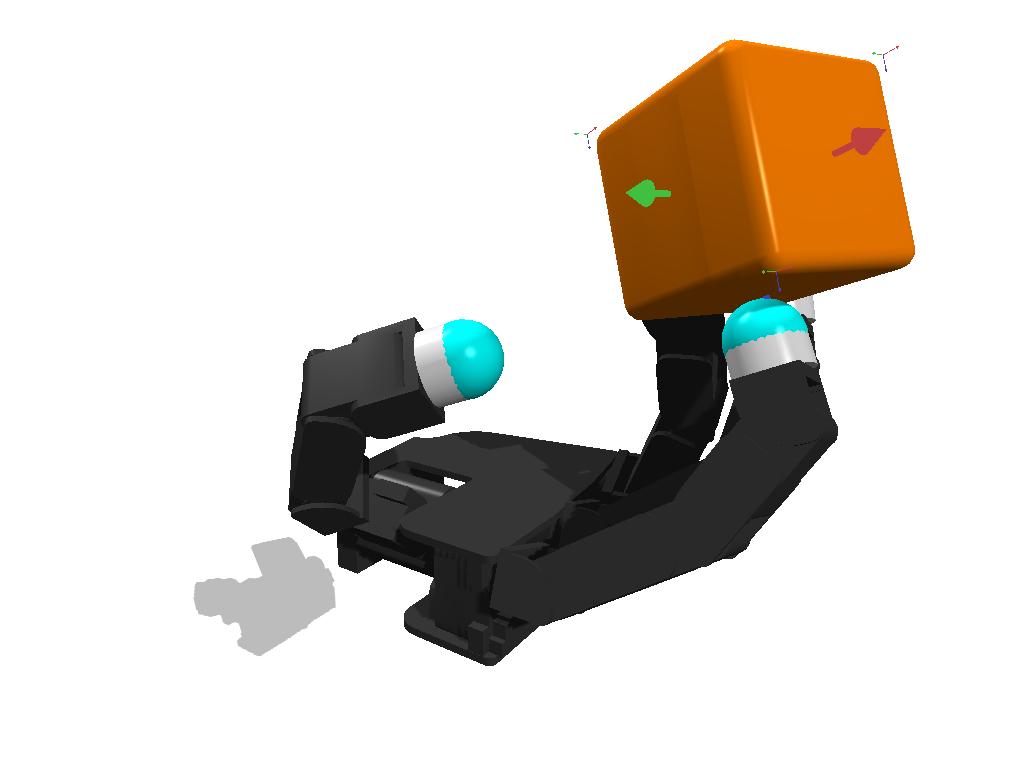}\pic{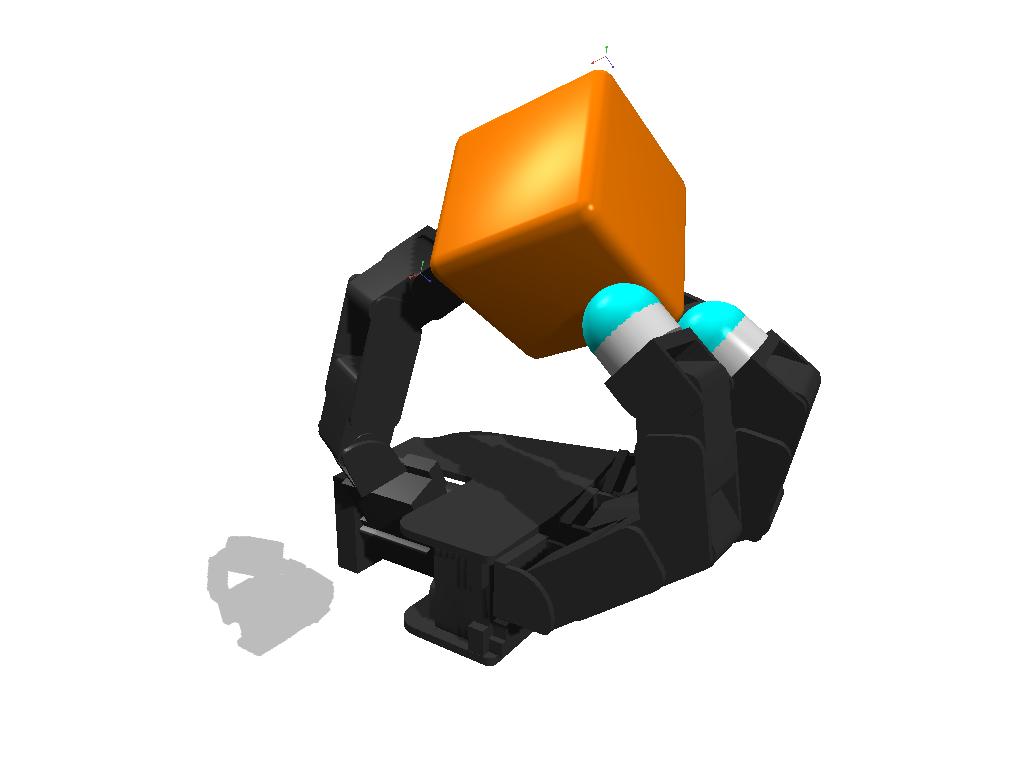}\pic{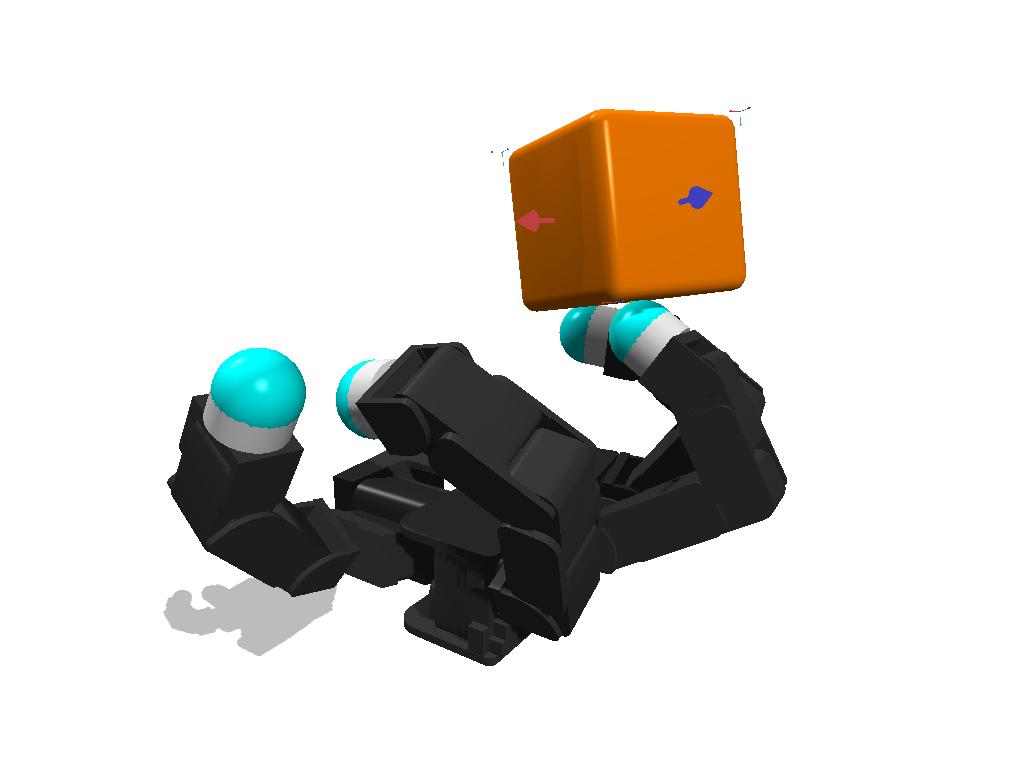}\pic{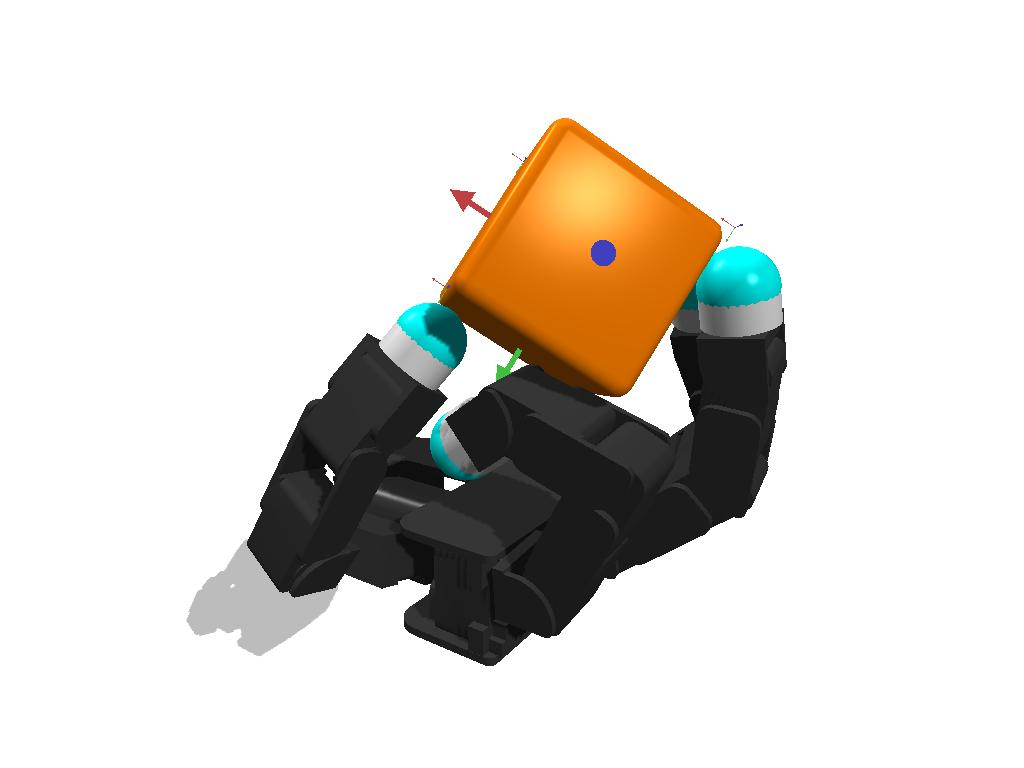}
\pic{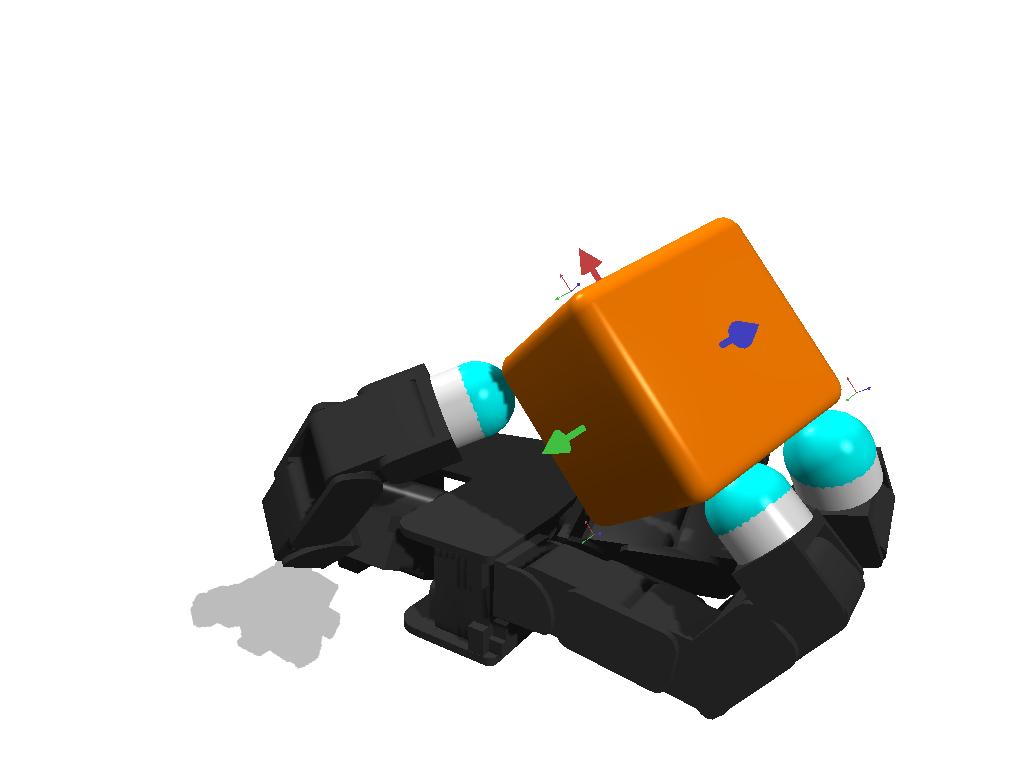}\pic{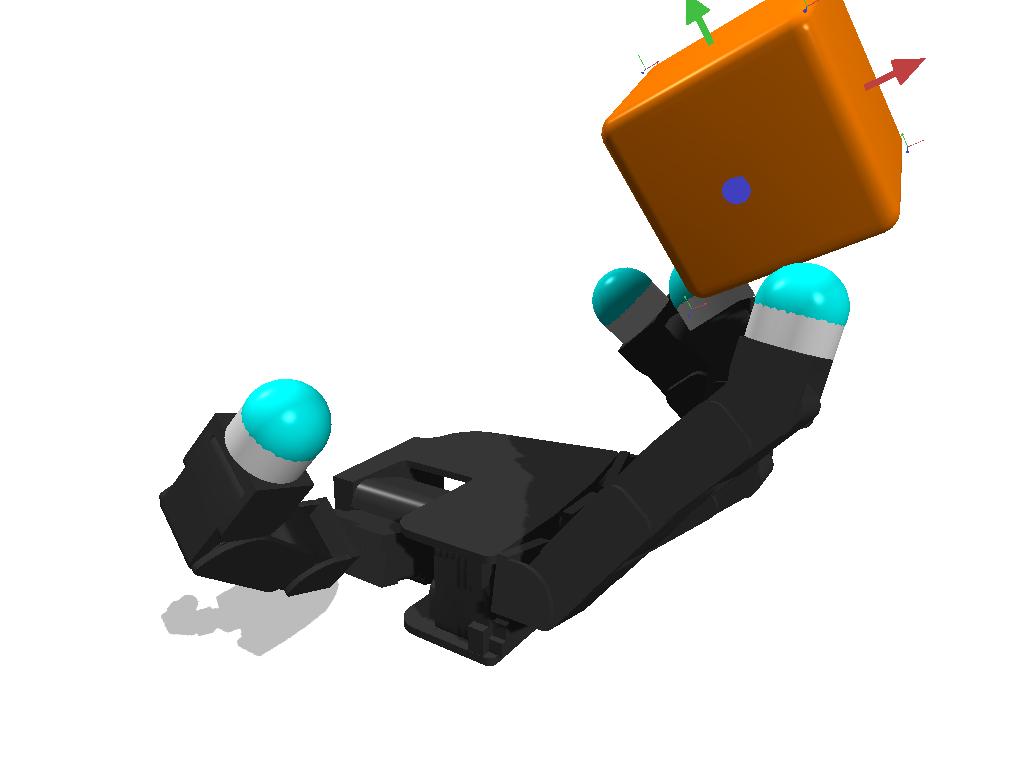}\pic{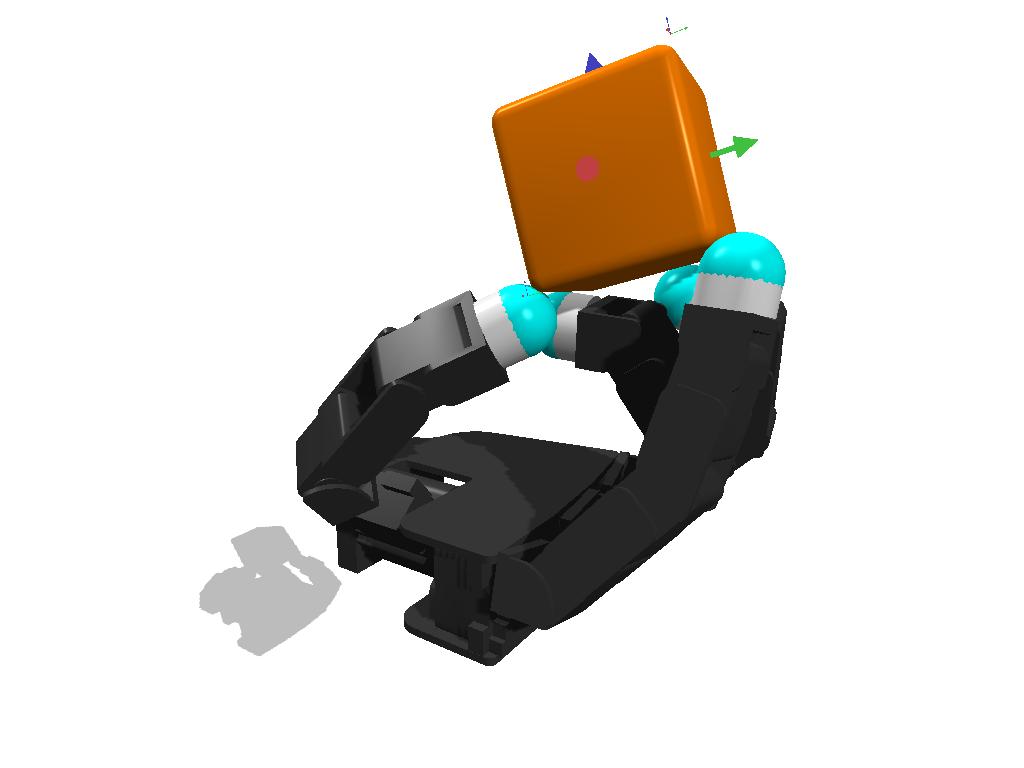}\pic{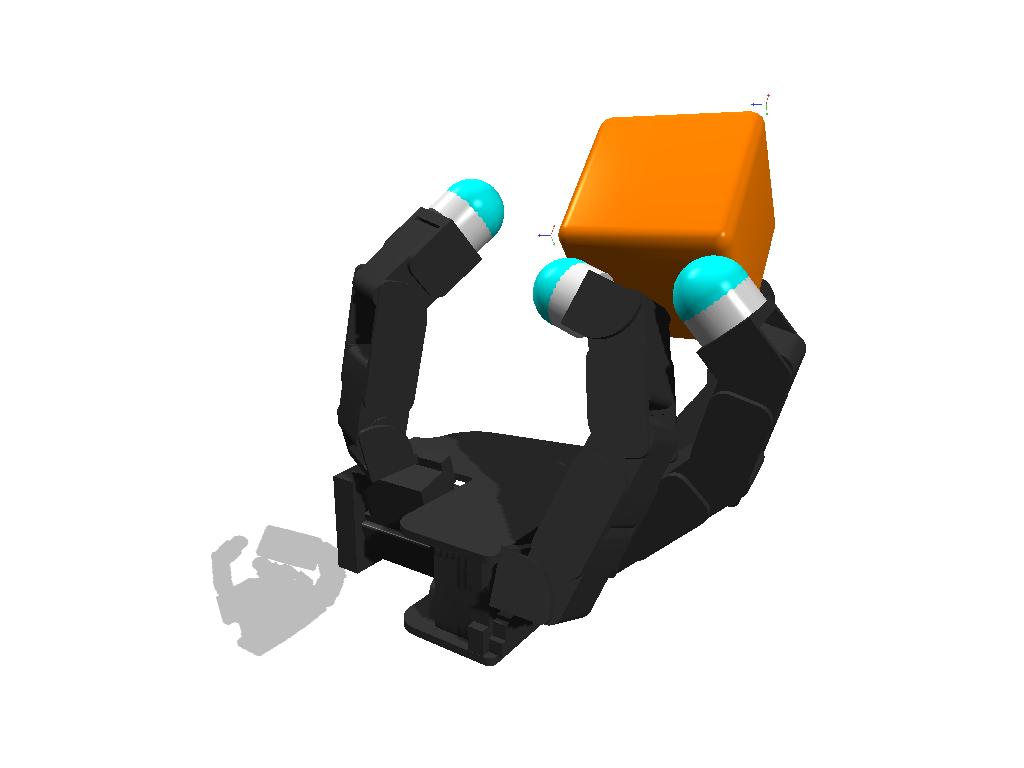}
\pic{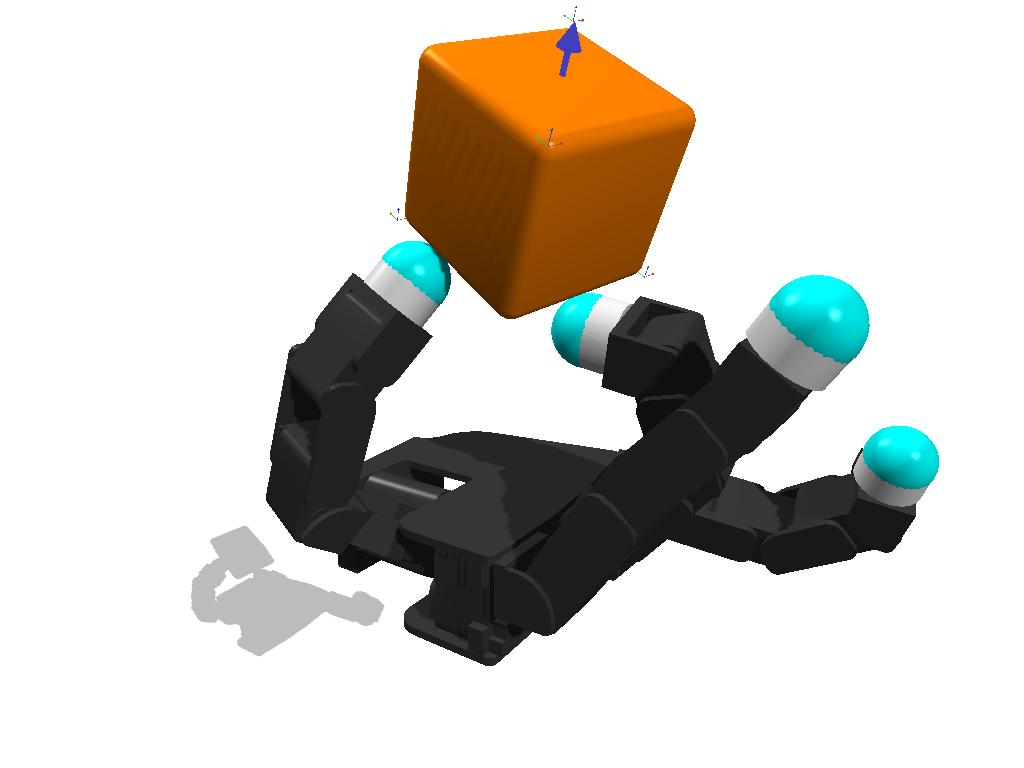}\pic{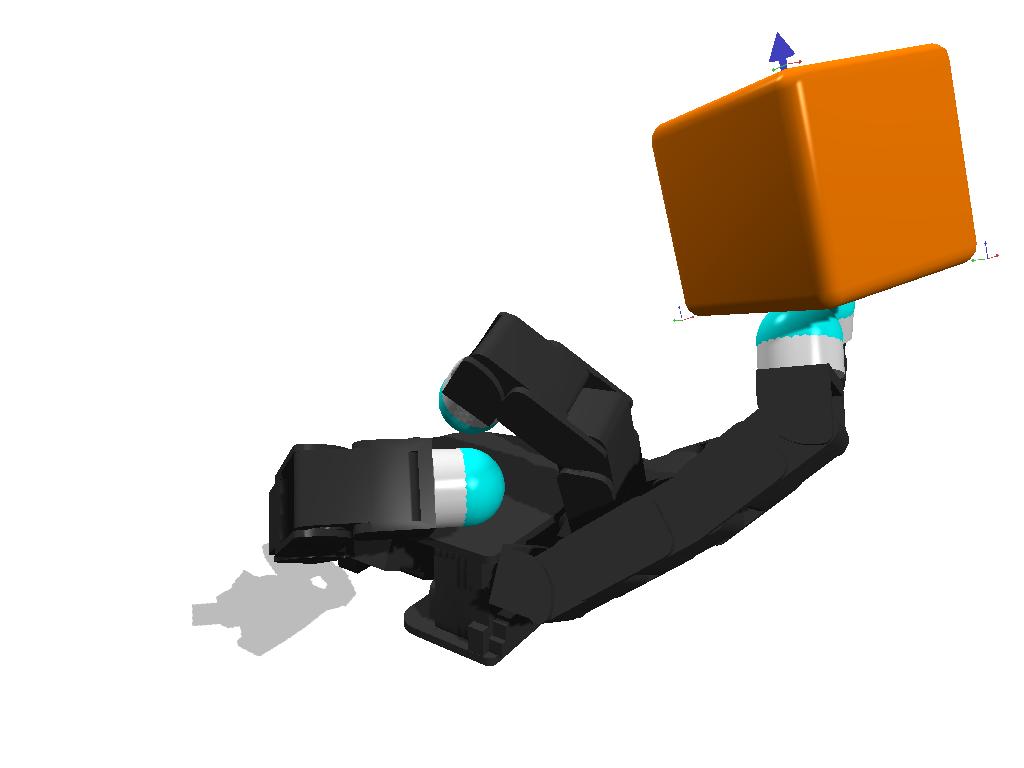}\pic{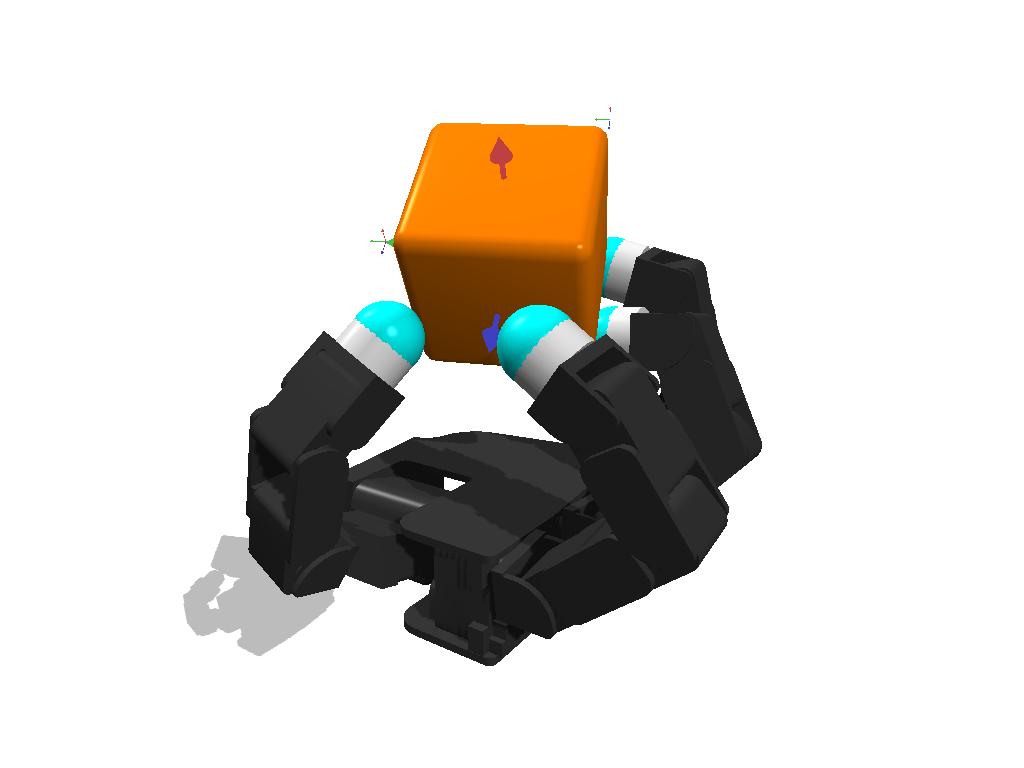}\pic{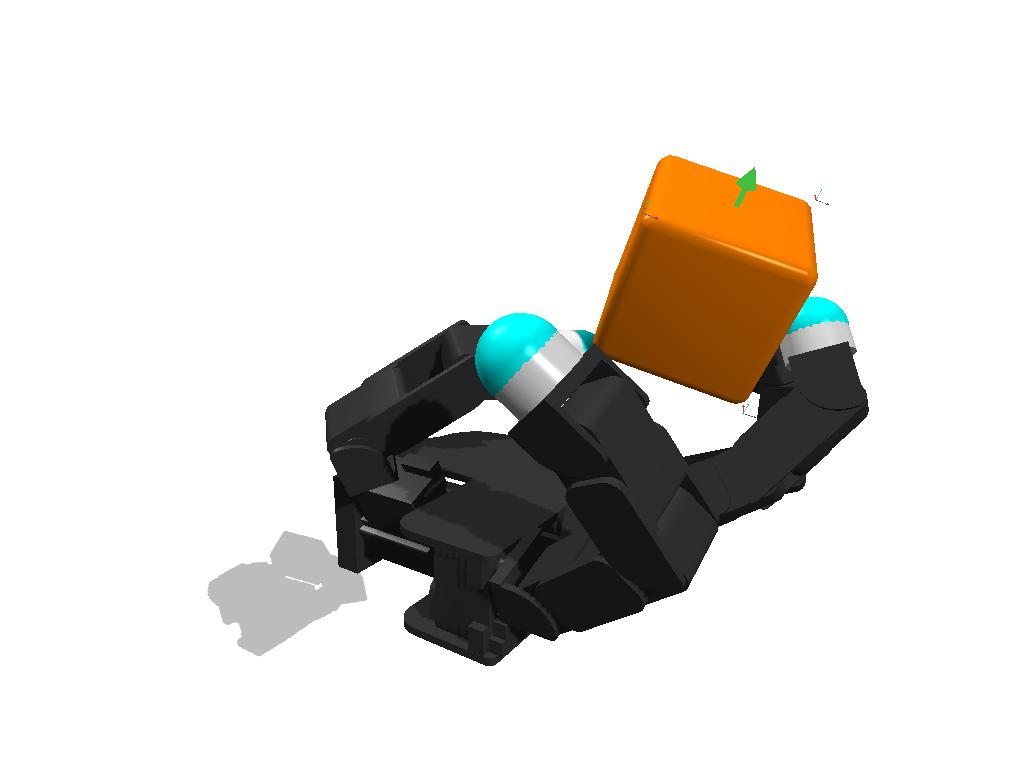}
\pic{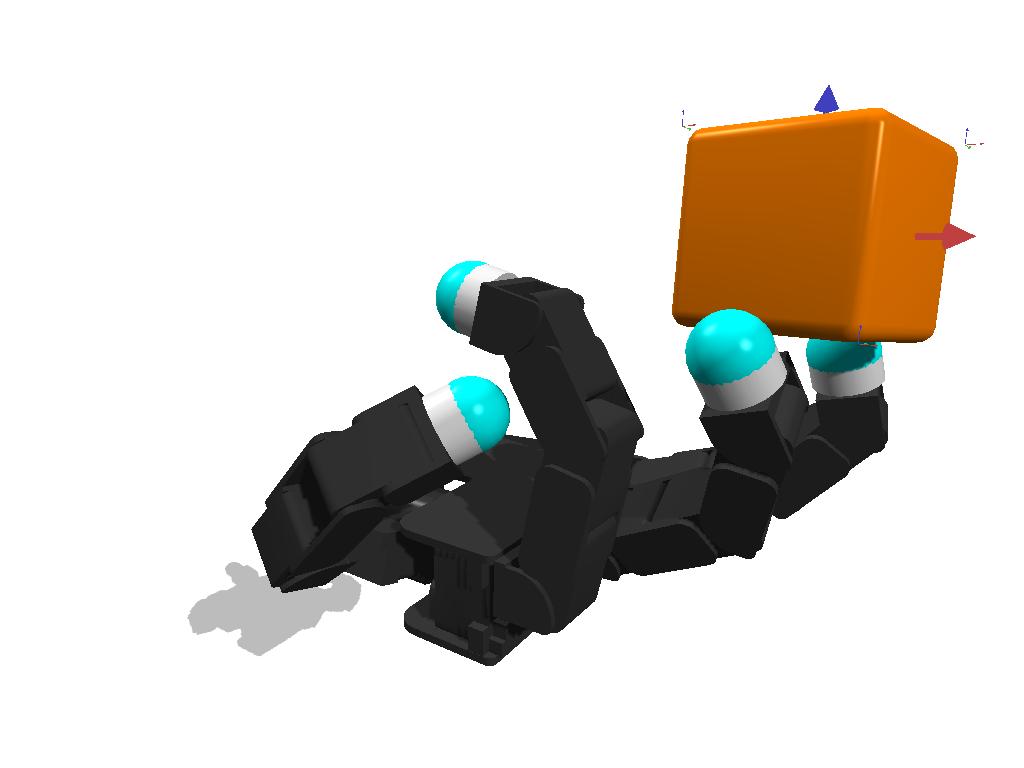}\pic{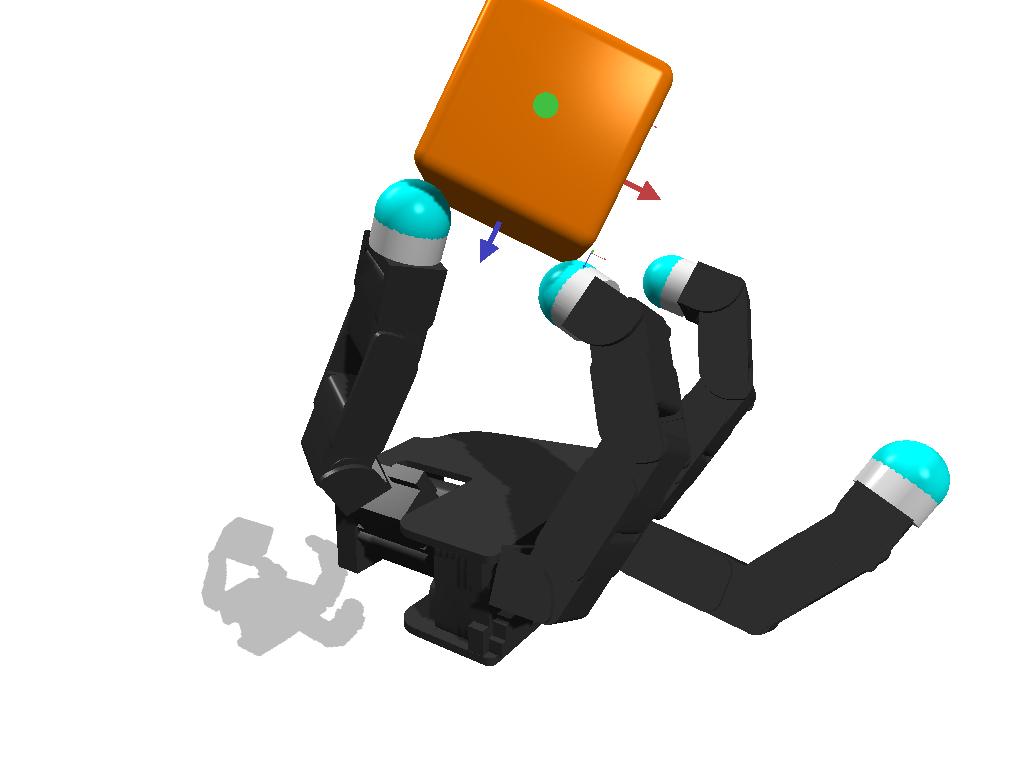}\pic{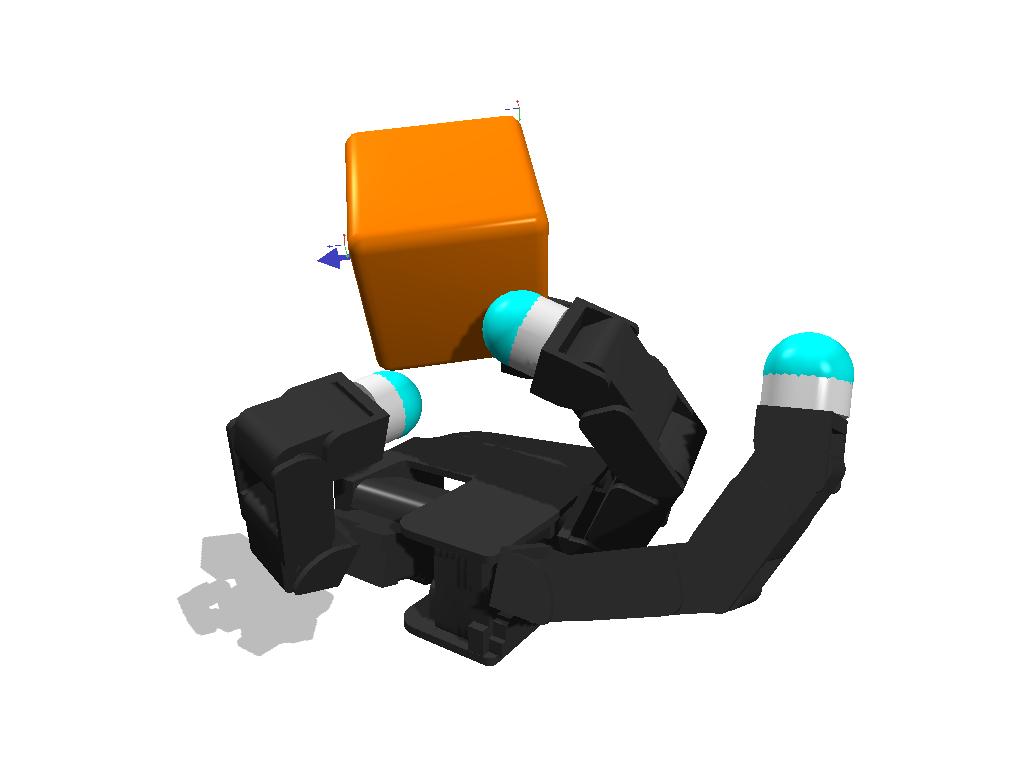}\pic{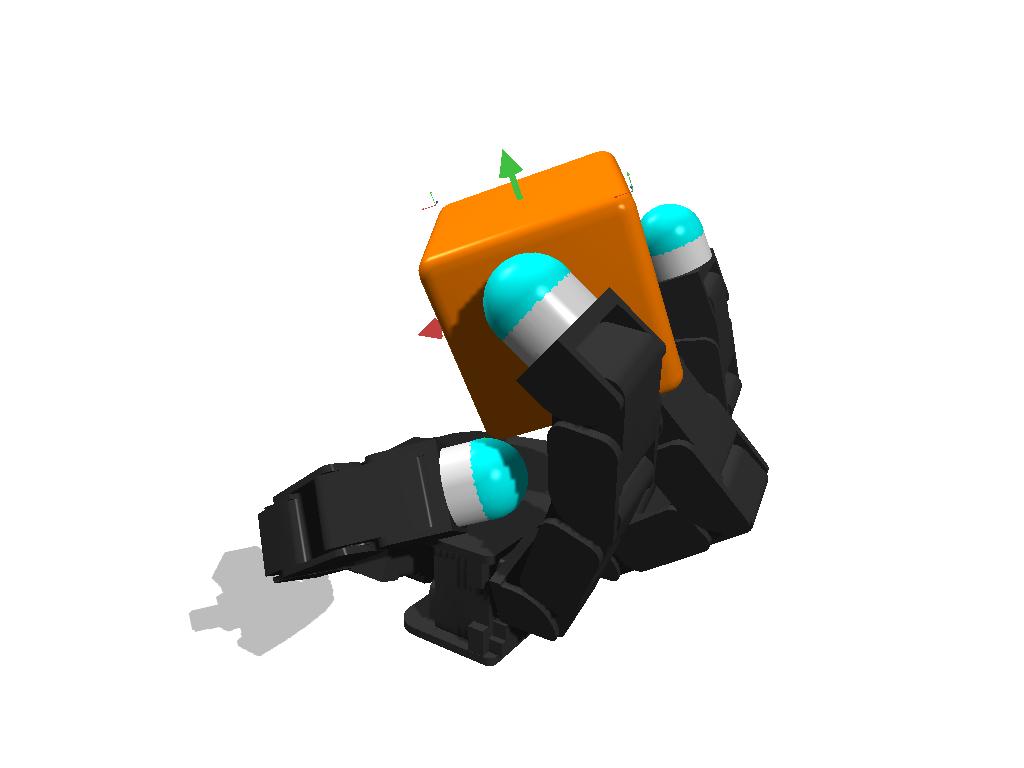}
\caption{\label{fig:allegro_cube}
  Constrained samples from the panda cube domain.}
\end{figure}

\begin{figure}[p]\centering
\newcommand{\pic}[1]{\includegraphics[width=.25\columnwidth,trim={0 0 0 25pt},clip]{data/configs/images/allegroDown_cube_sample_0000#1.jpg}}
\pic{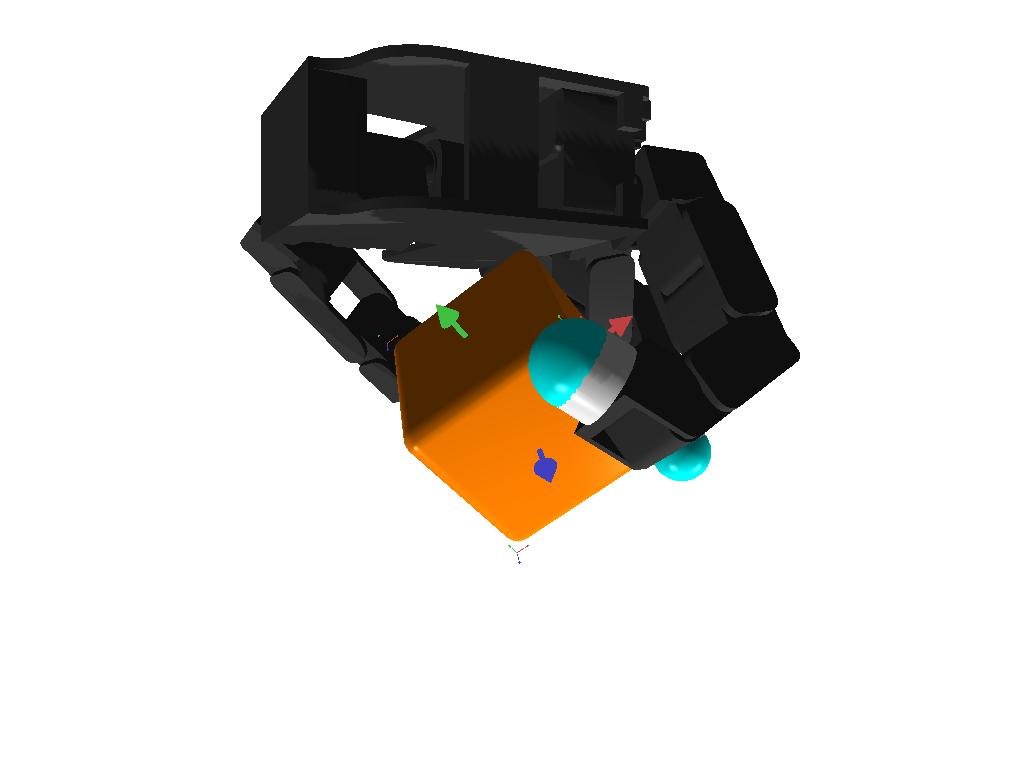}\pic{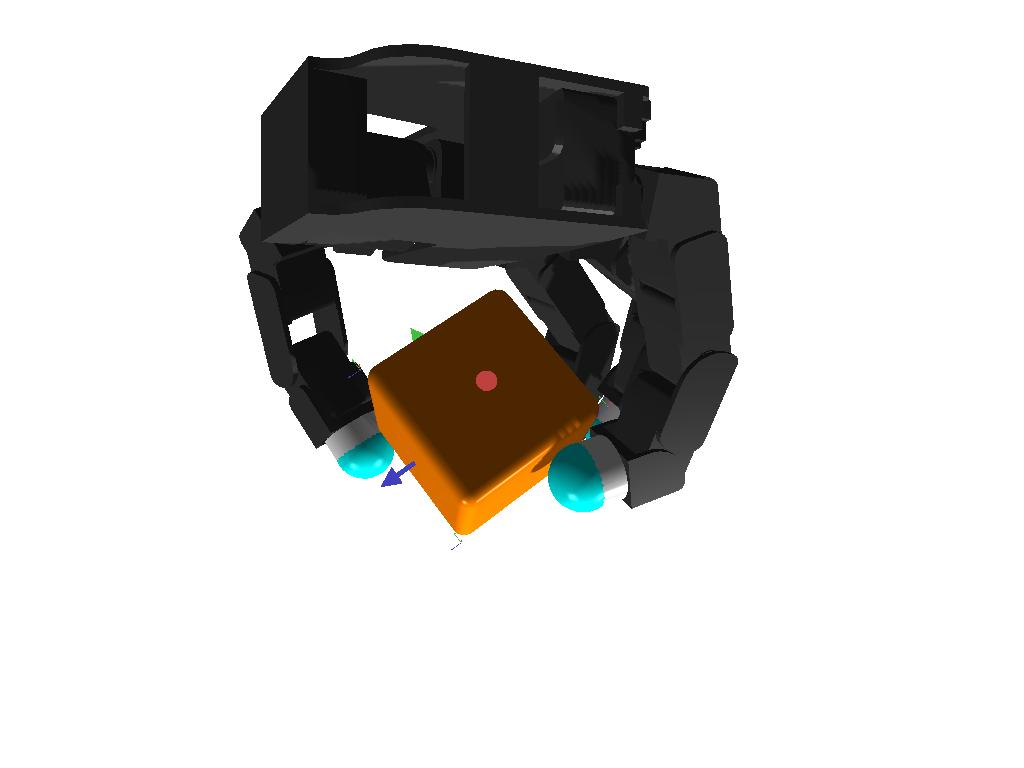}\pic{02}\pic{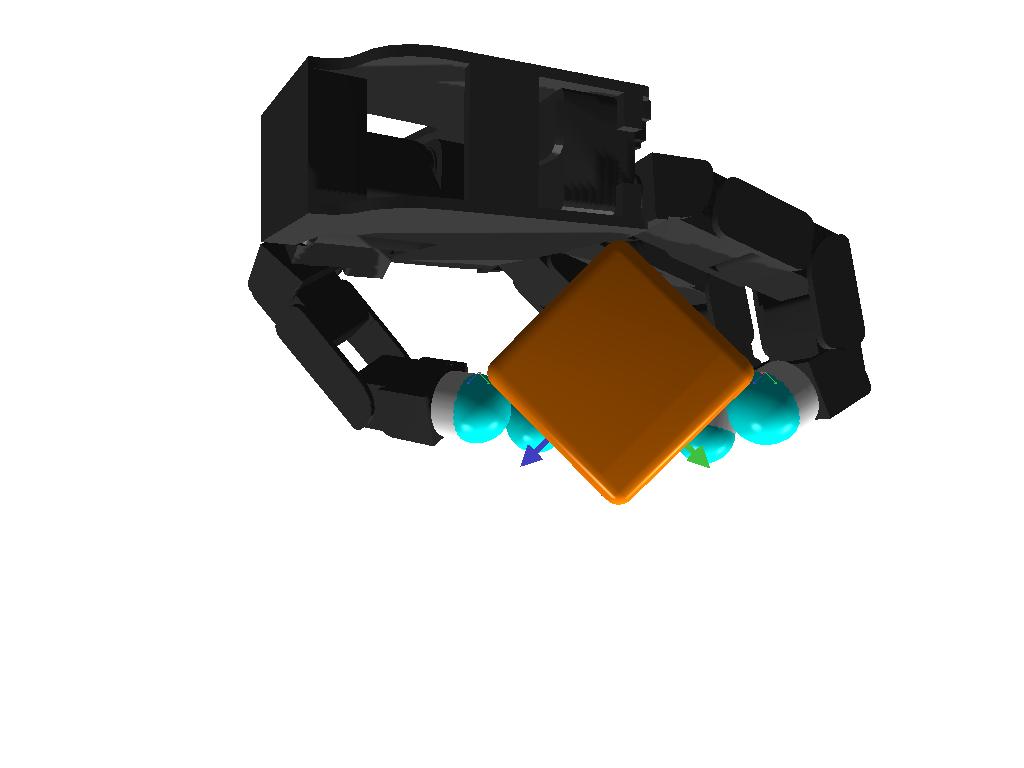}
\pic{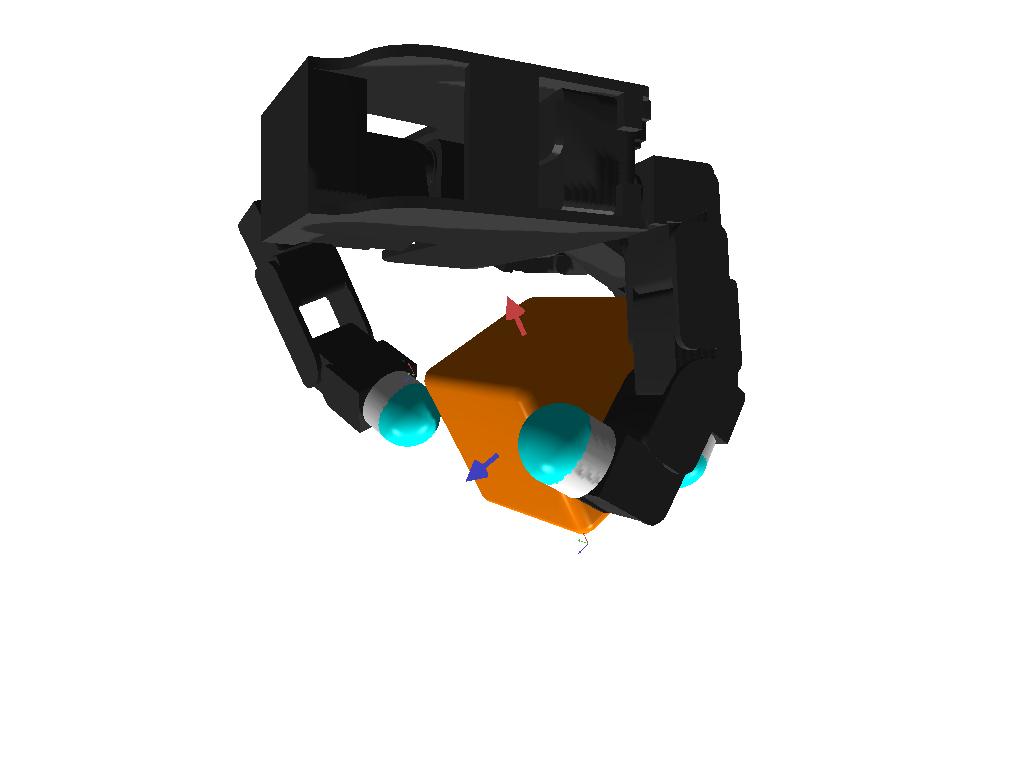}\pic{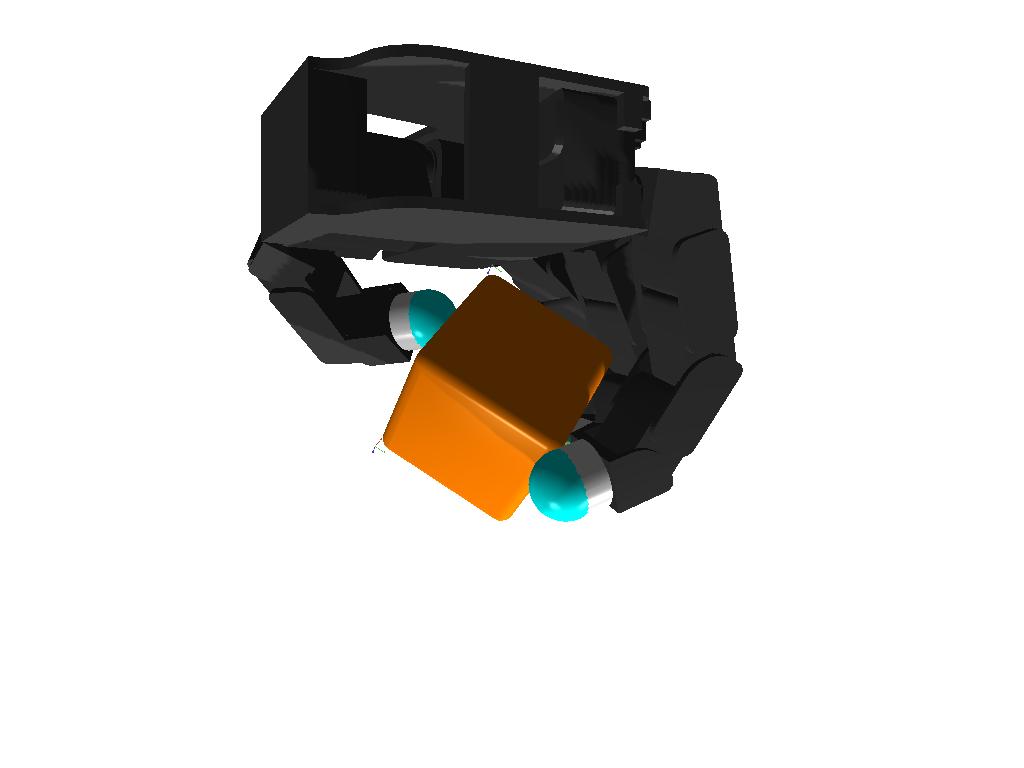}\pic{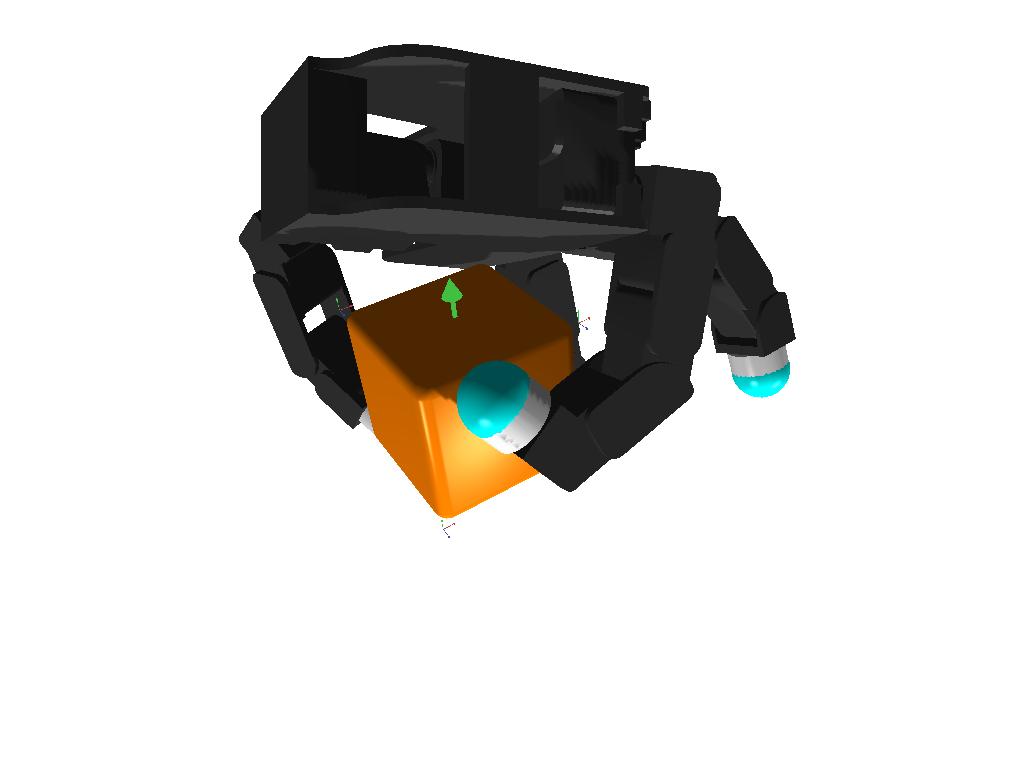}\pic{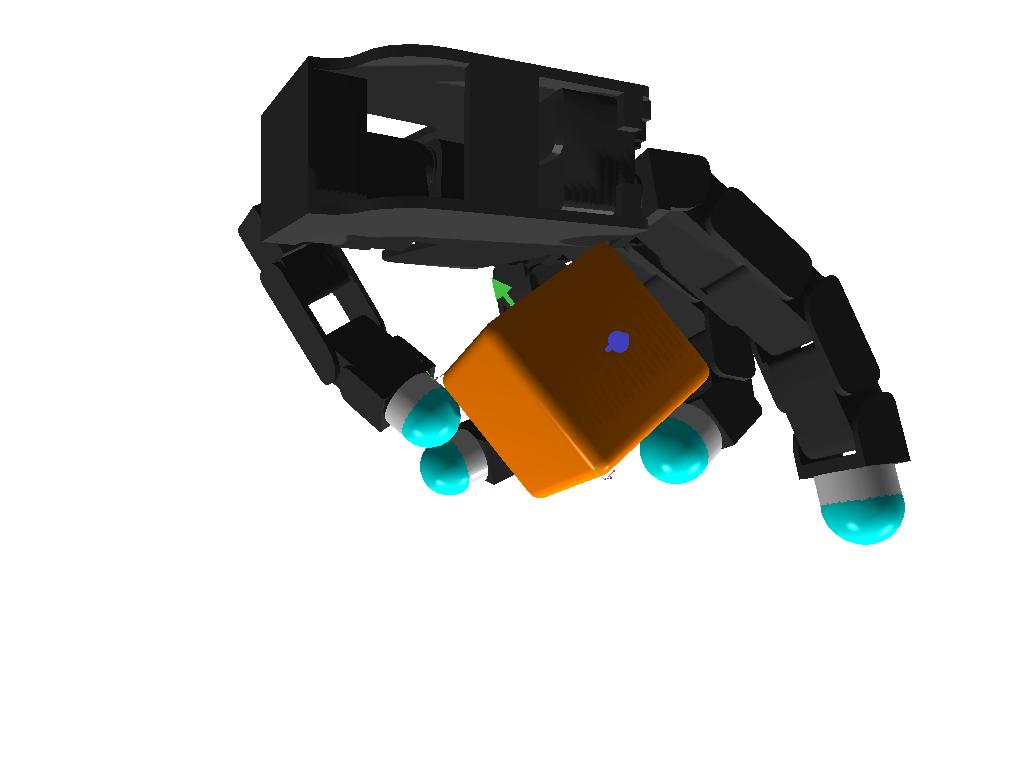}
\pic{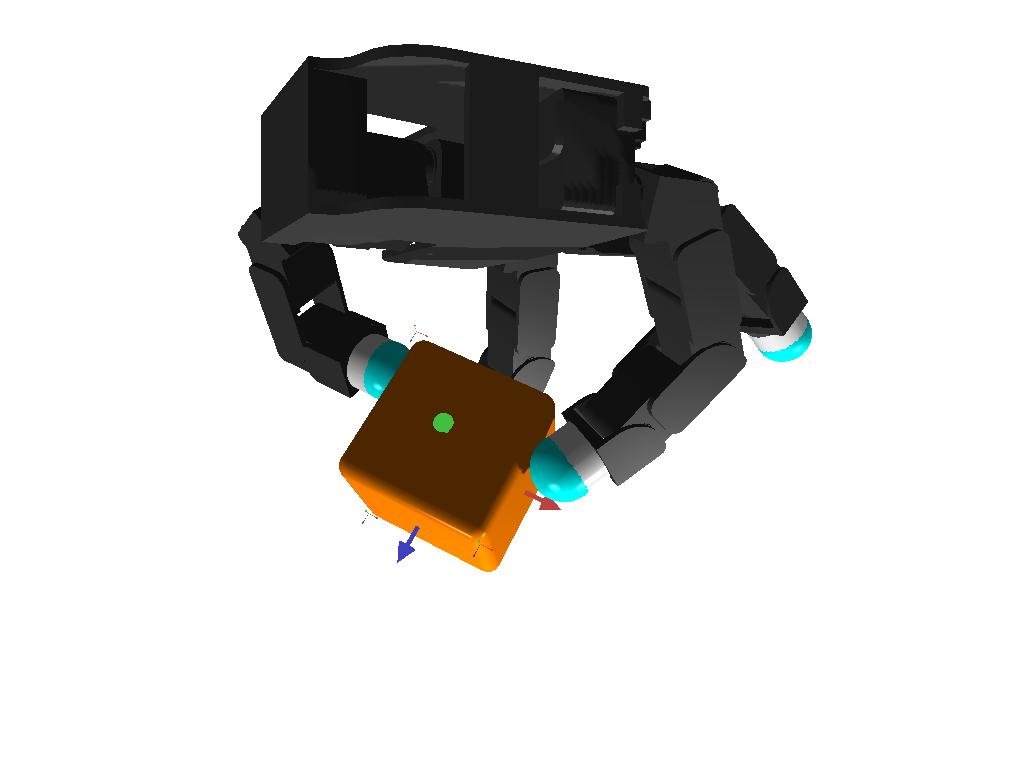}\pic{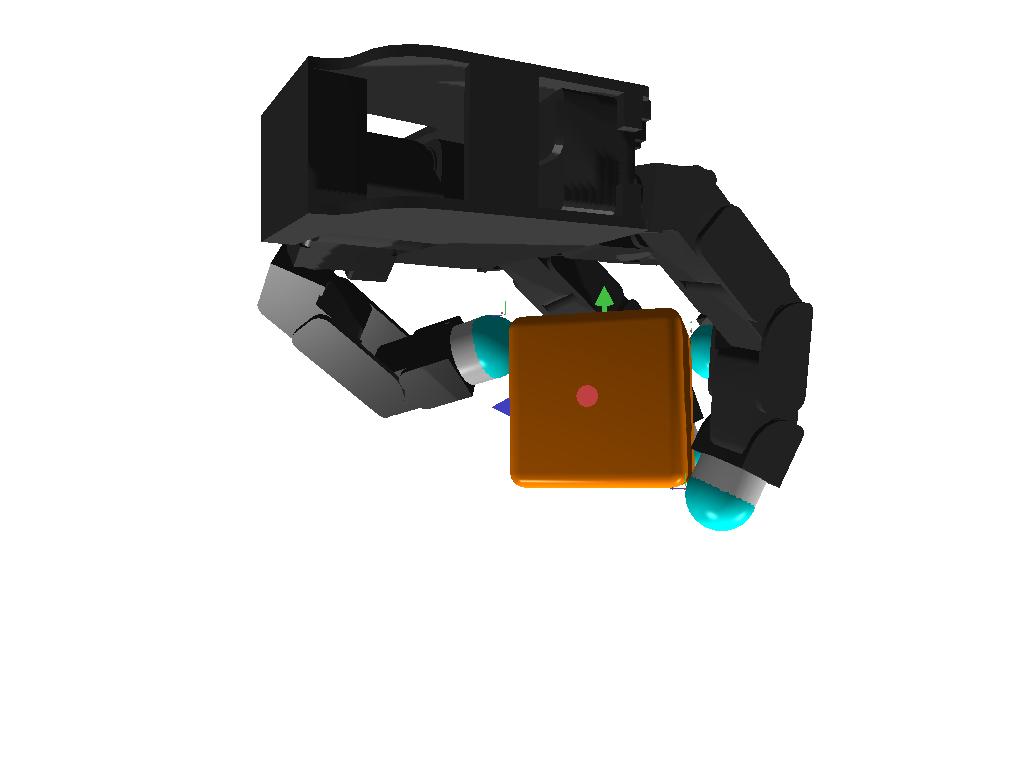}\pic{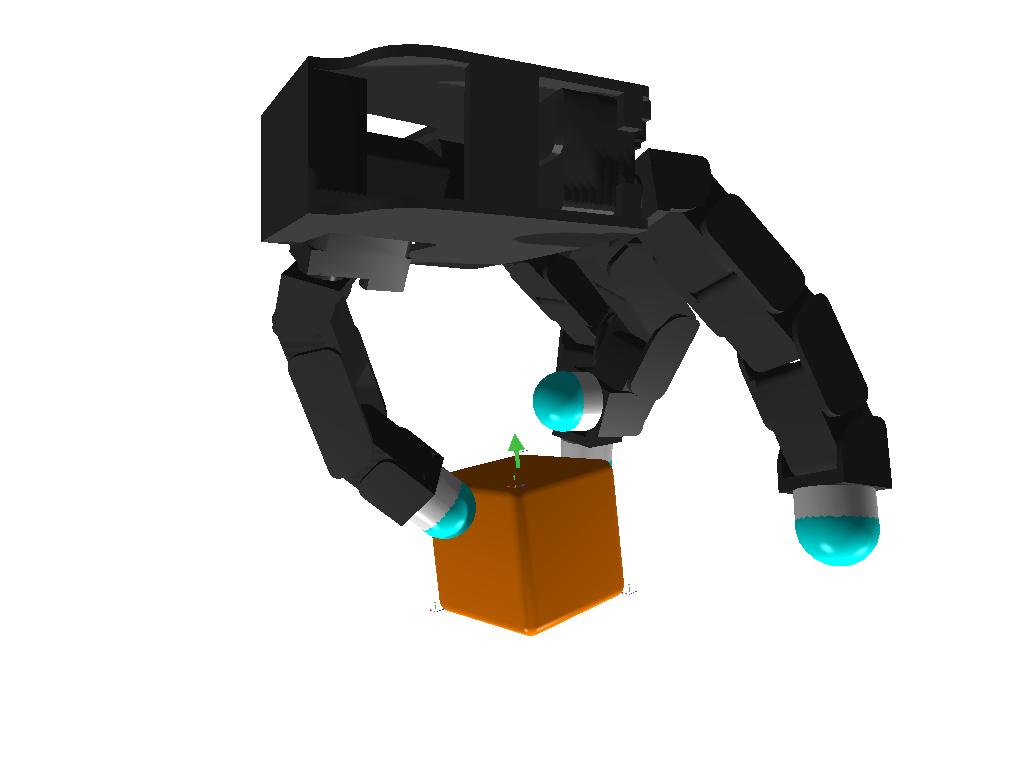}\pic{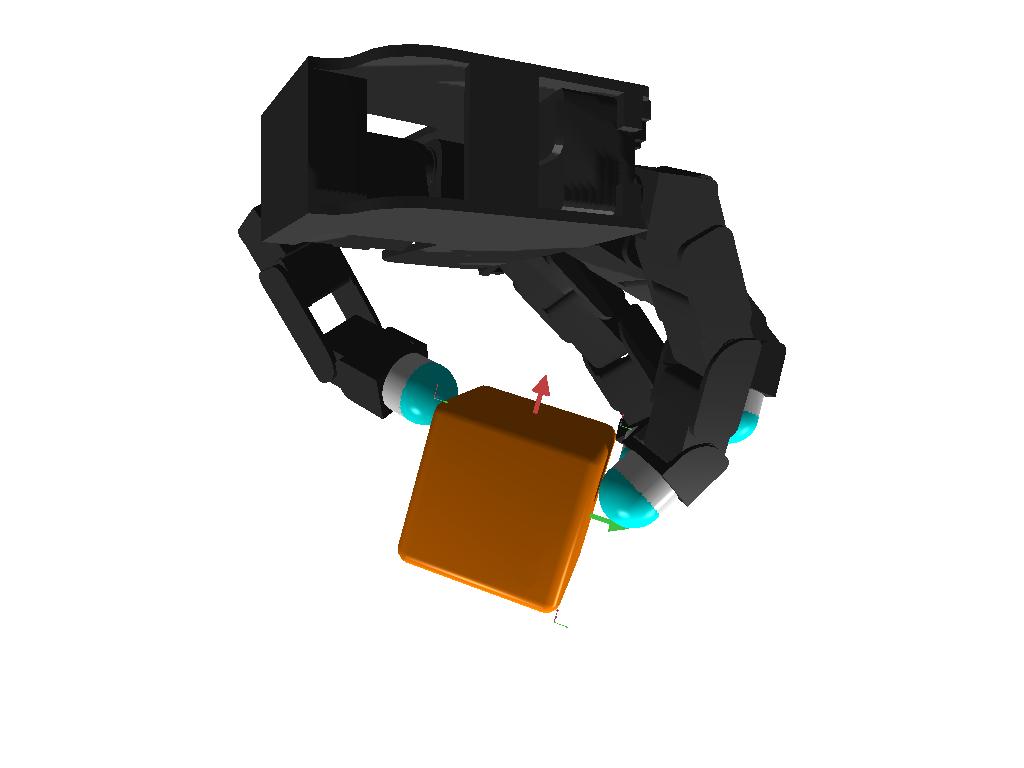}
\pic{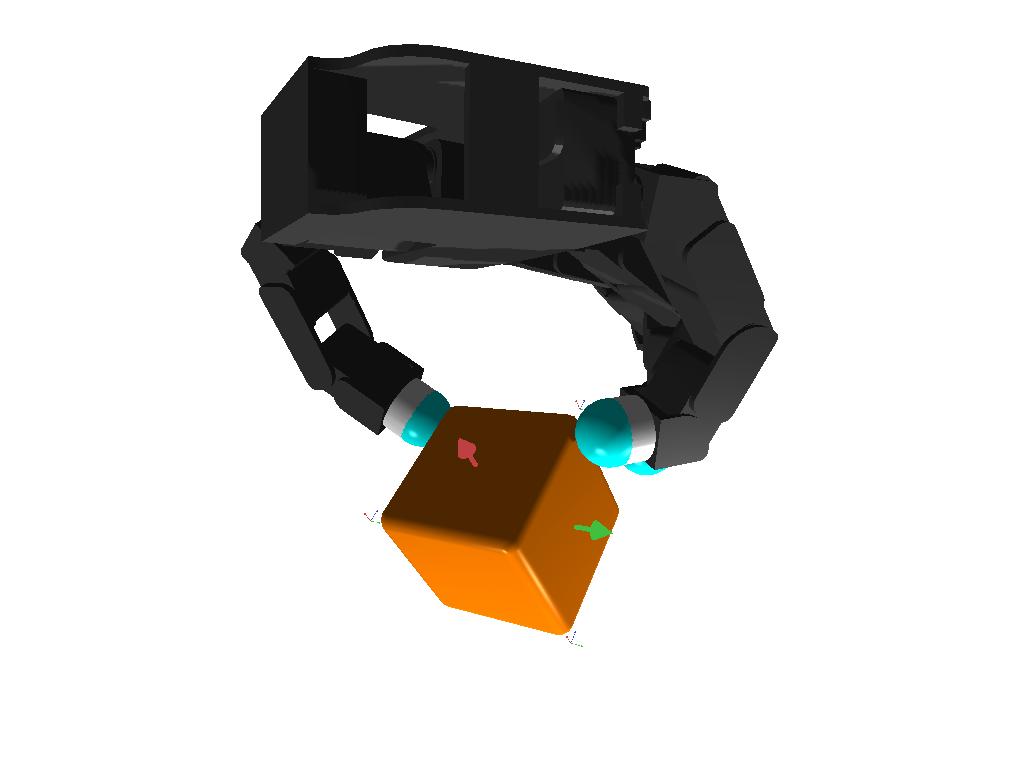}\pic{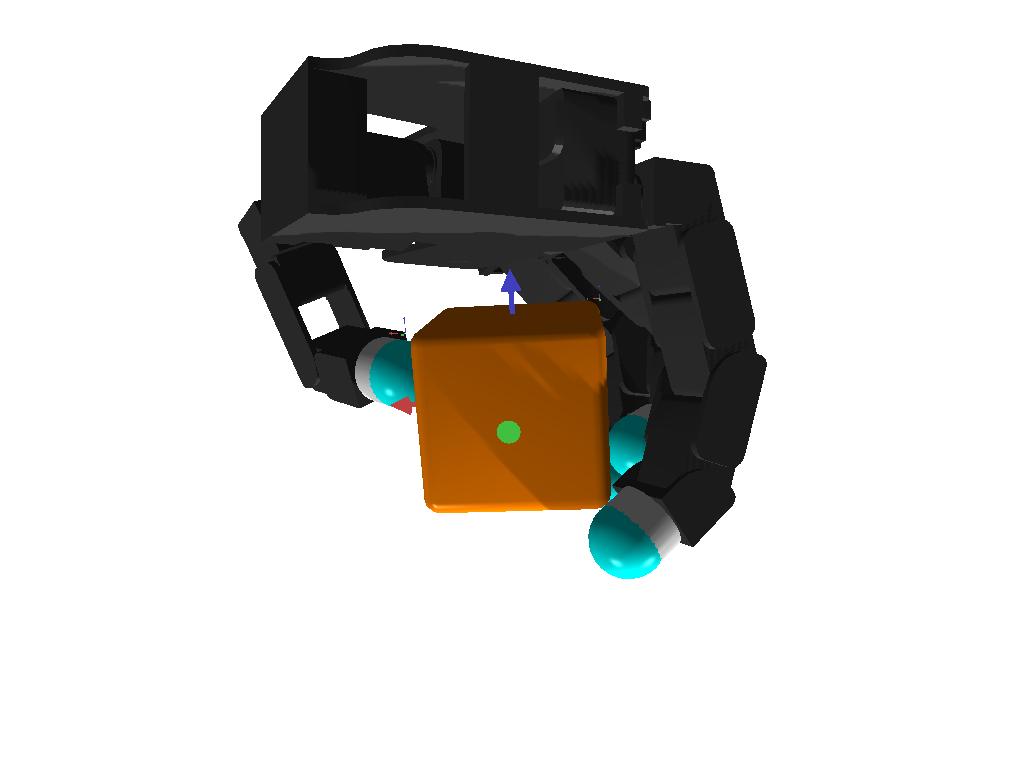}\pic{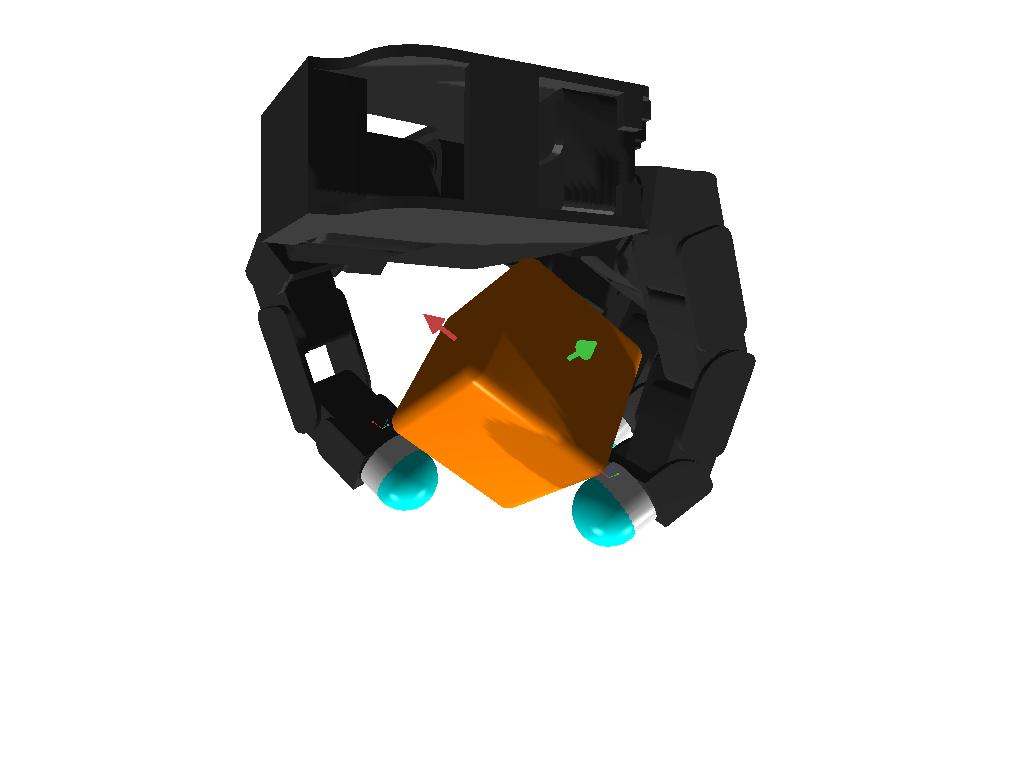}\pic{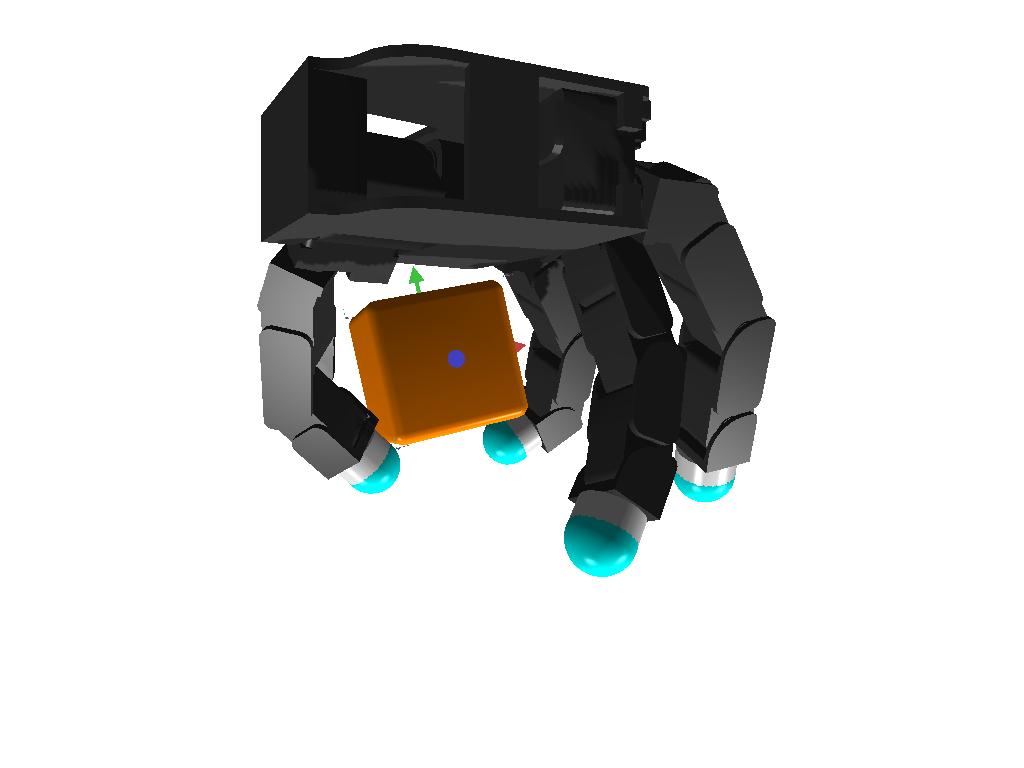}
\pic{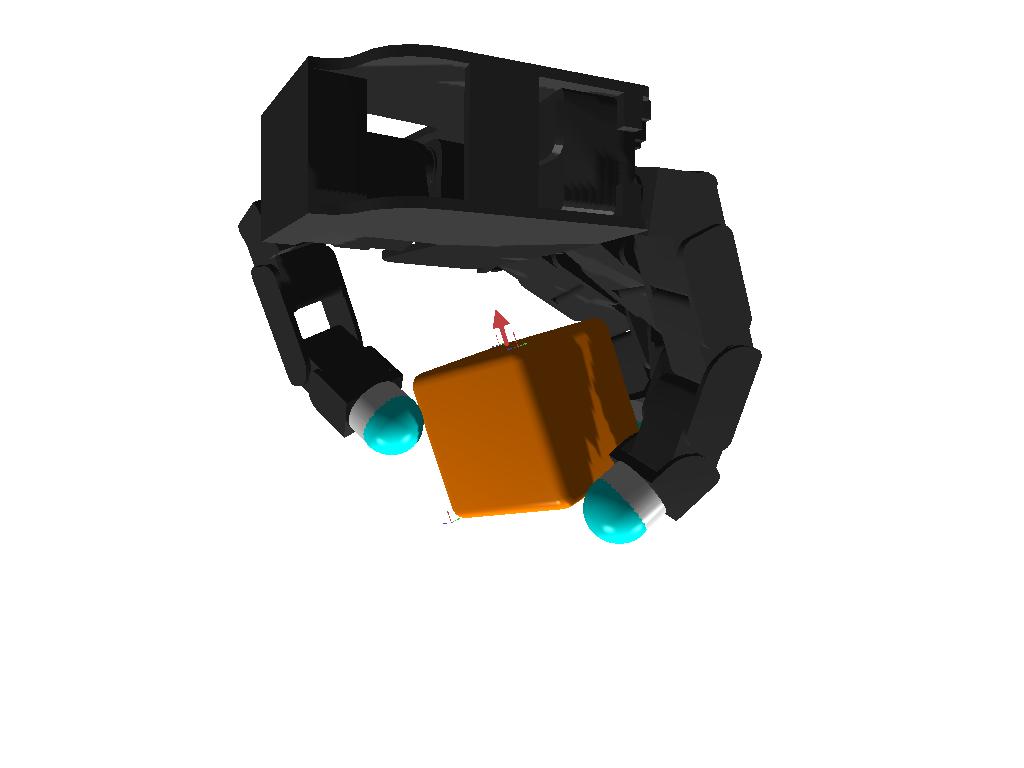}\pic{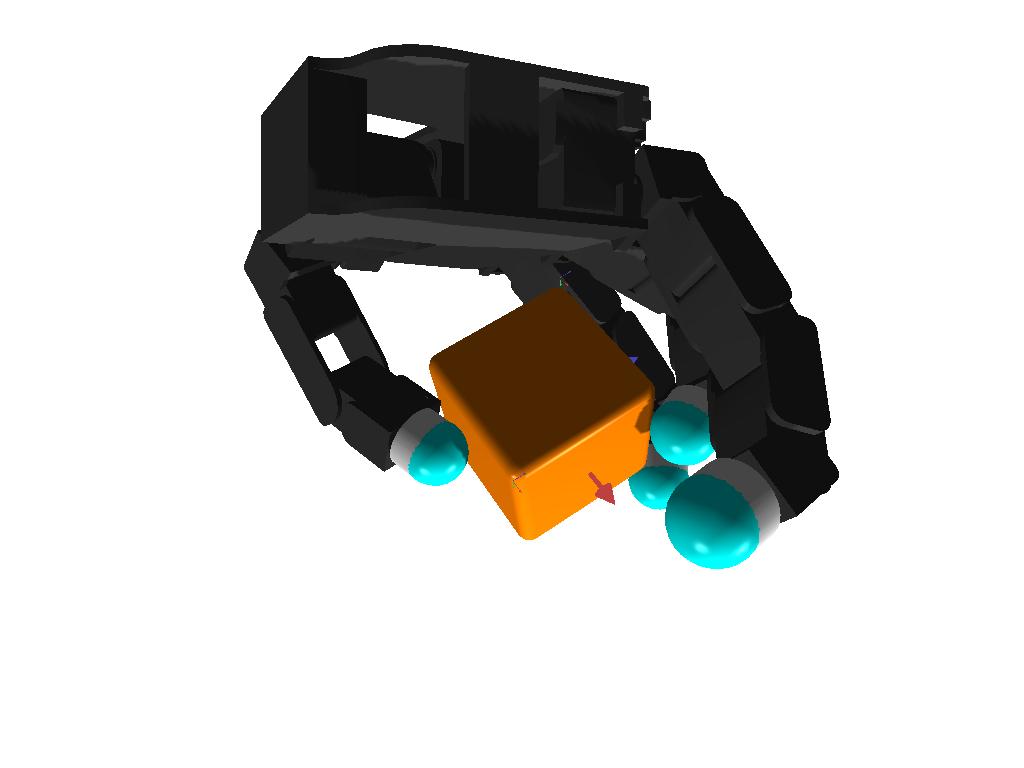}\pic{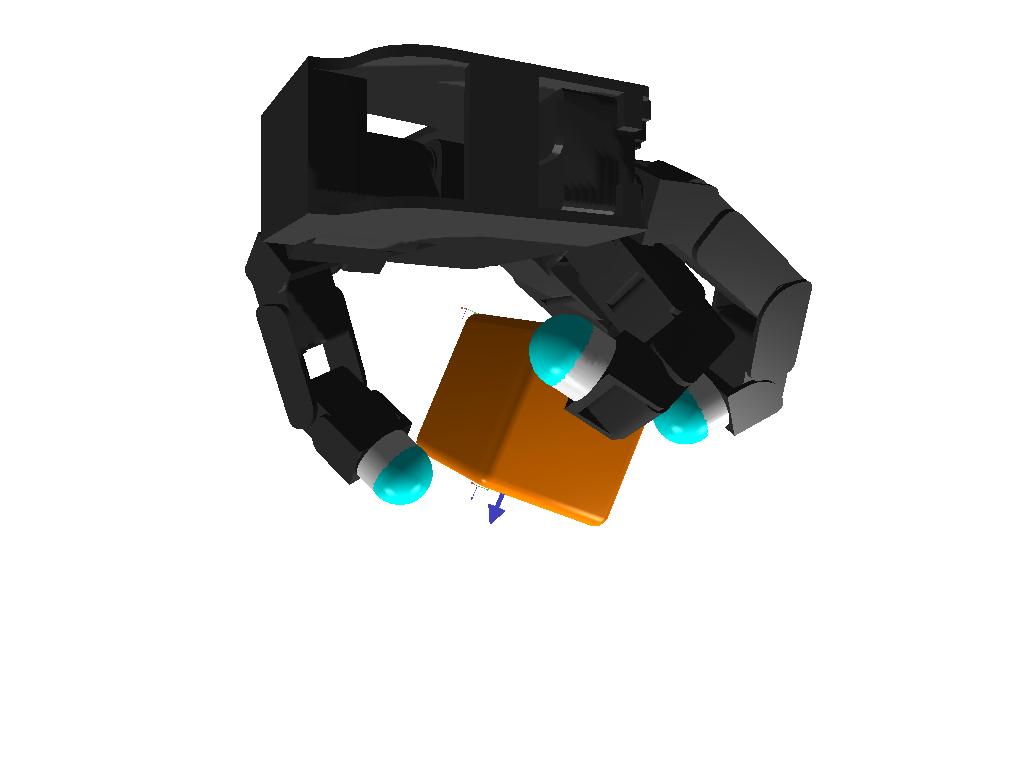}\pic{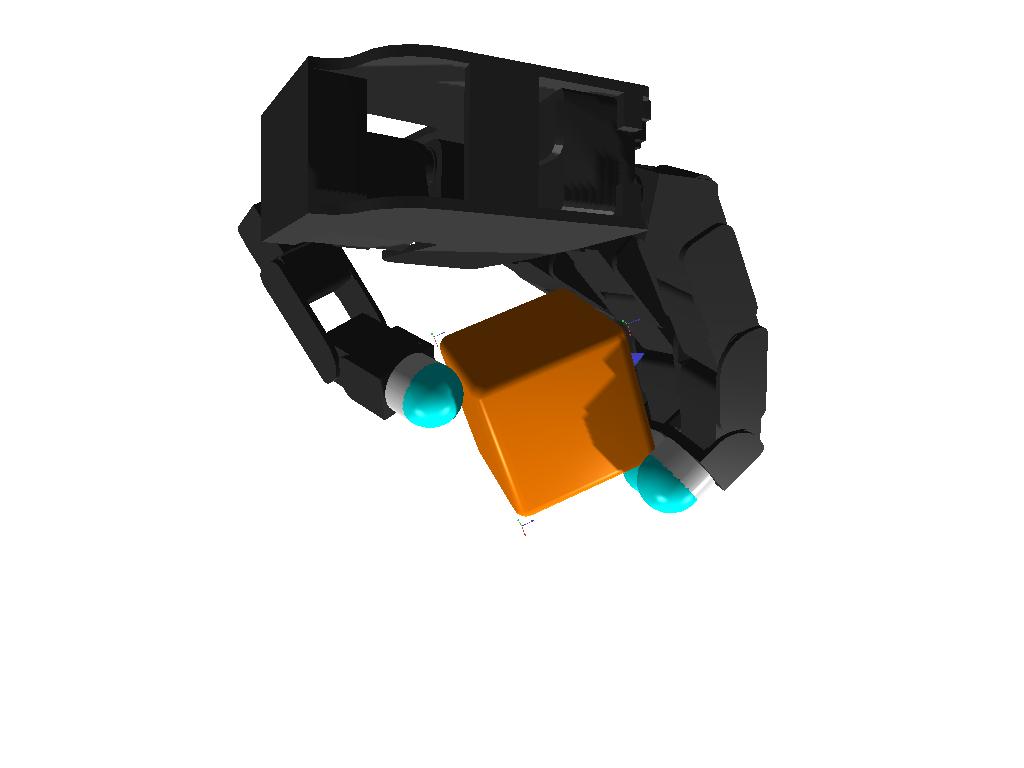}
\pic{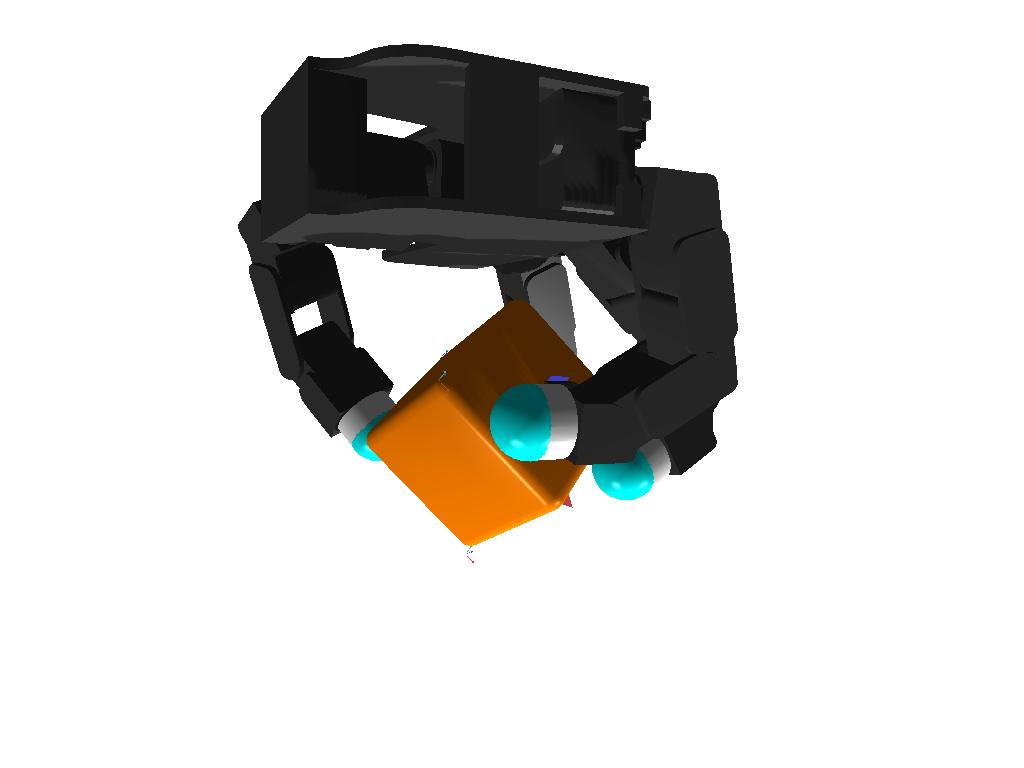}\pic{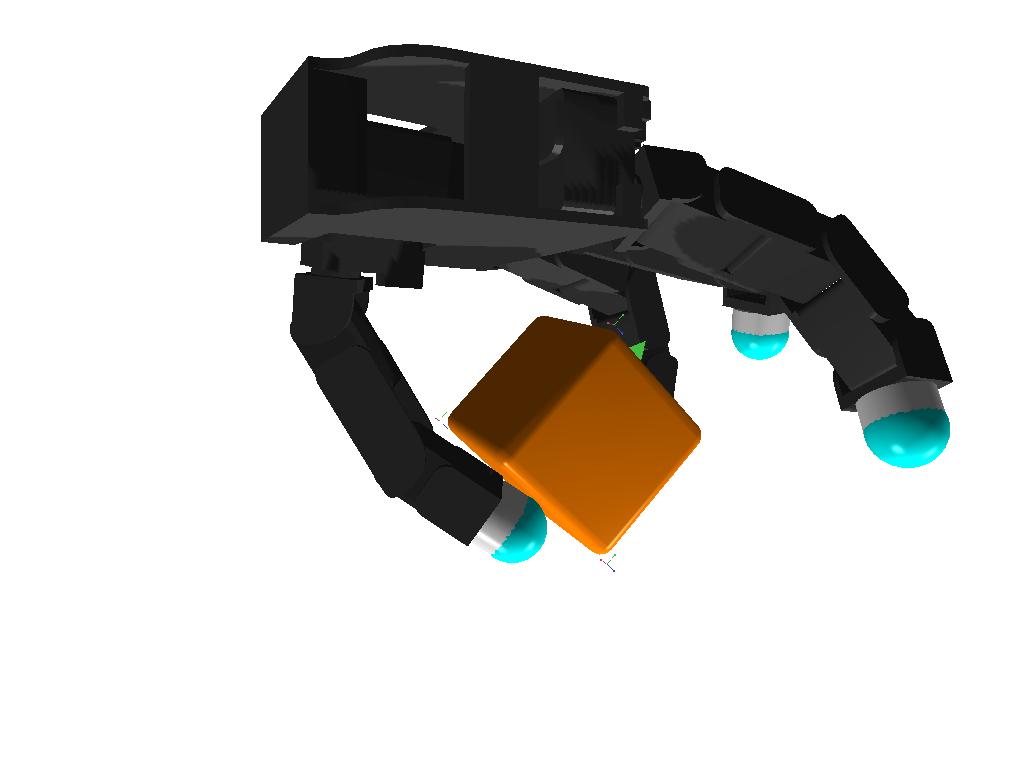}\pic{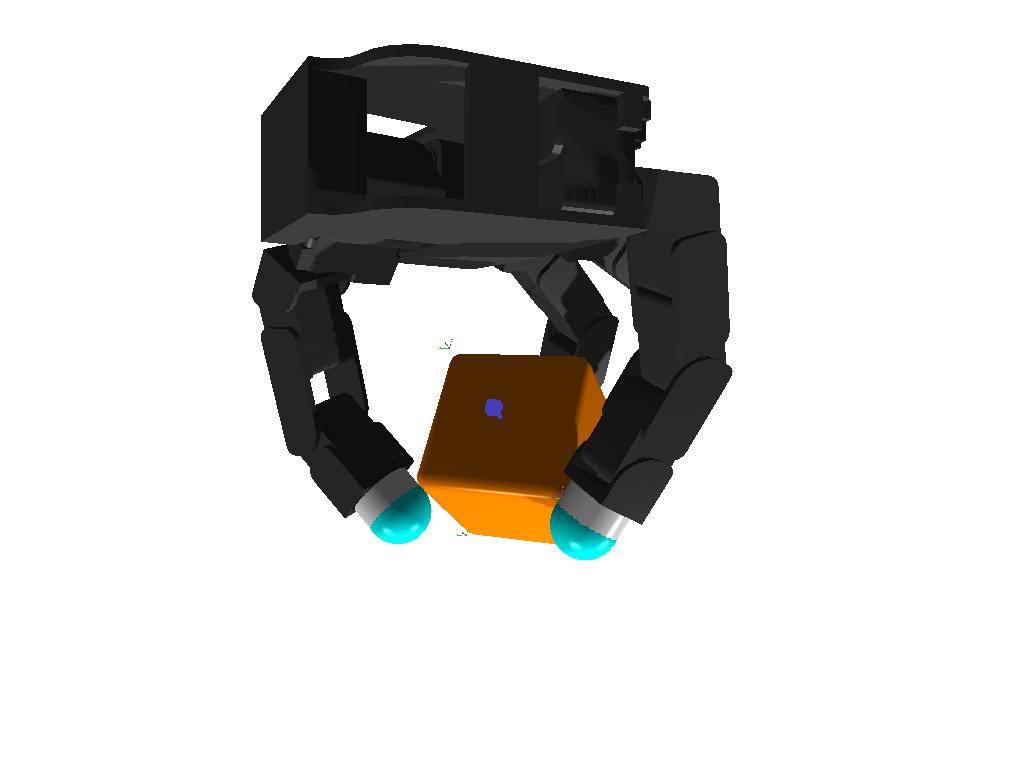}\pic{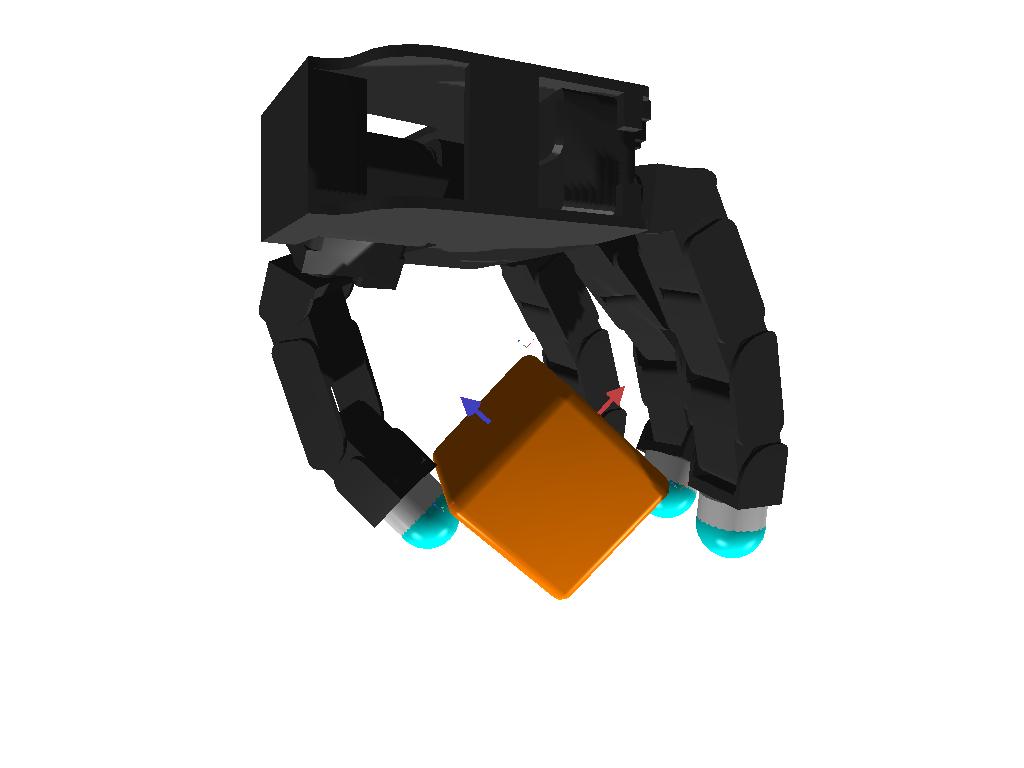}
\pic{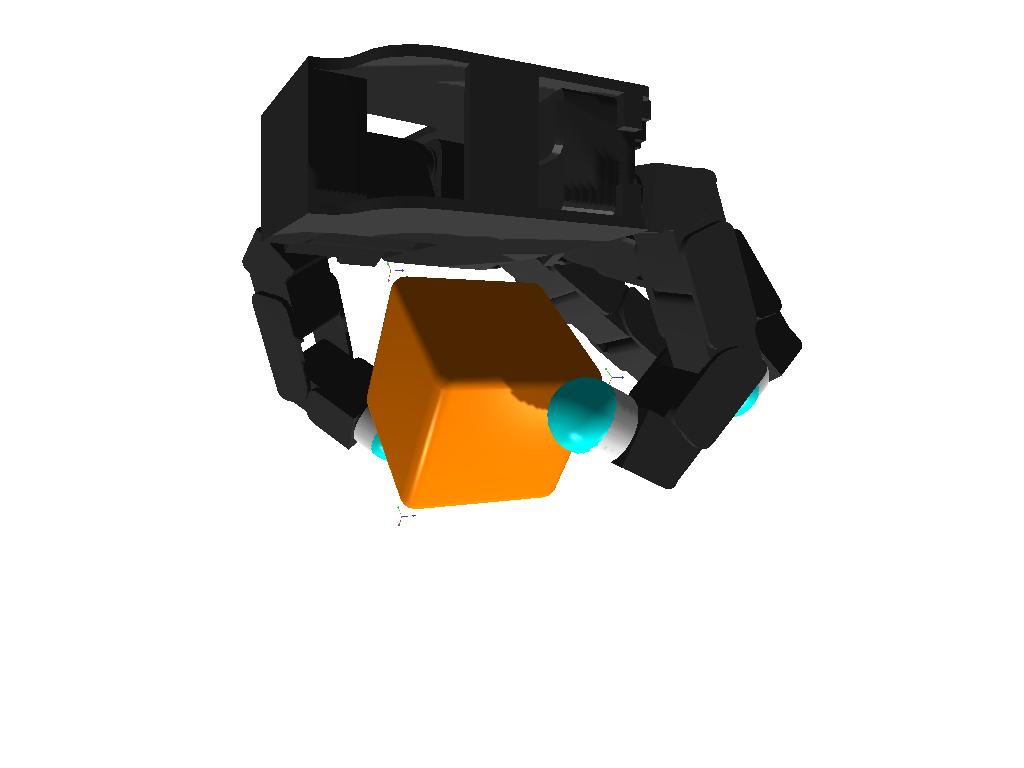}\pic{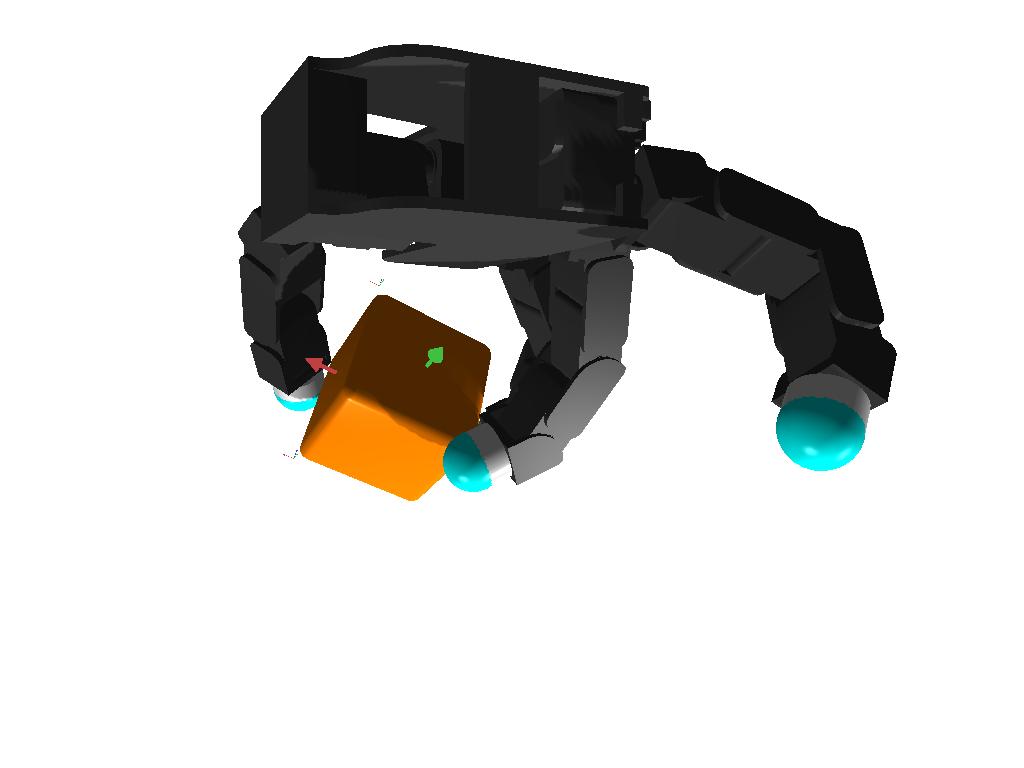}\pic{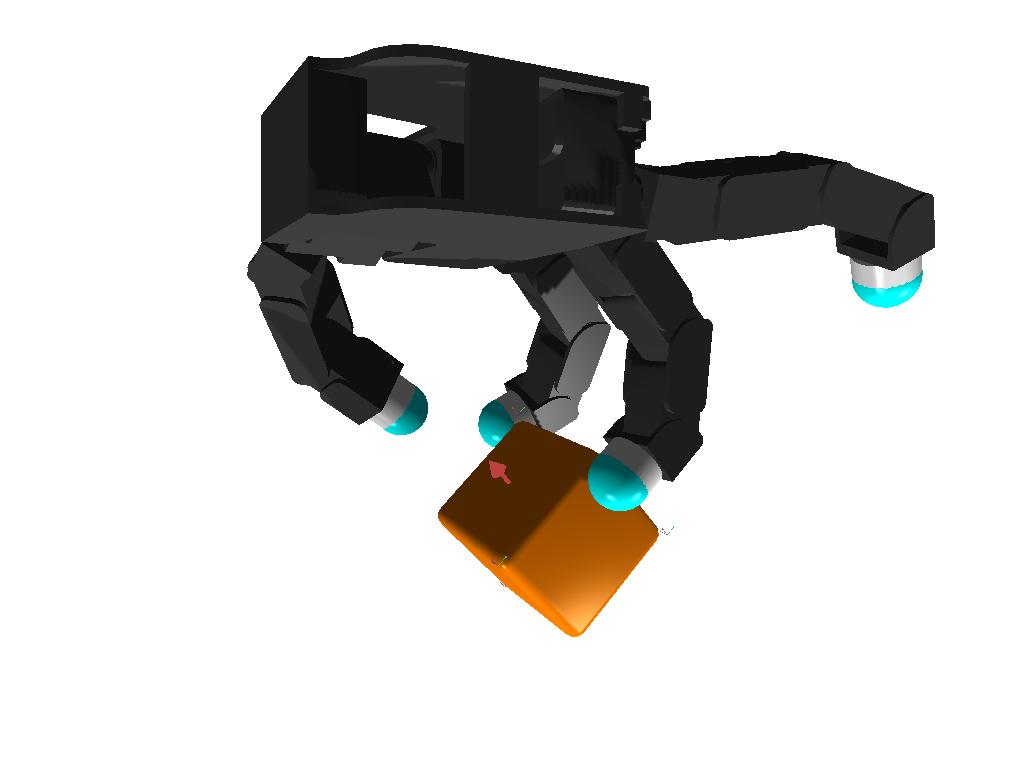}\pic{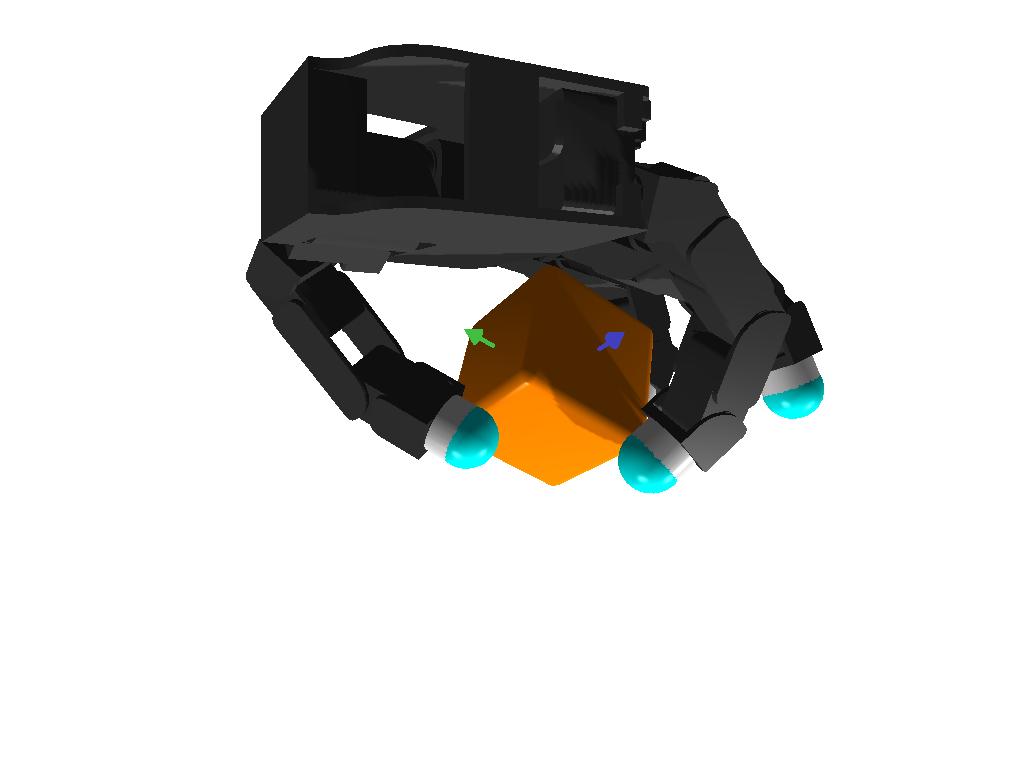}
\caption{\label{fig:allegroDown_cube}
  Constrained samples from the panda cube domain.}
\end{figure}

\end{document}